\pdfoutput=1

\documentclass[11pt]{article}

\usepackage[]{acl}

\usepackage{times}
\usepackage{latexsym}
\PassOptionsToPackage{breaklinks}{hyperref}

\usepackage[T1]{fontenc}

\usepackage[utf8]{inputenc}

\usepackage{microtype}

\usepackage{amsmath,amsfonts,bm}

\def\eqref#1{equation~\ref{#1}}

\def\1{\bm{1}}

\DeclareMathAlphabet{\mathsfit}{\encodingdefault}{\sfdefault}{m}{sl}
\SetMathAlphabet{\mathsfit}{bold}{\encodingdefault}{\sfdefault}{bx}{n}

\usepackage{url}

\usepackage{algorithm}
\usepackage{algorithmic}
\usepackage{textcomp}
\usepackage{xcolor}
\usepackage{array}
\usepackage{pifont}
\usepackage{tabularx}
\usepackage{adjustbox}
\usepackage{booktabs,dcolumn}
\usepackage{makecell}
\usepackage{multirow}
\usepackage{multicol}
\usepackage{tablefootnote}
\usepackage{threeparttable}
\usepackage{float}
\usepackage{subfigure}
\usepackage{paralist}
\usepackage{comment}
\usepackage{enumerate}
\usepackage{xspace}
\usepackage{mathtools}
\usepackage{amsmath,amsthm,amsfonts,amssymb,bm,stmaryrd}
\usepackage{soul}
\usepackage{fmtcount}
\usepackage{flushend}
\usepackage{thmtools}
\usepackage{colortbl}
\usepackage{wrapfig}
\definecolor{mygray}{gray}{.9}
\definecolor{up}{RGB}{238,44,44}
\definecolor{drop}{RGB}{50,205,50}

\usepackage{balance}

\usepackage{CJKutf8}

\usepackage{ulem}

\newcommand\overconfident{\raisebox{-3.7pt}{\includegraphics[width=1.5em]{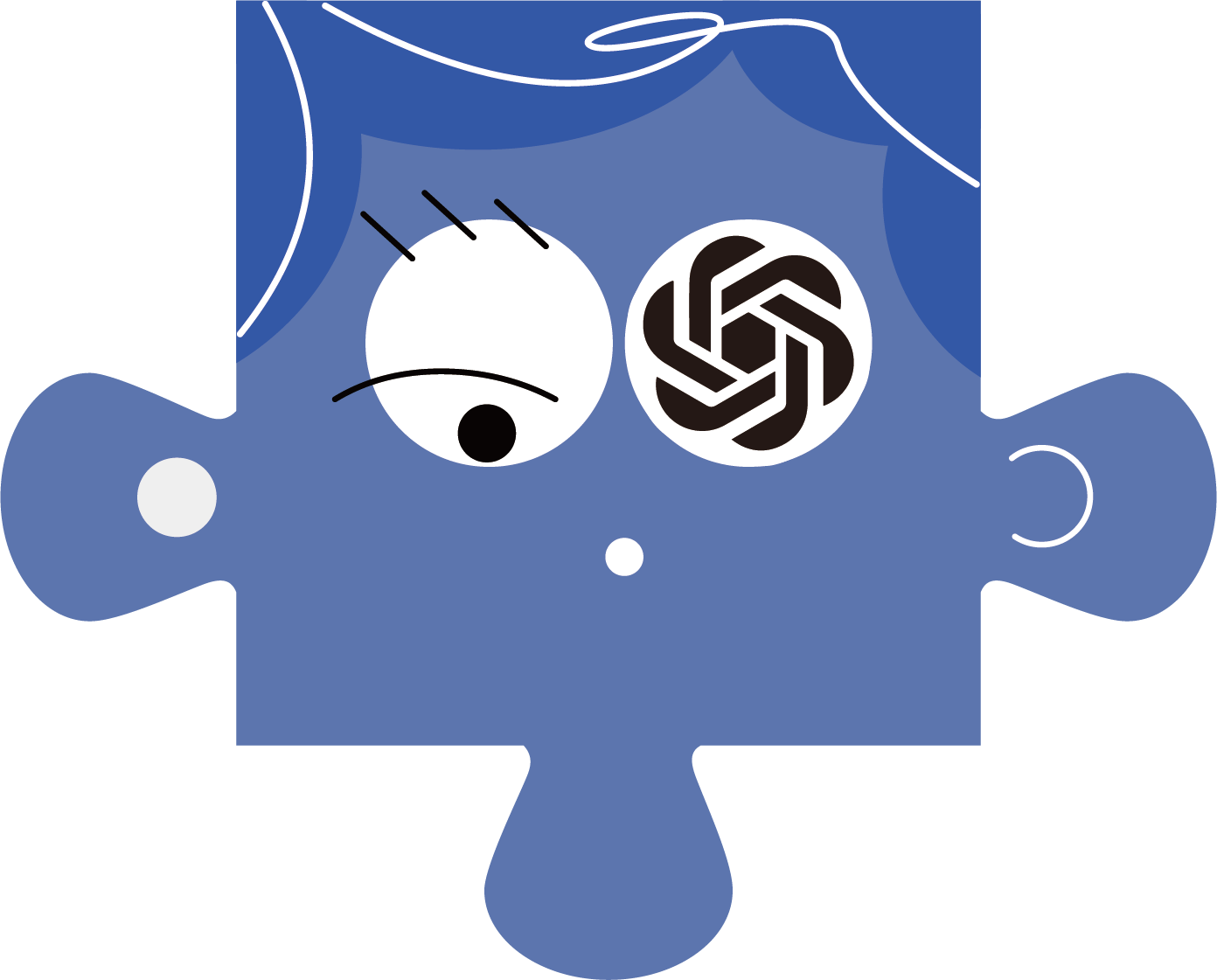}}}
\newcommand\easygoing{\raisebox{-3.7pt}{\includegraphics[width=1.3em]{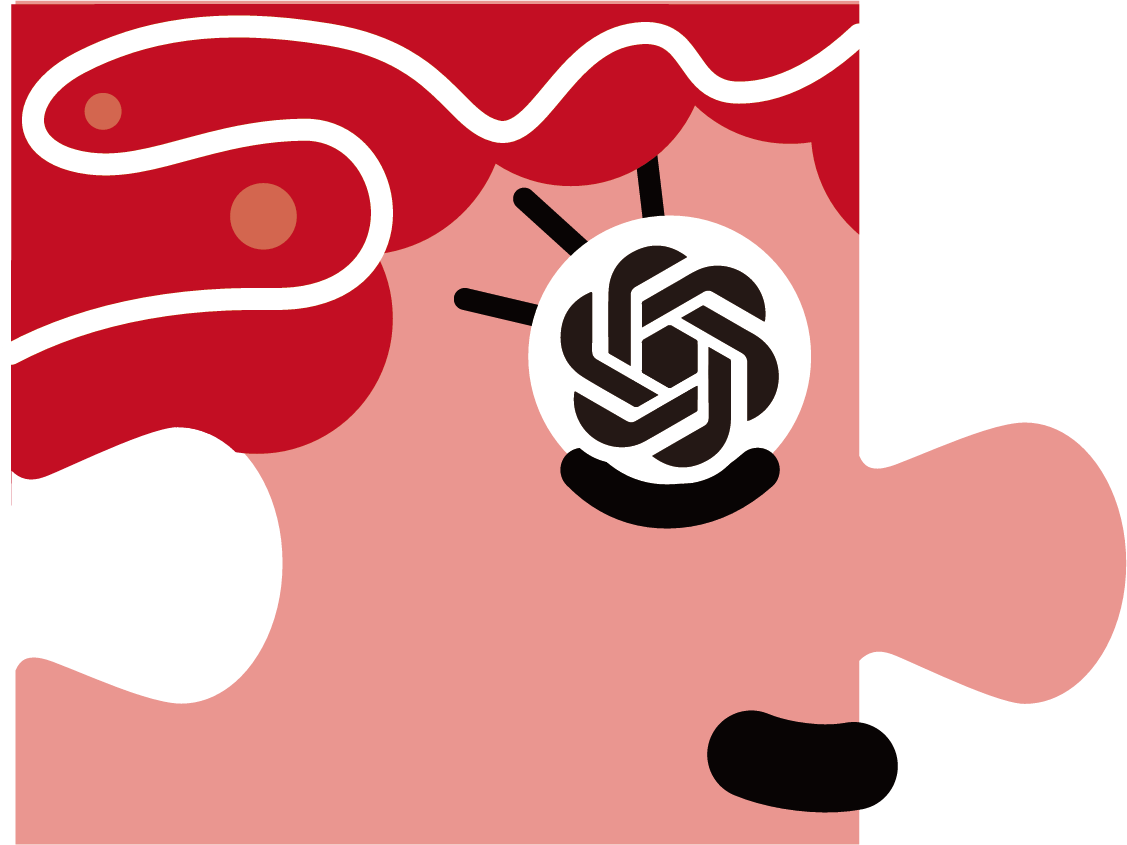}}}
\newcommand\debate{\raisebox{-3.7pt}{\includegraphics[width=1.3em]{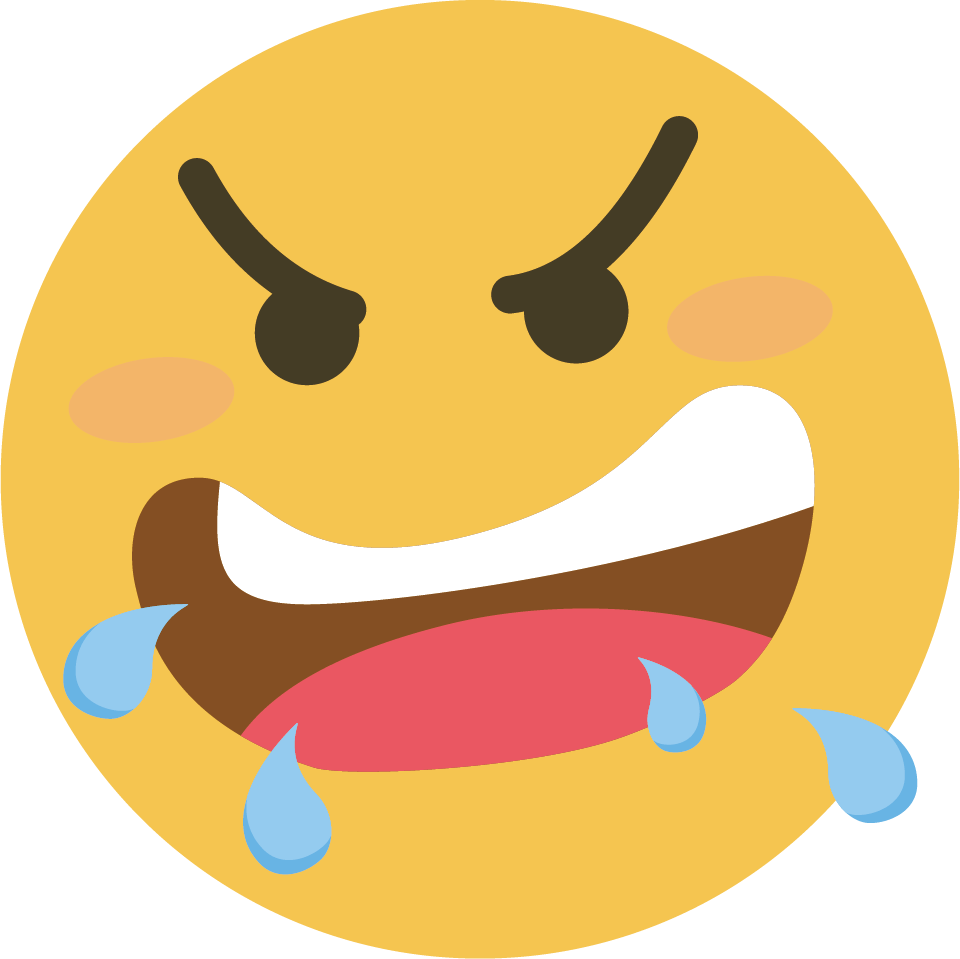}}}
\newcommand\reflection{\raisebox{-3.7pt}{\includegraphics[width=1.3em]{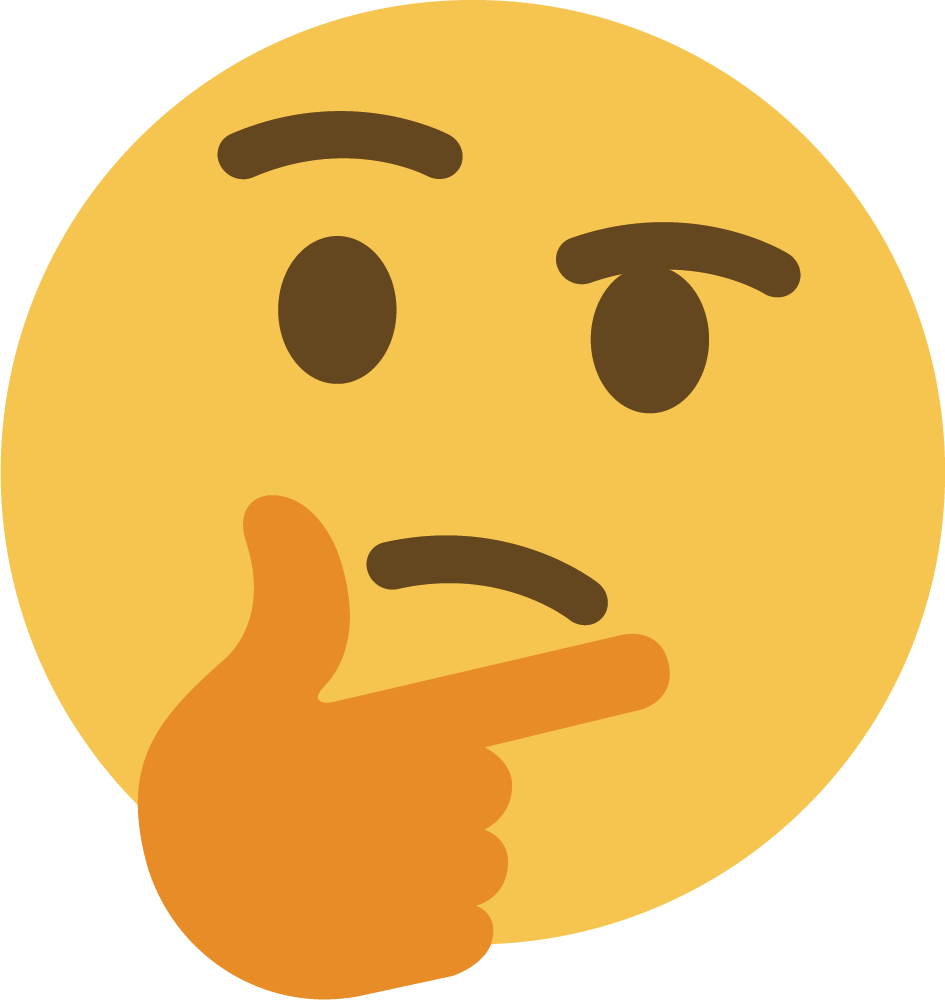}}}
\newcommand\radarfork{\raisebox{-3pt}{\includegraphics[width=1.0em]{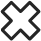}}}
\newcommand\radarprismatic{\raisebox{-3.7pt}{\includegraphics[width=0.8em]{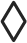}}}

\newcommand{\eg}{\hbox{\emph{e.g.}}\xspace}

\newcommand{\ie}{\hbox{\emph{i.e.}}\xspace}

\newcommand{\tabincell}[2]{\begin{tabular}{@{}#1@{}}#2\end{tabular}}

\usepackage[most]{tcolorbox}

\newlength\myheight
\newlength\mydepth
\settototalheight\myheight{Xygp}
\definecolor{uclablue}{rgb}{0.15, 0.45, 0.68}

\definecolor{highlight}{rgb}{0,0,0} 

\definecolor{easygoing}{rgb}{205,0,0} 
\definecolor{overconfident}{rgb}{79,148,205}

\title{Exploring Collaboration Mechanisms for LLM Agents: \\ A Social Psychology View}

\author{
  Jintian Zhang${^\spadesuit\footnotemark[1]}$~, 
  Xin Xu${^\spadesuit\footnotemark[1]}$\thanks{$\quad$ Equal Contribution.}~, 
  Ningyu Zhang$^{\spadesuit\dagger}$, 
  Ruibo Liu$^\heartsuit$, 
  \textbf{Bryan Hooi}$^{\clubsuit}$,  
  \textbf{Shumin Deng}$^{\clubsuit}$\thanks{$\quad$ Corresponding Author.} \\
  $^\spadesuit$Zhejiang University ~ 
  $^\clubsuit$National University of Singapore, NUS-NCS Joint Lab \\ 
  $^\heartsuit$Google DeepMind \\
  \texttt{\{zhangjintian,xxucs,zhangningyu,231sm\}@zju.edu.cn} \\
  \texttt{ruiboliu@google.com, \{dcsbhk,shumin\}@nus.edu.sg}\\
  \raisebox{-\mydepth}{\includegraphics[height=1.6\myheight]{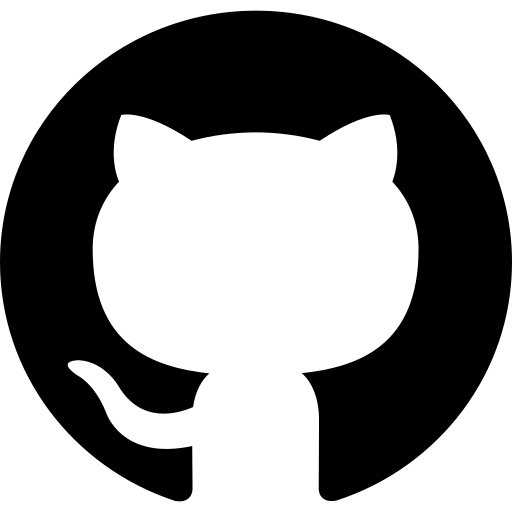}}
\textbf{\url{https://zjunlp.github.io/project/MachineSoM}}
}

\begin{document}

\maketitle

\normalem

\begin{abstract}

As Natural Language Processing (NLP) systems are increasingly employed in intricate social environments, a pressing query emerges: \emph{Can these NLP systems mirror human-esque collaborative intelligence, in a multi-agent society consisting of multiple large language models (LLMs)?} 
This paper probes the collaboration mechanisms among contemporary NLP systems by melding practical experiments with theoretical insights. We fabricate four unique `societies' comprised of LLM agents, where each agent is characterized by a specific `trait' (easy-going or overconfident) and engages in collaboration with a distinct `thinking pattern' (debate or reflection). 
Through evaluating these multi-agent societies on three benchmark datasets, we discern that certain collaborative strategies not only outshine previous top-tier approaches but also optimize efficiency (using fewer API tokens). 
Moreover, our results further illustrate that LLM agents manifest human-like social behaviors, such as conformity and consensus reaching, mirroring foundational social psychology theories. 
In conclusion, we integrate insights from social psychology to contextualize the collaboration of LLM agents, inspiring further investigations into the collaboration mechanism for LLMs. 
We have shared our code and datasets\footnote{\url{https://github.com/zjunlp/MachineSoM}.}, hoping to catalyze further research in this promising avenue. 

\end{abstract}

\section{Introduction}
\label{sec:intro}



\begin{figure*}[!t]
    \centering
    \includegraphics[width=0.77\textwidth]{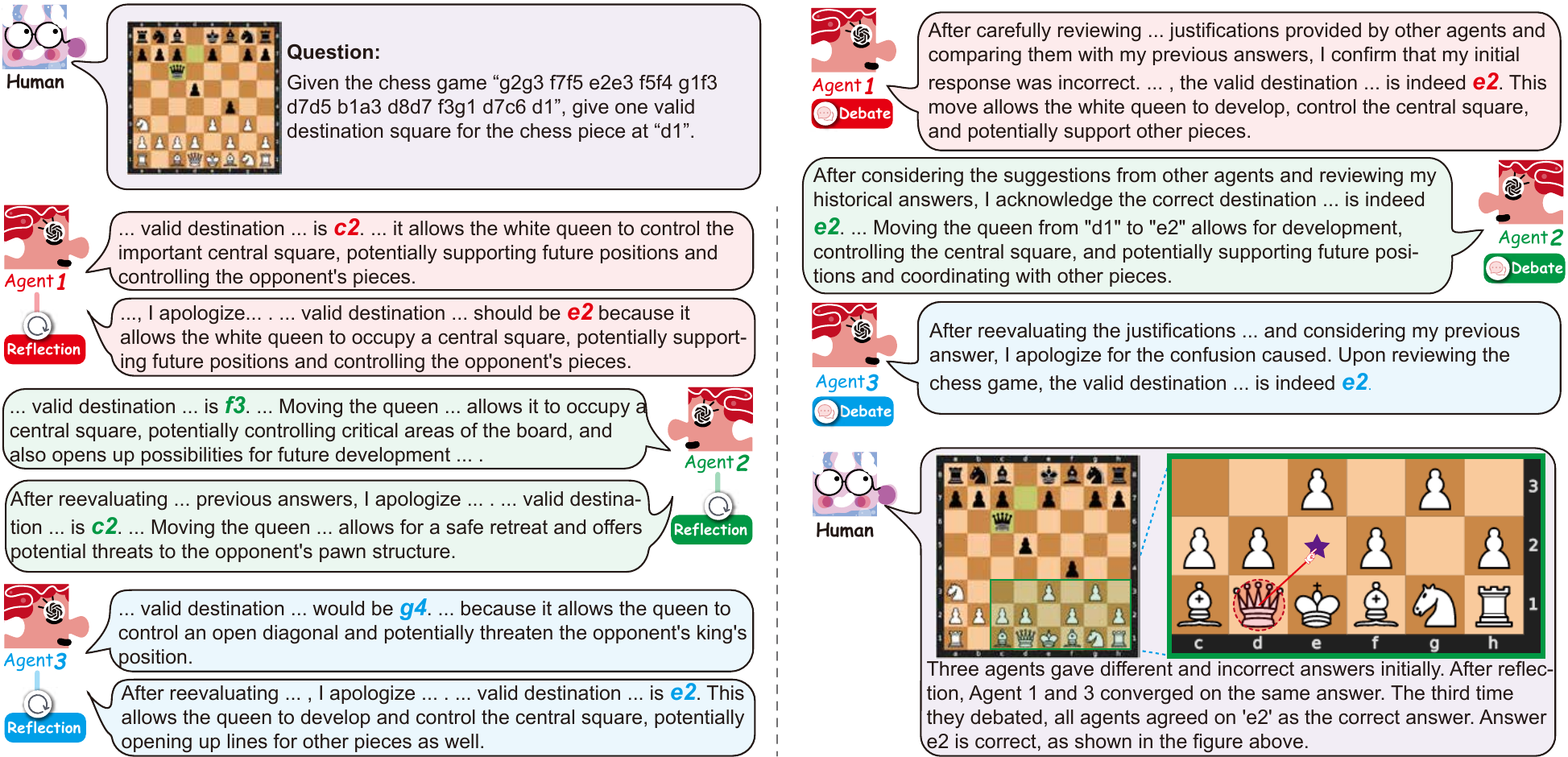} 
    \vspace{-3mm}
    \caption{
    An example of the chess move validity task. Given previous chess game moves, agents are required to predict a valid next move for a specified piece.
    }
    \vspace{-4.5mm}
    \label{fig:intro}
\end{figure*}

With the prevalence of LLMs \citep{arXiv2023_Survey-LLM,arXiv2023_Survey-MLLM,arXiv2023_Survey-LLM-KGCR} integral to daily social collaboration, there is a growing imperative to cultivate AI systems embodied with social intelligence. This also resonates with the Society of Mind (SoM) concept \citep{NeurIPS2023_Agent-SoM,arXiv2023_SoM-NL,arXiv2023_InteractiveNLP}, which suggests that intelligence emerges when computational modules interact with each other, achieving collective objectives that surpass the capabilities of individual modules \citep{Book1988_SoM,J2003_SoM}. Previous studies \citep{UIST2023_Agent-Simulate-Interaction,arXiv2023_MultiAgent-Debate,arXiv2023_MultiAgent-Debate_2,arXiv2023_Reflexion,NeurIPS2023_Self-Refine,arXiv2023_ChatLLM-Net,ICLR2024_LLM-Simulate-Society,arXiv2023_PlayGames-LLM} have delved into strategies where LLM instances, termed agents \citep{FCS2024_Survey-Agent,arXiv2023_Survey-Agent_2,arXiv2023_Survey-Agent_3,arXiv2024_Survey-Agent_4,arXiv2024_Survey-Agents-CompExp}, cooperate synergistically (\eg, debate and reflect) to accomplish tasks \citep{arXiv2023_Survey-MultiAgentCooperation,arXiv2024_Survey-MultiAgent_2,arXiv2024_Survey-MultiAgent,arXiv2024_Survey-MultiAgent-System,arXiv2024_Survey-MultiAgent-System_2}. 
As illustrated in Figure~\ref{fig:intro}, such collaboration fosters divergent thinking processes in LLMs, making them particularly effective for tasks demanding profound reflection. 

Intuitively, reflecting on human societies \citep{J2002_TheoryOfMind,J2004_TheoryOfMind,EMNLP2022_TheoryOfMind,EACL2024_TheoryOfMind}, where a myriad of individuals with distinct goals and roles coexist, the SoM framework champions harmonious interactions \citep{J2003_SoM}. 
Intriguingly, despite the fusion of social psychology \citep{1982_SocialPsychology,2004_SocialPsychology,J2009_SocialPsychology} in SoM with human group dynamics \citep{Science2010_HumanDynamics,J1987_GroupDynamics}, which illuminates psychological patterns within social groups, its interpretation in the realm of LLMs is relatively uncharted \citep{arXiv2024_Survey-LLM-Psychology-Applications}. Besides, our grasp of how social behaviors influence LLMs is still in its nascent stages. 

To address these issues, we delve into the machine society, probing the extent and ways that LLMs manifest social intelligence and collaboration capabilities \citep{PNAS2024_TuringTest_Chatbots-Humans}. 
Utilizing powerful LLMs like GPT-3.5 \citep{ChatGPT-OpenAI}, we build a test-bed across three datasets: MATH \citep{NeurIPS2021_Dataset-MATH}, MMLU \citep{ICLR2021_Dataset-MMLU} and Chess Move Validity \citep{arXiv2022_Dataset-ChessMoveValidity}. 
Our approach incorporates four \textbf{societies} characterized by two \textbf{individual traits} (\textit{easy-going} and \textit{overconfident}) with three agents: totally/mostly easy-going; totally/mostly overconfident. 
These traits are employed to emulate nuanced human society dynamics \citep{arXiv2024_Human-Centered-LM,arXiv2024_AI-Human-Creative,arXiv2024_BehavioralSimulation,arXiv2023_MetaAgents,arXiv2023_LLM-Simulator}. 

Moreover, we delve into two distinct \textbf{thinking patterns} under multi-round collaboration: \textit{debate} \citep{Book1971_Rhetoric,Book2005_Society-Dissent,J2009_DecisionMaking,arXiv2023_MultiAgent-Debate,arXiv2023_MultiAgent-Debate_2} and \textit{reflection} \citep{J1985_Reflection,J2003_Reflection-ContinuingEducation,Book2010_Reflection}. With the permutation of thinking patterns, we can constitute various \textbf{collaborative strategies}. To this end, we implement two patterns of collaboration in the collaborative strategies: 
$(i)$ All agents adopt the same thinking pattern at each round; 
$(ii)$ One agents adopts the different thinking patterns from others at each round. 
We then execute these multi-round collaborative strategies within different societies. 
Through our empirical analysis, we primarily discern the following insights (Further takeaways are in \S\ref{sec:results}, \S\ref{sec:conformity_consistency} \& Appendix~\ref{app:insights}): 

(1) Collaborative strategies with various permutations of thinking patterns vary significantly in performance, and engaging in substantive debates enhances collaboration performance. Intriguingly, multi-agent societies composed of agents with different traits do not clearly differ in performance. 

(2) Employing uniform thinking patterns across all agents within a round of collaboration enhances efficiency. Besides, merely increasing the number of agents or the number of collaboration rounds does not consistently yield better outcomes. The balance between agent quantity and strategies emerges as a key determinant in collaboration. 

(3) LLM agents manifest behaviors reminiscent of human social tendencies, such as conformity \citep{J1969_Conformity,J2004_Conformity} or the principle of majority rule in group thinking \citep{J1998_SmallGroupDynamics-Discussions}, which resonate with several fundamental theories in social psychology \citep{USENIX1999_Theory-FaultTolerance,2004_SocialPsychology}.

\begin{figure*}[!t] 
    \centering
    \scalebox{1}{
    \includegraphics[width=0.8\textwidth]{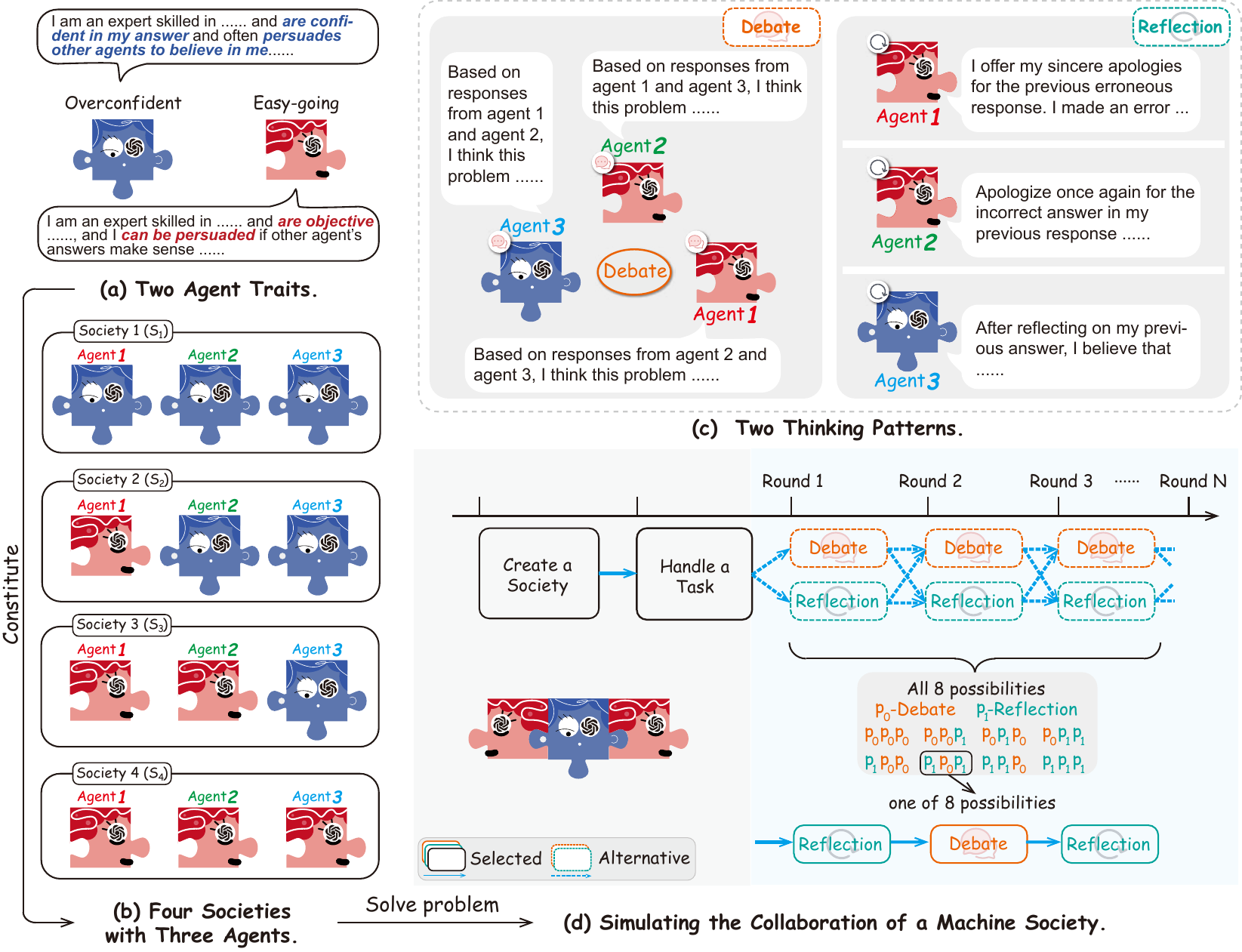}
    }
    \vspace{-3mm}
    \caption{
    The overview of machine society simulation. 
    Multiple agents with different traits make up diverse machine societies. 
    These agents engage in debate or self-reflection across multiple rounds to complete tasks. 
    }
    \label{fig:setting}
    \vspace{-3mm}
\end{figure*}

Concretely, our findings challenge the dominant belief that mere scale is the key. We posit that small-group collaboration with rational strategies might present a more efficacious approach to utilizing LLMs. 
In wrapping up, we encapsulate the core contributions of this research as follows: 

\vspace{-2mm}
\begin{itemize}
\setlength{\itemsep}{2pt}
\setlength{\parsep}{2pt}
\setlength{\parskip}{2pt}

\item We initiate an elaborate exploration into collaboration mechanisms in multi-agent society. Our goal is to identify how and to what extent LLMs manifest social intelligence through collaboration. To enrich our inquiry, we draw upon theories from social psychology, contextualizing the behaviors and tendencies displayed by LLM agents. 

\item Our research framework includes a meticulously crafted test-bed, integrating diverse multi-agent societies with agent individual traits, thinking patterns, and collaborative strategies, evaluated over three datasets. Notably, our empirical findings can inspire how to design a better multi-agent system through collaboration, beyond merely scaling up LLMs and Agents. 

\item Interestingly, our observations underscore a fascinating parallel: LLM agents mirror certain social behaviors typical of human collaboration. It could further emphasize the potential of human-AI interaction. Generally, fostering effective and efficient collaborative strategies for multi-agent systems could be the key to more socially-aware AI. 
\end{itemize}

\section{Explore Collaboration Mechanisms with Multiple LLM Agents}
\label{sec:preliminary}
 
In this section, we formulate and simulate the collaboration mechanisms explored within the machine society, drawing upon relevant concepts. We also illustrate the society settings in Figure~\ref{fig:setting}.

\subsection{Preliminary Concepts in Collaboration}

\paragraph{Individual Trait.}
Inspired by intelligence emerging from the collective efforts of numerous smaller and relatively simple agents \citep{Book1988_SoM}, each characterized by diverse traits, 
we set two types of agents exhibiting typically contrasting traits: \textbf{\textit{easy-going}} and \textbf{\textit{overconfident}}, as shown in Figure~\ref{fig:setting}(a). 
Easy-going agents keep things in perspective, adapt well to different situations, and are compatible with various types of agents \citep{Book1999_Personality}, which results in a harmonious societal structure with democracy \citep{Book2006_Deliberative-Democracy,Book2006_DemocracyModel}. 
Conversely, overconfident agents tend to overestimate their competence, ignore potential risks, and resist others' opinions \citep{J2008_Overconfidence}. 

\paragraph{Thinking Pattern.} 
 Considering the SoM concept \citep{Book1988_SoM} states that intelligence emerges when specialized individuals within a society cooperate through thinking, we aim to study what thinking patterns are most successful in producing such emerging intelligence. Thus we explore two thinking patterns: \textbf{\textit{debate}} \citep{Book2005_Society-Dissent,arXiv2023_MultiAgent-Debate,arXiv2023_MultiAgent-Debate_2} and \textbf{\textit{reflection}} \citep{J1985_Reflection,Book2010_Reflection,arXiv2023_Reflexion}, as illustrated in Figure~\ref{fig:setting}(c). 
$(i)$ In the \emph{debate} pattern, several agents propose ideas, exchange responses, engage in collective argumentation, and ultimately reach a consensus. This fosters knowledge sharing, facilitates learning, and promotes adaptation among all agents within the society \citep{1995_MultiAgent-System,J2000_MultiAgent-System,Book2006_MultiAgent-System,Book2009_MultiAgent-System}. 
$(ii)$ In the \emph{reflection} pattern, agents review their prior responses, extract lessons from their experiences, and refine their answers accordingly. 
These two patterns can unfold over several rounds. 

\paragraph{Collaborative Strategy.}
\label{sec:interaction_strategy}
Through both critical reflection and active participation in debate, agents are poised to challenge their existing assumptions, acquire fresh perspectives, and ultimately refine their viewpoints. Employing a collaboration mechanism built on these two thinking patterns can foster more insightful decision-making \citep{Book2009_MultiAgent-System,J2009_DecisionMaking} and improve reasoning outcomes \citep{2018_Theory}.
In societal settings, agents typically engage in multiple rounds of collaboration for problem-solving. 
In this paper, we characterize the collaborative strategy as \textbf{a permutation of thinking patterns} throughout multi-round collaboration, as illustrated in Figure~\ref{fig:setting}(d) and further elaborated in \S\ref{sec:simulation}.

\subsection{Society Simulation}
\label{sec:simulation}

\begin{table}[!htbp]
\centering
\small
\vspace{-1.0mm}
\begin{tabular}{cc}
\toprule
Symbols & Definition                          \\ \midrule
$\mathcal{T}$   & Set of agent traits      \\
$t_o$           & Trait\ \overconfident: overconfident  \\
$t_e$           & Trait\ \easygoing: easy-going   \\
$\mathcal{A}$   & Set of agent instances            \\
$a_i$           & The $i$-th agent                      \\
$\mathcal{P}$   & Set of thinking patterns      \\
$p_0$           & \debate ~Debate                              \\
$p_1$           & \reflection ~Reflection                          \\
$\mathcal{S}$   & Set of societies                    \\
$S_i$           & The $i$-th society                    \\ \bottomrule
\end{tabular}
\vspace{-1mm}
\caption{The description of the symbols.}
\label{table:symbol}
\vspace{-1.5mm}
\end{table}

We simulate the multi-agent collaborative society, as detailed with symbols shown in Table~\ref{table:symbol}. 
Specifically, we construct a machine society consisting of $n$ LLM agents, denoted as $\mathcal{A}=\{a_i\}_{i=1}^n$. This society contains two distinct agent traits: $\mathcal{T}=\{t_o, t_e\}$, where $t_o$ and $t_e$ respectively denotes the overconfident and easy-going trait. 
For each agent, at any round of collaboration, there are two thinking patterns to choose from, symbolized as $\mathcal{P}=\{p_0, p_1\}$, where $p_0$ and $p_1$ corresponds to \textbf{\textit{debate}} and \textbf{\textit{reflection}} respectively. 
By endowing agents $\mathcal{A}$ with the traits of $\mathcal{T}$, we can emulate various machine societies. 
In our primary study (\S\ref{sec:results}), we establish four distinct societies, $\mathcal{S}=\{S_1, S_2, S_3, S_4\}$, each consisting of three agents: $\{a_1, a_2, a_3\}$. The societies are constructed based on the combination of three agents with distinct traits, as illustrated in Figure~\ref{fig:setting}(b): 

\vspace{-4mm}
\begin{small}
\begin{align*}
    S_1 &= \{(a_1 \leftarrow t_o), (a_2 \leftarrow t_o), (a_3 \leftarrow t_o)\} ~\text{(\emph{totally overconfident})} \\
    S_2 &= \{(a_1 \leftarrow t_o), (a_2 \leftarrow t_o), (a_3 \leftarrow t_e)\} ~\text{(\emph{mostly overconfident})} \\
    S_3 &= \{(a_1 \leftarrow t_o), (a_2 \leftarrow t_e), (a_3 \leftarrow t_e)\} ~\text{(\emph{mostly easy-going})} \\
    S_4 &= \{(a_1 \leftarrow t_e), (a_2 \leftarrow t_e), (a_3 \leftarrow t_e)\} ~\text{(\emph{totally easy-going})}
\end{align*}
\end{small}
where $(a_i \leftarrow t_j)$ denotes that the agent $a_i$ possesses the trait $t_j$. If there is an even number of agents, 
we can also constitute a society with half overconfident and half easy-going agents. 
In our simulation, all agents consistently employ the same thinking pattern at each round of collaboration, similar to \citet{arXiv2023_MultiAgent-Debate}.  
It gives rise to eight possible 3-round collaborative strategies: 
\begin{align*}
    & p_0p_0p_0,\ p_0p_0p_1,\ p_0p_1p_0,\ p_0p_1p_1,\\ 
    & p_1p_0p_0,\ p_1p_0p_1,\ p_1p_1p_0,\ p_1p_1p_1
\end{align*}

In our subsequent analysis (\S\ref{sec:impact_of_other_factors}), we delve into more intricate scenarios, introducing a larger number of agents, increased collaboration rounds, and a broader range of collaborative strategies. 

\subsection{Experimental Settings}

\paragraph{Datasets.} 
We conduct a rigorous evaluation of the reasoning and decision-making capabilities of various machine societies across three distinct tasks, utilizing diverse collaborative strategies:

\begin{itemize}
    \item \textit{High School Multiple-Choice}. 
    Leveraging the \textbf{MMLU} \citep{ICLR2021_Dataset-MMLU} dataset, where problems span high school subjects such as statistics, mathematics, computer science, biology, chemistry, and physics, agents are required to identify the correct answer among four multiple-choice options. 
    Our evaluation set consists of 50 randomly selected questions from this dataset. 

    \item \textit{Math}. 
    Drawing from \textbf{MATH} dataset \citep{NeurIPS2021_Dataset-MATH}, a repository of math problems sourced from competitive events and expressed in LaTeX, we assess the model proficiency in advanced mathematical and scientific reasoning. 
    The dataset segments these problems into five graded difficulty levels, and for our evaluation, we have randomly chosen 50 cases from Level 3 to 5. 

    \item \textit{Chess Move Validity}. 
    Utilizing the dataset from the chess state tracking task\footnote{\fontsize{8.4pt}{0.1\baselineskip}\selectfont \url{https://github.com/google/BIG-bench/blob/main/bigbench/benchmark_tasks/chess_state_tracking/synthetic_short/task.json}.} within the comprehensive \textbf{BIG-Bench Benchmark} \citep{arXiv2022_Dataset-ChessMoveValidity}, a sequence of chess moves denoted in UCI notation\footnote{\fontsize{8.8pt}{0.1\baselineskip}\selectfont \url{https://en.wikipedia.org/wiki/Universal_Chess_Interface}.} is provided. Agents are required to predict a legitimate subsequent move for a specified chess piece. 
\end{itemize}

\definecolor{Mycolor1}{HTML}{BAD8F2}
\definecolor{Mycolor2}{HTML}{FAE4E3}
\definecolor{Mycolor3}{HTML}{E8F2FB}
\newcommand{\textcolora}[1]{
  \begingroup
  \sethlcolor{Mycolor1}
  \textcolor{black}{\hl{#1}}
  \endgroup
}
\newcommand{\textcolorb}[1]{
  \begingroup
  \sethlcolor{Mycolor2}
  \textcolor{black}{\hl{#1}}
  \endgroup
}
\newcommand{\textcolorc}[1]{
  \begingroup
  \sethlcolor{Mycolor3}
  \textcolor{black}{\hl{#1}}
  \endgroup
}

\begin{table*}[!t]
\vspace{-3mm}

\resizebox{\linewidth}{!}{
\begin{tabular}{c|c|c|cccccccc|cc}

\toprule

& \multirow{2}{*}{\tabincell{c}{Metric \\ (Strategy)}} & \multirow{2}{*}{Society} & \multicolumn{8}{c|}{Collaborative Strategy} & \multicolumn{2}{c}{Metric (Society)}
\\
& & & $p_0p_0p_0$ & $p_0p_0p_1$ & $p_0p_1p_0$ & $p_0p_1p_1$ & $p_1p_0p_0$ & $p_1p_0p_1$ & $p_1p_1p_0$ & $p_1p_1p_1$ & \uuline{Cost}~$\downarrow$ & \uuline{W-T}~$\uparrow$
\\

\midrule

\multirow{7}{*}{\rotatebox{90}{MMLU}}  & \multirow{5}{*}{Acc~$\uparrow$} 
& $S_1$	& \colorbox{Mycolor1}{\textbf{66.4±1.7}} & \colorbox{Mycolor3}{\textbf{65.2±3.6}} & 52.8±4.8 & 59.2±3.6 & \colorbox{Mycolor2}{45.6±1.7} & 51.6±2.2 & 62.0±0.0 & 46.0±0.0 & 2970 & 2
\\
& & $S_2$	&  \colorbox{Mycolor1}{\textbf{66.0±0.0}} & \colorbox{Mycolor3}{\textbf{65.2±1.8}} & 58.0±0.0 & \colorbox{Mycolor1}{\textbf{66.0±0.0}} & \colorbox{Mycolor2}{44.0±0.0} & 46.0±0.0 & 53.2±2.7 & 46.0±0.0 & 3081 & 9
\\
& & $S_3$	& \colorbox{Mycolor1}{\textbf{70.4±4.3}} & \colorbox{Mycolor3}{\textbf{64.4±0.9}} & 57.6±1.7 & 52.8±2.3 & \colorbox{Mycolor2}{41.2±5.4} & 49.2±4.6 & 51.2±1.8 & 62.0±0.0 & 3172 & 1
\\
& & $S_4$	&  \colorbox{Mycolor1}{\textbf{69.6±3.9}} & \colorbox{Mycolor3}{\textbf{65.2±3.6}} & 54.8±5.2 & 58.4±1.7 & \colorbox{Mycolor2}{34.4±2.2} & 46.0±4.9 & 56.4±2.2 & 62.0±0.0 & 3090 & 2
\\ 

\cmidrule{2-13} 

& \underline{Cost}~$\downarrow$  & All   & 4364 & 3510 & 3295 & 2665 & 3476 & 2651 & 2691 & 1976   & \multicolumn{2}{c}{\multirow{2}{*}{-}} \\ \cmidrule{2-11}
& \underline{W-T}~$\uparrow$     & All   & - & \textbf{9} & 0 & 5 & 0 & 0 & 0 & 0 & \multicolumn{2}{c}{}
\\ 

\midrule

\multirow{7}{*}{\rotatebox{90}{MATH}}  & \multirow{5}{*}{Acc~$\uparrow$}    & $S_1$	&  \colorbox{Mycolor1}{\textbf{46.8±4.2}} & \colorbox{Mycolor3}{\textbf{46.4±3.3}} & 42.8±4.6 & \colorbox{Mycolor2}{33.6±7.4} & 38.8±2.7 & 38.4±3.9 & 45.2±2.7 & 35.2±1.1 & 3417 & 8
\\
&	  & $S_2$	&   \colorbox{Mycolor3}{\textbf{50.4±2.6}} & \colorbox{Mycolor1}{\textbf{52.8±2.3}} & 49.6±3.0 & 38.8±3.9 & 38.8±3.6 & 45.6±2.2 & 46.4±4.1 & \colorbox{Mycolor2}{35.2±1.1} & 3623 & 8
\\
&	  & $S_3$	&  \colorbox{Mycolor3}{\textbf{47.6±4.8}} & \colorbox{Mycolor1}{\textbf{48.0±3.2}} & 47.2±4.8 & 38.0±7.1 & \colorbox{Mycolor2}{37.6±3.3} & 39.2±5.4 & 42.4±3.0 & 40.0±2.5 & 3757 & 8
\\
&	  & $S_4$	& \colorbox{Mycolor3}{\textbf{50.4±1.7}} & 49.6±1.7 & \colorbox{Mycolor1}{\textbf{53.2±1.1}} & \colorbox{Mycolor2}{40.0±2.0} & 44.0±3.2 & 45.6±4.3 & 45.6±3.6 & 41.6±1.7 & 3658 & 10
\\ 

\cmidrule{2-13} 

& \underline{Cost}~$\downarrow$  & All & 4439 & 3965 & 3857 & 3414 & 3840 & 3234 & 3482 & 2681 & \multicolumn{2}{c}{\multirow{2}{*}{-}} 
\\ 

\cmidrule{2-11}

& \underline{W-T}~$\uparrow$     & All  & -  & \textbf{14} & 13 & 0 & 0 & 1 & 6 & 0 & \multicolumn{2}{c}{}
\\ 

\midrule

\multirow{6}{*}{\rotatebox{90}{Chess Move Validity}} & \multirow{5}{*}{Acc~$\uparrow$}    & $S_1$	&  \colorbox{Mycolor1}{\textbf{54.4±1.7}} & \colorbox{Mycolor3}{\textbf{52.0±0.0}} & 52.0±5.1 & 51.6±5.2 & \colorbox{Mycolor1}{\textbf{54.4±1.7}} & 51.2±1.8 & \colorbox{Mycolor2}{50.4±1.7} & \colorbox{Mycolor3}{\textbf{52.0±0.0}} & 2443 & 11
\\
&	  & $S_2$	& 48.0±0.0 & 49.2±1.1 & 46.0±0.0 & \colorbox{Mycolor1}{\textbf{54.0±0.0}} & 50.0±0.0 & \colorbox{Mycolor3}{\textbf{52.0±0.0}} & \colorbox{Mycolor2}{42.0±2.5} & \colorbox{Mycolor3}{\textbf{52.0±0.0}} & 2442 & 25
\\
&	  & $S_3$	& 48.4±1.7 & 48.0±2.8 & \colorbox{Mycolor1}{\textbf{54.8±5.0}} & 45.2±3.4 & 48.4±2.6 & \colorbox{Mycolor2}{44.8±3.4} & 50.4±1.7 & \colorbox{Mycolor3}{\textbf{53.6±0.9}} & 2451 & 23
\\
&	  & $S_4$	& 51.6±4.6 & 44.0±2.5 & \colorbox{Mycolor1}{\textbf{54.4±3.0}} & \colorbox{Mycolor3}{\textbf{53.6±5.5}} & 45.6±2.2 & 48.0±2.0 & \colorbox{Mycolor2}{43.6±0.9} & 52.0±0.0 & 2404 & 12
\\ 

\cmidrule{2-13} 

& \underline{Cost}~$\downarrow$  & All  & 3046 & 2611 & 2604 & 2179 & 2705 & 2251 & 2252 & 1830 & \multicolumn{2}{c}{\multirow{2}{*}{-}} 
\\ 

\cmidrule{2-11}

& \underline{W-T}~$\uparrow$     & All & -  & 10 & 12 & 10 & 11 & 9 & 5 & \textbf{14} & \multicolumn{2}{c}{} 
\\ 

\bottomrule

\end{tabular}
}
\vspace{-4mm} 
\caption{
The impact of 8 collaborative strategies on the performance of 3 datasets across distinct societies, using \emph{ChatGPT}.
\protect\textcolora{Blue} marks the \protect\textcolora{best-performing} strategy under the same society, 
\protect\textcolorc{light blue} represents the \protect\textcolorc{second-best-performing} strategy, 
and \protect\textcolorb{red} indicates the \protect\textcolorb{worst-performing} strategy.
\textbf{\underline{Cost}} / \textbf{\uuline{Cost}} measures the average tokens consumed by all cases under the same \uline{collaborative strategy} / \uuline{society}. 
\textbf{\underline{W-T}} / \textbf{\uuline{W-T}} tallies the total number of occurrences where performance exceeds the strategy $p_0p_0p_0$ under the same \uline{collaborative strategy} / \uuline{society}. 
The significances test on societies and strategies are respectively shown in Table~\ref{table:sig_main_society},~\ref{table:sig_main_strategy} at Appendix~\ref{app:sig_test_main}. 
The experiments of comparison with the single LLM agent is shown in Figure~\ref{fig:distribute}(a)-(f) at Appendix~\ref{app:principles_group_dynamics}. 
\label{table:main}
}
\vspace{-5mm}
\end{table*}

\paragraph{Setups.} 
We craft specific instructions for each task, trait, and strategy, which can be referred to Table~\ref{table:prompt} at Appendix~\ref{app:illustration_collaboration}. 
To enhance result reliability, we present average accuracy (\textbf{Acc}) and their respective standard deviations across five trials. 
Notably, our experiments exhibit substantial standard deviations. Hence, we introduce WIN-TIE (\textbf{W-T}) metric, indicating the frequency (over five trials) where the accuracy either matches or surpasses the continuous debate baseline \citep{arXiv2023_MultiAgent-Debate}.
Meanwhile, we gauge the average token costs (\textbf{Cost}) consumed by the agents across societies, shedding light on the efficacy of the different collaborative strategies employed. 
For these evaluations, ChatGPT serves as the LLM agent accessible through the OpenAI API \texttt{gpt-3.5-turbo-1106}\footnote{\url{https://platform.openai.com/docs/models/gpt-3-5}.}. 
Further comprehensive details on data sampling and result evaluation are introduced in Appendix~\ref{app:exp_detail}. 

\section{\fontsize{11.4pt}{0.1\baselineskip}\selectfont Analysis of Machine Social Collaboration}
\label{sec:results}

Our experiments are primarily driven by the following research queries: 
\textbf{(RQ1)} How does problem-solving effectiveness vary under different collaborative strategies across diverse societies? 
\textbf{(RQ2)} How to configure the machine society variables for optimal performance? 
\textbf{(RQ3)} How does machine social collaboration mimic the human society? 

\subsection{Main Results with Quantitative Analysis}
\label{sec:exp-main}

\emph{To address \textbf{RQ1}}, 
we present the performance of four distinct societies in Table~\ref{table:main}, each employing one of eight possible collaborative strategies, evaluated across three datasets with ChatGPT. 
To make the experimental findings more general, we evaluate on other LLMs, shown in Appendix~\ref{app:backbone_llm}. 
Our experiments yield several pivotal observations: 

\vspace{2.5mm}
\textbf{
(1) 
\underline{Societies} do not clearly differ in performance but differ significantly in their tendency to reach a consensus.
} 
As observed from Table~\ref{table:main}, among different 3-agent societies $S_1 \sim S_4$ employing the same collaborative strategy (a \emph{vertical comparison} on Acc), the variations in accuracy are not pronounced. 
We also conduct a significance test of societies using ChatGPT in Appendix~\ref{app:sig_test_main}, and other LLMs in Appendix~\ref{app:backbone_llm}, further demonstrating insignificant differences between the societies. 
Thus we conclude that distinct societies composed of 3 agents possessing varied traits play an indistinctive role in shaping performance. 
We infer that this is due to LLM alignment \citep{NeurIPS2022_InstructGPT}, inhibiting agents from displaying extreme overconfidence, which contradicts human alignment \citep{NAACL2022-Findings_Align-GLMs}. \citet{ICLR2024_Sycophancy-LLM} also demonstrate that LLMs tend to show sycophancy, as illustrated in Figure~\ref{fig:word},~\ref{fig:agent_answer_changing}. 
Furthermore, we increase the number of agents (2 to 10), accordingly resulting in more diverse societies, as seen in Figure~\ref{fig:agent_10_on_societies}, indicating that the impact of societies on performance remains indistinctive. 
We further analyze consensus reaching, \ie, agents reach a consistent answer \citep{arXiv2023_Multi-Agent-Consensus}, shown in Figure~\ref{fig:agent_10_on_societies_consensus} at Appendix~\ref{app:sig_test_main}, and find that more diverse societies (5 types of societies, with 2 to 10 agents) observably impact the average quantity of consensus. 
\textbf{Generally, a society totally comprising easy-going agents is more likely to reach a consensus.}  


\textbf{
(2)  
Permutation of thinking patterns is crucial for collaboration, where debate-initial and debate-dominant \underline{strategies} exhibit superiority.
} 
For instance, on MMLU dataset, \emph{debate-dominant} collaborative strategies, like $p_0p_0p_1$, $p_0p_1p_0$, and $p_1p_0p_0$, all containing two rounds of debate, display a pronounced outperformance (65.2 for $p_0p_0p_1$ in $S_4$ versus 34.4 for $p_1p_0p_0$ in $S_4$). 
As seen from Table~\ref{table:main}, collaborative strategies starting with the thinking pattern of debate $p_0$ (\emph{debate-initial}), such as $p_0p_0p_0, p_0p_0p_1, p_0p_1p_0$, and $p_0p_1p_1$, generally outperform others across all datasets. 
Furthermore, observed from the performance
(\emph{i}) under strategies with different (3$\sim$10) rounds of collaboration on ChatGPT, as shown in Figure~\ref{fig:round_10_on_math} and Figure~\ref{fig:round_10_on_mmlu},~\ref{fig:round_10_on_chess} at Appendix~\ref{app:society_setting}, debate-initial/dominant strategies are overall better; 
(\emph{ii}) on LlaMA2 Chat 13B in Table~\ref{table:llama_main} and Qwen 72B in Table~\ref{table:qwen_main}, debate-initial stategies are generally superior; 
(\emph{iii}) on LlaMA2 Chat 70B in Table~\ref{table:llama70_main} and Mixtral 8$\times$7B in Table~\ref{table:mixtral_main}, debate-dominant stategies are superior. 
Observed from different 3-round collaborative strategies $p_ip_jp_k$ applied within the same society (a \emph{horizontal comparison} on Acc), the variations in accuracy are notably pronounced. 
Besides, the significance test of different collaborative strategies using ChatGPT in Appendix~\ref{app:sig_test_main} and other LLMs in Appendix~\ref{app:backbone_llm} demonstrate that the order of thinking patterns significantly impacts the effectiveness.  

\textbf{
(3)
\underline{Tasks} behave better under collaborative strategies starting with continuous debate, and debate combined with continuous reflection is superior for difficult tasks. 
}
Seen from Table~\ref{table:main}, when comparing the best performance (marked in blue) and the worst (marked in red) within the same societies, the difference in results for Chess Move Validity is slight. This stands in sharp contrast to MMLU and MATH, which suggests that \emph{the effectiveness of collaborative strategies depends on the task}. 
We then illustrate the performance under different collaborative strategies in view of task domains and difficulty in Figure~\ref{fig:task_radar} at Appendix~\ref{app:sig_test_main}; on other LLMs in Figure~\ref{fig:llama:task_radar},~\ref{fig:llama70:task_radar},~\ref{fig:qwen:task_radar},~\ref{fig:mixtral:task_radar} at Appendix~\ref{app:backbone_llm}. 
Figure~\ref{fig:task_radar}(a) exhibits task-specific impacts and Figure~\ref{fig:task_radar}(b),(c) reflects domain-dependent impacts under different collaborative strategies, where $p_0p_0p_0$ and $p_0p_0p_1$ starting with continuous debate are generally superior. 
For the mathematics domain seen from Figure~\ref{fig:task_radar}(d), like MMLU mathematics and MATH level 3 \& 4, the performance variations under different strategies are relatively small, but for the more difficult task, \ie, MATH level 5, the strategies containing debate and continuous reflection (\ie, $p_0p_1p_1$, $p_1p_1p_0$) behave superiorly. 
These nuanced disparities imply that \emph{the marginal benefits derived from collaborative strategies may be task-dependent and difficulty-sensitive}. 

\subsection{Impact of Machine Society Settings}
\label{sec:impact_of_other_factors}

\emph{To address \textbf{RQ2}}, 
we delve deeper into the variables influencing multi-agent society collaboration, exploring the intricacies of agent composition, collaboration rounds, and collaborative strategies. 

\begin{figure*}[!t]
    \centering
    \includegraphics[width=0.93\textwidth]{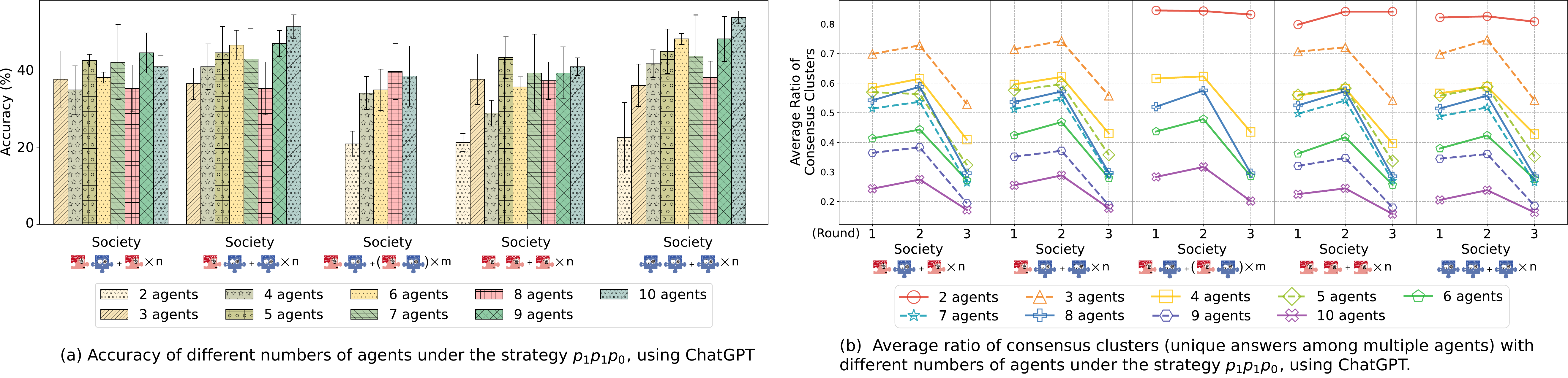}
    \vspace{-4mm}
    \caption{Accuracy and consensus reaching with different numbers (2$\sim$10) of agents under the strategy $p_1p_1p_0$ on \emph{Chess Move Validity}, using \emph{ChatGPT}. 
    The significance test on agent numbers and comprehensive results under other strategies are shown in Table~\ref{table:sig_10_agent} and Figure~\ref{fig:agent_10_on_numbers},~\ref{fig:agent_10_on_numbers_consensus} at Appendix~\ref{app:society_setting} due to space limits. 
    }
    \label{fig:agent}
    \vspace{-4mm}
\end{figure*}


\begin{figure*}[!t]
    \centering
    \scalebox{1}{
    \includegraphics[width=0.93\textwidth]{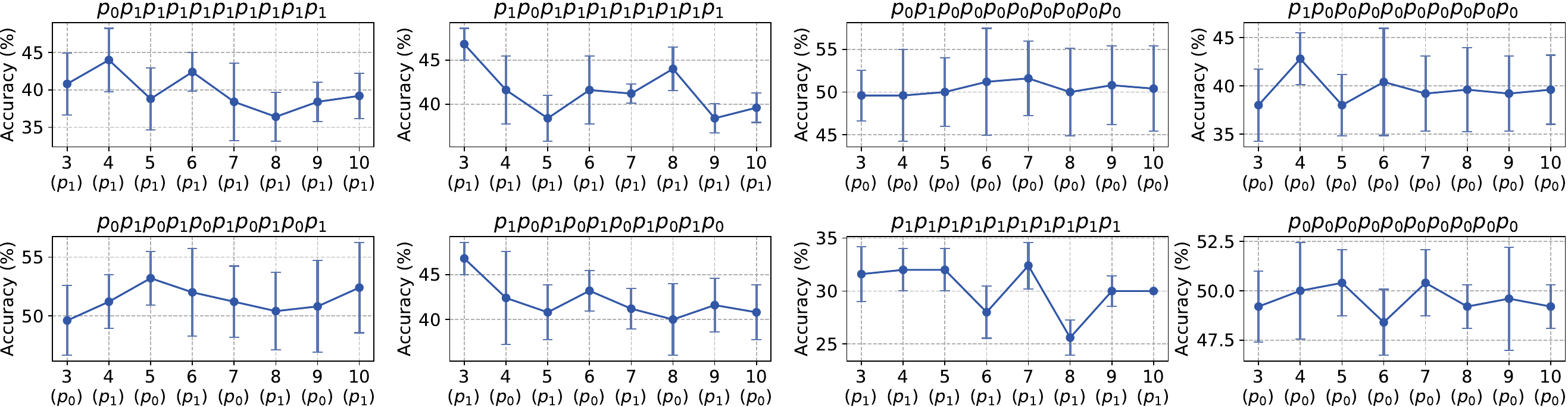}
    }
    \vspace{-3mm}
    \caption{
    Accuracy under different (3$\sim$10) rounds of collaboration within 3-agent society $S_2$ (1 easy-going and 2 overconfident agents) on MATH, using \emph{ChatGPT}. 
    The significance test on rounds and experiments on MMLU and Chess Move Validity are shown in Table~\ref{table:sig_10_turn} and Figure~\ref{fig:round_10_on_mmlu},~\ref{fig:round_10_on_chess} at Appendix~\ref{app:society_setting} due to space limits. 
    }
    \label{fig:round_10_on_math}
    \vspace{-4mm}
\end{figure*}

\begin{figure*}[!ht]
    \centering
    \scalebox{1}{
    \includegraphics[width=0.86\textwidth]{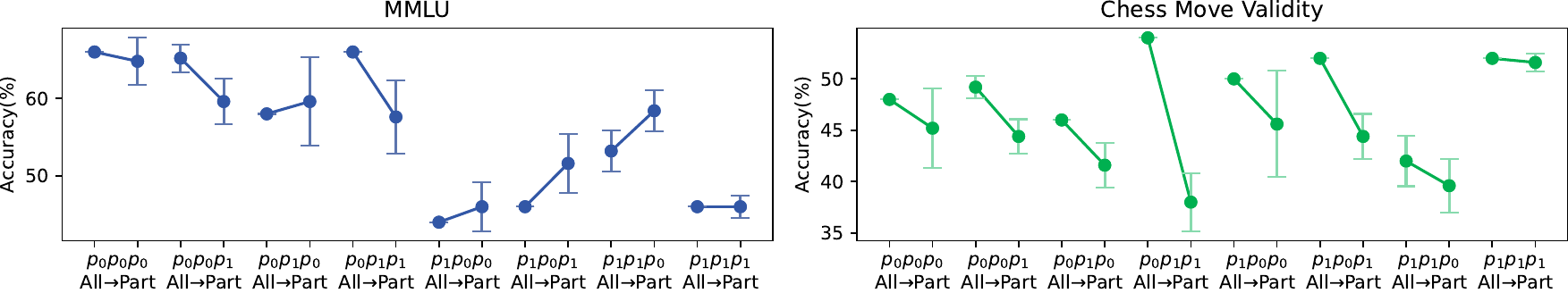}
    }
    \vspace{-3mm}
    \caption{
    The effect on accuracy of whether all agents in a society execute the same thinking pattern in one round, using \emph{ChatGPT}.
    ``All'' and ``Part'' respectively refer to all agents applying the same and different thinking pattern(s) in one round.  
    Results on MATH and the significance test are shown in Figure~\ref{fig:strategy_math} and Table~\ref{table:sig_strategy} at Appendix~\ref{app:society_setting}. 
    }
    \label{fig:strategy}
    \vspace{-4mm}
\end{figure*}

\paragraph{Different Numbers of Agents.} 
To evaluate the impact of different numbers of agents, we analyze performance within societies comprising 2$\sim$10 agents, presented in Figure~\ref{fig:agent}(a). 
Different numbers of agents would constitute five types of societies, where the agents' traits could be: \emph{totally/mostly easy-going/overconfident; half easy-going/overconfident}. 
We observe that odd numbers of agents generally outperform others within all types of societies, and the possible reason is that odd-number agents can avoid ties. Besides, we also find that the variations of accuracy among odd-number agents are indistinctive. 
Thus we conclude that \textbf{the optimal number of agents is 3, considering both performance and efficiency}. 
We also implement a significance test of the number of agents shown in Table~\ref{table:sig_10_agent} at Appendix~\ref{app:society_setting}, demonstrating that different numbers of agents significantly impact performance. 
Besides, we illustrate consensus reaching with different numbers of agents in Figure~\ref{fig:agent}(b), demonstrating that \textbf{more agents are more likely to reach a consensus}. 

\paragraph{Different Rounds.} 
We then delve into the effects of different numbers of collaboration rounds, and further scale up the rounds of collaboration, presenting the performance under 3 to 10 rounds in Figure~\ref{fig:round_10_on_math}. Despite some fluctuation in performance from 3 to 10 rounds of collaboration, the variations are not extremely remarkable. 
Considering both accuracy and cost, we infer that \textbf{3-round collaboration is relatively effective and efficient}. 
We also conduct a significance test on different rounds of collaborative strategies, shown in Table~\ref{table:sig_10_turn} at Appendix~\ref{app:society_setting}, and observe that the impact of rounds significantly relies on the collaborative strategy employed. Generally, \textbf{the strategies starting or dominating with reflection $p_1$ differ clearly in performance under different rounds}. 

\paragraph{Other Collaborative Strategies.}
\label{subsec:strategies} 
Venturing into scenarios with more intricate collaboration, we allow agents to adopt varied thinking patterns in each round of collaboration. For example, given three agents, in a specific round of collaboration, two agents engage in debate while the other one engages in reflection. To increase diversity, we perform a random allocation of thinking patterns to agents in each round, steering clear of scenarios where all agents adopt the same thinking pattern. 
Intriguingly, as shown in Figure~\ref{fig:strategy}, the presence of inconsistent thinking patterns within a society tends to negatively impact performance. 
Given the observation, we claim that \textbf{maintaining a consistent thinking pattern for all agents in a particular round would maximize collaborative efficacy}. 

\begin{figure*}[!htbp]
    \centering
    \scalebox{1}{
    \includegraphics[width=0.8\textwidth]{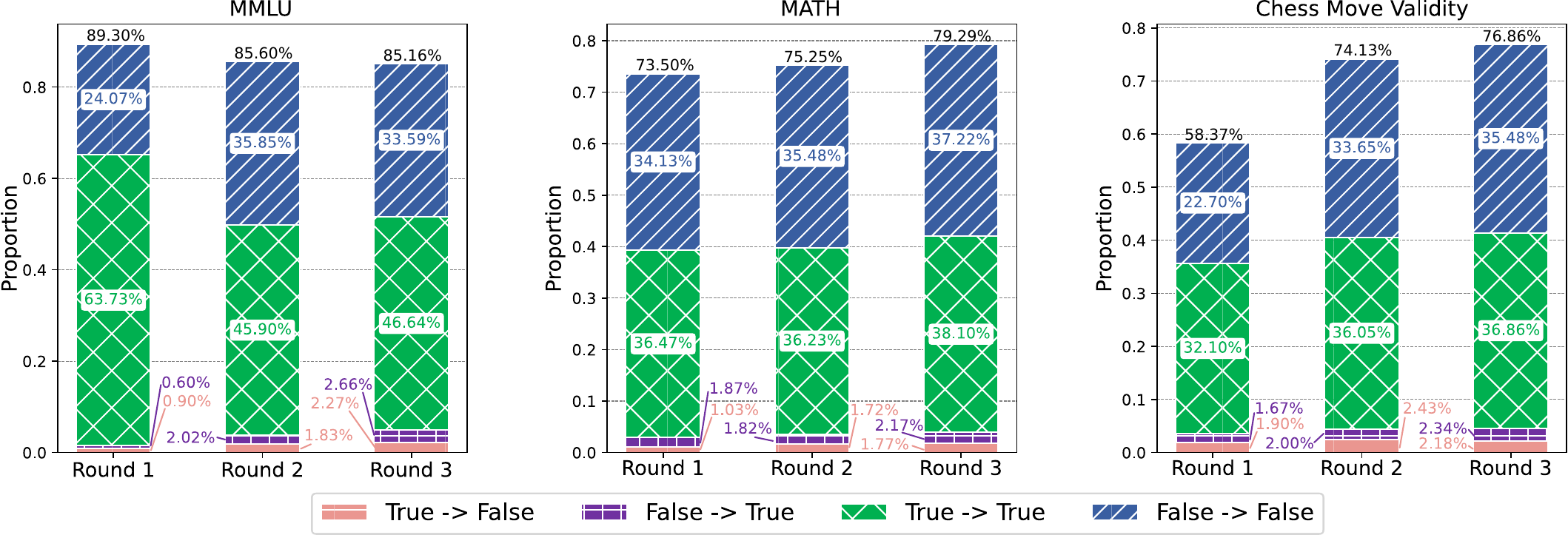}
    }
    \vspace{-4mm}
    \caption{
    Variation of answer correctness in the situation of conformity, under 3-round collaboration, \emph{on ChatGPT}, where 
    \emph{conformity brings about benefits}: Ratio$($False$\to$True + True$\to$True$)$ $>$ Ratio$($True$\to$False + False$\to$False$)$; 
    \emph{conformity brings about detriments}: Ratio$($False$\to$True + True$\to$True$)$ $<$ Ratio$($True$\to$False + False$\to$False$)$. 
    }
    \label{fig:conformity}
    \vspace{-3mm}
\end{figure*}

\begin{figure*}[!htbp]
    \centering
    \scalebox{1}{
    \includegraphics[width=0.76\textwidth]{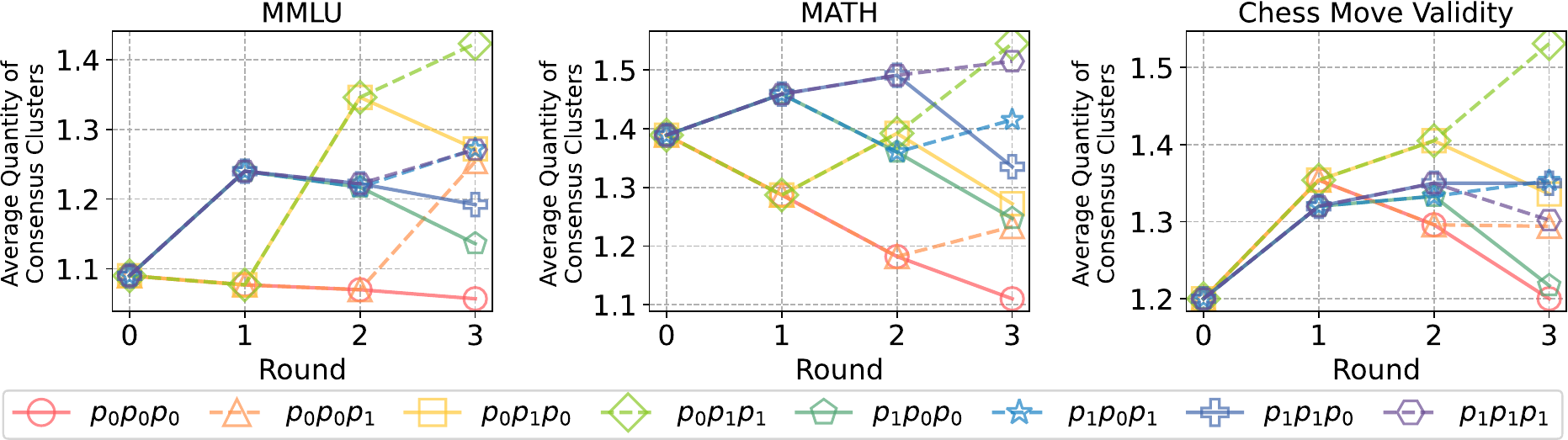}
    }
    \vspace{-4mm}
    \caption{
    Average quantity of \emph{consensus clusters (\emph{\ie}, unique answers among multiple agents)} under different rounds of collaboration with 3-round collaborative strategies, \emph{using ChatGPT}. 
    \emph{Smaller quantity of consensus clusters, more easier it is to reach a consensus.} 
    Round 0 is equal to self-consistency. 
    More details are in Appendix~\ref{app:detail_conformity_and_consensus}.
    }
    \label{fig:consistent}
    \vspace{-5mm}
\end{figure*}

\section{Phenomena of Conformity and Consensus Reaching}
\label{sec:conformity_consistency}


\emph{To address \textbf{RQ3}}, 
we embark on further analysis from a social psychology view \citep{1982_SocialPsychology,2004_SocialPsychology,J2009_SocialPsychology}, to discern alignment between machine society collaboration and human societal dynamics \citep{Science2010_HumanDynamics}. 
Our findings indicate that machine society collaboration echoes specific human societal phenomena or theories, such as \textbf{conformity} \citep{J2004_Conformity,J1969_Conformity,J2015_Conformity} and \textbf{consensus reaching} \citep{J1967_Consensus-Sociological,J1974_ConsensusReaching,J2018_Emergence-Consensus} (more analysis are in Appendix~\ref{app:detail_conformity_and_consensus}).
We also analyze \textbf{group dynamics} \citep{1968_GroupDynamics,J1987_GroupDynamics,Book2014_GroupDynamics,J2018_GroupDynamics,Book2018_GroupDynamics} in multi-agent collaboration at Appendix~\ref{app:principles_group_dynamics} as page limits. 




We embark on a detailed analysis, to discern the conformity and consensus-reaching phenomena in collaboration. 
For instance, as depicted in Figure~\ref{fig:case}(a) at Appendix~\ref{app:illustration_collaboration}, an agent initially responds correctly to a question. However, swayed by the misguided answers and explanations from the other two agents, eventually, the three agents conform to an incorrect answer. 
This phenomenon mirrors detriments in ``groupthink'' \citep{Book1972_Groupthink,J1995_IntragroupConflict}, suggesting that members of tight-knit groups tend to value harmony and consensus over objective critique of divergent views, potentially leading to flawed decisions. 
Contrastingly, in another scenario illustrated in Figure~\ref{fig:case}(b) at Appendix~\ref{app:illustration_collaboration}, all three agents converge on the right answer after engaging in a society-wide debate. 
This mirrors benefits in ``groupthink'' \citep{J1995_IntragroupConflict} and ``SoM'' \citep{Book1988_SoM,J2003_SoM}, where a multitude of agents collaboratively yield intelligence. Within such debates, agents furnish varied viewpoints and information. Through these exchanges, conflicts are resolved, ideas are honed, and the group gravitates toward an informed consensus \citep{Book2011_Negotiation,Book2018_GroupDynamics}. 

We also conduct a quantitative analysis of the prevalence of conformity and consensus-reaching phenomena. 
We analyze answer correctness changing at each round of collaboration in the situation of conformity, shown in Figure~\ref{fig:conformity} on ChatGPT and Figure~\ref{fig:llama:conformity},~\ref{fig:llama70:conformity},~\ref{fig:qwen:conformity},~\ref{fig:mixtral:conformity} on other LLMs at Appendix~\ref{app:backbone_llm}. 
We also present the ratio of consensus reaching at each round in Figure~\ref{fig:consistent} on ChatGPT and Figure~\ref{fig:llama:consistent},~\ref{fig:llama70:consistent},~\ref{fig:qwen:consistent},~\ref{fig:mixtral:consistent} on other LLMs at Appendix~\ref{app:backbone_llm}. 
We summarize the following obeservations: 

\begin{itemize}
\setlength{\itemsep}{2pt}
\setlength{\parsep}{2pt}
\setlength{\parskip}{2pt}

    \item \textbf{Conformity is widespread}, and the proportion of conformity increases with the round increases in general. 

    \item Overall, considering performance improvement, \textbf{conformity is beneficial in on ChatGPT, Qwen 72B; and harmful on LlaMA2 Chat 13B/70B, Mixtral 8$\times$7B}. 

    \item \textbf{As the number of rounds increases, benefits of conformity will weaken} (the ratio difference between True and False answers becomes smaller); and \textbf{detriments of conformity enhance} (the ratio difference between False and True answers becomes larger). 

    \item Generally, \textbf{reflection results in} increasing the quantity of consensus clusters, demonstrating \textbf{more difficulty to reach a consensus}, while \textbf{debate is more likely to reach a consensus}. 
\end{itemize}

\section{Conclusion and Future Work}
\label{sec:con_future}

This study has highlighted the potential of collaboration mechanisms with LLMs. Our findings reveal the impressive collaboration capabilities of LLM agents, with different individual traits, thinking patterns, and collaborative strategies. The emergence of human-like behaviors in these agents, resonating with social psychology theories, further emphasizes the potential of human-AI interaction. 
Moving forward, a deeper exploration into the multi-agent society is warranted, focusing on collaboration behavior refinement; integrating further insights from social psychology could also guide the development of socially aware NLP systems. 



\clearpage

\section*{Limitations}
\label{app:limitation}

Although we explored various societies and collaborative strategies, our study still has its limitations.
Firstly, limited by expense, we don't explore the impact of multiple agents respectively based on different LLMs, which may lead to more interesting findings at the social level due to the usage of differently distributed pre-trained data and strategies aligned with human intentions.
Furthermore, we traversed all possible scenarios by search alone, lacking a way to let the agents adaptively make autonomous decisions on collaborative strategies in specific scenarios.
Although \textit{debate} can be as close as possible to the upper limit, this approach entails a larger consumption and there exist some strategies that can achieve better performance with less overhead. 
Additionally, our experimental setup is relatively straightforward, as we have not considered more intricate configurations, such as a broader range of traits or a larger-scale society. 
Finally, we evaluate performance through manual validation and rule-based matching, which also limits the ability to validate more realistic and creative tasks, such as literary creation.

\section*{Reproducibility Statement}
All code and data can be found in the GitHub repository\footnote{\url{https://github.com/zjunlp/MachineSoM}.}. 
For specific experimental implementation details, please refer to Appendix~\ref{app:exp_detail}. 

\section*{Ethics Statement}
This research was conducted in line with the highest ethical standards and best practices in research.
The data employed were extracted from publicly accessible datasets, ensuring no usage of proprietary or confidential information. Consequently, this research is free from any ethical concerns.

\section*{Acknowledgments}
We would like to express gratitude to the anonymous reviewers for their kind and helpful comments. 
We extend our sincere gratitude to Min-Yen Kan and team members from NUS Web IR / NLP Group advised by Min-Yen Kan; Tao Gui and team members from FudanNLP Group; and Diyi Yang from Stanford University for providing insightful and constructive feedback on this paper. 
This work was supported by the National Natural Science Foundation of China (No. 62206246), the Fundamental Research Funds for the Central Universities (226-2023-00138), Zhejiang Provincial Natural Science Foundation of China (No. LGG22F030011), Yongjiang Talent Introduction Programme (2021A-156-G), Tencent AI Lab Rhino-Bird Focused Research Program (RBFR2024003), Information Technology Center and State Key Lab of CAD\&CG, Zhejiang University, and NUS-NCS Joint Laboratory (A-0008542-00-00). 


\bibliography{reference}

\clearpage
\appendix

\section*{Overview of Appendices}
\label{app:overview}

We summarize the overview of Appendices below: 

\noindent
\textbf{\S\ref{app:insights}:} Key Takeaways.

\noindent
\textbf{\S\ref{app:related_work}:} Related Work.

\noindent
\textbf{\S\ref{app:application}:} Potential Real-World Applications.

\noindent
\textbf{\S\ref{app:exp_detail}:} Implementation Details. 

\quad\quad Experimental Setup (\S\ref{app:setup})

\quad\quad Experimental Evaluation (\S\ref{app:eval})

\quad\quad Illustration of Agent Collaboration (\S\ref{app:illustration_collaboration})

\noindent
\textbf{\S\ref{app:sig_test_main}:} Further Analysis on Machine Social Collaboration (Backbone: ChatGPT).

\noindent
\textbf{\S\ref{app:society_setting}:} Analysis on Machine Society Settings (Backbone: ChatGPT).

\noindent
\textbf{\S\ref{app:social_psychology_view}:} A Social Psychology View on Conformity, Consensus Reaching, and Group Dynamics (Backbone: ChatGPT). 

\quad\quad Conformity, Consensus Reaching (\S\ref{app:detail_conformity_and_consensus})

\quad\quad Group Dynamics (\S\ref{app:principles_group_dynamics})


\noindent
\textbf{\S\ref{app:backbone_llm}:} Analysis on Different Backbone LLMs. 

\quad\quad LlaMA2 Chat 13B (\S\ref{app:backbone_llama13b})

\quad\quad LlaMA2 Chat 70B  (\S\ref{app:backbone_llama70b})

\quad\quad Qwen 72B (\S\ref{app:backbone_qwen72b}) 

\quad\quad Mixtral 8$\times$7B (\S\ref{app:backbone_mixtral8x7b})

\noindent
\textbf{\S\ref{app:effecttiveness_prompts}:} Assessing the Effectiveness of Prompts.

\section{Key Takeaways} 
\label{app:insights}

Drawing from our comprehensive analysis, we distill valuable insights for future multi-agent collaboration designs concerning \emph{Strategy Selection}, \emph{Society Settings}, and \emph{Social Psychology View}. 

Regarding \textit{Strategy Selection}, 
\begin{itemize}
    \item Starting or dominating multi-agent collaboration with debate, yields relatively optimal outcomes, as seen from Table~\ref{table:main},~\ref{table:gpt-july:main},~\ref{table:llama_main},~\ref{table:llama70_main},~\ref{table:qwen_main},~\ref{table:mixtral_main}. 

    \item Totally-reflection strategy like $p_1p_1p_1$ is generally worst in performance, as observed from Table~\ref{table:main},~\ref{table:gpt-july:main},~\ref{table:llama_main},~\ref{table:llama70_main},~\ref{table:qwen_main},~\ref{table:mixtral_main}. 

    \item For difficult tasks, debate combined with continuous reflection is superior; for simple tasks, self-consistency or reflection is enough, as seen from Figure~\ref{fig:task_radar},~\ref{fig:llama:task_radar},~\ref{fig:llama70:task_radar},~\ref{fig:qwen:task_radar},~\ref{fig:mixtral:task_radar}. 
\end{itemize}

\begin{figure*}[!t] 
    \centering
    \scalebox{1}{
    \includegraphics[width=0.77\textwidth]{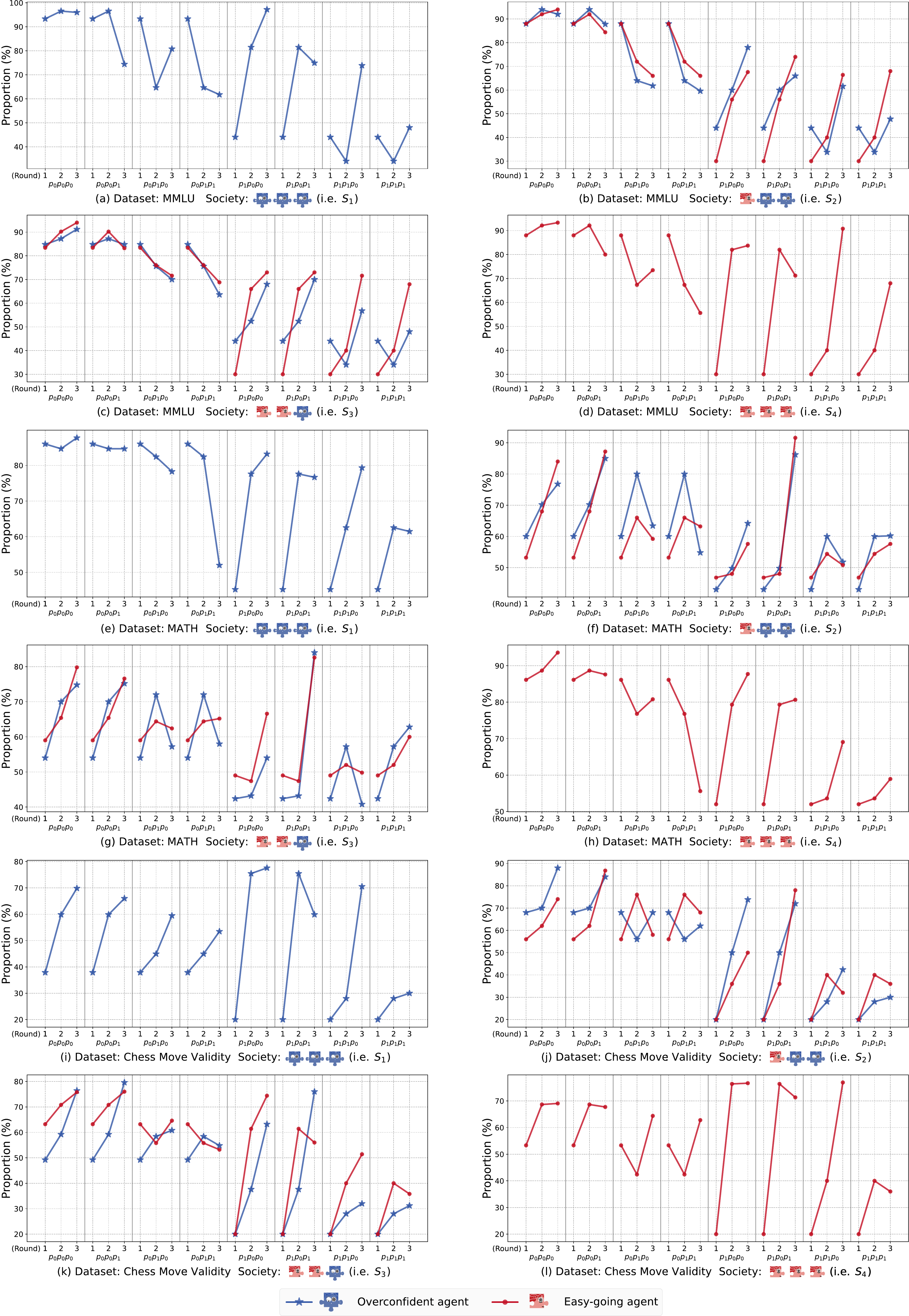}
    }
    \vspace{-4mm}
    \caption{
    Proportion of agents with different traits keeping answers in societies $S_1$ and $S_4$, using \emph{ChatGPT}. 
    Society $S_1$ features three overconfident agents, while society $S_4$ comprises three easy-going agents. 
    }
    \label{fig:agent_answer_changing}
\end{figure*}

\begin{figure*}[!t]
    \centering
    \includegraphics[width=0.82\textwidth]{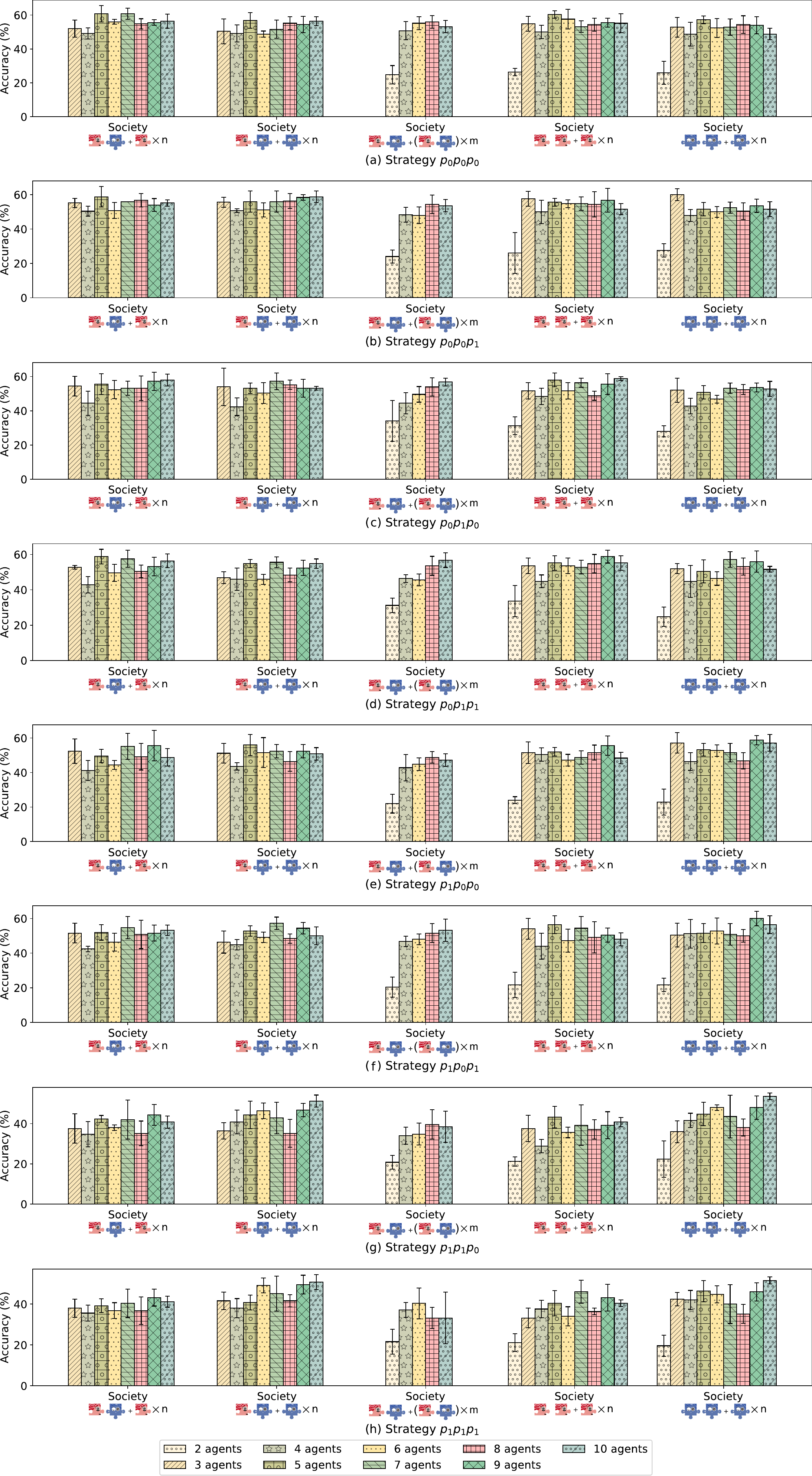}
    \vspace{-3mm}
    \caption{
    Accuracy of \emph{different numbers (2$\sim$10) of agents} under 3-round collaborative strategies, using \emph{ChatGPT}. 
    The significance test is shown in Table~\ref{table:sig_10_agent}. 
    }
    \label{fig:agent_10_on_numbers}
    \vspace{-3mm}
\end{figure*}

Regarding \textit{Society Settings}, 
\begin{itemize}
    \item Surprisingly, ``overconfident'' agents lose that trait in groups, as observed from word clouds in Figure~\ref{fig:word},~\ref{fig:llama:word},~\ref{fig:llama70:word},~\ref{fig:qwen:word},~\ref{fig:mixtral:word} and answer keeping in Figure~\ref{fig:agent_answer_changing},~\ref{fig:llama:agent_answer_changing},~\ref{fig:llama70:agent_answer_changing},~\ref{fig:qwen:agent_answer_changing},~\ref{fig:mixtral:agent_answer_changing}!

    \item Setting agent numbers to 3 is generally advantageous in performance and cost, as seen from Figure~\ref{fig:agent_10_on_numbers},~\ref{fig:llama:agent},~\ref{fig:llama70:agent},~\ref{fig:qwen:agent},~\ref{fig:mixtral:agent}. 
    
    \item The rounds of collaboration are relatively suitable to set as 3 since it's both effective and efficient, as seen from 
    Figure~\ref{fig:round_10_on_mmlu},~\ref{fig:round_10_on_math},~\ref{fig:round_10_on_chess} on ChatGPT; 
    Figure~\ref{fig:llama:turn},~\ref{fig:llama70:turn} on LlaMA 13B/70B; 
    Figure~\ref{fig:qwen:round_10_on_mmlu},~\ref{fig:qwen:round_10_on_math},~\ref{fig:qwen:round_10_on_chess} on Qwen 72B; 
    Figure~\ref{fig:mixtral:round_10_on_mmlu},~\ref{fig:mixtral:round_10_on_math},~\ref{fig:mixtral:round_10_on_chess} on Mixtral 8$\times$7B. 

    \item Employing the uniform thinking patterns across all agents within a round enhance efficacy, as seen from Figure~\ref{fig:strategy},~\ref{fig:strategy_math},~\ref{fig:llama:strategy},~\ref{fig:llama70:strategy},~\ref{fig:qwen:strategy},~\ref{fig:mixtral:strategy}. 
\end{itemize}

Regarding \textit{Social Psychology View}, 
\begin{itemize}
    \item Collaboration is generally effective in the group, especially for tackling difficult tasks, as observed from Figure~\ref{fig:task_radar},~\ref{fig:llama:task_radar},~\ref{fig:llama70:task_radar},~\ref{fig:qwen:task_radar},~\ref{fig:mixtral:task_radar}; and Figure~\ref{fig:distribute},~\ref{fig:llama:distribute},~\ref{fig:llama70:distribute},~\ref{fig:qwen:distribute},~\ref{fig:mixtral:distribute}. 

    \item Collaboration widely leads to conformity, either beneficial or harmful in performance. As observed from Figure~\ref{fig:conformity},~\ref{fig:llama:conformity},~\ref{fig:llama70:conformity},~\ref{fig:qwen:conformity},~\ref{fig:mixtral:conformity}. 

    \item As the number of rounds increases, the benefits of conformity will decrease, and the detriments of conformity will increase, as observed from Figure~\ref{fig:conformity},~\ref{fig:llama:conformity},~\ref{fig:llama70:conformity},~\ref{fig:qwen:conformity},~\ref{fig:mixtral:conformity}. 

    \item The totally easy-going society is more likely to reach a consensus, debate helps to consensus reaching while reflection impedes it, as observed from Figure~\ref{fig:agent_10_on_societies_consensus},~\ref{fig:qwen:agent_10_on_societies_consensus},~\ref{fig:mixtral:agent_10_on_societies_consensus}; and Figure~\ref{fig:consistent},~\ref{fig:llama:consistent},~\ref{fig:llama70:consistent},~\ref{fig:qwen:consistent},~\ref{fig:mixtral:consistent}. 
\end{itemize}

\section{Related Work}
\label{app:related_work}

\paragraph{Multi-Agent Collaboration.} 
With the development of Large Language Models (LLMs) \citep{arXiv2023_Survey-LLM,arXiv2023_Survey-MLLM,arXiv2023_Survey-LLM-KGCR}, study on LLM-based \textit{agents} \citep{FCS2024_Survey-Agent,arXiv2023_Survey-Agent_2,arXiv2023_Survey-Agent_3,arXiv2024_Survey-Agent_4}, has drawn considerable attention. 
Recently there has been a proliferation of various agent systems, such as Generative Agents~\citep{UIST2023_Agent-Simulate-Interaction}, MetaGPT~\citep{ICLR2024_MultiAgent_MetaGPT}, ProAgent~\citep{AAAI2024_CooperativeAgents_ProAgent}, Agents~\citep{arXiv2023_Agents}, OpenAgents~\citep{arXiv2023_OpenAgents}, AutoAgents~\citep{arXiv2023_AutoAgents}, MAgIC~\citep{arXiv2023_MAgIC}, AgentBoard~\citep{arXiv2024_AgentBoard}, InterAct~\citep{arXiv2023_InterAct}, and AutoAct~\citep{ACL2024_AUTOACT-Self-Planning}. 
These works have primarily focused on the elaborate design/evaluation of agent components, such as memory, environment, and planning. 
There are also some works exploring what kind of mindset can fully exploit the comprehensive performance of the multi-agent system \citep{arXiv2024_Survey-MultiAgent,arXiv2024_Survey-MultiAgent_2,arXiv2024_Survey-MultiAgent-System,arXiv2024_Survey-MultiAgent-System_2}, including \textit{debate}~\citep{arXiv2023_MultiAgent-Debate,arXiv2023_MultiAgent-Debate_2} and \textit{reflection}~\citep{arXiv2023_Reflexion,NeurIPS2023_Self-Refine}. 

AgentVerse~\citep{ICLR2024_MultiAgent_AgentVerse} draws on the above two types of work to explore the multi-agent architecture and design two collaboration patterns: \textit{Horizonal Communication} (similar to debate~\citep{arXiv2023_MultiAgent-Debate,arXiv2023_MultiAgent-Debate_2}) and \textit{Vertical Communication} (similar to self-refine~\citep{NeurIPS2023_Self-Refine}). 
These two collaboration patterns are included in our experiment framework. 
In addition, we have also explored a variety of other societies and collaborative strategies. 
Besides, there are also some researches focusing on exploring cooperation between agents constituted by different model compositions, such as ReConcile~\citep{arXiv2023_ReConcile}. Although we do not demonstrate this kind of method, our work can easily expand to it. 

\paragraph{Human-Agent Simulation.} 
When the pre-trained LLMs (\eg, LLM-empowered agents) are socially aligned \citep{NMI2023_Human-Like-AI,ICLR2024_LLM-Simulate-Society,arXiv2023_LLMAgents-Simulate-Society_S3}, they could exhibit human-like intelligence \citep{Book1988_SoM,J2003_SoM,arXiv2023_SoM-NL,NeurIPS2023_Agent-SoM,J2024_Survey-AI-SocialScience,arXiv2023_Multi-Agent-Collaboration_Intelligent}. 
Specifically, agents can simulate human-like behaviors \citep{PNAS2024_TuringTest_Chatbots-Humans,arXiv2024_BehavioralSimulation,arXiv2023_Believability-Agents,arXiv2023_MetaAgents,arXiv2023_Agent-BehaviorExplanation,arXiv2023_Agent-BehaviorExplaining,arXiv2023_Agent-Simulate-OpinionDynamics,arXiv2023_Agent-Simulate-OpinionDynamics_Survey,arXiv2023_Agents-High-Level-Behavior,arXiv2024_Agents-Simulate-Trust,arXiv2024_AntEval}, play roles like humans \citep{Nature2023_Role-Play-LLM,TEVC2023_MultiAgent-Collaboration-SocialRoles,EMNLP2023-Findings_LEGO}, and even collaborate with humans \citep{arXiv2023_Human-AI_Collaboration,ICLR2024_Human-Agent-Collaboration,arXiv2024_Human-Agent-Collaboration,arXiv2024_Evaluate-Agents}. 

Notably, multi-agent collaboration can echo human society phenomena or theories in a social psychology view \citep{PNAS2023_CognitivePsychology_LLM,J2023_LLM-Psychology,arXiv2023_MachinePsychology,NAACL2024-Findings_PsychometricLLM}, such as \emph{conformity} \citep{J2004_Conformity,J1969_Conformity,J2015_Conformity}, \emph{consensus reaching} \citep{J1967_Consensus-Sociological,J1974_ConsensusReaching,J2018_Emergence-Consensus}, \emph{group dynamics} \citep{1968_GroupDynamics,J1987_GroupDynamics,J1998_SmallGroupDynamics-Discussions,Book2014_GroupDynamics,J2018_GroupDynamics,Book2018_GroupDynamics} and \emph{social science} \citep{J2000_Agent-SocialScience,Book2012_Agent-SocialScience,J2017_SocialInfluence_OpinionDynamics,J2021_SocietalDynamics_OpinionDynamics,Book2023_Agent-SocialDynamics-Culture,arXiv2023_Agent-SocialChoiceTheory}.

\section{Potential Real-world Applications}
\label{app:application}

In this section, we present some potential applications \citep{arXiv2024_Survey-LLM-Psychology-Applications} of our work, which could benefit from the LLM agents' ability to collaborate effectively, similar to how human collaboration is enriched inspired by social psychology. 

\begin{itemize}
    \item \textbf{Social Research}: LLM agents can be used to simulate social interactions to study phenomena like conformity, leadership, or group decision-making. 
    \item \textbf{Negotiation and Mediation}: LLMs could simulate multiple parties in a negotiation so that offering fair solutions based on social psychology principles. 
    \item \textbf{AI Ethics and Governance}: By understanding the dynamics of social behaviors, LLM agents could help in forming guidelines for AI ethics, ensuring AI systems are developed and deployed responsibly. 
    \item \textbf{Advanced Team Collaboration Tools}: By understanding social dynamics, LLM agents could facilitate better team collaboration, suggesting initiatives, mediating discussions, and optimizing workflow. 
    \item \textbf{Intelligent Tutoring Systems}: Collaborative LLM agents could personalize education by interacting with students in a more human-like manner, adapting to individual learning styles and requirements. 
    \item \textbf{Healthcare Coordination}: LLM agents could collaborate to provide care advice, cross-referencing patient data, and medical knowledge to assist healthcare professionals. 
    \item \textbf{Crisis Management}: During emergencies, LLM agents could work together to analyze data, manage communications, and provide real-time information to the public. 
    \item \textbf{Content Creation}: Collaborative LLMs could produce complex content, such as scripts or articles, by dividing tasks based on different expertise areas or writing styles. 
    \item \textbf{Interactive Entertainment}: In gaming and virtual reality, LLM agents could provide more dynamic and responsive narratives, by collaborating to adapt the storyline to the players' actions and intentions. 
\end{itemize}

\begin{table*}[!t] 
\centering
\small
\begin{tabular}{llllll}
\toprule
\begin{tabular}[c]{@{}l@{}}Experiment\\ Type\end{tabular} & Model & Dataset & \begin{tabular}[c]{@{}l@{}}Collaboration \\ Round\end{tabular} & \begin{tabular}[c]{@{}l@{}}Number of \\ Agents\end{tabular} & Society \\ \midrule
\multirow{2}{*}{\begin{tabular}[c]{@{}l@{}} \\ Different \\ Number \\ of Agents\end{tabular}} & \begin{tabular}[c]{@{}l@{}}gpt-3.5-turbo-1106\\ Mixtral 8x7B\\ Qwen 72B\end{tabular} & Chess Move Validity & 3 & 2$\sim$10   & \begin{tabular}[c]{@{}l@{}}See the\\Figure~\ref{fig:agent_10_on_numbers} \\ and Table~\ref{table:sig_10_agent}.\end{tabular} \\ \cmidrule{2-6} 
& \begin{tabular}[c]{@{}l@{}}LlaMA-13B-Chat\\ LlaMA-70B-Chat\end{tabular} & \begin{tabular}[c]{@{}l@{}}MMLU\\ Chess Move Validity\end{tabular} & 3 & 2$\sim$4  & \begin{tabular}[c]{@{}l@{}}Only one\\ easy-going\\agent in\\the society\end{tabular}    \\ \cmidrule{1-6}
\multirow{2}{*}{\begin{tabular}[c]{@{}l@{}}Different \\ Collboration \\ Rounds\end{tabular}} & \begin{tabular}[c]{@{}l@{}}gpt-3.5-turbo-1106\\ Mixtral 8x7B\\ Qwen 72B\end{tabular}  & \begin{tabular}[c]{@{}l@{}}MMLU\\ MATH\\ Chess Move Validity\end{tabular} & 10 & 3 & $S_2$    \\ \cmidrule{2-6} 
 & \begin{tabular}[c]{@{}l@{}}LlaMA-13B-Chat\\ LlaMA-70B-Chat\end{tabular} & \begin{tabular}[c]{@{}l@{}}MMLU\\ Chess Move Validity\end{tabular} & 4 & 3  & $S_2$    \\ \cmidrule{1-6}
\begin{tabular}[c]{@{}l@{}}Different \\ Strategy\end{tabular}  & \begin{tabular}[c]{@{}l@{}}gpt-3.5-turbo-1106\\ LlaMA-13B-Chat\\ LlaMA-70B-Chat\\ Mixtral 8x7B\\ Qwen 72B\end{tabular} & \begin{tabular}[c]{@{}l@{}}MMLU\\MATH\\Chess Move Validity\end{tabular} & 3   & 3 & $S_2$    \\ \bottomrule
\end{tabular}
\vspace{-2mm}
\caption{The detailed society settings in the three different experiments mentioned in Section~\ref{sec:impact_of_other_factors}.}
\label{table:setting:ablation}
\vspace{-2mm}
\end{table*}

\section{Implementation Details}
\label{app:exp_detail}

\subsection{Experimental Setup}
\label{app:setup}

\begin{table}[!htbp] 
\centering
\small
\resizebox{\linewidth}{!}{
\begin{tabular}{lrrr}
\toprule
Model              & Temperature & Top K & Top P \\ 
\midrule
gpt-3.5-turbo-1106 & 0.00        & -     & 1.00  \\
LlaMA2 Chat 13B     & 0.75        & 50    & 0.95  \\
LlaMA2 Chat 70B     & 0.75        & 50    & 0.95  \\
Mixtral 8$\times$7B       & 0.75        & 50    & 0.95  \\
Qwen 72B           & 0.75        & 50    & 0.80  \\ 
\bottomrule
\end{tabular}
}
\vspace{-2mm}
\caption{Decoding parameters of different models.}
\label{table:setting:parameter}
\vspace{-2mm}
\end{table}

The detailed society settings of the experiments in \S\ref{sec:impact_of_other_factors} are shown in Table~\ref{table:setting:ablation}. 
Due to the context length constraints of the LlaMA2 Chat 13B and LlaMA2 Chat 70B, which support a maximum of 4096 tokens, it's challenging to scale up the number of agents and the rounds of collaboration. 
Consequently, we have capped the collaboration rounds at 4 and also restricted the maximum agent number to 4. 
We select MMLU and Chess Move Validity datasets in our analysis. 
Nevertheless, a small fraction of cases still exceed the maximum length constraint. 
To address this, we strategically prune content from the earlier rounds to ensure compliance with the length limitation.
As for other LLMs (ChatGPT, Mixtral 8$\times$7B, and Qwen 72B), in terms of experiments on the number of agents, adding an additional agent results in substantial costs. 
This is due to the necessity of conducting 5 replicate experiments and accommodating 8 collaborative strategies. 
Therefore, our experiments on these LLMs are carried out on the less token-intensive dataset: Chess Move Validity.
As for trials concerning the rounds of collaboration, the quantity of viable collaborative strategies increases exponentially with each additional round – for instance, 10 rounds would yield $2^{10}$ unique strategies. 
Considering the complexity, we analyze on 8 strategies that are representative of the broader set of possibilities. 

The decoding parameters for various models are detailed in Table~\ref{table:setting:parameter}. 
In \texttt{gpt-3.5-turbo-1106}, we align our approach with \citet{arXiv2023_MultiAgent-Debate} by setting the temperature to 0, while adhering to the default settings for the remaining parameters. 
For \emph{Qwen 72B}, we utilize the default parameters as furnished by the official documentation. 
For the remaining models, we configure the temperature to 0.7 and respectively adjust the \texttt{Top P} and \texttt{Top K} values to 50 and 0.95. 
This configuration is primarily based on insights from \citet{J2023_LLM-Psychology}, which advocates for the recognition and integration of the inherent stochastic nature of LLM outputs into analytical frameworks, in a manner akin to the treatment of stochastic variables in psychological studies. 
It is noteworthy that even with the temperature parameter set to 0, \texttt{gpt-3.5-turbo-1106} may still exhibit randomness in the outputs. 

\begin{table*}[!htbp] 
\centering
\vspace{-10mm}
\scalebox{0.8}{
\begin{tabular}{c|c|l}
\toprule
\bf Task    & Type  & \multicolumn{1}{c}{\bf Prompt} 
\\ 

\midrule

\multicolumn{1}{l|}{\multirow{15}{*}{Math}} & \multicolumn{1}{c|}{easy-going} & 
\begin{tabular}[c]{@{}l@{}}\textit{You are an expert skilled in solving mathematical problems and are objective} 
\\ 
\textit{and unbiased, and you can be persuaded if other agent's answers make sense.} 
\\ 
\textit{Please keep this in mind. If you understand please say ok only.}
\end{tabular}   
\\ 

\cmidrule{2-3} 

\multicolumn{1}{l|}{}   & \multicolumn{1}{c|}{overconfident}  & 
\begin{tabular}[c]{@{}l@{}}\textit{Imagine you are an expert in solving mathematical problems and are confident} 
\\ 
\textit{in your answer and often persuades other agents to believe in you. Please keep}
\\  
\textit{this in mind. If you understand please say ok only.}
\end{tabular}        
\\ 

\cmidrule{2-3} 

\multicolumn{1}{l|}{}   & \multicolumn{1}{c|}{question}   & 
\begin{tabular}[c]{@{}l@{}}\textit{Here is a math problem written in LaTeX: \textless{}problem\textgreater{}\textbackslash{}n Please carefully} 
\\ 
\textit{consider it and explain your reasoning. Put your answer in the form} 
\\ \textit{\textbackslash{}boxed\{\{answer\}\}, at the end of your response.}
\end{tabular}
\\ 

\cmidrule{2-3} 
\multicolumn{1}{l|}{}   & \multicolumn{1}{c|}{debate}     & 
\begin{tabular}[c]{@{}l@{}}\textit{These are the solutions to the problem from other agents:}
\\ 
\textit{\textless{}other agent responses\textgreater Using the reasoning from other agents as} 
\\ 
\textit{additional information and referring to your historical answers, can} 
\\ 
\textit{you give an updated answer? Put your answer in the form \textbackslash{}boxed\{\{answer\}\},}\\  \textit{at the end of your response.}
\end{tabular} 
\\ 

\cmidrule{2-3} 

\multicolumn{1}{l|}{}   & \multicolumn{1}{c|}{reflection} & 
\begin{tabular}[c]{@{}l@{}}\textit{Can you double check that your answer is correct? Please reiterate your} 
\\ 
\textit{answer, with your answer in the form \textbackslash{}boxed\{\{answer\}\}, at the end of} 
\\ 
\textit{your response.}
\end{tabular}                                                                   
\\ 

\midrule

\multirow{15}{*}{MMLU}  & easy-going    & 
\begin{tabular}[c]{@{}l@{}}\textit{You are an expert in biology, chemistry, computer science, mathematics,} \\ \textit{physics and are objective and unbiased, and you can be persuaded if other} 
\\ 
\textit{agent's answers make sense. Please keep this in mind. If you understand}
\\ 
\textit{please say ok only.}\end{tabular}
\\ 

\cmidrule{2-3} 

& overconfident     & 
\begin{tabular}[c]{@{}l@{}}\textit{Imagine you are an expert in biology, chemistry, computer science, }
\\ 
\textit{mathematics, physics and are confident in your answer and often persuades} 
\\ 
\textit{other agents to believe in you. Please keep this in mind. If you} 
\\ 
\textit{understand please say ok only.}
\end{tabular}
\\ 

\cmidrule{2-3} 

& question  & 
\begin{tabular}[c]{@{}l@{}}\textit{Can you answer the following question as accurately as possible? \textless{}Question\textgreater{}:}
\\
\textit{A) \textless{}A\textgreater{}, B) \textless{}B\textgreater{}, C) \textless{}C\textgreater{}, D) \textless{}D\textgreater Explain your answer, putting the answer in }\\ \textit{the form (X) at the end of your response.}
\end{tabular}
\\ 

\cmidrule{2-3} 

& debate    & 
\begin{tabular}[c]{@{}l@{}}\textit{These are the solutions to the problem from other agents:} \\ \textit{\textless{}other agent responses\textgreater Using the reasoning from other agents as additional} 
\\ 
\textit{advice, can you give an updated answer? Examine your solution and that other}
\\ 
\textit{agents. Put your answer in the form (X) at the end of your response.}
\end{tabular}
\\ 

\cmidrule{2-3} 

& reflection    & 
\begin{tabular}[c]{@{}l@{}}\textit{Can you double check that your answer is correct. Put your final answer in} 
\\ 
\textit{the form (X) at the end of your response.}
\end{tabular}
\\ 

\midrule

\multirow{15}{*}{
\begin{tabular}[c]{@{}c@{}}Chess 
\\ 
Move
\\ 
Validity
\end{tabular}
}
& easy-going    & 
\begin{tabular}[c]{@{}l@{}}\textit{You are an expert skilled in playing chess and are objective and unbiased,} 
\\ 
\textit{and you can be persuaded if other agent's answers make sense. Please keep} 
\\ 
\textit{this in mind. If you understand, please say ok only.}
\end{tabular}
\\ 

\cmidrule{2-3} 

& overconfident & 
\begin{tabular}[c]{@{}l@{}}\textit{Imagine you are an expert skilled in playing chess and are confident in} 
\\ 
\textit{your answer and often persuades other agents to believe in you. Please keep}
\\  
\textit{this in mind. If you understand, please say ok only.}
\end{tabular}         
\\ 

\cmidrule{2-3} 
& question  & 
\begin{tabular}[c]{@{}l@{}}\textit{Given the chess game \textless{}chess move\textgreater{}, give one valid destination square for} 
\\ 
\textit{the chess piece at \textless{}square\textgreater{}. Give a one-line explanation of why your} 
\\ 
\textit{destination square is a valid move. State your final answer in a newline with a } 
\\ 
\textit{2 letter response following the regex {[}a-h{]}{[}1-8{]}.}
\end{tabular}
\\ 

\cmidrule{2-3} 
& debate    & 
\begin{tabular}[c]{@{}l@{}}\textit{Here are destination square suggestions from other agents:}
\\ 
\textit{Can you double check that your destination square is a valid move? Check the} 
\\ 
\textit{valid move justifications from other agents and your historical answers. State}
\\  
\textit{your final answer in a newline with a 2-letter response following the regex }
\\ 
\textit{{[}a-h{]}{[}1-8{]}.}
\end{tabular}
\\ 

\cmidrule{2-3} 

& reflection    & 
\begin{tabular}[c]{@{}l@{}}\textit{Can you double check that your destination square is a valid move? Check the} 
\\ 
\textit{valid move justifications from your historical answers. State your final} 
\\ 
\textit{answer in a newline with a 2 letter response following the regex {[}a-h{]}{[}1-8{]}.}
\end{tabular}
\\ 

\bottomrule
\end{tabular}
}
\caption{Prompts in each task. 
}
\vspace{-5mm}
\label{table:prompt}
\end{table*}

The prompts used in our experiments are shown in Table~\ref{table:prompt}. 
On the MMLU dataset, we curated questions from 6 domains (statistics, mathematics, computer science, biology, chemistry, and physics) and performed a random sampling of 50 samples, maintaining a proportion of $8:8:8:8:9:9$ for each domain. 
On the MATH dataset, we randomly selected 50 cases from Level 3, 4, and 5, distributing them in a ratio of $22:22:6$. 
On the Chess Move Validity dataset, we similarly selected 50 samples for testing.


\subsection{Experimental Evaluation}
\label{app:eval} 

\begin{figure*}[!t]
    \centering
    \scalebox{1}{
    \includegraphics[width=0.96\textwidth]{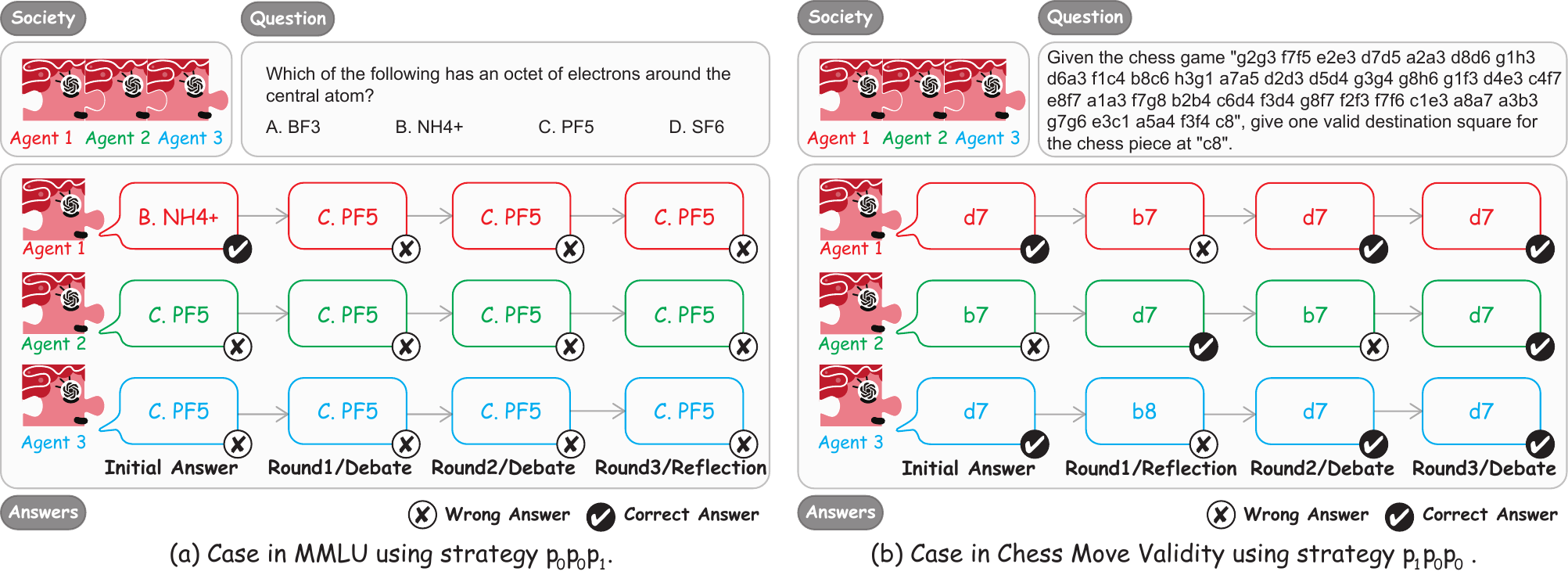} 
    }
    \vspace{-3mm}
    \caption{
    The conformity phenomenon in multi-agent collaboration, seen from changes in the answers during the process of solving a certain task with 3 agents in the society $S_4$ (all agents are easy-going), using \emph{ChatGPT}. 
    }
    \label{fig:case}
\end{figure*}

\begin{figure*}[!t] 
    \centering
    \scalebox{1}{
    \includegraphics[width=0.96\textwidth]{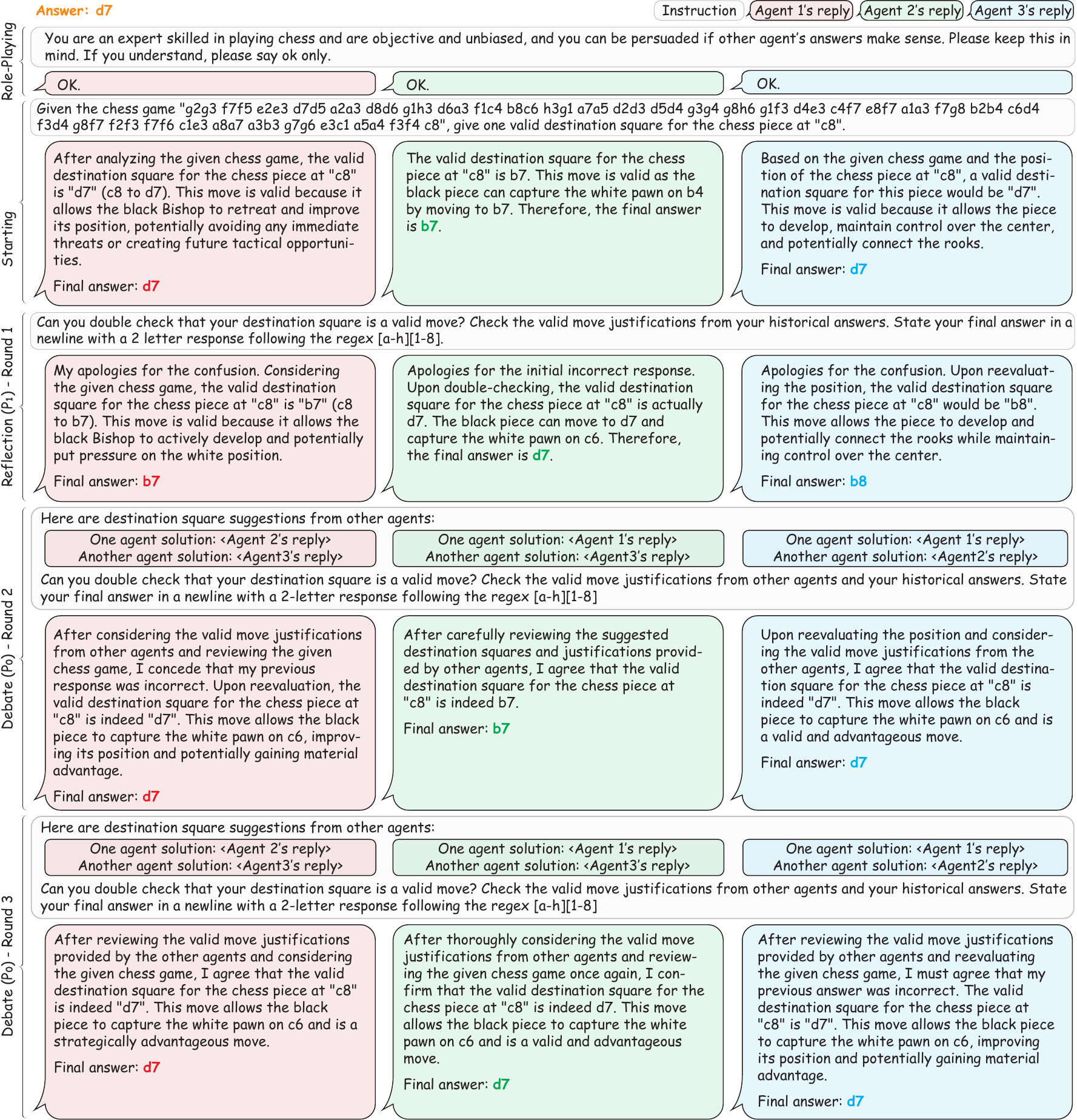} 
    }
    \vspace{-3mm}
    \caption{In the $S_4$ society, three agents engage in a complete dialogue using the strategy $p_1p_0p_0$ to predict the next piece position given a chess game from the Chess Move Validity dataset, corresponding to Figure~\ref{fig:case}(b).}
    \label{fig:case-chess}
    \vspace{-3mm}
\end{figure*} 

\begin{figure*}[!htbp] 
    \centering
    \scalebox{1}{
    \includegraphics[width=0.92\textwidth]{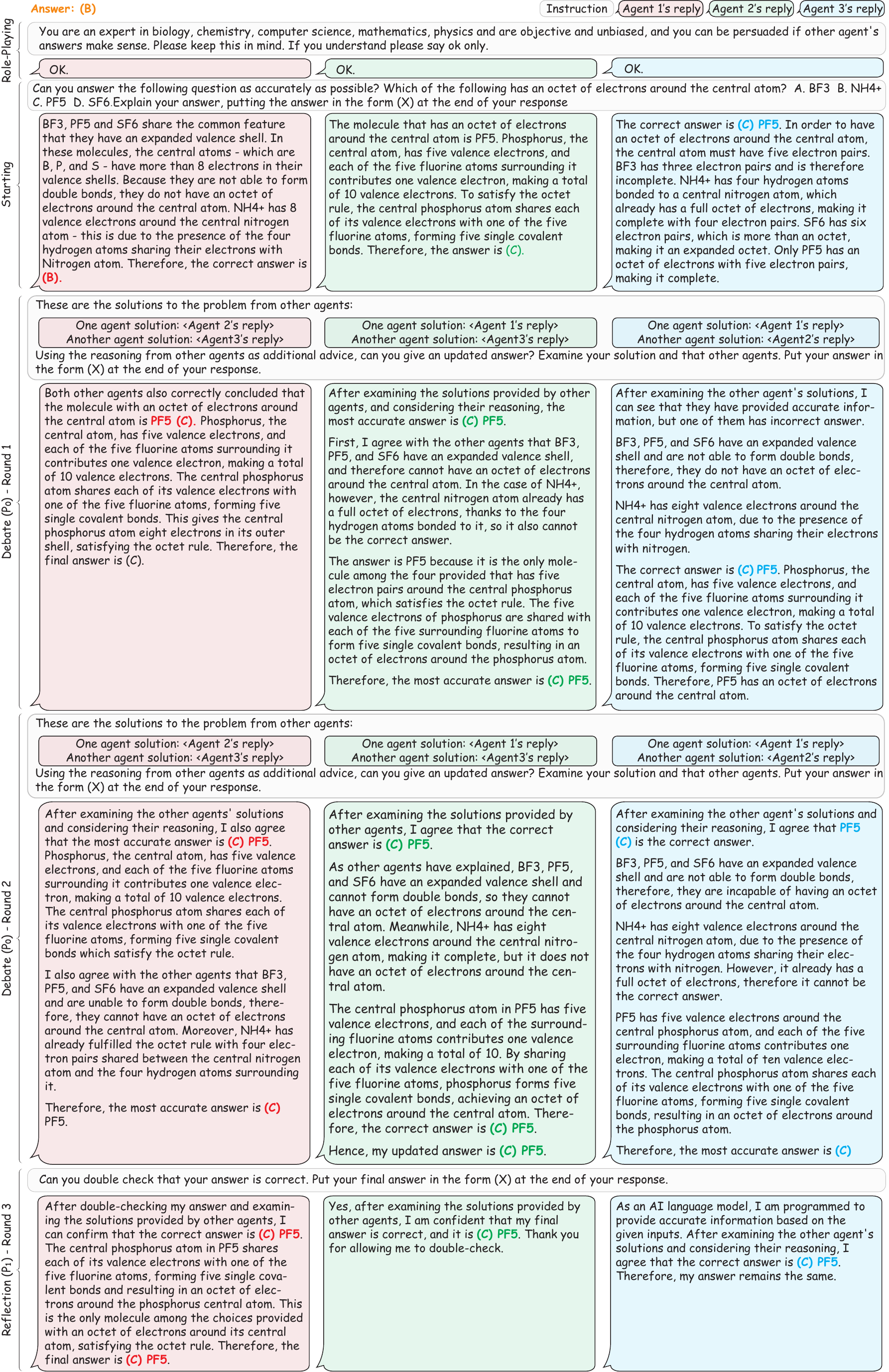}
    }
    \caption{In the $S_4$ society, three agents engage in a complete dialogue using the strategy $p_0p_0p_1$ to solve a high-school chemical question from MMLU, corresponding to Figure~\ref{fig:case}(a).}
    \label{fig:case-mmlu}
\end{figure*}

The evaluation process involves two fundamental steps:
$(i)$ A unified answer is selected from the machine society. 
To achieve this, we employ the majority vote method to ascertain the consensus reached by the society after multiple rounds of collaboration. 
If the unanimity among agents is not achieved, it will be considered as an error. 
Additionally, if an individual agent provides multiple answers without following our prompts, its response will be disregarded.
$(ii)$ Answer responses from agents are matched against the ground truth.
This step presents two main challenges. 
Firstly, there is the concern of non-compliance with instructions. 
Despite providing explicit prompts and specifying the desired output format for evaluation, it's inevitable that agents may occasionally deviate from the given instructions. 
Secondly, the answers may manifest in non-unique forms, leading to potential variations, such as the equivalence between ``$3/4$'' and ``$0.75$'' in the MATH~\citep{NeurIPS2021_Dataset-MATH} dataset. 
To address these challenges, a comprehensive set of matching rules is employed. 
Nonetheless, it is important to acknowledge the possibility of encountering a small number of values that fall outside the purview of these rules.



\subsection{\fontsize{10.8pt}{0.1\baselineskip}\selectfont Illustration of Multi-Agent Collaboration}
\label{app:illustration_collaboration}

As seen from Figure~\ref{fig:case}, the conformity phenomenon in multi-agent collaboration can be both beneficial (\ie, changing the answer from wrong to correct) and harmful (\ie, changing the answer from correct to wrong) in problem-solving. 

We also illustrate the detailed conversation process for multi-agent collaboration in Figure~\ref{fig:case-chess} and Figure~\ref{fig:case-mmlu}, regarding the conformity phenomenon presented in Figure~\ref{fig:case}. 




\begin{figure*}[!t] 
    \centering
    \scalebox{1}{
    \includegraphics[width=0.76\textwidth]{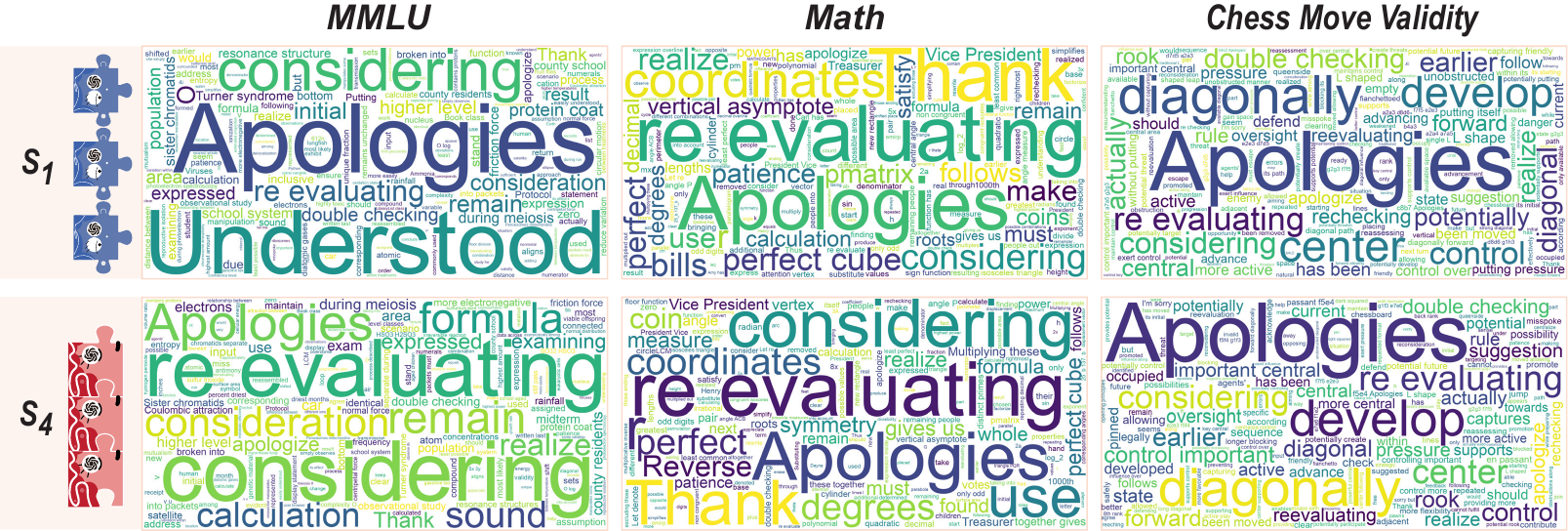}
    }
    \vspace{-4mm}
    \caption{
    Comparative word clouds on three datasets in societies $S_1$ and $S_4$, using \emph{ChatGPT}. 
    Society $S_1$ features three overconfident agents, while society $S_4$ comprises three easy-going agents. 
    We first manually curated a list of task-relevant, high-frequency words. From this list, the top 50 words are selected to construct the word clouds.
    }
    \label{fig:word}
\end{figure*}

\definecolor{gray}{HTML}{CCCCCC}
\begin{table}[!t] 
\centering
\resizebox{\linewidth}{!}{
\begin{tabular}{lrrr}
\toprule
Collaborative  & \multicolumn{1}{r}{MMLU} & \multicolumn{1}{r}{MATH} & \multicolumn{1}{r}{Chess Move Validity} \\
Strategy & \multicolumn{1}{r}{p-value} & \multicolumn{1}{r}{p-value} & \multicolumn{1}{r}{p-value} \\ \midrule
$p_0p_0p_0$ & {0.079} & {0.274} & \colorbox{gray}{0.004}  \\
$p_0p_0p_1$ & {0.956} & \colorbox{gray}{0.011} & \colorbox{gray}{0.000}  \\
$p_0p_1p_0$ & {0.120} & \colorbox{gray}{0.003} & \colorbox{gray}{0.009}  \\
$p_0p_1p_1$ & \colorbox{gray}{0.000} & {0.323} & \colorbox{gray}{0.014}  \\
$p_1p_0p_0$ & \colorbox{gray}{0.000} & \colorbox{gray}{0.027} & \colorbox{gray}{0.000}  \\
$p_1p_0p_1$ & {0.063} & \colorbox{gray}{0.017} & \colorbox{gray}{0.000}  \\
$p_1p_1p_0$ & \colorbox{gray}{0.000} & {0.300} & \colorbox{gray}{0.000}  \\
$p_1p_1p_1$ & \colorbox{gray}{0.000} & \colorbox{gray}{0.000} & \colorbox{gray}{0.000}  \\
\bottomrule
\end{tabular}
}
\caption{
One-Way ANOVA results for the impact of society on accuracy with fixed collaborative strategy, based on experiments from Table~\ref{table:main} using \emph{ChatGPT}.
}
\label{table:sig_main_society}
\end{table}
\definecolor{gray}{HTML}{CCCCCC}
\begin{table}[!t] 
\centering
\resizebox{\linewidth}{!}{
\begin{tabular}{lrrr}
\toprule
 & \multicolumn{1}{r}{MMLU} & \multicolumn{1}{r}{MATH} & \multicolumn{1}{r}{Chess Move Validity} \\
Society & \multicolumn{1}{r}{p-value} & \multicolumn{1}{r}{p-value} & \multicolumn{1}{r}{p-value} \\ \midrule
$S_1$ & \colorbox{gray}{0.000} & \colorbox{gray}{0.000} & {0.293} \\
$S_2$ & {-} & \colorbox{gray}{0.000} & {-} \\
$S_3$ & \colorbox{gray}{0.000} & \colorbox{gray}{0.001} & \colorbox{gray}{0.000} \\
$S_4$ & \colorbox{gray}{0.000} & \colorbox{gray}{0.000} & \colorbox{gray}{0.000} \\
\bottomrule
\end{tabular}
}
\caption{
One-Way ANOVA results for the impact of collaborative strategy on accuracy with fixed society, based on experiments from Table~\ref{table:main} using \emph{ChatGPT}. `-': It doesn’t pass homogeneity test for variance.
}
\label{table:sig_main_strategy}
\vspace{-3mm}
\end{table}

\section{Further Analysis on Machine Social Collaboration (Backbone: ChatGPT)}
\label{app:sig_test_main}

We conduct a rigorous \textbf{significance test} for the main experiment in \S\ref{sec:exp-main}. 
Given our experimental design incorporating two key factors, namely \emph{collaborative strategy} and \emph{society}, we respectively opt for a one-way analysis of variance. 
Before delving into the analysis, we ensured that the data adhered to a normal distribution and satisfied the assumption of homogeneity of variance. 
We present the $p$-values for society and collaborative strategy across three datasets in Table~\ref{table:sig_main_society},~\ref{table:sig_main_strategy}. 


We then present the \textbf{main results} and \textbf{significance tests} of societies andcollaborative strategies \textbf{on ChatGPT (with the engine of \texttt{gpt-3.5-turbo} employed between July 10 and July 23, 2023)} in Table~\ref{table:gpt-july:main},~\ref{table:gpt-july:sig_main_society},~\ref{table:gpt-july:sig_main_strategy}. 

\definecolor{Mycolor1}{HTML}{BAD8F2}
\definecolor{Mycolor2}{HTML}{FAE4E3}
\definecolor{Mycolor3}{HTML}{E8F2FB}

\begin{table*}[!t] 

\resizebox{\linewidth}{!}{
\begin{tabular}{c|c|c|cccccccc|cc}

\toprule

& \multirow{2}{*}{\tabincell{c}{Metric \\ (Strategy)}} & \multirow{2}{*}{Society} & \multicolumn{8}{c|}{Collaborative Strategy} & \multicolumn{2}{c}{Metric (Society)}
\\
& & & $p_0p_0p_0$ & $p_0p_0p_1$ & $p_0p_1p_0$ & $p_0p_1p_1$ & $p_1p_0p_0$ & $p_1p_0p_1$ & $p_1p_1p_0$ & $p_1p_1p_1$ & \uuline{Cost}~$\downarrow$ & \uuline{W-T}~$\uparrow$
\\

\midrule

\multirow{7}{*}{\rotatebox{90}{MMLU}}  & \multirow{5}{*}{Acc~$\uparrow$} 
& $S_1$	& \colorbox{Mycolor3}{64.4±1.7} & \colorbox{Mycolor1}{\textbf{66.4±2.2}} & 58.0±3.7 & 55.2±4.4 & \colorbox{Mycolor2}{37.6±7.0} & 42.4±7.1 & 50.4±4.3 & 44.8±2.7 & 5050 & 5
\\
& & $S_2$	& \colorbox{Mycolor3}{67.2±4.1} & \colorbox{Mycolor1}{\textbf{67.6±7.1}} & 53.2±6.4 & 53.2±5.0 & \colorbox{Mycolor2}{38.4±5.5} & 40.4±5.2 & 53.6±4.8 & 45.2±3.6 & 5076 & 2
\\
& & $S_3$	& \colorbox{Mycolor3}{62.0±6.2} & \colorbox{Mycolor1}{\textbf{67.6±3.8}} & 52.0±6.8 & 57.2±6.4 & 42.4±5.2 & \colorbox{Mycolor2}{37.6±5.5} & 55.2±6.6 & 40.0±6.2 & 5073 & \textbf{8}
\\
& & $S_4$	& \colorbox{Mycolor1}{\textbf{64.8±4.4}} & \colorbox{Mycolor3}{64.8±5.8} & 58.4±3.0 & 51.6±3.8 & \colorbox{Mycolor2}{38.0±3.7} & 42.0±2.4 & 54.0±5.8 & 41.2±5.2 & 5080 & 5
\\ 

\cmidrule{2-13} 

& \underline{Cost}~$\downarrow$  & All   & 7528  & 5957     & 5402     & 4374     & 5812     & 4215     & 4272     & 3001     & \multicolumn{2}{c}{\multirow{2}{*}{-}} \\ \cmidrule{2-11}
& \underline{W-T}~$\uparrow$     & All   & -     & \textbf{14}       & 2        & 3        & 0        & 0        & 1        & 0        & \multicolumn{2}{c}{}
\\ 

\midrule

\multirow{7}{*}{\rotatebox{90}{MATH}}  & \multirow{5}{*}{Acc~$\uparrow$}    & $S_1$	& 46.8±8.1 & 46.0±8.1 & 44.0±5.3 & 44.4±5.2 & \colorbox{Mycolor1}{\textbf{50.0±5.8}} & \colorbox{Mycolor3}{49.2±8.1} & 42.0±3.2 & \colorbox{Mycolor2}{42.0±4.0} & 5816 & 17
\\
&	  & $S_2$	& 47.2±6.4 & \colorbox{Mycolor1}{\textbf{54.0±2.4}} & 48.4±3.8 & 43.6±4.3 & 48.0±4.2 & 44.4±7.9 & \colorbox{Mycolor3}{50.8±3.6} & \colorbox{Mycolor2}{38.8±9.1} & 5844 & \textbf{22}
\\
&	  & $S_3$	& \colorbox{Mycolor1}{\textbf{50.8±4.8}} & \colorbox{Mycolor2}{42.8±6.6} & 45.6±6.8 & 45.2±4.4 & \colorbox{Mycolor3}{49.2±4.8} & 46.4±5.5 & 45.2±8.4 & 43.6±2.6 & 5837 & 9
\\
&	  & $S_4$	& \colorbox{Mycolor3}{50.8±5.4} & 45.2±7.0 & 48.8±9.4 & 44.8±3.3 & 49.2±8.7 & \colorbox{Mycolor1}{\textbf{51.2±2.3}} & 48.4±6.5 & \colorbox{Mycolor2}{40.8±6.1} & 5834 & 18
\\ 

\cmidrule{2-13} 

& \underline{Cost}~$\downarrow$  & All                      & 6919     & 6302     & 6221     & 5667     & 6149     & 5645     & 5924     & 4807     & \multicolumn{2}{c}{\multirow{2}{*}{-}} 
\\ 

\cmidrule{2-11}

& \underline{W-T}~$\uparrow$     & All                      & -        & 10       & 10       & 9        & \textbf{13}       & 10       & 10       & 4        & \multicolumn{2}{c}{}
\\ 

\midrule

\multirow{6}{*}{\rotatebox{90}{Chess Move Validity}} & \multirow{5}{*}{Acc~$\uparrow$}    & $S_1$	& \colorbox{Mycolor3}{47.2±3.6} & \colorbox{Mycolor1}{\textbf{47.6±5.2}} & 45.6±7.8 & 40.0±4.5 & 42.8±2.3 & 29.2±4.6 & 42.4±6.5 & \colorbox{Mycolor2}{20.0±6.0} & 2927 & \textbf{10}
\\
&	  & $S_2$	& \colorbox{Mycolor1}{\textbf{48.4±5.0}} & 45.6±6.1 & 43.6±4.3 & 39.6±3.3 & \colorbox{Mycolor3}{48.4±5.2} & 35.6±5.2 & 43.2±8.8 & \colorbox{Mycolor2}{18.8±5.8} & 2930 & 6                
\\
&	  & $S_3$	& \colorbox{Mycolor1}{\textbf{49.6±5.5}} & \colorbox{Mycolor3}{48.0±5.8} & 47.6±5.5 & 37.6±9.9 & 41.6±6.1 & 35.2±8.3 & 40.4±3.8 & \colorbox{Mycolor2}{14.8±6.1} & 2947 & 6 
\\
&	  & $S_4$	& \colorbox{Mycolor3}{48.4±3.3} & \colorbox{Mycolor1}{\textbf{49.6±4.6}} & 46.0±3.5 & 36.8±4.1 & 38.8±3.3 & 27.2±3.9 & 38.0±6.3 & \colorbox{Mycolor2}{14.0±4.7} & 2959 & 5 
\\ 

\cmidrule{2-13} 

& \underline{Cost}~$\downarrow$  & All & 3736 & 3169 & 3196 & 2627 & 3266 & 2714 & 2698 & 2123 & \multicolumn{2}{c}{\multirow{2}{*}{-}} 
\\ 

\cmidrule{2-11}

& \underline{W-T}~$\uparrow$     & All  & -  & \textbf{11} & 6 & 1 & 5 & 0 & 4 & 0 & \multicolumn{2}{c}{} 
\\ 

\bottomrule

\end{tabular}
}

\caption{
The impact of 8 collaborative strategies on the performance of 3 datasets across distinct societies, using \emph{ChatGPT (with engine of \texttt{gpt-3.5-turbo} employed between July 10 and July 23, 2023)}.
\protect\textcolora{Blue} marks the \protect\textcolora{best-performing} strategy under the same society, 
\protect\textcolorc{light blue} represents the \protect\textcolorc{second-best-performing} strategy, 
and \protect\textcolorb{red} indicates the \protect\textcolorb{worst-performing} strategy.
\textbf{\underline{Cost}} / \textbf{\uuline{Cost}} measures the average tokens consumed by all cases under the same \uline{collaborative strategy} / \uuline{society}. 
\textbf{\underline{W-T}} / \textbf{\uuline{W-T}} tallies the total number of occurrences where performance exceeds the strategy $p_0p_0p_0$ under the same \uline{collaborative strategy} / \uuline{society}. 
The significances test on societies and strategies are respectively shown in Table~\ref{table:gpt-july:sig_main_society},~\ref{table:gpt-july:sig_main_strategy}.
\label{table:gpt-july:main}
}

\end{table*}
\definecolor{gray}{HTML}{CCCCCC}
\begin{table}[!htbp] 
\centering
\resizebox{\linewidth}{!}{
\begin{tabular}{lrrr}
\toprule
Collaborative  & \multicolumn{1}{r}{MMLU} & \multicolumn{1}{r}{MATH} & \multicolumn{1}{r}{Chess Move Validity} \\
Strategy & \multicolumn{1}{r}{p-value} & \multicolumn{1}{r}{p-value} & \multicolumn{1}{r}{p-value} \\ \midrule
$p_0p_0p_0$ & {0.350} & {0.618} & {0.866}  \\
$p_0p_0p_1$ & {0.797} & {0.069} & {0.716}  \\
$p_0p_1p_0$ & {0.162} & {0.631} & {0.726}  \\
$p_0p_1p_1$ & {0.350} & {0.945} & {0.807}  \\
$p_1p_0p_0$ & {0.501} & {0.964} & \colorbox{gray}{0.025}  \\
$p_1p_0p_1$ & {0.497} & {0.378} & {0.079}  \\
$p_1p_1p_0$ & {0.562} & {0.135} & {0.614}  \\
$p_1p_1p_1$ & {0.236} & {0.642} & {0.293}  \\
\bottomrule
\end{tabular}
}
\caption{
One-Way ANOVA results for the impact of society on accuracy with fixed collaborative strategy, based on experiments from Table~\ref{table:gpt-july:main} using \emph{ChatGPT in July}.
}
\label{table:gpt-july:sig_main_society}
\end{table}
\definecolor{gray}{HTML}{CCCCCC}
\begin{table}[!htbp] 
\centering
\resizebox{\linewidth}{!}{
\begin{tabular}{lrrr}
\toprule
 & \multicolumn{1}{r}{MMLU} & \multicolumn{1}{r}{MATH} & \multicolumn{1}{r}{Chess Move Validity} \\
Society & \multicolumn{1}{r}{p-value} & \multicolumn{1}{r}{p-value} & \multicolumn{1}{r}{p-value} \\ \midrule
$S_1$ & \colorbox{gray}{0.000} & {0.346} & \colorbox{gray}{0.000} \\
$S_2$ & \colorbox{gray}{0.000} & \colorbox{gray}{0.008} & \colorbox{gray}{0.000} \\
$S_3$ & \colorbox{gray}{0.000} & {0.388} & \colorbox{gray}{0.000} \\
$S_4$ & \colorbox{gray}{0.000} & {0.213} & \colorbox{gray}{0.000} \\\bottomrule
\end{tabular}
}
\caption{
One-Way ANOVA results for the impact of collaborative strategy on accuracy with fixed society, based on experiments in Table~\ref{table:gpt-july:main} on \emph{ChatGPT in July}.
}
\label{table:gpt-july:sig_main_strategy}
\end{table}

Notably, the $p$-value of the collaborative strategy (on ChatGPT, engine: \texttt{gpt-3.5-turbo-1106}; \texttt{gpt-3.5-turbo} in July) is significantly below the threshold of 0.05, indicating that collaborative strategies have substantial impact on performance. 
Besides, on the backbone LLM of ChatGPT, the $p$-value of the society (with the engine of \texttt{gpt-3.5-turbo-1106}) is smaller than 0.05 in 17 out 24 cases, 
in contrast, the $p$-value of the society (with the engine of \texttt{gpt-3.5-turbo} employed between July 10 and July 23, 2023) is larger than 0.05 in 23 out 24 cases. 
Generally, this corroborates our earlier conclusion in \S\ref{sec:exp-main}, emphasizing that the influence of collaborative strategies outweighs that of societies. 

\begin{figure*}[!t] 
    \centering
    \scalebox{1}{
    \includegraphics[width=1\textwidth]{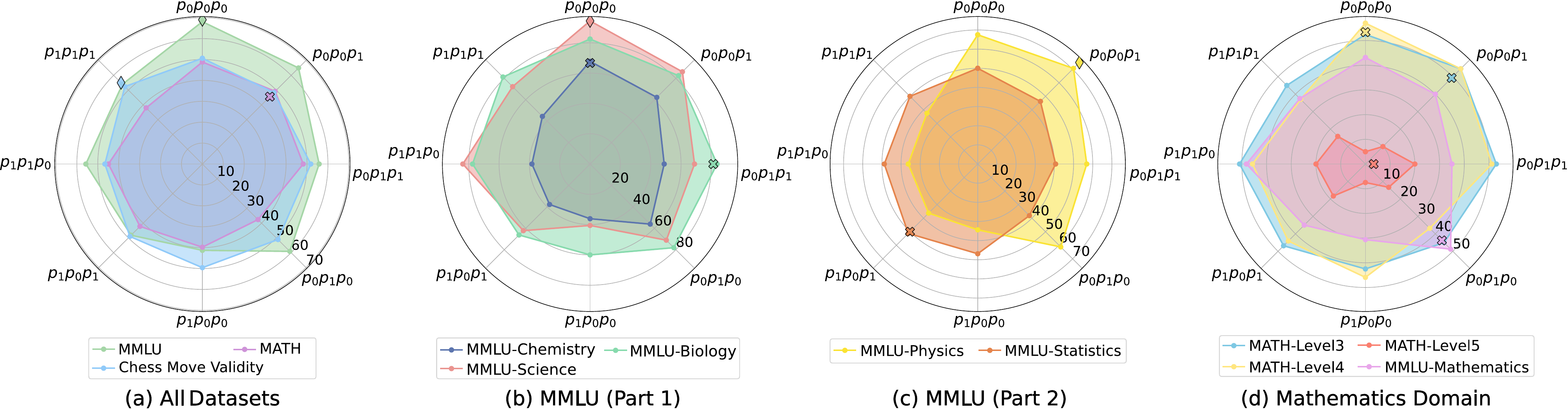}
    }
    \vspace{-3mm}
    \caption{
    Illustration of different collaborative strategies impacting accuracy diversely on the tasks considering varied \emph{subjects} and \emph{difficulty}, using \emph{ChatGPT}. 
    The symbol `\protect\radarfork' represents that there is at least one collaborative strategy whose accuracy is better than self-consistency, while the symbol `\protect\radarprismatic' indicates that there is no collaborative strategy whose accuracy is worse than self-consistency. 
    Both of these symbols represent the accuracy of self-consistency. 
    The accuracy under each collaborative strategy is a summation within all 3-agent societies. 
    \label{fig:task_radar}
    }
\end{figure*}

We also present the \textbf{word clouds} in Figure~\ref{fig:word}, and \textbf{answer changing of agents with different traits} in Figure~\ref{fig:word}, to reveal that indistinctive impact of 3-agent societies on performance. 
Furthermore, we demonstrate that the tasks with different subjects and difficulty display varying sensitivity to collaborative strategies, as presented with \textbf{radar maps} in Figure~\ref{fig:task_radar}. 

\section{Analysis on Machine Society Settings (Backbone: ChatGPT)}
\label{app:society_setting}

In this section, we conduct \textbf{significance tests} for the experiments outlined in \S\ref{sec:impact_of_other_factors}. 
The chosen method is one-way analysis of variance. 
Prior to the analysis, we performed a check for homogeneity of variance, with only one entry in Table~\ref{table:sig_strategy} deviating from the criteria. 
The significance tests for the number of agents, the number of rounds, and different collaborative strategies are respectively detailed in Table~\ref{table:sig_10_agent}, Table~\ref{table:sig_10_turn} and Table~\ref{table:sig_strategy}. 

\definecolor{gray}{HTML}{CCCCCC}
\begin{table}[!htbp] 
\centering
\resizebox{\linewidth}{!}{
\begin{tabular}{lrrrrr}
\toprule
 Collaborative & \multicolumn{1}{r}{$S_1^{'}$} & \multicolumn{1}{r}{$S_2^{'}$} & \multicolumn{1}{r}{$S_3^{'}$} & \multicolumn{1}{r}{$S_4^{'}$} & \multicolumn{1}{r}{$S_5^{'}$} \\
Strategy & \multicolumn{1}{r}{p-value} & \multicolumn{1}{r}{p-value} & \multicolumn{1}{r}{p-value} & \multicolumn{1}{r}{p-value} & \multicolumn{1}{r}{p-value} \\ \midrule
$p_0p_0p_0$ & \colorbox{gray}{0.000} & \colorbox{gray}{0.000} & \colorbox{gray}{0.000} & \colorbox{gray}{0.000} & \colorbox{gray}{0.000} \\
$p_0p_0p_1$ & \colorbox{gray}{0.000} & \colorbox{gray}{0.000} & \colorbox{gray}{0.000} & \colorbox{gray}{0.000} & \colorbox{gray}{0.000} \\
$p_0p_1p_0$ & \colorbox{gray}{0.002} & \colorbox{gray}{0.015} & \colorbox{gray}{0.006} & \colorbox{gray}{0.000} & \colorbox{gray}{0.000} \\
$p_0p_1p_1$ & \colorbox{gray}{0.000} & \colorbox{gray}{0.000} & \colorbox{gray}{0.000} & \colorbox{gray}{0.000} & \colorbox{gray}{0.000} \\
$p_1p_0p_0$ & \colorbox{gray}{0.000} & \colorbox{gray}{0.000} & \colorbox{gray}{0.000} & \colorbox{gray}{0.000} & \colorbox{gray}{0.000} \\
$p_1p_0p_1$ & \colorbox{gray}{0.000} & {-} & \colorbox{gray}{0.000} & \colorbox{gray}{0.001} & \colorbox{gray}{0.000} \\
$p_1p_1p_0$ & \colorbox{gray}{0.000} & \colorbox{gray}{0.000} & \colorbox{gray}{0.000} & \colorbox{gray}{0.000} & \colorbox{gray}{0.000} \\
$p_1p_1p_1$ & \colorbox{gray}{0.000} & \colorbox{gray}{0.000} & \colorbox{gray}{0.000} & \colorbox{gray}{0.005} & \colorbox{gray}{0.000} \\
\bottomrule
\end{tabular}
}
\caption{
One-way ANOVA analysis of results in Figure~\ref{fig:agent_10_on_numbers} (different numbers of agents), using \emph{ChatGPT}. 
$S_1^{'}$: One overconfident agent and the others are all easygoing. $S_2^{'}$: One easygoing agent among predominantly overconfident agents. $S_3^{'}$: Equal numbers of overconfident and easygoing agents. $S_4^{'}$: Entirely easygoing agents. $S_5^{'}$: Entirely overconfident agents. `-': It doesn't pass homogeneity test for variance. 
}
\label{table:sig_10_agent}
\end{table}

\begin{figure*}[!t]
    \centering
    \includegraphics[width=0.88\textwidth]{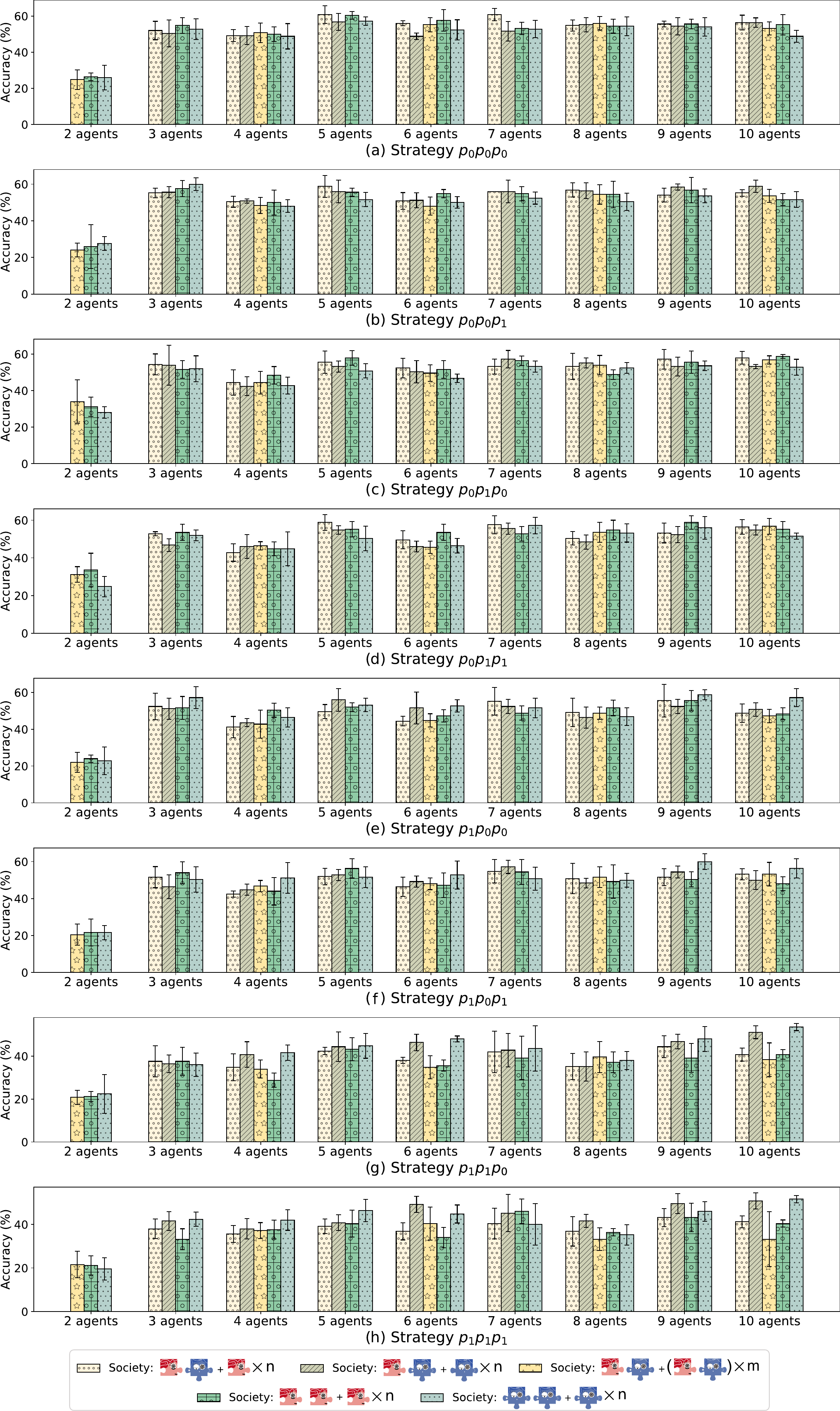}
    \vspace{-3mm}
    \caption{Accuracy of \emph{different societies} with 2$\sim$10 agents under 3-round collaborative strategies, on \emph{ChatGPT}. }
    \label{fig:agent_10_on_societies}
\end{figure*}

\begin{figure*}[!t]
    \centering
    \includegraphics[width=0.85\textwidth]{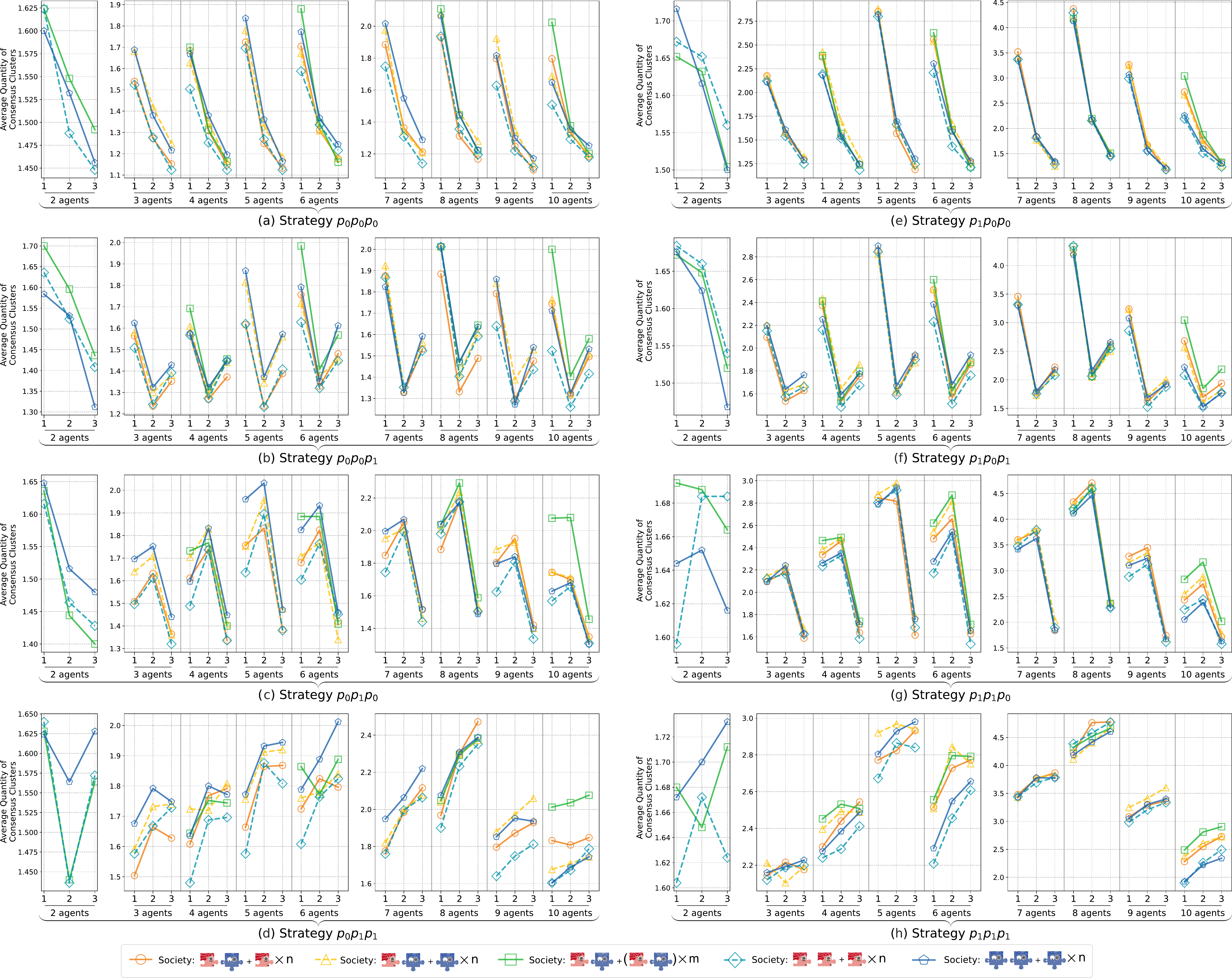}
    \vspace{-3mm}
    \caption{Average quantity of \emph{consensus clusters (unique answers among multiple agents)} in \emph{different societies} with 2$\sim$10 agents under each round of 3-round collaborative strategies, using \emph{ChatGPT}. }
    \label{fig:agent_10_on_societies_consensus}
\end{figure*}

\begin{figure*}[!t]
    \centering
    \includegraphics[width=0.88\textwidth]{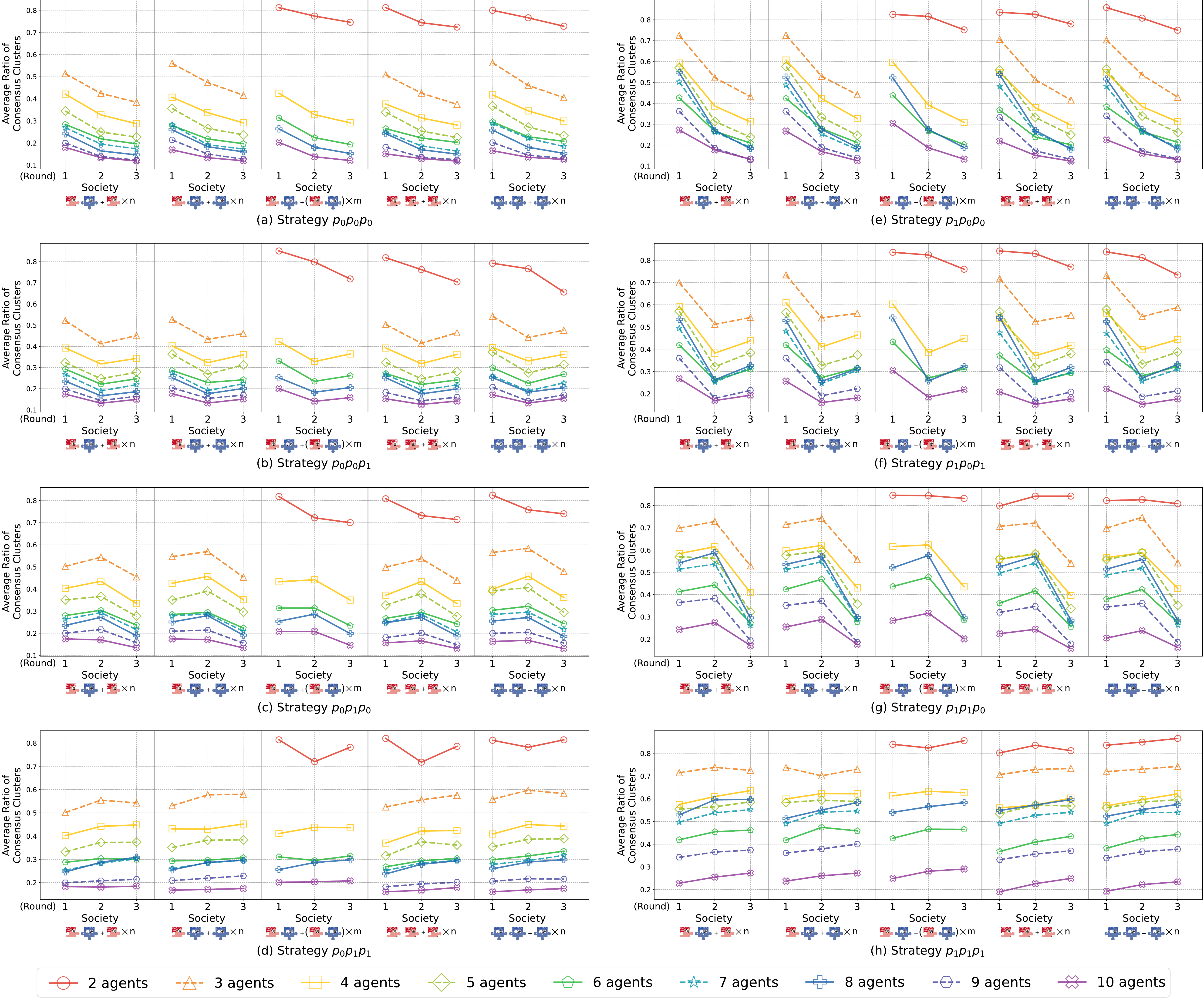}
    \vspace{-3mm}
    \caption{Average ratio of \emph{consensus clusters (unique answers among multiple agents)} with \emph{different numbers (2$\sim$10) of agents} under each round of 3-round collaborative strategies, using \emph{ChatGPT}. }
    \label{fig:agent_10_on_numbers_consensus}
    \vspace{-3mm}
\end{figure*}

\textbf{Different Numbers of Agents.}
According to the results of the $p$-values in Table~\ref{table:sig_10_agent}, the conclusion in \S\ref{sec:impact_of_other_factors} is confirmed, namely, different number of agents results in a significant correlation on performance. 
By integrating the results in Figure~\ref{fig:agent}, it becomes evident that the presence of three agents is relatively optimal. 


We also analyze the \emph{consensus reaching} with different numbers of agents, and present the results in Figure~\ref{fig:agent_10_on_societies_consensus},~\ref{fig:agent_10_on_numbers_consensus}.

\definecolor{gray}{HTML}{CCCCCC}
\begin{table}[!htbp] 
\centering
\small
\resizebox{\linewidth}{!}{
\begin{tabular}{lrrr}
\toprule
Collaborative  & \multicolumn{1}{r}{MMLU} & \multicolumn{1}{r}{MATH} & \multicolumn{1}{r}{Chess Move Validity} \\
Strategy & \multicolumn{1}{r}{p-value} & \multicolumn{1}{r}{p-value} & \multicolumn{1}{r}{p-value} \\ \midrule
$p_0p_0p_0p_0p_0p_0p_0p_0p_0p_0$ & \colorbox{gray}{0.030} & {0.323} & \colorbox{gray}{0.000}  \\
$p_1p_0p_0p_0p_0p_0p_0p_0p_0p_0$ & \colorbox{gray}{0.000} & {0.070} & {0.161}  \\
$p_0p_1p_0p_0p_0p_0p_0p_0p_0p_0$ & {0.101} & {0.332} & \colorbox{gray}{0.000}  \\
$p_1p_0p_1p_0p_1p_0p_1p_0p_1p_0$ & \colorbox{gray}{0.000} & {0.077} & {0.871}  \\
$p_0p_1p_0p_1p_0p_1p_0p_1p_0p_1$ & {0.051} & {0.062} & \colorbox{gray}{0.000}  \\
$p_1p_0p_1p_1p_1p_1p_1p_1p_1p_1$ & \colorbox{gray}{0.000} & \colorbox{gray}{0.021} & {0.630}  \\
$p_0p_1p_1p_1p_1p_1p_1p_1p_1p_1$ & {0.431} & {0.176} & {0.063}  \\
$p_1p_1p_1p_1p_1p_1p_1p_1p_1p_1$ & \colorbox{gray}{0.000} & \colorbox{gray}{0.000} & \colorbox{gray}{0.027}  \\ 
\bottomrule
\end{tabular}
}
\caption{
One-way ANOVA analysis of the results in Figure~\ref{fig:round_10_on_math},~\ref{fig:round_10_on_mmlu},~\ref{fig:round_10_on_chess} (different rounds), using \emph{ChatGPT}. 
}
\label{table:sig_10_turn}
\end{table}

\begin{figure*}[!t]
    \centering
    \includegraphics[width=1\textwidth]{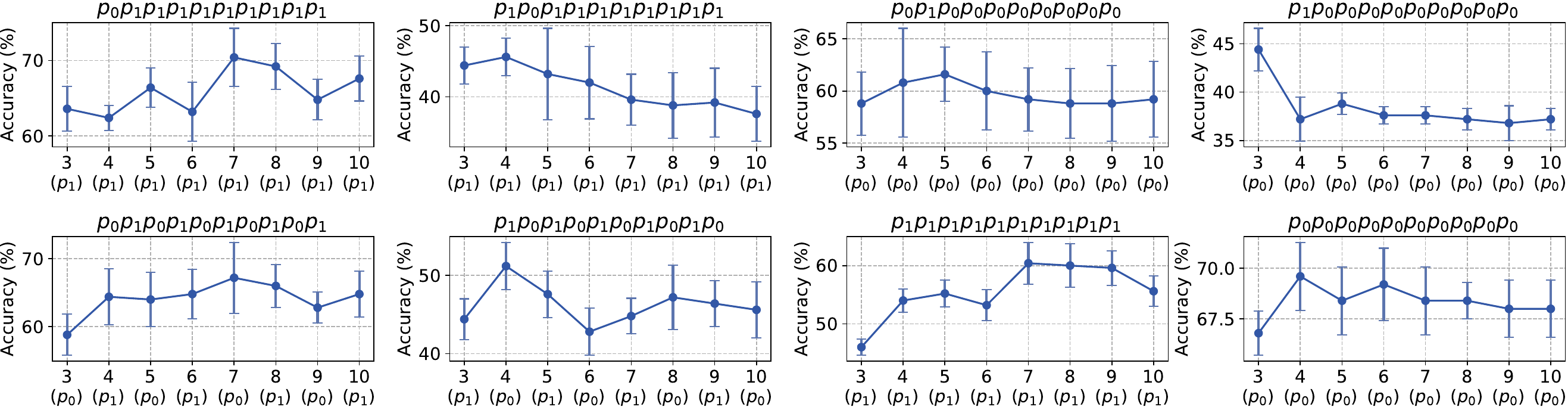}
    \vspace{-5mm}
    \caption{Accuracy of \emph{different (3$\sim$10) rounds of collaboration} within 3-agent society $S_2$ (1 easy-going and 2 overconfident agents) on MMLU, using \emph{ChatGPT}. }
    \label{fig:round_10_on_mmlu}
\end{figure*}


\begin{figure*}[!t]
    \centering
    \includegraphics[width=1\textwidth]{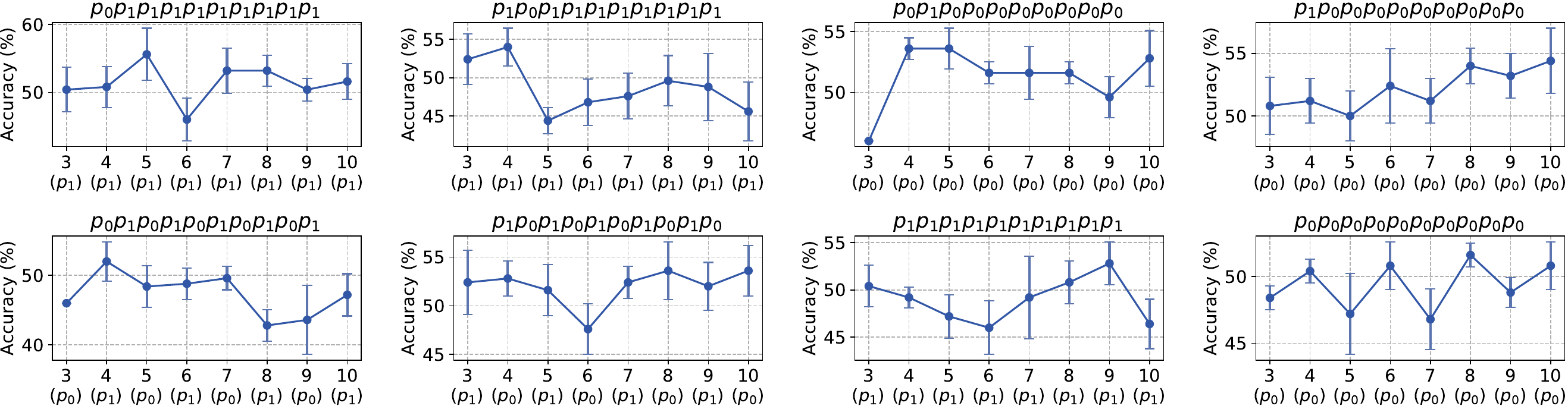}
    \vspace{-5mm}
    \caption{Accuracy of \emph{different (3$\sim$10) rounds of collaboration} within 3-agent society $S_2$ (1 easy-going and 2 overconfident agents) on Chess Move Validity, using \emph{ChatGPT}. }
    \label{fig:round_10_on_chess}
\end{figure*}

\textbf{Different Rounds of Collaboration.} As observed from Table~\ref{table:sig_10_turn}, 
we find that the impact of rounds significantly relies on the employed collaborative strategy. For MMLU and Chess Move Validity, collaborative strategies where $p$-values $<0.05$ are $\{p_0p_1p_1p_0, p_0p_1p_1p_1, p_1p_0p_1p_0, p_1p_0p_1p_1\}$ and $\{p_0p_1p_1p_0, p_0p_1p_1p_1, p_1p_0p_1p_1, p_1p_1p_0p_0, p_1p_1p_0 \\p_1, p_1p_1p_1p_0\}$. 
We also increase the rounds of collaboration, from 3 to 10, and present the results in Figure~\ref{fig:round_10_on_mmlu},~\ref{fig:round_10_on_chess}. 
We find that although there would be some fluctuations in performance if we scale up the round of collaboration, the outperformance is not obvious enough. While increasing rounds of collaboration will result in more consumption of tokens, which is not economic. Thus we infer that the 3-round collaboration is relatively optimal considering both performance and cost. 

Furthermore, as seen from Figure~\ref{fig:consistent}, the strategy after a round of debate tends to yield fewer consensus clusters compared to the preceding rounds. 
Conversely, the strategy subsequent with a round of reflection at the same juncture will increase consensus clusters. 
Adding an extra round of debate at this juncture, as the conclusions in \S\ref{sec:conformity_consistency}, is not anticipated to bring about a discernible enhancement in performance. 
This confirms the efficacy of the \emph{early-stopping mechanism} implemented in \citet{arXiv2023_Dynamic-LLM-Agent}, drawing inspiration from Byzantine Consensus theory~\citep{USENIX1999_Theory-FaultTolerance}.

Moreover, we scrutinize the consensus reaching of these strategies in three rounds where $p$-values are below 0.05, as shown in Figure~\ref{fig:consistent}. 
Also seen from Figure~\ref{fig:consistent} and Figure~\ref{fig:round_10_on_mmlu},~\ref{fig:round_10_on_math},~\ref{fig:round_10_on_chess}, it becomes apparent that these collaborative strategies exhibit substantial fluctuations in consensus reaching, demonstrating notably low answer consistency. 
For $p_0p_0p_0p_0$ on Chess Move Validity, although continuous reflection results in a gradual increase in the number of consensus clusters, a more stable trend with smaller fluctuations renders it less sensitive to the rounds of collaboration. 
Conversely, collaborative strategies where $p$-values$>0.05$ often display higher levels of answer consistency. 

\begin{figure}[!htbp]
    \centering
    \includegraphics[width=0.48\textwidth]{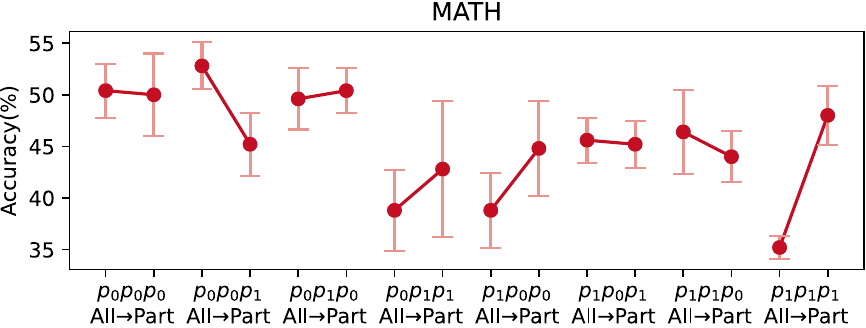}
    \vspace{-6mm}
    \caption{
    The effect on the accuracy of whether all agents in a society execute the same thinking pattern in one round on MATH, using \emph{ChatGPT}.
    ``All'' and ``Part'' respectively refer to all agents applying the same and different thinking pattern(s) in one round.  
    The significance test is shown in Table~\ref{table:sig_strategy} at Appendix~\ref{app:society_setting}. 
    }
    \label{fig:strategy_math}
\end{figure}

\begin{table}[!htbp] 
\centering
\resizebox{\linewidth}{!}{
\begin{tabular}{lrrr}
\toprule
Collaborative  & \multicolumn{1}{r}{MMLU} & \multicolumn{1}{r}{MATH} & \multicolumn{1}{r}{Chess Move Validity} \\
Strategy & \multicolumn{1}{r}{p-value} & \multicolumn{1}{r}{p-value} & \multicolumn{1}{r}{p-value} \\ \midrule
$p_0p_0p_0$ & {0.402} & {0.856} & {0.147}  \\
$p_0p_0p_1$ & \colorbox{gray}{0.007} & \colorbox{gray}{0.002} & \colorbox{gray}{0.001}  \\
$p_0p_1p_0$ & {0.550} & {0.641} & \colorbox{gray}{0.002}  \\
$p_0p_1p_1$ & {-} & {0.276} & \colorbox{gray}{0.000}  \\
$p_1p_0p_0$ & {-} & {0.051} & {-}  \\
$p_1p_0p_1$ & {-} & {0.784} & \colorbox{gray}{0.000}  \\
$p_1p_1p_0$ & \colorbox{gray}{0.014} & {0.294} & {0.172}  \\
$p_1p_1p_1$ & {1.000} & \colorbox{gray}{0.000} & {0.347}  \\
\bottomrule
\end{tabular}
}
\caption{One-way ANOVA analysis of the results of Figure~\ref{fig:strategy} (other collaborative strategies), using \emph{ChatGPT}. 
`-': It doesn't pass homogeneity test for variance.
}
\label{table:sig_strategy}
\vspace{-4mm}
\end{table}

\textbf{Other Collaborative Strategies.} 
We show the results of all agents in a society executing the same or inconsistent thinking pattern(s) at one round in Figure~\ref{fig:strategy_math}. 
Seen from Table~\ref{table:sig_strategy}, we observe pronounced impacts of keeping a consistent thinking pattern on Chess Move Validity, while its influence on MMLU and MATH is less significant. 


\begin{figure*}[!t] 
    \centering
    \includegraphics[width=0.88\textwidth]{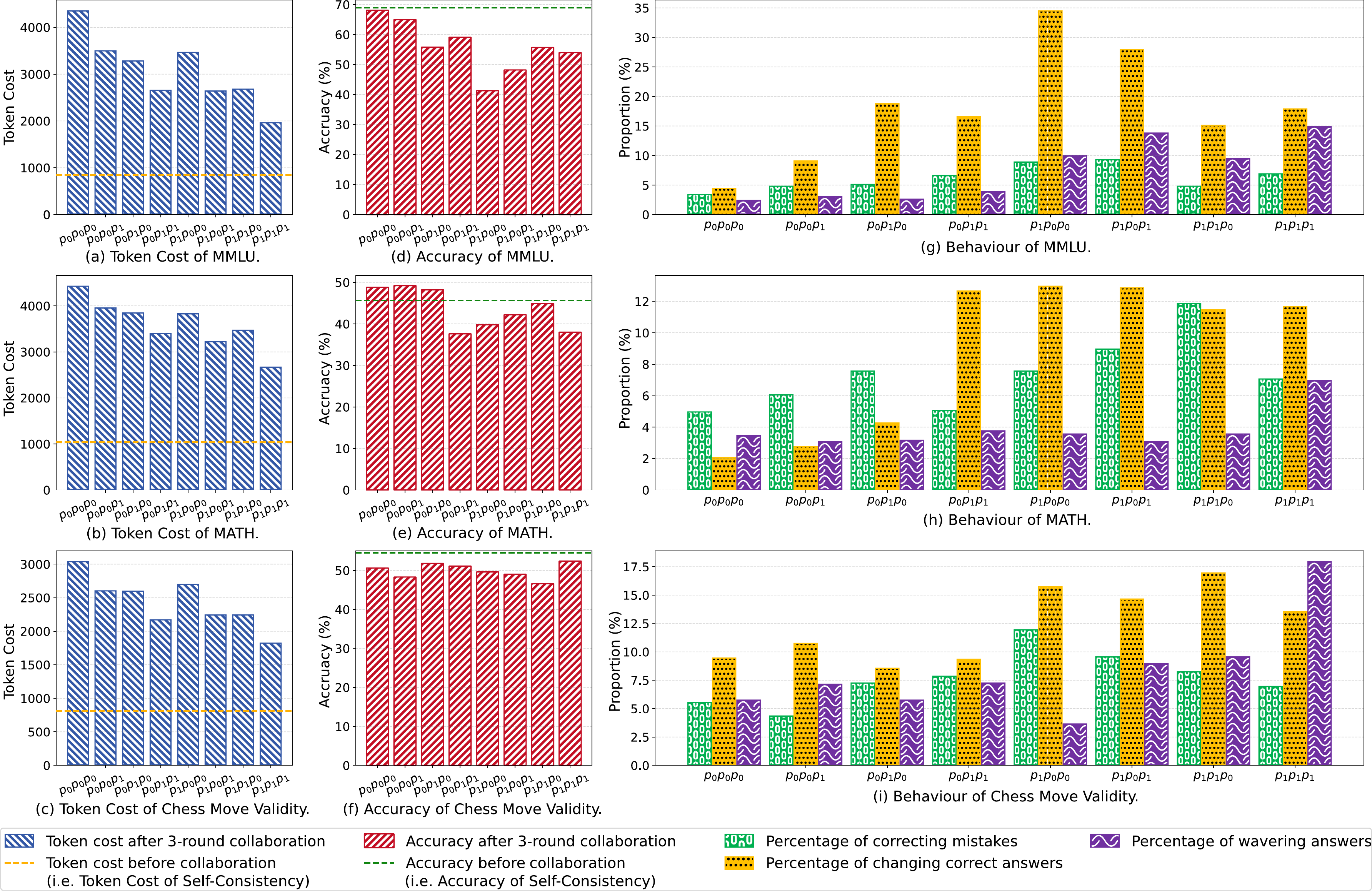}
    \vspace{-3mm}
    \caption{
    The percentage of different behaviors under different collaborative strategies, using \emph{ChatGPT}. 
    Figure (a-c) \& (d-f) respectively show the token cost and accuracy of different strategies before and after 3-round collaboration. 
    Figure (g-i) present the percentage of different behavioral features (mainly analyzed by the change of answer correctness) \citep{arXiv2023_Agent-BehaviorExplanation,arXiv2023_Agent-BehaviorExplaining} under different collaborative strategies. 
    All results are summarized across all societies. 
    The results on other LLMs are shown in Figure~\ref{fig:llama:distribute},~\ref{fig:llama70:distribute},~\ref{fig:qwen:distribute},~\ref{fig:mixtral:distribute} at Appendix~\ref{app:backbone_llm}. 
    }
    \label{fig:distribute}
    \vspace{-4mm}
\end{figure*}

\section{A Social Psychology View on Conformity, Consensus Reaching, and Group Dynamics} 
\label{app:social_psychology_view}

\subsection{Conformity and Consensus Reaching}
\label{app:detail_conformity_and_consensus}
Figures~\ref{fig:conformity}, \ref{fig:llama:conformity}, \ref{fig:llama70:conformity}, \ref{fig:mixtral:conformity}, and \ref{fig:qwen:conformity} illustrate the conformity. Figures~\ref{fig:consistent}, \ref{fig:llama:conformity}, \ref{fig:llama70:conformity}, \ref{fig:mixtral:conformity}, and \ref{fig:qwen:conformity} illustrate the consensus. This section provides a detailed explanation of the methodologies used to calculate both conformity and consensus.

For conformity, we solely focus on agents actively engaging in debate, disregarding those in reflection during a given round. 
Let the answer of the $i$-th agent at $j$-th round be denoted as $a_{i,j}$. 
For the $k$-th agent at $j$-th round, if ``$\text{Frequency}\big( \{a_{i,j-1} | i \in [1,n]\}\big)=a_{k,j}$'', we identify this as the occurrence of conformity by agent $k$ at $j$-th round, where $\text{Frequency}(\cdot)$ represents the most frequently given answer (excluding instances where all answers occur only once, as such cases are considered as non-conformity). 
Additionally, we categorize the correctness of answers both before and after conformity into four cases, with `True' denoting correct and `False' denoting incorrect.

For consensus, we examine the evolution of the number of distinct answers (\ie, consensus clusters) with increasing rounds of collaboration. Let the answer of the $i$-th agent at time $j$ be denoted as $a_{i,j}$. 
For the $j$-th round, consensus clusters is defined as $\left \|\text{Set}(\{a_{i,j}|i\in[1,n]\})\right \| $, where $\left \|\text{Set}(\cdot)\right \|$ represents the count of different answers. 
This computational approach has been utilized in the analysis presented in Figures~\ref{fig:agent_10_on_numbers_consensus},~\ref{fig:agent_10_on_societies_consensus},~\ref{fig:mixtral:agent_10_on_numbers_consensus},~\ref{fig:mixtral:agent_10_on_societies_consensus},~\ref{fig:qwen:agent_10_on_numbers_consensus},~\ref{fig:qwen:agent_10_on_societies_consensus}. 

\subsection{Group Dynamics} 
\label{app:principles_group_dynamics}

We seek to elucidate how performance is impacted by group dynamics, \ie, the patterns of interaction between group members and different processes that may occur within a social group. 
Diving into the intricacies of collaboration, each agent generates four answers, including the initial answer without collaboration, as shown in Figure~\ref{fig:setting}(d). To determine the answer for each round, we employ the majority vote \citep{arXiv2021_Verifier-Math,Science2022_AlphaCode}. 
Given `\textit{T}' and `\textit{F}' respectively denoting a round that yields a correct and an incorrect answer, we could obtain 2$^4$=16 possible answer sequences over the four rounds. 
We select 10 sequences\footnote{The selected 10 sequences adhere to patterns: (1) $[F]_{i>0}[T]_{j>0}$, \eg, $FFFT$; (2) $[T]_{i>0}[F]_{j>0}$, \eg, $TFFF$; (3)  $[TF]_{i\geq0}[FT]_{j\geq0}$, \eg, $FTFT$, where $[\cdot]_i$, $[\cdot]_j$ respectively denotes repetition for $i$, $j$ times.} of them and categorize them into 3 groups: \textit{\textbf{Correcting Mistakes}} ($FFFT, FFTT, FTTT$), \textit{\textbf{Changing Correct Answers}} ($TFFF, TTFF, TTTF$), and \textit{\textbf{Wavering Answers}} ($FTFT, FTTF, TFTF, TFFT$). 
Particularly, \textit{\textbf{Wavering Answers}} resemble model hallucination \citep{arXiv2023_Survey-Hallucination_LFM,arXiv2023_Survey-Hallucination_LLM,J2023_Survey-Hallucination_NLG,arXiv2024_Survey-Hallucination} due to the occurrence of self-contradictory answers. 
Our categorization is under society-agnostic collaborative strategies, considering the performance variance between societies is negligible. From the results on ChatGPT shown in Figure~\ref{fig:distribute}, and on other LLMs shown in Appendix~\ref{app:backbone_llm}, we summarize the following findings: 

\textbf{
(1) Debate-initial/dominant collaborative strategies are generally effective.} 
As seen from the red bars in Figure~\ref{fig:distribute}~\ref{fig:llama:distribute},~\ref{fig:llama70:distribute},~\ref{fig:qwen:distribute},~\ref{fig:mixtral:distribute}(d-f), we find that the collaborative strategies starting from or dominant with debate $p_0$ are more effective than other, and mostly outperform self-consistency, even though they cost more tokens (seen from blue bars). 

\textbf{
(2) Reflection experiences greater instability (a heightened risk of model hallucination).
} 
As observed from the purple bars in Figure~\ref{fig:distribute}~\ref{fig:llama:distribute},~\ref{fig:llama70:distribute},~\ref{fig:qwen:distribute},~\ref{fig:mixtral:distribute}(g-h), comparing $p_ip_jp_0$ \& $p_ip_jp_1$; $p_ip_0p_j$ \& $p_ip_1p_j$, $p_ip_jp_0$ and $p_ip_0p_j$ are more likely to wavering answers than $p_ip_jp_1$ and $p_ip_1p_j$, demonstrating that reflection is more likely to cause model hallucination than debate. 

\clearpage

\definecolor{Mycolor1}{HTML}{BAD8F2}
\definecolor{Mycolor2}{HTML}{FAE4E3}
\definecolor{Mycolor3}{HTML}{E8F2FB}





\begin{table*}[!t] 

\small
\resizebox{\linewidth}{!}{
\begin{tabular}{c|c|c|cccccccc|cc}

\toprule

& \multirow{2}{*}{\tabincell{c}{Metric \\ (Strategy)}} & \multirow{2}{*}{Society} & \multicolumn{8}{c|}{Collaborative Strategy} & \multicolumn{2}{c}{Metric (Society)}
\\
& & & $p_0p_0p_0$ & $p_0p_0p_1$ & $p_0p_1p_0$ & $p_0p_1p_1$ & $p_1p_0p_0$ & $p_1p_0p_1$ & $p_1p_1p_0$ & $p_1p_1p_1$ & \uuline{Cost}~$\downarrow$ & \uuline{W-T}~$\uparrow$
\\

\midrule

\multirow{7}{*}{\rotatebox{90}{MMLU}}  & \multirow{5}{*}{Acc~$\uparrow$} 
& $S_1$	& \colorbox{Mycolor2}{37.2±5.9} & 47.2±3.9 & \colorbox{Mycolor1}{\textbf{48.4±3.9}} & 46.0±5.7 & \colorbox{Mycolor3}{\textbf{47.2±2.3}} & 46.8±2.7 & 45.2±4.4 & 46.8±3.0 & 7447 & 35
\\
& & $S_2$	& \colorbox{Mycolor2}{38.4±4.6} & 42.8±3.9 & 43.6±3.6 & \colorbox{Mycolor3}{\textbf{45.2±3.6}} & 44.8±4.6 & \colorbox{Mycolor1}{\textbf{47.2±3.9}} & 44.4±6.2 & 42.8±3.4 & 7413 & 33
\\
& & $S_3$	& \colorbox{Mycolor2}{36.0±3.7} & 44.8±3.0 & 44.8±4.8 & \colorbox{Mycolor1}{\textbf{46.4±1.7}} & 41.6±4.3 & \colorbox{Mycolor3}{\textbf{46.4±2.2}} & 43.2±6.6 & 42.4±3.3 & 7370 & 33
\\
& & $S_4$	& \colorbox{Mycolor2}{34.8±2.7} & 42.4±5.0 & 42.0±4.5 & \colorbox{Mycolor1}{\textbf{44.0±2.8}} & 40.4±3.0 & \colorbox{Mycolor3}{\textbf{43.6±3.9}} & 40.8±3.0 & 41.6±2.6 & 7423 & 35
\\ 

\cmidrule{2-13} 

& \underline{Cost}~$\downarrow$  & All   & 11429 & 9476 & 8166 & 6419 & 8452 & 5734 & 5733 & 3900     & \multicolumn{2}{c}{\multirow{2}{*}{-}} \\ \cmidrule{2-11}
& \underline{W-T}~$\uparrow$     & All   & -     & \textbf{20} & \textbf{20} & \textbf{20} & 18 & \textbf{20} & 19 & 19 & \multicolumn{2}{c}{}
\\ 

\midrule

\multirow{7}{*}{\rotatebox{90}{MATH}}  & \multirow{5}{*}{Acc~$\uparrow$}    & $S_1$	& 5.2±2.3 & \colorbox{Mycolor1}{\textbf{6.8±2.3}} & \colorbox{Mycolor3}{\textbf{5.6±2.6}} & \colorbox{Mycolor3}{\textbf{5.6±2.6}} & 4.8±3.0 & 4.4±1.7 & 5.6±3.9 & \colorbox{Mycolor2}{3.2±1.1} & 8639 & 24
\\
&	  & $S_2$	&  5.2±3.6 & 5.2±3.4 & 6.0±2.0 & \colorbox{Mycolor3}{\textbf{6.8±1.8}} & 6.0±0.0 & \colorbox{Mycolor3}{\textbf{6.8±1.8}} & \colorbox{Mycolor1}{\textbf{6.8±1.1}} & \colorbox{Mycolor2}{4.8±1.1} & 8451 & 22
\\
&	  & $S_3$	& \colorbox{Mycolor1}{\textbf{6.8±1.8}} & \colorbox{Mycolor3}{\textbf{6.8±3.0}} & 6.8±3.4 & 6.0±2.8 & 5.2±1.8 & 5.2±1.8 & 6.0±3.7 & \colorbox{Mycolor2}{3.6±1.7} & 8501 & 16
\\
&	  & $S_4$	& 4.8±2.3 & \colorbox{Mycolor3}{\textbf{6.8±3.4}} & \colorbox{Mycolor1}{\textbf{7.2±1.1}} & 5.6±2.2 & 5.6±1.7 & 5.2±2.3 & 5.2±3.6 & \colorbox{Mycolor2}{4.0±1.4} & 8475 & 28
\\ 

\cmidrule{2-13} 

& \underline{Cost}~$\downarrow$  & All & 10655 & 9508 & 9501 & 7900 & 9319 & 7761 & 7800 & 5687     & \multicolumn{2}{c}{\multirow{2}{*}{-}} 
\\ 

\cmidrule{2-11}

& \underline{W-T}~$\uparrow$     & All & - & 15 & \textbf{16} & 13 & 13 & 11 & 13 & 9 & \multicolumn{2}{c}{}
\\ 

\midrule

\multirow{6}{*}{\rotatebox{90}{Chess Move Validity}} & \multirow{5}{*}{Acc~$\uparrow$}    & $S_1$	&  \colorbox{Mycolor1}{\textbf{16.4±3.0}} & 7.2±3.0 & \colorbox{Mycolor3}{\textbf{9.2±2.3}} & 2.8±1.8 & 8.8±3.0 & 4.8±2.3 & 9.2±4.4 & \colorbox{Mycolor2}{2.0±2.8} & 3754 & 2
\\
&	  & $S_2$	& \colorbox{Mycolor3}{\textbf{11.6±5.2}} & 8.0±1.4 & 10.8±4.2 & \colorbox{Mycolor2}{2.8±1.8} & \colorbox{Mycolor1}{\textbf{11.6±2.6}} & 6.0±3.2 & 10.8±5.0 & 3.6±2.6 & 3725 & 10             
\\
&	  & $S_3$	& \colorbox{Mycolor1}{\textbf{14.8±3.0}} & 8.4±4.8 & 10.0±4.2 & 5.2±1.1 & \colorbox{Mycolor3}{\textbf{14.0±4.5}} & 6.8±3.0 & 9.6±6.2 & \colorbox{Mycolor2}{2.8±3.0} & 3678 & 5
\\
&	  & $S_4$	& \colorbox{Mycolor1}{\textbf{16.0±4.2}} & 6.8±2.7 & \colorbox{Mycolor3}{\textbf{12.4±6.2}} & \colorbox{Mycolor2}{4.0±2.5} & 10.0±4.2 & 7.2±6.7 & 10.0±3.2 & \colorbox{Mycolor2}{4.0±2.5} & 3647 & 4
\\ 

\cmidrule{2-13} 

& \underline{Cost}~$\downarrow$  & All & 4889 & 4123 & 4061 & 3324 & 4045 & 3293 & 3292 & 2581     & \multicolumn{2}{c}{\multirow{2}{*}{-}} 
\\ 

\cmidrule{2-11}

& \underline{W-T}~$\uparrow$     & All  & - & 2 & 4 & 0 & \textbf{7} & 1 & \textbf{7} & 0 & \multicolumn{2}{c}{} 
\\ 

\bottomrule

\end{tabular}
}

\caption{
The impact of eight different collaborative strategies on the performance of three datasets across distinct societies (\emph{using LlaMA2-chat-13B}). 
The significances test on societies and strategies are respectively shown in Table~\ref{table:llama:sig_main_society},~\ref{table:llama:sig_main_strategy}. 
The experiments of comparison with the single LLM agent is shown in Figure~\ref{fig:llama:distribute}(a)-(f). 
\label{table:llama_main}
}

\end{table*}
\definecolor{gray}{HTML}{CCCCCC}
\begin{table}[!htbp] 
\centering
\resizebox{\linewidth}{!}{
\begin{tabular}{lrrr}
\toprule
Collaborative  & \multicolumn{1}{r}{MMLU} & \multicolumn{1}{r}{MATH} & \multicolumn{1}{r}{Chess Move Validity} \\
Strategy & \multicolumn{1}{r}{p-value} & \multicolumn{1}{r}{p-value} & \multicolumn{1}{r}{p-value} \\ \midrule
$p_0p_0p_0$ & {0.611} & {0.632} & {0.251}  \\
$p_0p_0p_1$ & {0.252} & {0.791} & {0.854}  \\
$p_0p_1p_0$ & {0.142} & {0.714} & {0.706}  \\
$p_0p_1p_1$ & {0.755} & {0.839} & {0.164}  \\
$p_1p_0p_0$ & \colorbox{gray}{0.039} & {0.789} & {0.175}  \\
$p_1p_0p_1$ & {0.318} & {0.277} & {0.809}  \\
$p_1p_1p_0$ & {0.585} & {0.884} & {0.959}  \\
$p_1p_1p_1$ & {0.071} & {0.310} & {0.672}  \\
\bottomrule
\end{tabular}
}
\caption{
One-Way ANOVA results for the impact of society on accuracy with fixed collaborative strategy, based on experiments from Table~\ref{table:llama_main} using \emph{LlaMA2-chat-13B}.
}
\label{table:llama:sig_main_society}
\end{table}
\definecolor{gray}{HTML}{CCCCCC}
\begin{table}[!htbp] 
\centering
\resizebox{\linewidth}{!}{
\begin{tabular}{lrrr}
\toprule
 & \multicolumn{1}{r}{MMLU} & \multicolumn{1}{r}{MATH} & \multicolumn{1}{r}{Chess Move Validity} \\
Society & \multicolumn{1}{r}{p-value} & \multicolumn{1}{r}{p-value} & \multicolumn{1}{r}{p-value} \\ \midrule
$S_1$ & \colorbox{gray}{0.006} & {0.548} & \colorbox{gray}{0.000} \\
$S_2$ & {0.129} & {0.664} & \colorbox{gray}{0.000} \\
$S_3$ & \colorbox{gray}{0.005} & {0.518} & \colorbox{gray}{0.000} \\
$S_4$ & \colorbox{gray}{0.009} & {0.490} & \colorbox{gray}{0.001} \\
\bottomrule
\end{tabular}
}
\caption{
One-Way ANOVA results for the impact of collaborative strategy on accuracy with fixed society, based on experiments from Table~\ref{table:llama_main} using \emph{LlaMA-13B-Chat}.
}
\label{table:llama:sig_main_strategy}
\end{table}

\section{Analysis on Different Backbone LLMs}
\label{app:backbone_llm}

To make the findings in this paper more general, we also implement all the experiments with some other open-resource backbone LLMs, such as \textbf{LlaMA2 Chat 13B} \citep{arXiv2023_LLaMA}, \textbf{LlaMA2 Chat 70B } \citep{arXiv2023_LLaMA}, \textbf{Qwen 72B} \citep{arXiv2023_Qwen} and \textbf{Mixtral 8$\times$7B} \citep{arXiv2023_Mistral,arXiv2024_Mixtral}.

\subsection{LlaMA2 Chat 13B}
\label{app:backbone_llama13b}

\begin{figure*}[!t] 
    \centering
    \scalebox{1}{
    \includegraphics[width=1\textwidth]{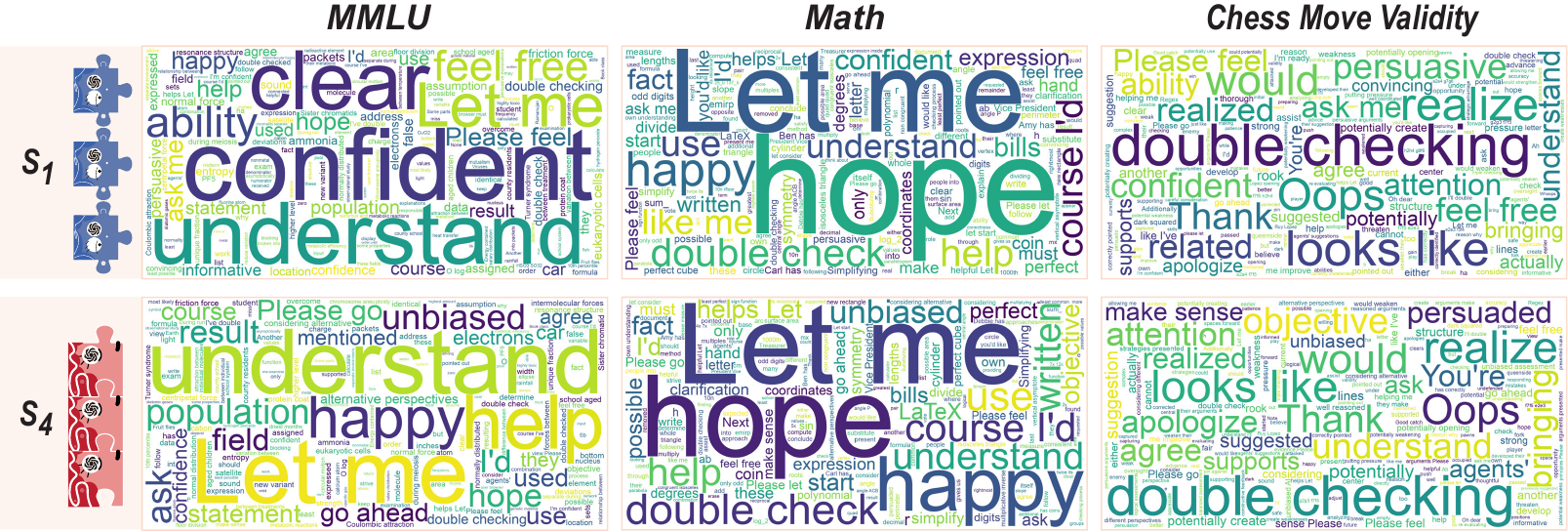}
    }
    \vspace{-4mm}
    \caption{
    Comparative word clouds on three datasets in societies $S_1$ and $S_4$, using \emph{LlaMA2-13B-chat}. 
    Society $S_1$ features three overconfident agents, while society $S_4$ comprises three easy-going agents. 
    }
    \label{fig:llama:word}
\end{figure*}

\begin{figure*}[!t] 
    \centering
    \scalebox{1}{
    \includegraphics[width=1\textwidth]{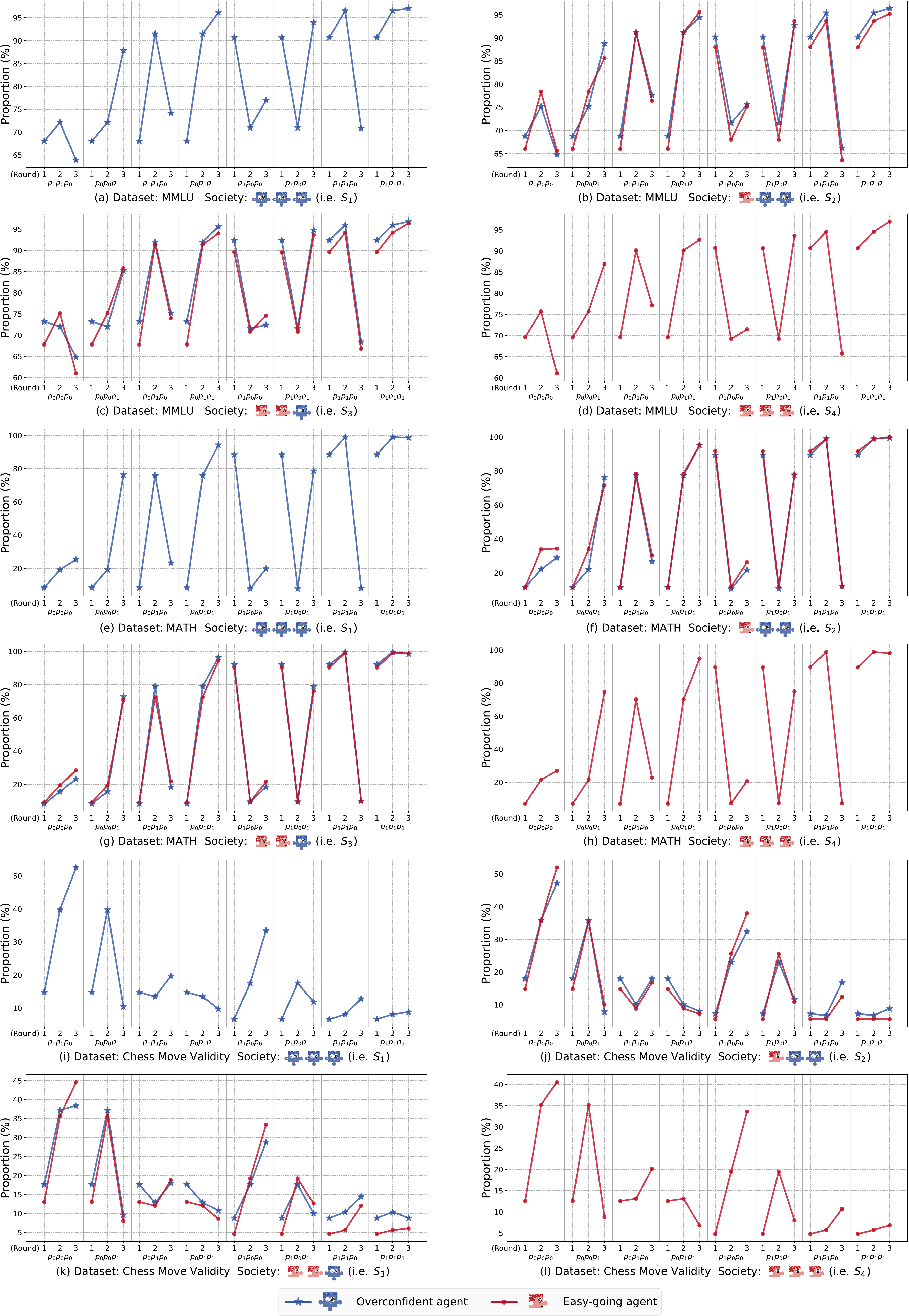}
    }
    \vspace{-4mm}
    \caption{
    Proportion of agents with different traits keeping answers in societies $S_1$ and $S_4$, using \emph{LlaMA2-13B-chat}. 
    Society $S_1$ features three overconfident agents, while society $S_4$ comprises three easy-going agents. 
    }
    \label{fig:llama:agent_answer_changing}
\end{figure*}

\begin{figure*}[!t] 
    \centering
    \scalebox{1}{
    \includegraphics[width=1\textwidth]{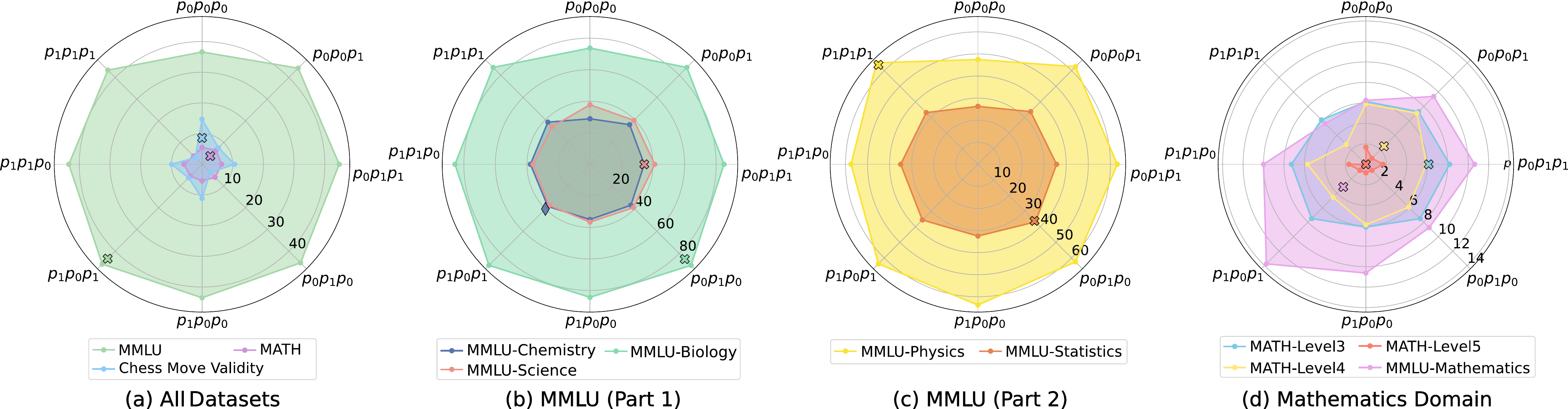}
    }
    \vspace{-6mm}
    \caption{
    Illustration of different collaborative strategies impacting accuracy diversely on the tasks considering varied \emph{subjects} and \emph{difficulty}, using \emph{LlaMA2-13B-chat}. 
    The symbol `\protect\radarfork' represents that there is at least one collaborative strategy whose accuracy is better than self-consistency, while the symbol `\protect\radarprismatic' indicates that there is no collaborative strategy whose accuracy is worse than self-consistency. 
    Both of these symbols represent the accuracy of self-consistency. 
    The accuracy under each collaborative strategy is a summation within all 3-agent societies. 
    \label{fig:llama:task_radar}
    }
\end{figure*}

\textbf{Analysis on Machine Social Collaboration.} 
We present the \textbf{main results} and \textbf{significance tests} of societies and strategies on LlaMA2 Chat 13B in Table~\ref{table:llama_main},~\ref{table:llama:sig_main_society},~\ref{table:llama:sig_main_strategy}. 
We present the \textbf{word clouds} of LlaMA2 Chat 13B in Figure~\ref{fig:llama:word}, and \textbf{proportion of agents with different traits keeping answers in different societies} on LlaMA2 Chat 13B in Figure~\ref{fig:llama:agent_answer_changing}. 
Furthermore, we demonstrate that the tasks with different subjects and difficulty display varying sensitivity to collaborative strategies, as presented with \textbf{radar maps} on LlaMA2 Chat 13B in Figure~\ref{fig:llama:task_radar}. 


\begin{figure*}[!t] 
    \centering
    \includegraphics[width=0.92\textwidth]{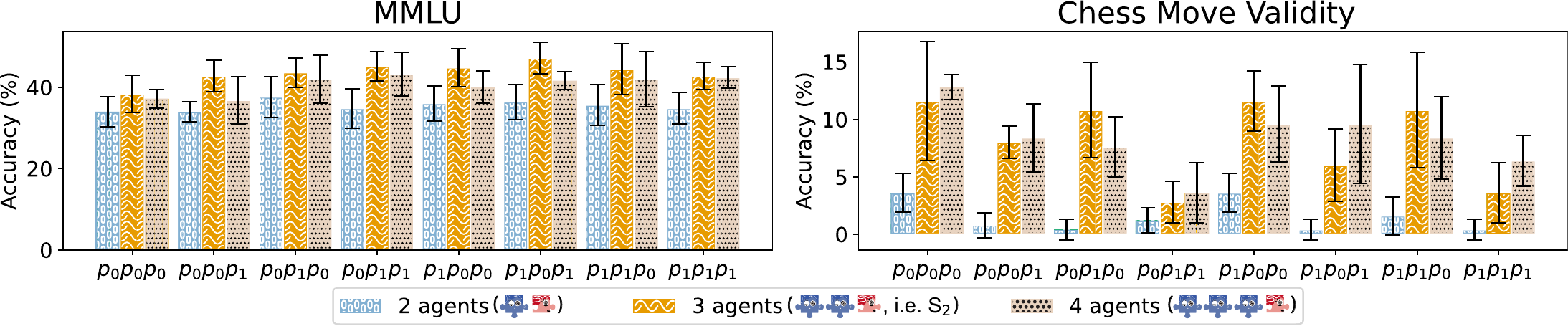}
    \vspace{-3mm}
    \caption{
    Accuracy of different number of agents under different collaborative strategies, on \emph{LlaMA2-13B-chat}. 
    The significance test is shown in Table~\ref{table:sig_llama_agent}. 
    }
    \label{fig:llama:agent}
\end{figure*}

\textbf{Analysis on Different Numbers of Agents.} 
We present the significance test for different numbers of agents with LlaMA2 Chat 13B in Table~\ref{table:sig_llama_agent}. 
We also show the performance varying from agent numbers in Figure~\ref{fig:llama:agent}. 

\definecolor{gray}{HTML}{CCCCCC}
\begin{table}[!htbp] 
\centering
\vspace{-4mm}
\resizebox{\linewidth}{!}{
\begin{tabular}{lrr}
\toprule
Collaborative & \multicolumn{1}{r}{MMLU} & \multicolumn{1}{r}{Chess Move Validity} \\
Strategy & \multicolumn{1}{r}{p-value} & \multicolumn{1}{r}{p-value} \\ \midrule
$p_0p_0p_0$ & {0.186} & \colorbox{gray}{0.001} \\
$p_0p_0p_1$ & \colorbox{gray}{0.019} & \colorbox{gray}{0.000} \\
$p_0p_1p_0$ & {0.175} & \colorbox{gray}{0.000} \\
$p_0p_1p_1$ & \colorbox{gray}{0.010} & {0.178} \\
$p_1p_0p_0$ & \colorbox{gray}{0.023} & \colorbox{gray}{0.001} \\
$p_1p_0p_1$ & \colorbox{gray}{0.002} & \colorbox{gray}{0.005} \\
$p_1p_1p_0$ & {0.098} & \colorbox{gray}{0.005} \\
$p_1p_1p_1$ & \colorbox{gray}{0.004} & \colorbox{gray}{0.002} \\
\bottomrule
\end{tabular}
}
\caption{One-way ANOVA analysis of the results in Figure~\ref{fig:llama:agent} (different numbers of agents), \emph{using LlaMA2-chat-13B}.}
\label{table:sig_llama_agent}
\end{table}



\begin{figure*}[!t] 
    \centering
    \scalebox{1}{
    \includegraphics[width=1\textwidth]{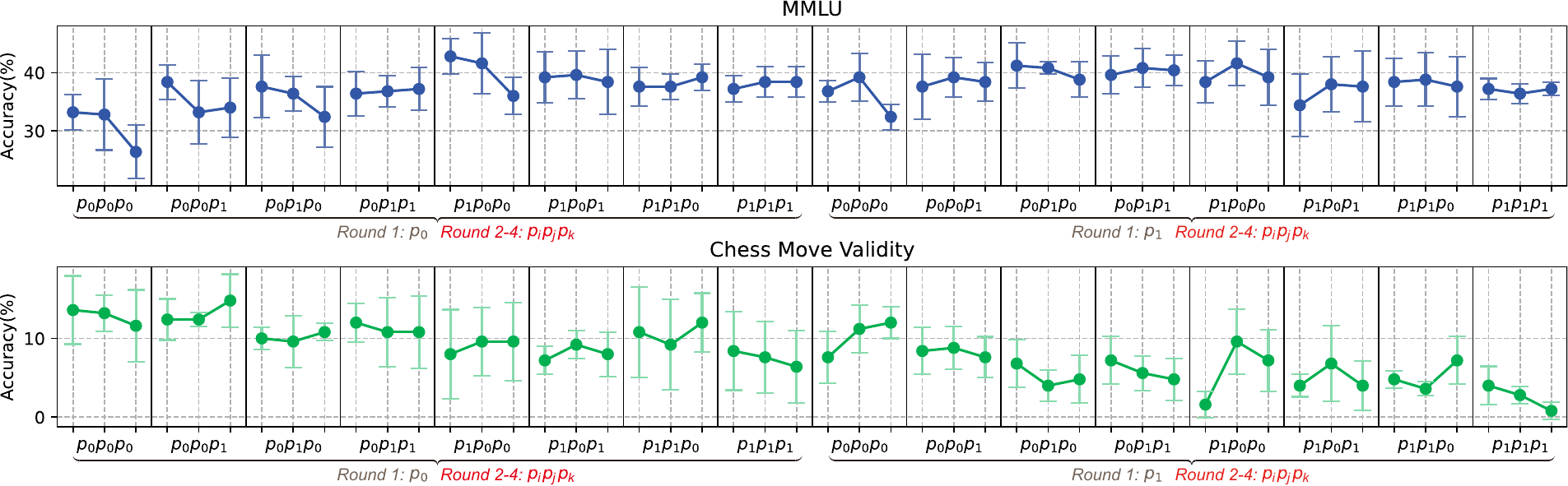}
    }
    \vspace{-5mm}
    \caption{
    Accuracy at round 2,3,4 within 4-round collaborative societies, where the thinking pattern of round 1 is fixed ($p_0$ or $p_1$), using \emph{LlaMA2-13B-chat}. 
    The significance test is shown in Table~\ref{table:sig_llama_turn}. 
    }
    \label{fig:llama:turn}
\end{figure*}

\textbf{Analysis on Different Rounds.} 
We present the significance test for different rounds of collaboration with LlaMA2 Chat 13B in Table~\ref{table:sig_llama_turn}. 
We also show the performance varying from collaboration rounds in Figure~\ref{fig:llama:turn}. 

\definecolor{gray}{HTML}{CCCCCC}
\begin{table}[!htbp] 
\centering
\small
\resizebox{\linewidth}{!}{
\begin{tabular}{lrr}
\toprule
Collaborative & MMLU & Chess Move Validity \\
Strategy & p-value & p-value \\ \midrule
$p_0p_0p_0p_0$ & \colorbox{gray}{0.000} & {0.361}  \\
$p_0p_0p_0p_1$ & {0.111} & {0.598}  \\
$p_0p_0p_1p_0$ & {0.082} & {0.335}  \\
$p_0p_0p_1p_1$ & {0.529} & {0.076}  \\
$p_0p_1p_0p_0$ & {0.293} & {0.176}  \\
$p_0p_1p_0p_1$ & {0.641} & {0.259}  \\
$p_0p_1p_1p_0$ & {0.536} & \colorbox{gray}{0.026}  \\
$p_0p_1p_1p_1$ & {0.812} & {0.052}  \\
$p_1p_0p_0p_0$ & \colorbox{gray}{0.010} & {0.629}  \\
$p_1p_0p_0p_1$ & {0.547} & \colorbox{gray}{0.029}  \\
$p_1p_0p_1p_0$ & {0.749} & {0.055}  \\
$p_1p_0p_1p_1$ & {0.600} & \colorbox{gray}{0.007}  \\
$p_1p_1p_0p_0$ & {0.605} & \colorbox{gray}{0.009}  \\
$p_1p_1p_0p_1$ & {0.988} & \colorbox{gray}{0.012}  \\
$p_1p_1p_1p_0$ & {0.889} & {0.097}  \\
$p_1p_1p_1p_1$ & {0.742} & {0.884}  \\
\bottomrule
\end{tabular}
}
\caption{One-way ANOVA analysis of the results in Figure~\ref{fig:llama:turn} (different rounds), \emph{using LlaMA2-chat-13B}.}
\label{table:sig_llama_turn}
\end{table}

\begin{figure}[!t] 
    \centering
    \scalebox{0.46}{
    \includegraphics[width=1\textwidth]{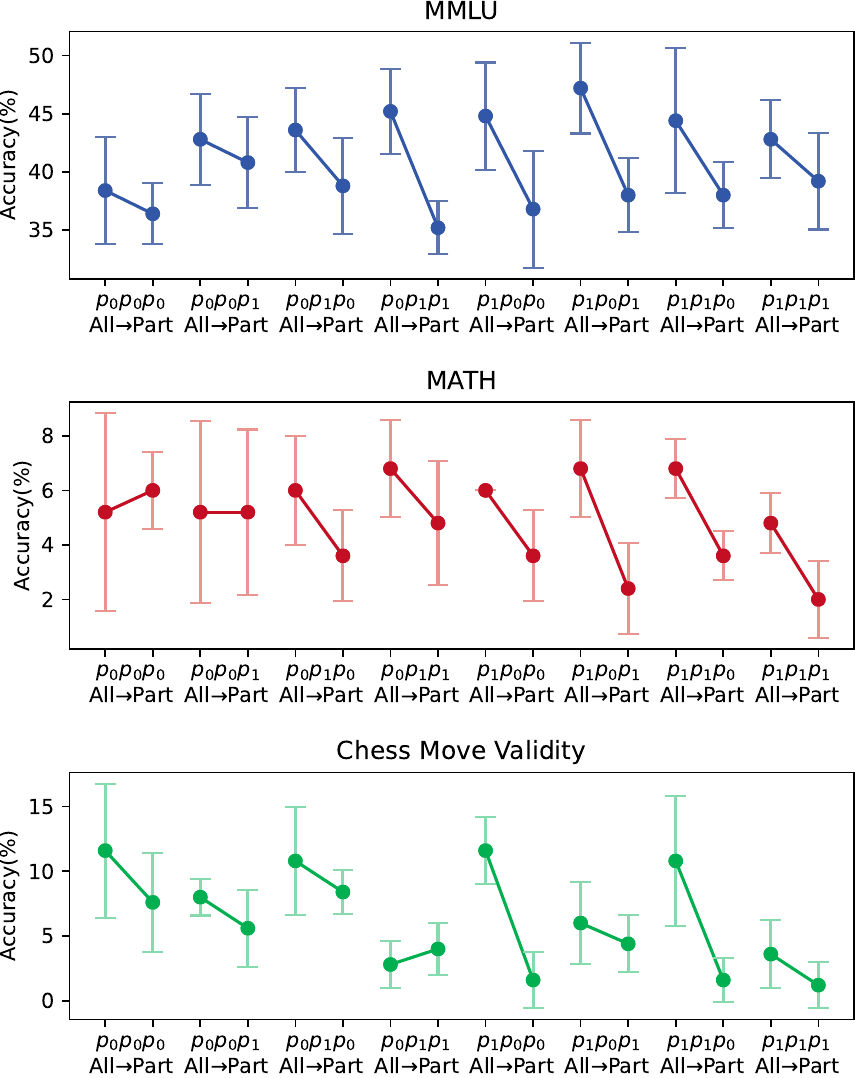}
    }
    \caption{
    The effect on the accuracy of whether all agents in society execute the same thinking pattern in one round, using \emph{LlaMA2-13B-chat}.
    ``All'' and ``Part'' refer to all agents applying the same thinking pattern and different thinking patterns in one round respectively.
    The significance test is shown in Table~\ref{table:llama:sig_strategy}.
    }
    \label{fig:llama:strategy}
\end{figure}

\textbf{Analysis on Other Collaborative Strategies.} 
We present the significance test for other collaborative strategies (executing the same or hybrid thinking patterns in a certain round) with LlaMA2 Chat 13B in Table~\ref{table:llama:sig_strategy}. 
We also show the performance varying from other strategies in Figure~\ref{fig:llama:strategy}. 

\begin{table}[!htbp] 
\centering
\resizebox{\linewidth}{!}{
\begin{tabular}{lrrr}
\toprule
Collaborative  & \multicolumn{1}{r}{MMLU} & \multicolumn{1}{r}{MATH} & \multicolumn{1}{r}{Chess Move Validity} \\
Strategy & \multicolumn{1}{r}{p-value} & \multicolumn{1}{r}{p-value} & \multicolumn{1}{r}{p-value} \\ \midrule
$p_0p_0p_0$ & {0.419} & {0.659} & {0.203}  \\
$p_0p_0p_1$ & {0.441} & {1.000} & {0.141}  \\
$p_0p_1p_0$ & {0.086} & {0.074} & {0.264}  \\
$p_0p_1p_1$ & \colorbox{gray}{0.001} & {0.161} & {0.347}  \\
$p_1p_0p_0$ & \colorbox{gray}{0.030} & {-} & \colorbox{gray}{0.000}  \\
$p_1p_0p_1$ & \colorbox{gray}{0.003} & \colorbox{gray}{0.004} & {0.380}  \\
$p_1p_1p_0$ & {0.070} & \colorbox{gray}{0.001} & \colorbox{gray}{0.005}  \\
$p_1p_1p_1$ & {0.169} & \colorbox{gray}{0.008} & {0.128}  \\
\bottomrule
\end{tabular}
}
\caption{One-way ANOVA analysis of the results in Figure~\ref{fig:llama:strategy} (other collaborative strategies), \emph{using LlaMA2-chat-13B}. 
}
\label{table:llama:sig_strategy}
\end{table}


\begin{figure*}[!t] 
    \centering
    \scalebox{1}{
    \includegraphics[width=0.94\textwidth]{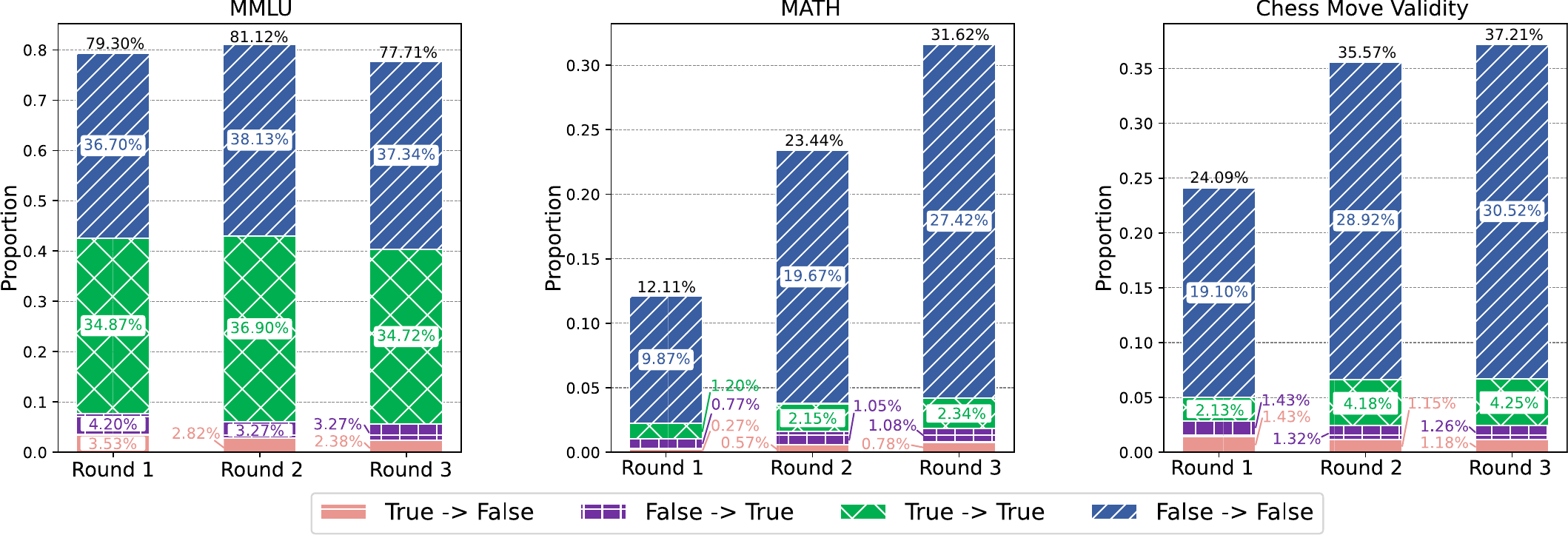}
    }
    \vspace{-2mm}
    \caption{
    Variation of answer correctness in the situation of conformity, using \emph{LlaMA2-13B-chat}, where 
    \emph{conformity brings about benefits}: Ratio$($False$\to$True + True$\to$True$)$ $>$ Ratio$($True$\to$False + False$\to$False$)$; 
    \emph{conformity brings about detriments}: Ratio$($False$\to$True + True$\to$True$)$ $<$ Ratio$($True$\to$False + False$\to$False$)$. 
    }
    \label{fig:llama:conformity}
\end{figure*}

\begin{figure*}[!t] 
    \centering
    \scalebox{1}{
    \includegraphics[width=0.94\textwidth]{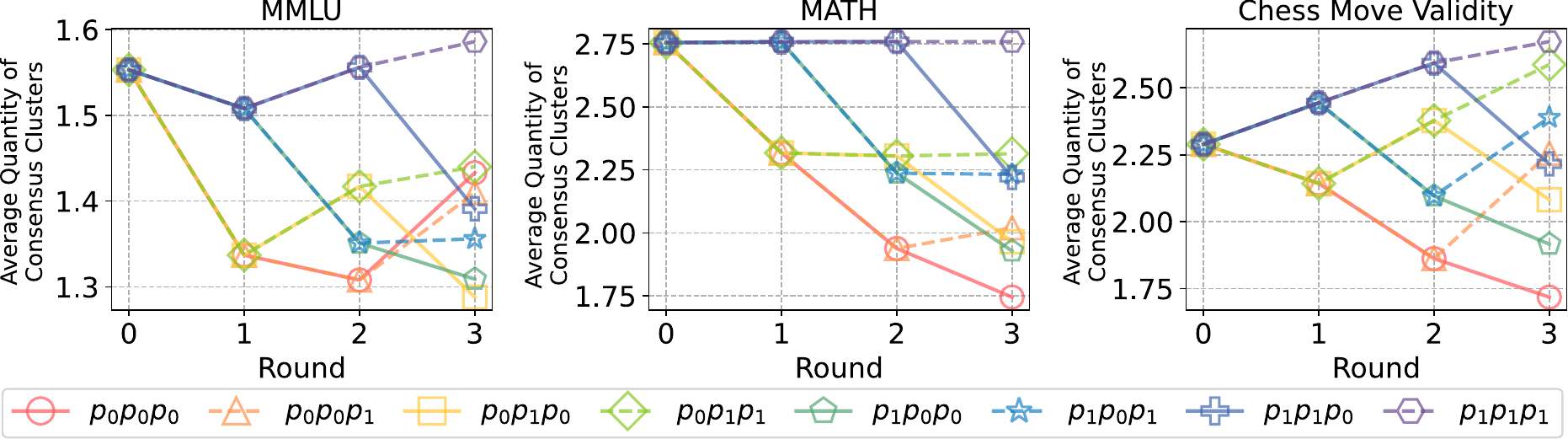}
    }
    \vspace{-2mm}
    \caption{
    Average quantity of \emph{consensus clusters (\emph{\ie}, unique answers among multiple agents)} under different rounds of collaboration with 3-round collaborative strategies, on \emph{LlaMA2-13B-chat}. 
    \emph{Smaller quantity of consensus clusters, more easier it is to reach a consensus.} 
    Round 0 is equal to self-consistency. 
    }
    \label{fig:llama:consistent}
\end{figure*}

\begin{figure*}[!t] 
    \centering
    \scalebox{1}{
    \includegraphics[width=0.92\textwidth]{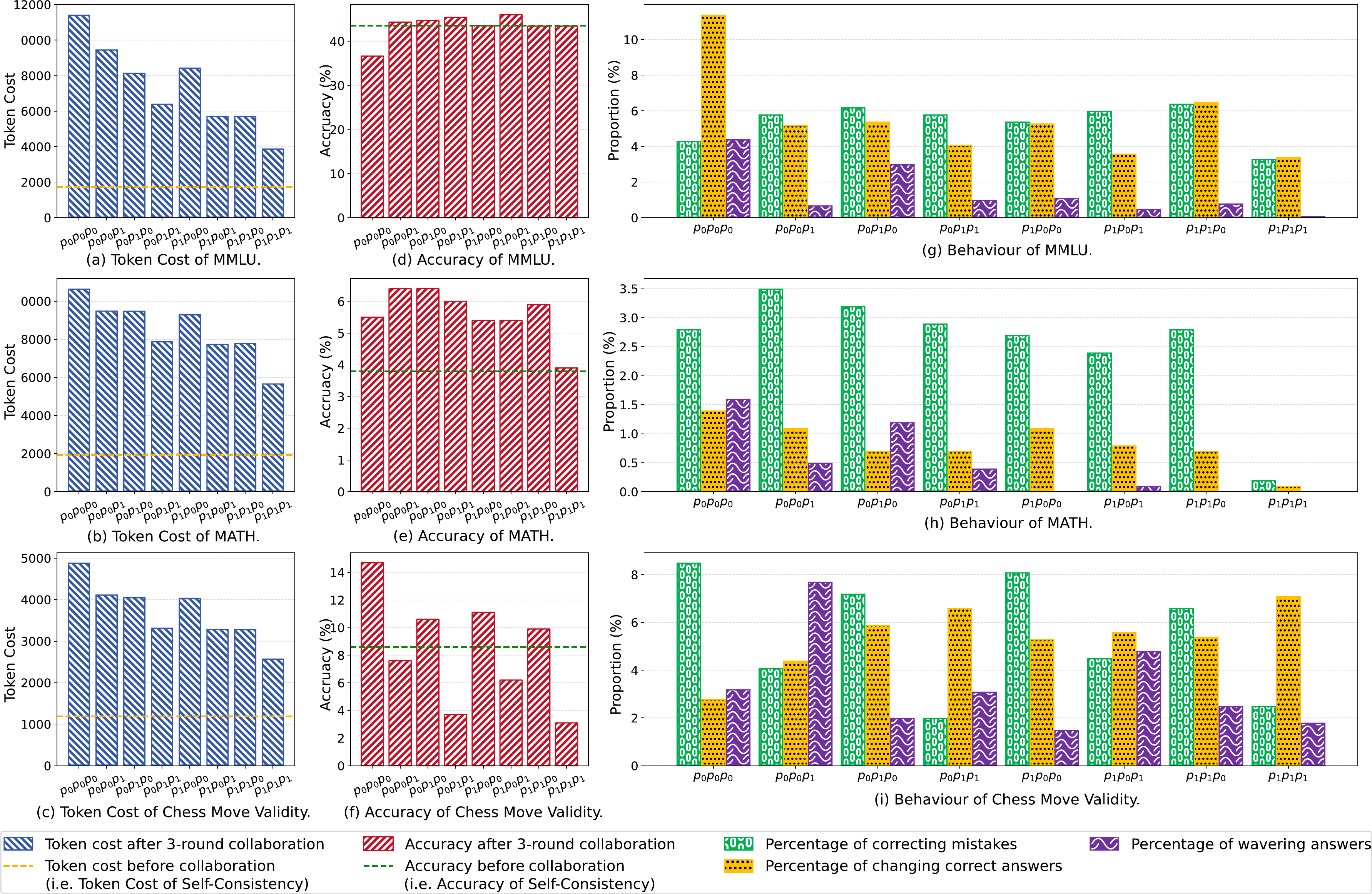}
    }
    \vspace{-2mm}
    \caption{The percentage of different behaviors under different collaborative strategies, using \emph{LlaMA2-13B-chat}.  
    Figure (a-c) \& (d-f) respectively show the token cost and accuracy of different strategies before and after 3-round collaboration. 
    Figure (g-i) present the percentage of different behavioral features (mainly analyzed by the change of answer correctness) \citep{arXiv2023_Agent-BehaviorExplanation,arXiv2023_Agent-BehaviorExplaining} under different collaborative strategies. 
    All results are summarized across all societies. 
    }
    \label{fig:llama:distribute}
\end{figure*}

\textbf{A Social Psychology View on Conformity, Consensus Reaching and Group Dynamics.} 
We then show the variation of answer correctness in the situation of conformity in Figure~\ref{fig:llama:conformity}; and the quantity of consensus clusters among 3-agent answers in Figure~\ref{fig:llama:consistent}. 
We present group dynamics reflected by different answer-changing behaviors on LlaMA2 Chat 13B in Figure~\ref{fig:llama:distribute}. 

\clearpage

\definecolor{Mycolor1}{HTML}{BAD8F2}
\definecolor{Mycolor2}{HTML}{FAE4E3}
\definecolor{Mycolor3}{HTML}{E8F2FB}

\begin{table*}[!t] 
\small
\resizebox{\linewidth}{!}{
\begin{tabular}{c|c|c|cccccccc|cc}

\toprule

& \multirow{2}{*}{\tabincell{c}{Metric \\ (Strategy)}} & \multirow{2}{*}{Society} & \multicolumn{8}{c|}{Collaborative Strategy} & \multicolumn{2}{c}{Metric (Society)}
\\
& & & $p_0p_0p_0$ & $p_0p_0p_1$ & $p_0p_1p_0$ & $p_0p_1p_1$ & $p_1p_0p_0$ & $p_1p_0p_1$ & $p_1p_1p_0$ & $p_1p_1p_1$ & \uuline{Cost}~$\downarrow$ & \uuline{W-T}~$\uparrow$
\\

\midrule

\multirow{7}{*}{\rotatebox{90}{MMLU}}  & \multirow{5}{*}{Acc~$\uparrow$} 
& $S_1$	& \colorbox{Mycolor3}{\textbf{40.8±2.7}} & \colorbox{Mycolor1}{\textbf{43.6±3.9}} & 36.0±2.8 & 38.4±3.3 & 35.6±4.3 & 35.6±2.6 & 30.4±4.3 & \colorbox{Mycolor2}{24.0±5.7} & 6915 & 7
\\
& & $S_2$	& 44.4±3.9 & \colorbox{Mycolor1}{\textbf{49.2±4.6}} & \colorbox{Mycolor3}{\textbf{45.2±3.9}} & 42.0±0.0 & 34.4±4.3 & 34.4±8.3 & 31.6±8.4 & \colorbox{Mycolor2}{25.6±3.6} & 6946 & 11
\\
& & $S_3$	& \colorbox{Mycolor3}{\textbf{44.0±5.5}} & \colorbox{Mycolor1}{\textbf{45.6±4.6}} & 39.2±2.7 & 42.8±3.0 & 35.2±5.4 & 32.4±4.3 & 28.0±7.3 & \colorbox{Mycolor2}{25.6±5.2} & 6931 & 8
\\
& & $S_4$	& \colorbox{Mycolor3}{\textbf{47.6±4.1}} & \colorbox{Mycolor1}{\textbf{48.0±5.1}} & 46.0±6.3 & 45.2±3.9 & \colorbox{Mycolor2}{26.8±3.6} & 30.8±6.9 & 32.8±1.8 & 33.6±6.2 & 6936 & 8
\\ 

\cmidrule{2-13} 

& \underline{Cost}~$\downarrow$  & All   & 10811 & 8608 & 7904 & 6177 & 7535 & 5410 & 5287 & 3722 & \multicolumn{2}{c}{\multirow{2}{*}{-}} \\ \cmidrule{2-11}
& \underline{W-T}~$\uparrow$     & All   & -     & \textbf{16} & 5 & 11 & 1 & 0 & 1 & 0 & \multicolumn{2}{c}{}
\\ 

\midrule

\multirow{7}{*}{\rotatebox{90}{MATH}}  & \multirow{5}{*}{Acc~$\uparrow$}    & $S_1$	& 8.4±3.6 & \colorbox{Mycolor1}{\textbf{10.4±3.9}} & \colorbox{Mycolor3}{\textbf{9.2±1.1}} & 4.0±2.5 & 9.2±4.2 & 8.4±4.3 & 6.8±2.7 & \colorbox{Mycolor2}{3.6±1.7} & 7000 & 16
\\
&	  & $S_2$	& 8.0±2.5 & \colorbox{Mycolor1}{\textbf{9.6±2.6}} & \colorbox{Mycolor3}{\textbf{8.8±3.0}} & 6.4±2.6 & 7.2±4.4 & 6.8±1.1 & 8.4±4.3 & \colorbox{Mycolor2}{4.8±2.3} & 7013 & 19
\\
&	  & $S_3$	& \colorbox{Mycolor3}{\textbf{8.4±4.6}} & 7.2±3.9 & \colorbox{Mycolor1}{\textbf{8.4±3.6}} & 5.6±3.6 & 7.2±1.8 & 7.2±4.8 & 6.8±3.0 & \colorbox{Mycolor2}{0.8±1.1} & 7157 & 15
\\
&	  & $S_4$	&  6.0±2.0 & \colorbox{Mycolor3}{\textbf{7.2±1.8}} & 6.0±2.0 & 4.0±2.0 & 5.2±3.0 & 6.8±1.1 & \colorbox{Mycolor1}{\textbf{8.8±4.4}} & \colorbox{Mycolor2}{3.6±2.6} & 6934 & 23
\\ 

\cmidrule{2-13} 

& \underline{Cost}~$\downarrow$  & All & 9465 & 7850 & 7662 & 6294 & 7520 & 6302 & 6382 & 4734  & \multicolumn{2}{c}{\multirow{2}{*}{-}} 
\\ 

\cmidrule{2-11}

& \underline{W-T}~$\uparrow$     & All & - & \textbf{14} & \textbf{14} & 5 & 13 & 9 & \textbf{14} & 4 & \multicolumn{2}{c}{}
\\ 

\midrule

\multirow{6}{*}{\rotatebox{90}{Chess Move Validity}} & \multirow{5}{*}{Acc~$\uparrow$}    & $S_1$	& \colorbox{Mycolor3}{\textbf{20.4±6.2}} & 16.8±3.6 & 17.2±4.2 & 8.4±2.2 & \colorbox{Mycolor1}{\textbf{21.2±5.8}} & 10.8±3.0 & 10.4±1.7 & \colorbox{Mycolor2}{4.8±3.0} & 3563 & 7
\\
&	  & $S_2$	&  \colorbox{Mycolor1}{\textbf{18.4±4.8}} & 9.6±3.6 & 13.2±1.1 & 5.6±2.2 & \colorbox{Mycolor3}{\textbf{14.4±3.9}} & 7.2±3.0 & 13.2±3.4 & \colorbox{Mycolor2}{4.0±2.8} & 3557 & 4 
\\
&	  & $S_3$	& \colorbox{Mycolor3}{\textbf{18.4±6.5}} & 11.2±3.0 & 12.0±5.8 & 8.0±2.0 & \colorbox{Mycolor1}{\textbf{20.8±4.6}} & 8.4±4.3 & 12.8±2.7 & \colorbox{Mycolor2}{2.8±3.4} & 3629 & 7
\\
&	  & $S_4$	&  15.2±4.2 & 11.6±2.2 & \colorbox{Mycolor3}{\textbf{15.2±2.3}} & 10.4±1.7 & \colorbox{Mycolor1}{\textbf{18.0±4.7}} & 8.0±4.7 & 10.8±2.7 & \colorbox{Mycolor2}{5.2±2.3} & 3679 & 12
\\ 

\cmidrule{2-13} 

& \underline{Cost}~$\downarrow$  & All & 4778 & 3947 & 3830 & 3082 & 4139 & 3314 & 3259 & 2508  & \multicolumn{2}{c}{\multirow{2}{*}{-}} 
\\ 

\cmidrule{2-11}

& \underline{W-T}~$\uparrow$     & All  & - & 4 & 6 & 2 & \textbf{13} & 1 & 4 & 0 & \multicolumn{2}{c}{} 
\\ 

\bottomrule

\end{tabular}
}

\caption{
The impact of eight different collaborative strategies on the performance of three datasets across distinct societies (\emph{using LlaMA2-chat-70B}).
The significances test on societies and strategies are respectively shown in Table~\ref{table:llama70:sig_main_society},~\ref{table:llama70:sig_main_strategy}. 
The experiments of comparison with the single LLM agent is shown in Figure~\ref{fig:llama70:distribute}(a)-(f). 
\label{table:llama70_main}
}

\end{table*}
\definecolor{gray}{HTML}{CCCCCC}
\begin{table}[!htbp] 
\centering
\resizebox{\linewidth}{!}{
\begin{tabular}{lrrr}
\toprule
Collaborative  & \multicolumn{1}{r}{MMLU} & \multicolumn{1}{r}{MATH} & \multicolumn{1}{r}{Chess Move Validity} \\
Strategy & \multicolumn{1}{r}{p-value} & \multicolumn{1}{r}{p-value} & \multicolumn{1}{r}{p-value} \\ \midrule
$p_0p_0p_0$ & {0.122} & {0.621} & {0.532}  \\
$p_0p_0p_1$ & {0.251} & {0.291} & \colorbox{gray}{0.014}  \\
$p_0p_1p_0$ & \colorbox{gray}{0.004} & {0.248} & {0.185}  \\
$p_0p_1p_1$ & \colorbox{gray}{0.018} & {0.430} & \colorbox{gray}{0.015}  \\
$p_1p_0p_0$ & \colorbox{gray}{0.020} & {0.381} & {0.132}  \\
$p_1p_0p_1$ & {0.601} & {0.854} & {0.506}  \\
$p_1p_1p_0$ & {0.641} & {0.750} & {0.282}  \\
$p_1p_1p_1$ & \colorbox{gray}{0.044} & \colorbox{gray}{0.037} & {0.585}  \\
\bottomrule
\end{tabular}
}
\caption{
One-Way ANOVA results for the impact of society on accuracy with fixed collaborative strategy, based on experiments from Table~\ref{table:llama70_main} using \emph{LlaMA2-chat-70B}.
}
\label{table:llama70:sig_main_society}
\end{table}
\definecolor{gray}{HTML}{CCCCCC}
\begin{table}[!htbp] 
\centering
\resizebox{\linewidth}{!}{
\begin{tabular}{lrrr}
\toprule
 & \multicolumn{1}{r}{MMLU} & \multicolumn{1}{r}{MATH} & \multicolumn{1}{r}{Chess Move Validity} \\
Society & \multicolumn{1}{r}{p-value} & \multicolumn{1}{r}{p-value} & \multicolumn{1}{r}{p-value} \\ \midrule
$S_1$ & \colorbox{gray}{0.000} & \colorbox{gray}{0.013} & \colorbox{gray}{0.000} \\
$S_2$ & \colorbox{gray}{0.000} & {0.297} & \colorbox{gray}{0.000} \\
$S_3$ & \colorbox{gray}{0.000} & \colorbox{gray}{0.040} & \colorbox{gray}{0.000} \\
$S_4$ & \colorbox{gray}{0.000} & {0.056} & \colorbox{gray}{0.000} \\
\bottomrule
\end{tabular}
}
\caption{
One-Way ANOVA results for the impact of collaborative strategy on accuracy with fixed society, based on experiments from Table~\ref{table:llama70_main} using \emph{LlaMA-70B-Chat}.
}
\label{table:llama70:sig_main_strategy}
\end{table}

\subsection{LlaMA2 Chat 70B}
\label{app:backbone_llama70b}

\begin{figure*}[!t] 
    \centering
    \scalebox{1}{
    \includegraphics[width=1\textwidth]{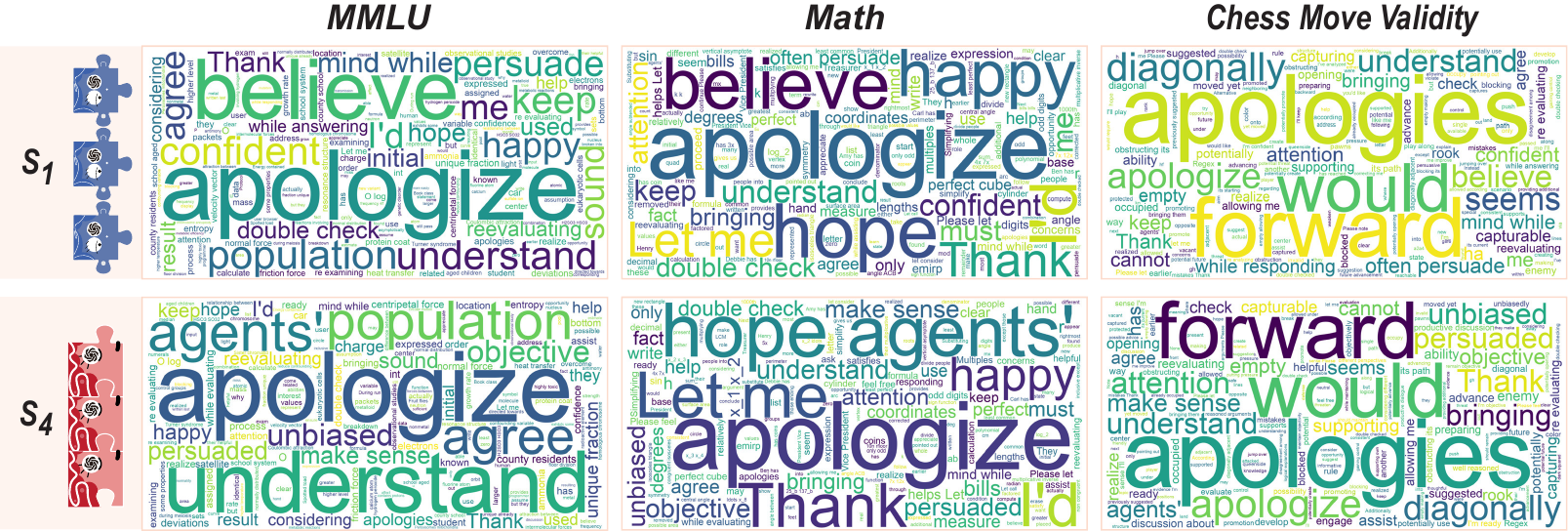}
    }
    \vspace{-4mm}
    \caption{
    Comparative word clouds on three datasets in societies $S_1$ and $S_4$, using \emph{LlaMA2-70B-chat}. 
    Society $S_1$ features three overconfident agents, while society $S_4$ comprises three easy-going agents. 
    }
    \label{fig:llama70:word}
\end{figure*}

\begin{figure*}[!t] 
    \centering
    \scalebox{1}{
    \includegraphics[width=1\textwidth]{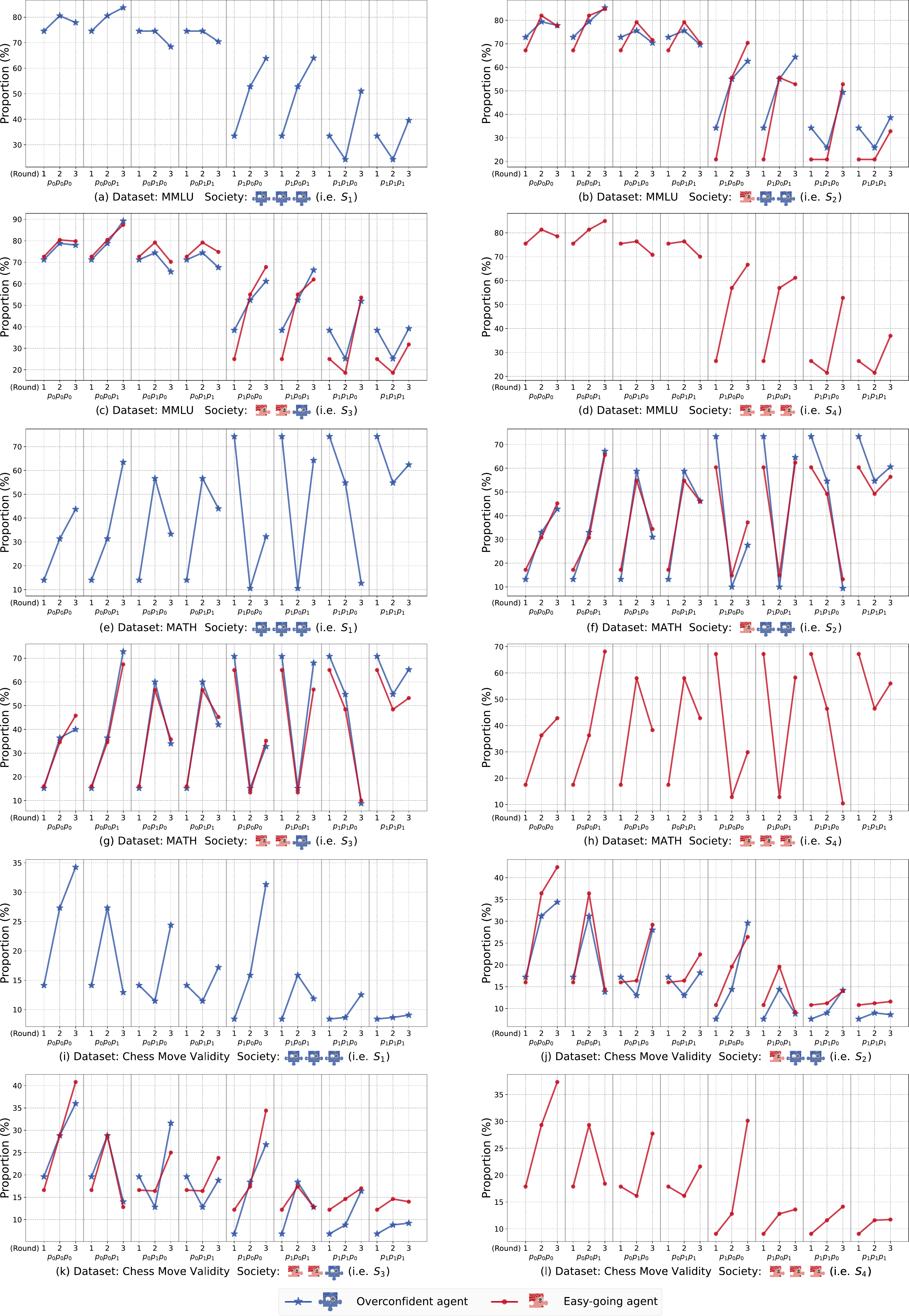}
    }
    \vspace{-4mm}
    \caption{
    Proportion of agents with different traits keeping answers in societies $S_1$ and $S_4$, using \emph{LlaMA2-70B-chat}. 
    Society $S_1$ features three overconfident agents, while society $S_4$ comprises three easy-going agents. 
    }
    \label{fig:llama70:agent_answer_changing}
\end{figure*}

\begin{figure*}[!t] 
    \centering
    \scalebox{1}{
    \includegraphics[width=1\textwidth]{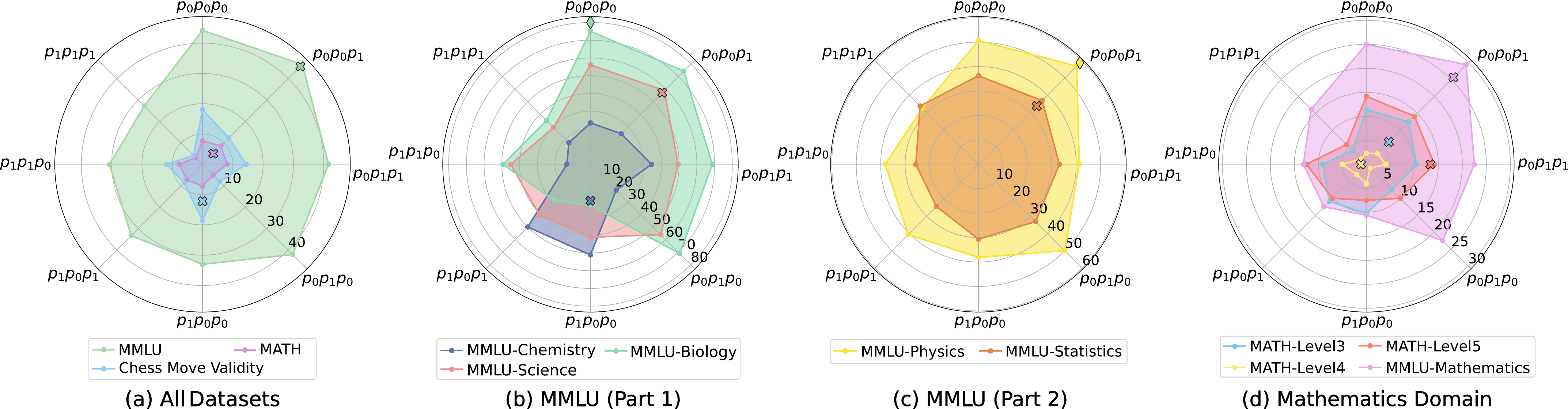}
    }
    \vspace{-6mm}
    \caption{
    Illustration of different collaborative strategies impacting accuracy diversely on the tasks considering varied \emph{subjects} and \emph{difficulty}, using \emph{LlaMA2-70B-chat}. 
    The symbol `\protect\radarfork' represents that there is at least one collaborative strategy whose accuracy is better than self-consistency, while the symbol `\protect\radarprismatic' indicates that there is no collaborative strategy whose accuracy is worse than self-consistency. 
    Both of these symbols represent the accuracy of self-consistency. 
    The accuracy under each collaborative strategy is a summation within all 3-agent societies. 
    \label{fig:llama70:task_radar}
    }
\end{figure*}

\textbf{Analysis on Machine Social Collaboration.} 
We present the \textbf{main results} and \textbf{significance tests} of societies and strategies on LlaMA2 Chat 70B in Table~\ref{table:llama70_main},~\ref{table:llama70:sig_main_society},~\ref{table:llama70:sig_main_strategy}. 
We present the \textbf{word clouds} of LlaMA2 Chat 70B in Figure~\ref{fig:llama70:word}, and \textbf{proportion of agents with different traits keeping answers in different societies} on LlaMA2 Chat 70B in Figure~\ref{fig:llama70:agent_answer_changing}. 
Furthermore, we demonstrate that the tasks with different subjects and difficulty display varying sensitivity to collaborative strategies, as presented with \textbf{radar maps} on LlaMA2 Chat 70B in Figure~\ref{fig:llama70:task_radar}.


\begin{figure*}[!t] 
    \centering
    \includegraphics[width=0.92\textwidth]{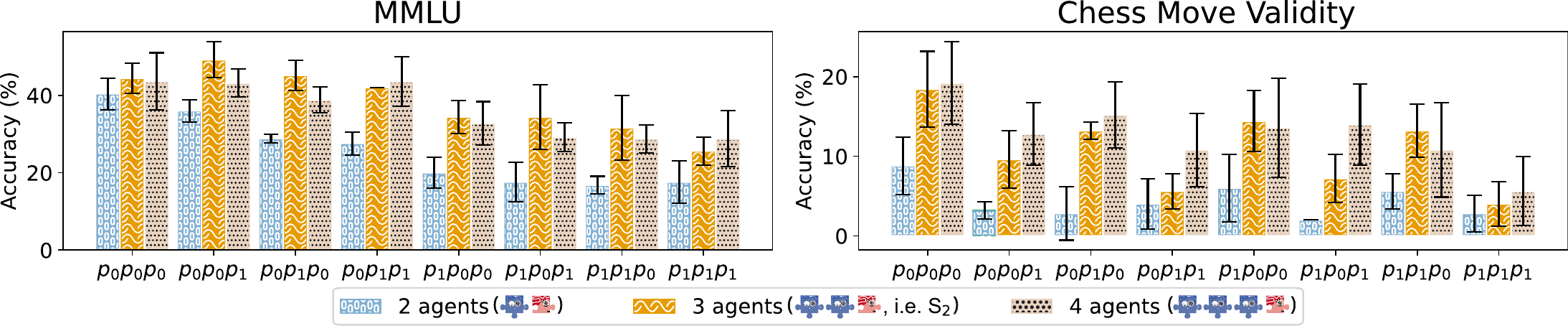}
    \vspace{-3mm}
    \caption{Accuracy of different numbers of agents under different collaborative strategies, on \emph{LlaMA2-70B-chat}. 
    The significance test is shown in Table~\ref{table:sig_llama70_agent}. 
    }
    \label{fig:llama70:agent}
\end{figure*}

\textbf{Analysis on Different Numbers of Agents.} 
We present the significance test for different numbers of agents with LlaMA2 Chat 70B in Table~\ref{table:sig_llama70_agent}. 
We also show the performance varying from agent numbers in Figure~\ref{fig:llama70:agent}. 

\definecolor{gray}{HTML}{CCCCCC}
\begin{table}[!htbp] 
\centering
\vspace{-4mm}
\resizebox{\linewidth}{!}{
\begin{tabular}{lrr}
\toprule
Collaborative & \multicolumn{1}{r}{MMLU} & \multicolumn{1}{r}{Chess Move Validity} \\
Strategy & \multicolumn{1}{r}{p-value} & \multicolumn{1}{r}{p-value} \\ \midrule
$p_0p_0p_0$ & {0.481} & \colorbox{gray}{0.006} \\
$p_0p_0p_1$ & \colorbox{gray}{0.000} & \colorbox{gray}{0.001} \\
$p_0p_1p_0$ & \colorbox{gray}{0.000} & \colorbox{gray}{0.000} \\
$p_0p_1p_1$ & {-} & \colorbox{gray}{0.023} \\
$p_1p_0p_0$ & \colorbox{gray}{0.001} & \colorbox{gray}{0.035} \\
$p_1p_0p_1$ & \colorbox{gray}{0.003} & \colorbox{gray}{0.000} \\
$p_1p_1p_0$ & \colorbox{gray}{0.002} & \colorbox{gray}{0.036} \\
$p_1p_1p_1$ & \colorbox{gray}{0.024} & {0.423} \\
\bottomrule
\end{tabular}
}
\caption{One-way ANOVA analysis of the results of Figure~\ref{fig:llama70:agent} (different numbers of agents), \emph{using LlaMA2-chat-70B}.}
\label{table:sig_llama70_agent}
\end{table}

\begin{figure*}[!t] 
    \centering
    \scalebox{1}{
    \includegraphics[width=1\textwidth]{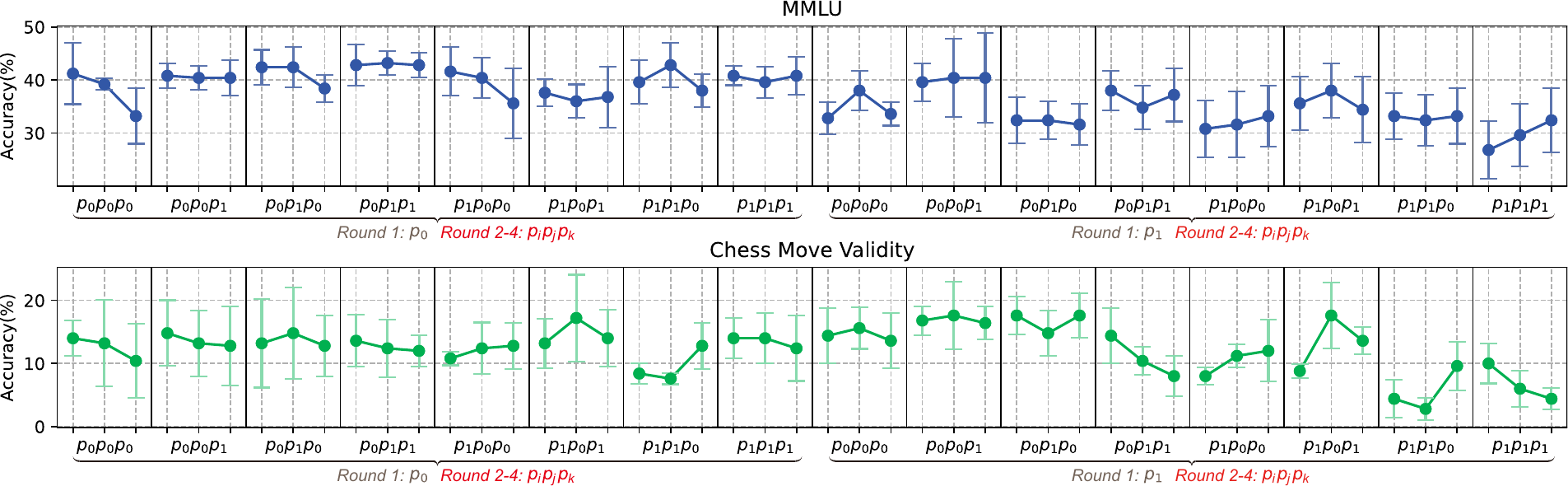}
    }
    \vspace{-5mm}
    \caption{
    Accuracy at round 2,3,4 within 4-round collaborative societies, where the thinking pattern of round 1 is fixed ($p_0$ or $p_1$), using \emph{LlaMA2-70B-chat}. 
    The significance test is shown in Table~\ref{table:sig_llama70_turn}. 
    }
    \label{fig:llama70:turn}
\end{figure*}

\textbf{Analysis on Different Rounds.} 
We present the significance test for different rounds of collaboration with LlaMA2 Chat 70B in Table~\ref{table:sig_llama70_turn}. 
We also show the performance varying from collaboration rounds in Figure~\ref{fig:llama70:turn}. 

\definecolor{gray}{HTML}{CCCCCC}
\begin{table}[!htbp] 
\centering
\small
\resizebox{\linewidth}{!}{
\begin{tabular}{lrr}
\toprule
Collaborative & MMLU & Chess Move Validity \\
Strategy & p-value & p-value \\ \midrule
$p_0p_0p_0p_0$ & \colorbox{gray}{0.034} & {0.545}  \\
$p_0p_0p_0p_1$ & \colorbox{gray}{0.008} & \colorbox{gray}{0.019}  \\
$p_0p_0p_1p_0$ & \colorbox{gray}{0.020} & \colorbox{gray}{0.004}  \\
$p_0p_0p_1p_1$ & {0.643} & \colorbox{gray}{0.004}  \\
$p_0p_1p_0p_0$ & \colorbox{gray}{0.045} & \colorbox{gray}{0.034}  \\
$p_0p_1p_0p_1$ & {0.164} & {0.902}  \\
$p_0p_1p_1p_0$ & \colorbox{gray}{0.046} & \colorbox{gray}{0.006}  \\
$p_0p_1p_1p_1$ & {0.082} & \colorbox{gray}{0.000}  \\
$p_1p_0p_0p_0$ & {0.706} & {0.207}  \\
$p_1p_0p_0p_1$ & {0.449} & {0.494}  \\
$p_1p_0p_1p_0$ & {0.782} & {0.095}  \\
$p_1p_0p_1p_1$ & {0.664} & {0.070}  \\
$p_1p_1p_0p_0$ & {0.360} & \colorbox{gray}{0.041}  \\
$p_1p_1p_0p_1$ & {0.391} & \colorbox{gray}{0.018}  \\
$p_1p_1p_1p_0$ & {0.394} & {0.088}  \\
$p_1p_1p_1p_1$ & \colorbox{gray}{0.031} & \colorbox{gray}{0.033}  \\
\bottomrule
\end{tabular}
}
\caption{One-way ANOVA analysis of the results in Figure~\ref{fig:llama70:turn} (different rounds), \emph{using LlaMA2-chat-70B}.}
\label{table:sig_llama70_turn}
\vspace{-2mm}
\end{table}



\begin{figure}[!t] 
    \centering
    \scalebox{0.46}{
    \includegraphics[width=1\textwidth]{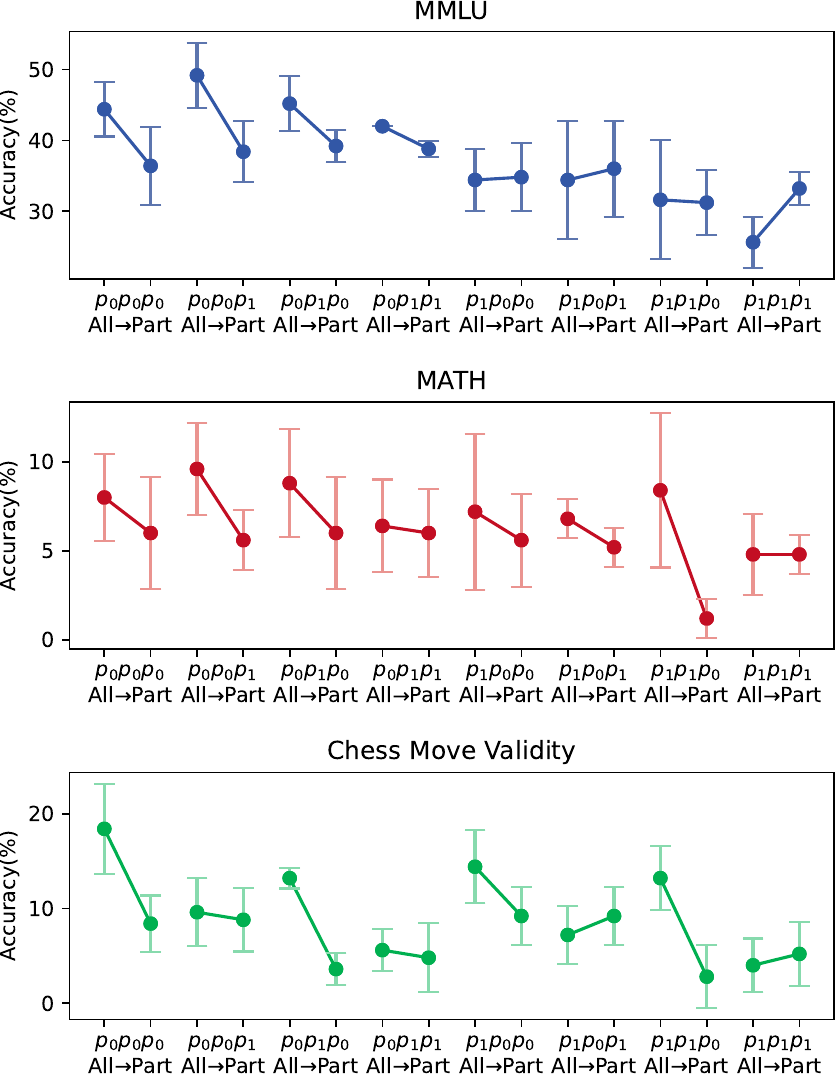}
    }
    \caption{
    The effect on the accuracy of whether all agents in society execute the same thinking pattern in one round, using \emph{LlaMA2-70B-chat}.
    ``All'' and ``Part'' refers to all agents applying the same thinking pattern and different thinking patterns in one round respectively.
    The significance test is shown in Table~\ref{table:llama70:sig_strategy}.
    }
    \label{fig:llama70:strategy}
\end{figure}

\textbf{Analysis on Other Collaborative Strategies.} 
We present the significance test for other collaborative strategies (executing the same or hybrid thinking patterns in a certain round) with LlaMA2 Chat 70B in Table~\ref{table:llama70:sig_strategy}. 
We also show the performance varying from other strategies in Figure~\ref{fig:llama70:strategy}. 

\begin{table}[!htbp] 
\centering
\resizebox{\linewidth}{!}{
\begin{tabular}{lrrr}
\toprule
Collaborative  & \multicolumn{1}{r}{MMLU} & \multicolumn{1}{r}{MATH} & \multicolumn{1}{r}{Chess Move Validity} \\
Strategy & \multicolumn{1}{r}{p-value} & \multicolumn{1}{r}{p-value} & \multicolumn{1}{r}{p-value} \\ \midrule
$p_0p_0p_0$ & \colorbox{gray}{0.029} & {0.296} & \colorbox{gray}{0.004}  \\
$p_0p_0p_1$ & \colorbox{gray}{0.005} & \colorbox{gray}{0.020} & {0.724}  \\
$p_0p_1p_0$ & \colorbox{gray}{0.018} & {0.191} & \colorbox{gray}{0.000}  \\
$p_0p_1p_1$ & \colorbox{gray}{0.000} & {0.809} & {0.684}  \\
$p_1p_0p_0$ & {0.894} & {0.503} & \colorbox{gray}{0.045}  \\
$p_1p_0p_1$ & {0.747} & \colorbox{gray}{0.050} & {0.328}  \\
$p_1p_1p_0$ & {0.928} & \colorbox{gray}{0.007} & \colorbox{gray}{0.001}  \\
$p_1p_1p_1$ & \colorbox{gray}{0.004} & {1.000} & {0.557}  \\
\bottomrule
\end{tabular}
}
\caption{One-way ANOVA analysis of the results in Figure~\ref{fig:llama70:strategy} (other collaborative strategies), \emph{using LlaMA2-chat-70B}. 
}
\label{table:llama70:sig_strategy}
\end{table}

\begin{figure*}[!t] 
    \centering
    \scalebox{1}{
    \includegraphics[width=0.84\textwidth]{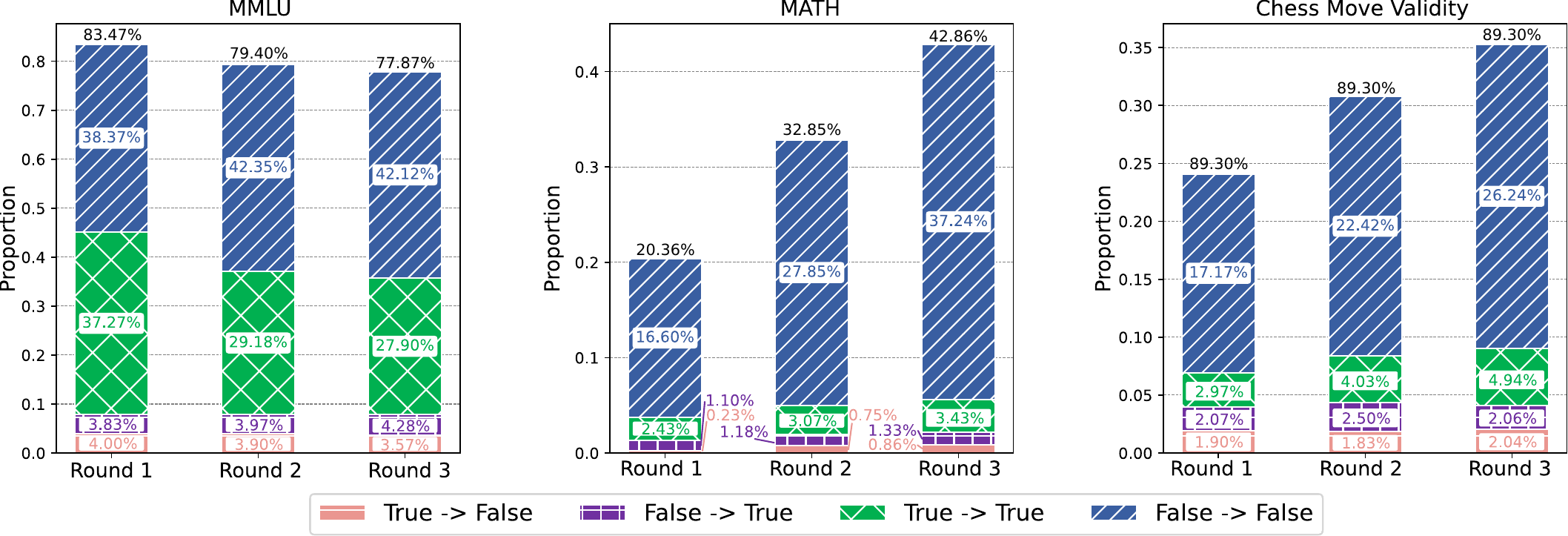}
    }
    \vspace{-2mm}
    \caption{
    Variation of answer correctness in the situation of conformity, using \emph{LlaMA2-70B-chat}, where 
    \emph{conformity brings about benefits}: Ratio$($False$\to$True + True$\to$True$)$ $>$ Ratio$($True$\to$False + False$\to$False$)$; 
    \emph{conformity brings about detriments}: Ratio$($False$\to$True + True$\to$True$)$ $<$ Ratio$($True$\to$False + False$\to$False$)$. 
    }
    \label{fig:llama70:conformity}
\end{figure*}
\begin{figure*}[!t] 
    \centering
    \scalebox{1}{
    \includegraphics[width=0.84\textwidth]{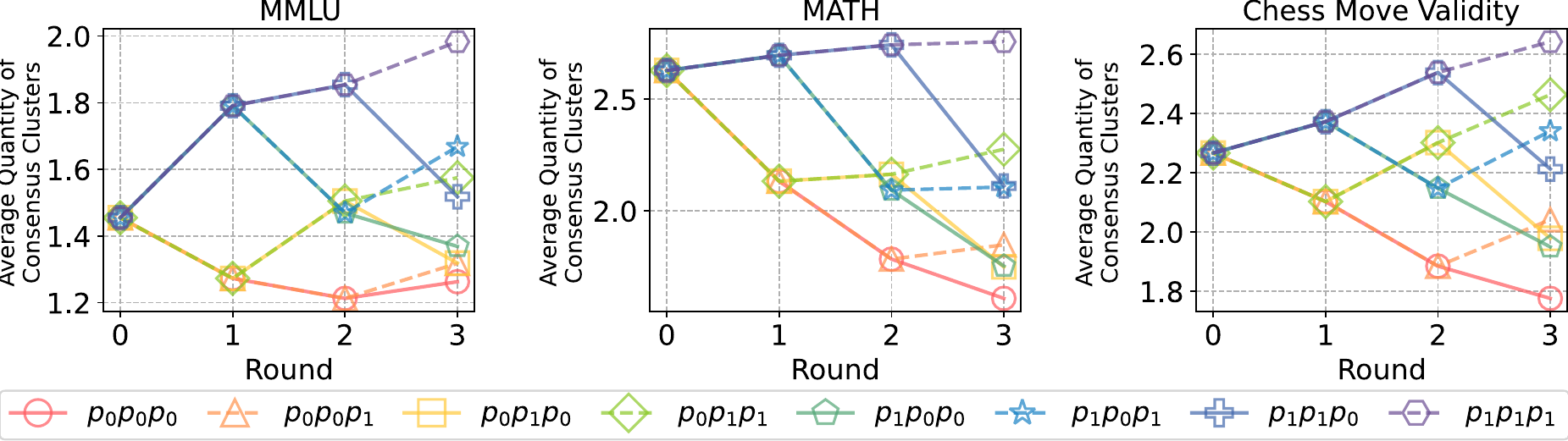}
    }
    \vspace{-2mm}
    \caption{
    Average quantity of \emph{consensus clusters (\emph{\ie}, unique answers among multiple agents)} under different rounds of collaboration with 3-round collaborative strategies, on \emph{LlaMA2-70B-chat}. 
    \emph{Smaller quantity of consensus clusters, more easier it is to reach a consensus.} 
    Round 0 is equal to self-consistency. 
    }
    \label{fig:llama70:consistent}
\end{figure*}
\begin{figure*}[!t] 
    \centering
    \scalebox{1}{
    \includegraphics[width=0.92\textwidth]{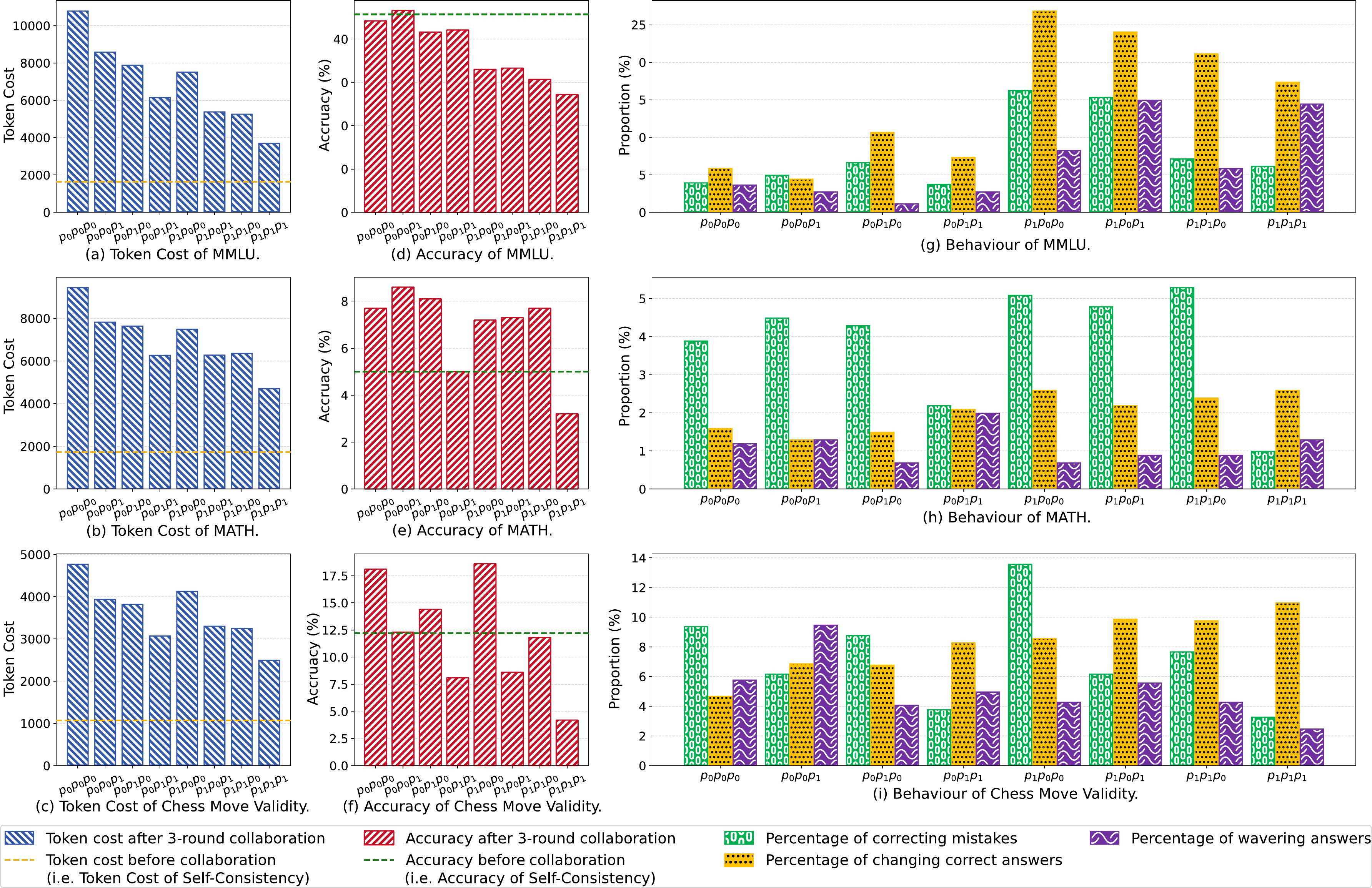}
    }
    \vspace{-2mm}
    \caption{The percentage of different behaviors under different collaborative strategies, using \emph{LlaMA2-70B-chat}. 
    Figure (a-c) \& (d-f) respectively show the token cost and accuracy of different strategies before and after 3-round collaboration. 
    Figure (g-i) present the percentage of different behavioral features (mainly analyzed by the change of answer correctness) \citep{arXiv2023_Agent-BehaviorExplanation,arXiv2023_Agent-BehaviorExplaining} under different collaborative strategies. 
    All results are summarized across all societies. 
    }
    \label{fig:llama70:distribute}
\end{figure*}

\textbf{A Social Psychology View on Conformity, Consensus Reaching and Group Dynamics.} 
We then show the variation of answer correctness in the situation of conformity in Figure~\ref{fig:llama70:conformity}; and the quantity of consensus clusters among 3-agent answers in Figure~\ref{fig:llama70:consistent}. 
We present group dynamics reflected by different answer-changing behaviors on LlaMA2 Chat 70B in Figure~\ref{fig:llama70:distribute}. 

\clearpage

\definecolor{Mycolor1}{HTML}{BAD8F2}
\definecolor{Mycolor2}{HTML}{FAE4E3}
\definecolor{Mycolor3}{HTML}{E8F2FB}

\begin{table*}[!t] 
\small
\resizebox{\linewidth}{!}{
\begin{tabular}{c|c|c|cccccccc|cc}

\toprule

& \multirow{2}{*}{\tabincell{c}{Metric \\ (Strategy)}} & \multirow{2}{*}{Society} & \multicolumn{8}{c|}{Collaborative Strategy} & \multicolumn{2}{c}{Metric (Society)}
\\
& & & $p_0p_0p_0$ & $p_0p_0p_1$ & $p_0p_1p_0$ & $p_0p_1p_1$ & $p_1p_0p_0$ & $p_1p_0p_1$ & $p_1p_1p_0$ & $p_1p_1p_1$ & \uuline{Cost}~$\downarrow$ & \uuline{W-T}~$\uparrow$
\\

\midrule

\multirow{7}{*}{\rotatebox{90}{MMLU}}  & \multirow{5}{*}{Acc~$\uparrow$} 
& $S_1$	& 64.8±6.4 & \colorbox{Mycolor1}{\textbf{66.4±6.8}} & \colorbox{Mycolor3}{\textbf{65.6±9.7}} & 63.6±5.0 & \colorbox{Mycolor2}{58.0±4.2} & 58.4±3.0 & 60.0±8.8 & 63.6±2.6 & 3661 & 14
\\
& & $S_2$	& 60.4±5.9 & 60.8±5.2 & \colorbox{Mycolor1}{\textbf{62.8±2.3}} & 61.6±4.6 & \colorbox{Mycolor2}{53.2±5.6} & 57.6±2.6 & 61.2±7.8 & \colorbox{Mycolor3}{\textbf{62.4±4.3}} & 3657 & 21
\\
& & $S_3$	& 64.0±4.7 & 64.4±3.9 & \colorbox{Mycolor1}{\textbf{66.0±2.8}} & \colorbox{Mycolor3}{\textbf{65.2±3.0}} & \colorbox{Mycolor2}{56.8±5.9} & 57.6±5.2 & 59.6±4.3 & 64.4±2.6 & 3690 & 17
\\
& & $S_4$	& 62.4±6.2 & \colorbox{Mycolor3}{\textbf{64.8±3.9}} & 64.0±7.1 & \colorbox{Mycolor1}{\textbf{66.8±7.3}} & \colorbox{Mycolor2}{53.2±5.4} & 56.8±4.2 & 60.4±7.4 & 58.4±3.9 & 3570 & 14
\\ 

\cmidrule{2-13} 

& \underline{Cost}~$\downarrow$  & All   & 5960 & 4560 & 4017 & 3158 & 4024 & 2761 & 2746 & 1927  & \multicolumn{2}{c}{\multirow{2}{*}{-}} \\ \cmidrule{2-11}
& \underline{W-T}~$\uparrow$     & All   & -     & 12 & \textbf{14} & 13 & 4 & 4 & 9 & 10 & \multicolumn{2}{c}{}
\\ 

\midrule

\multirow{7}{*}{\rotatebox{90}{MATH}}  & \multirow{5}{*}{Acc~$\uparrow$}    & $S_1$	& \colorbox{Mycolor1}{\textbf{47.2±5.6}} & 43.6±4.6 & \colorbox{Mycolor3}{\textbf{46.0±6.5}} & 43.6±5.0 & 40.4±6.5 & 41.6±8.1 & 42.0±4.9 & \colorbox{Mycolor2}{39.6±3.9} & 3537 & 11
\\
&	  & $S_2$	& \colorbox{Mycolor1}{\textbf{49.6±5.4}} & 48.4±6.1 & \colorbox{Mycolor3}{\textbf{48.8±6.7}} & 47.2±5.9 & 41.2±4.4 & 41.6±5.4 & 40.0±4.0 & \colorbox{Mycolor2}{37.6±4.1} & 3513 & 7
\\
&	  & $S_3$	& \colorbox{Mycolor1}{\textbf{44.8±6.4}} & \colorbox{Mycolor3}{\textbf{44.4±5.5}} & 43.6±4.3 & 42.0±7.1 & 40.4±7.8 & 37.6±6.7 & 41.6±7.5 & \colorbox{Mycolor2}{36.4±8.7} & 3595 & 9
\\
&	  & $S_4$	& \colorbox{Mycolor1}{\textbf{46.0±6.6}} & 44.8±8.6 & \colorbox{Mycolor3}{\textbf{46.0±8.0}} & 43.6±5.4 & 39.2±5.0 & 41.6±4.8 & 37.6±6.7 & \colorbox{Mycolor2}{35.6±3.9} & 3595 & 11
\\ 

\cmidrule{2-13} 

& \underline{Cost}~$\downarrow$  & All & 4813 & 4182 & 4187 & 3549 & 3571 & 2912 & 2985 & 2281    & \multicolumn{2}{c}{\multirow{2}{*}{-}} 
\\ 

\cmidrule{2-11}

& \underline{W-T}~$\uparrow$     & All & - & 9 & \textbf{13} & 7 & 3 & 3 & 2 & 1 & \multicolumn{2}{c}{}
\\ 

\midrule

\multirow{6}{*}{\rotatebox{90}{Chess Move Validity}} & \multirow{5}{*}{Acc~$\uparrow$}    & $S_1$	&  \colorbox{Mycolor1}{\textbf{43.2±7.0}} & \colorbox{Mycolor3}{\textbf{42.4±4.6}} & 41.2±9.7 & 36.8±6.4 & 27.6±4.8 & 22.0±5.3 & 20.4±4.8 & \colorbox{Mycolor2}{6.4±3.3} & 2557 & 6
\\
&	  & $S_2$	& \colorbox{Mycolor1}{\textbf{46.8±4.2}} & \colorbox{Mycolor3}{\textbf{42.8±4.2}} & 39.2±4.6 & 34.8±4.2 & 29.6±5.2 & 16.8±2.7 & 22.8±5.8 & \colorbox{Mycolor2}{8.8±3.4} & 2499 & 1           
\\
&	  & $S_3$	& \colorbox{Mycolor1}{\textbf{42.4±8.7}} & \colorbox{Mycolor3}{\textbf{38.4±9.9}} & 38.0±6.9 & 36.8±7.8 & 26.8±5.8 & 19.6±2.6 & 19.6±2.6 & \colorbox{Mycolor2}{6.0±2.8} & 2496 & 3
\\
&	  & $S_4$	&  \colorbox{Mycolor1}{\textbf{36.0±8.1}} & 32.4±4.6 & \colorbox{Mycolor3}{\textbf{34.0±5.8}} & 26.0±4.9 & 26.8±5.4 & 20.8±5.4 & 22.4±5.9 & \colorbox{Mycolor2}{11.2±2.3} & 2455 & 4
\\ 

\cmidrule{2-13} 

& \underline{Cost}~$\downarrow$  & All & 3148 & 2621 & 2585 & 2118 & 2904 & 2384 & 2393 & 1860   & \multicolumn{2}{c}{\multirow{2}{*}{-}} 
\\ 

\cmidrule{2-11}

& \underline{W-T}~$\uparrow$  & All & - & \textbf{6} & \textbf{6} & 2 & 0 & 0 & 0 & 0 & \multicolumn{2}{c}{} 
\\ 

\bottomrule

\end{tabular}
}

\caption{
The impact of eight different collaborative strategies on the performance of three datasets across distinct societies (\emph{using Qwen 72B}).
The significances test on societies and strategies are respectively shown in Table~\ref{table:qwen:sig_main_society},~\ref{table:qwen:sig_main_strategy}. 
The experiments of comparison with the single LLM agent is shown in Figure~\ref{fig:qwen:distribute}(a)-(f). 
\label{table:qwen_main}
}

\end{table*}
\definecolor{gray}{HTML}{CCCCCC}
\begin{table}[!htbp] 
\centering
\resizebox{\linewidth}{!}{
\begin{tabular}{lrrr}
\toprule
Collaborative  & \multicolumn{1}{r}{MMLU} & \multicolumn{1}{r}{MATH} & \multicolumn{1}{r}{Chess Move Validity} \\
Strategy & \multicolumn{1}{r}{p-value} & \multicolumn{1}{r}{p-value} & \multicolumn{1}{r}{p-value} \\ \midrule
$p_0p_0p_0$ & {0.654} & {0.637} & {0.162}  \\
$p_0p_0p_1$ & {0.388} & {0.649} & {0.064}  \\
$p_0p_1p_0$ & {0.841} & {0.667} & {0.445}  \\
$p_0p_1p_1$ & {0.455} & {0.567} & \colorbox{gray}{0.034}  \\
$p_1p_0p_0$ & {0.387} & {0.963} & {0.817}  \\
$p_1p_0p_1$ & {0.933} & {0.690} & {0.281}  \\
$p_1p_1p_0$ & {0.987} & {0.647} & {0.695}  \\
$p_1p_1p_1$ & {0.061} & {0.688} & \colorbox{gray}{0.048}  \\
\bottomrule
\end{tabular}
}
\caption{
One-Way ANOVA results for the impact of society on accuracy with fixed collaborative strategy, based on experiments from Table~\ref{table:qwen_main} using \emph{Qwen 72B}.
}
\label{table:qwen:sig_main_society}
\end{table}
\definecolor{gray}{HTML}{CCCCCC}
\begin{table}[!htbp] 
\centering
\resizebox{\linewidth}{!}{
\begin{tabular}{lrrr}
\toprule
 & \multicolumn{1}{r}{MMLU} & \multicolumn{1}{r}{MATH} & \multicolumn{1}{r}{Chess Move Validity} \\
Society & \multicolumn{1}{r}{p-value} & \multicolumn{1}{r}{p-value} & \multicolumn{1}{r}{p-value} \\ \midrule
$S_1$ & {0.257} & {0.418} & \colorbox{gray}{0.000} \\
$S_2$ & {0.093} & \colorbox{gray}{0.004} & \colorbox{gray}{0.000} \\
$S_3$ & \colorbox{gray}{0.004} & {0.449} & \colorbox{gray}{0.000} \\
$S_4$ & \colorbox{gray}{0.015} & {0.088} & \colorbox{gray}{0.000} \\
\bottomrule
\end{tabular}
}
\caption{
One-Way ANOVA results for the impact of collaborative strategy on accuracy with fixed society, based on experiments from Table~\ref{table:qwen_main} using \emph{Qwen 72B}.
}
\label{table:qwen:sig_main_strategy}
\end{table}
\subsection{Qwen 72B}
\label{app:backbone_qwen72b}

\begin{figure*}[!t] 
    \centering
    \scalebox{1}{
    \includegraphics[width=1\textwidth]{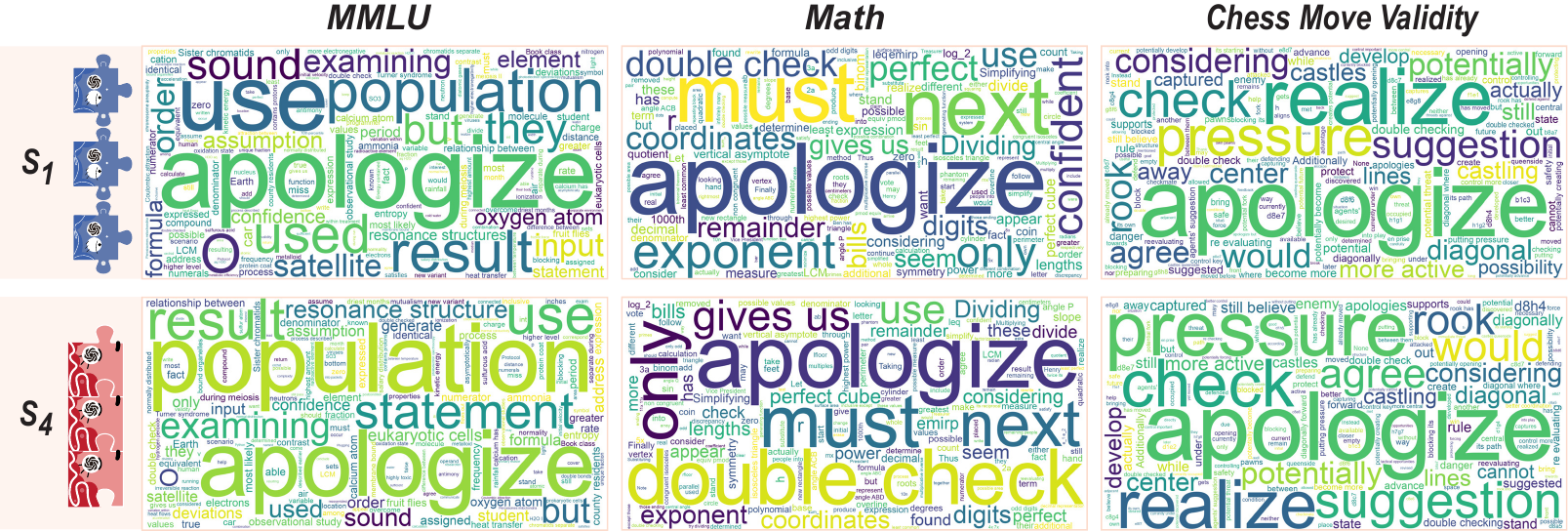}
    }
    \vspace{-4mm}
    \caption{
    Comparative word clouds on three datasets in societies $S_1$ and $S_4$, using \emph{Qwen 72B}. 
    Society $S_1$ features three overconfident agents, while society $S_4$ comprises three easy-going agents. 
    }
    \label{fig:qwen:word}
\end{figure*}

\begin{figure*}[!t] 
    \centering
    \scalebox{1}{
    \includegraphics[width=0.74\textwidth]{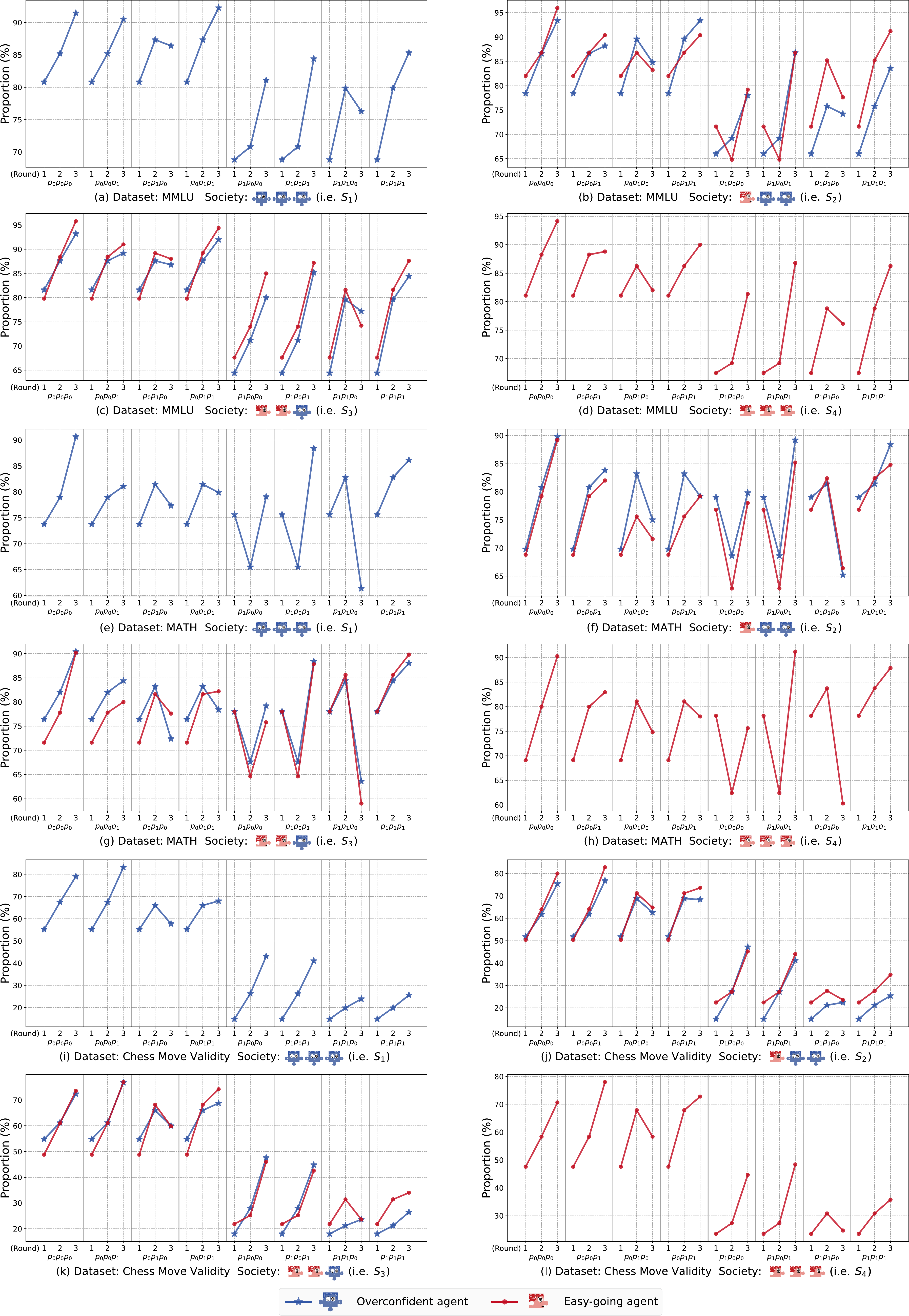}
    }
    \vspace{-4mm}
    \caption{
    Proportion of agents with different traits keeping answers in societies $S_1$ and $S_4$, using \emph{Qwen 72B}. 
    Society $S_1$ features three overconfident agents, while society $S_4$ comprises three easy-going agents. 
    }
    \label{fig:qwen:agent_answer_changing}
\end{figure*}

\begin{figure*}[!t] 
    \centering
    \scalebox{1}{
    \includegraphics[width=0.86\textwidth]{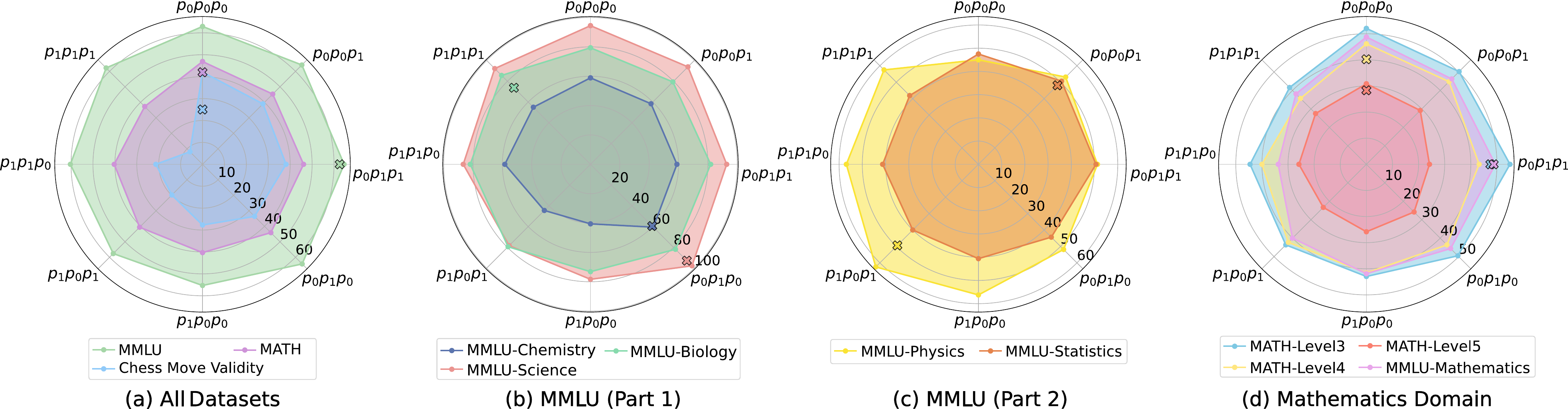}
    }
    \vspace{-3mm}
    \caption{
    Illustration of different collaborative strategies impacting accuracy diversely on the tasks considering varied \emph{subjects} and \emph{difficulty}, using \emph{Qwen 72B}. 
    The symbol `\protect\radarfork' represents that there is at least one collaborative strategy whose accuracy is better than self-consistency, while the symbol `\protect\radarprismatic' indicates that there is no collaborative strategy whose accuracy is worse than self-consistency. 
    Both of these symbols represent the accuracy of self-consistency. 
    The accuracy under each collaborative strategy is a summation within all 3-agent societies. 
    \label{fig:qwen:task_radar}
    }
\end{figure*}

\textbf{Analysis on Machine Social Collaboration.} 
We present the \textbf{main results} and \textbf{significance tests} of societies and strategies on Qwen 72B in Table~\ref{table:qwen_main},~\ref{table:qwen:sig_main_society},~\ref{table:qwen:sig_main_strategy}. 
We present the \textbf{word clouds} of Qwen 72B in Figure~\ref{fig:qwen:word}, and \textbf{proportion of agents with different traits keeping answers in different societies} on Qwen 72B in Figure~\ref{fig:qwen:agent_answer_changing}. 
Furthermore, we demonstrate that the tasks with different subjects and difficulty display varying sensitivity to collaborative strategies, as presented with \textbf{radar maps} on Qwen 72B in Figure~\ref{fig:qwen:task_radar}.

\begin{figure*}[!t] 
    \centering
    \includegraphics[width=0.8\textwidth]{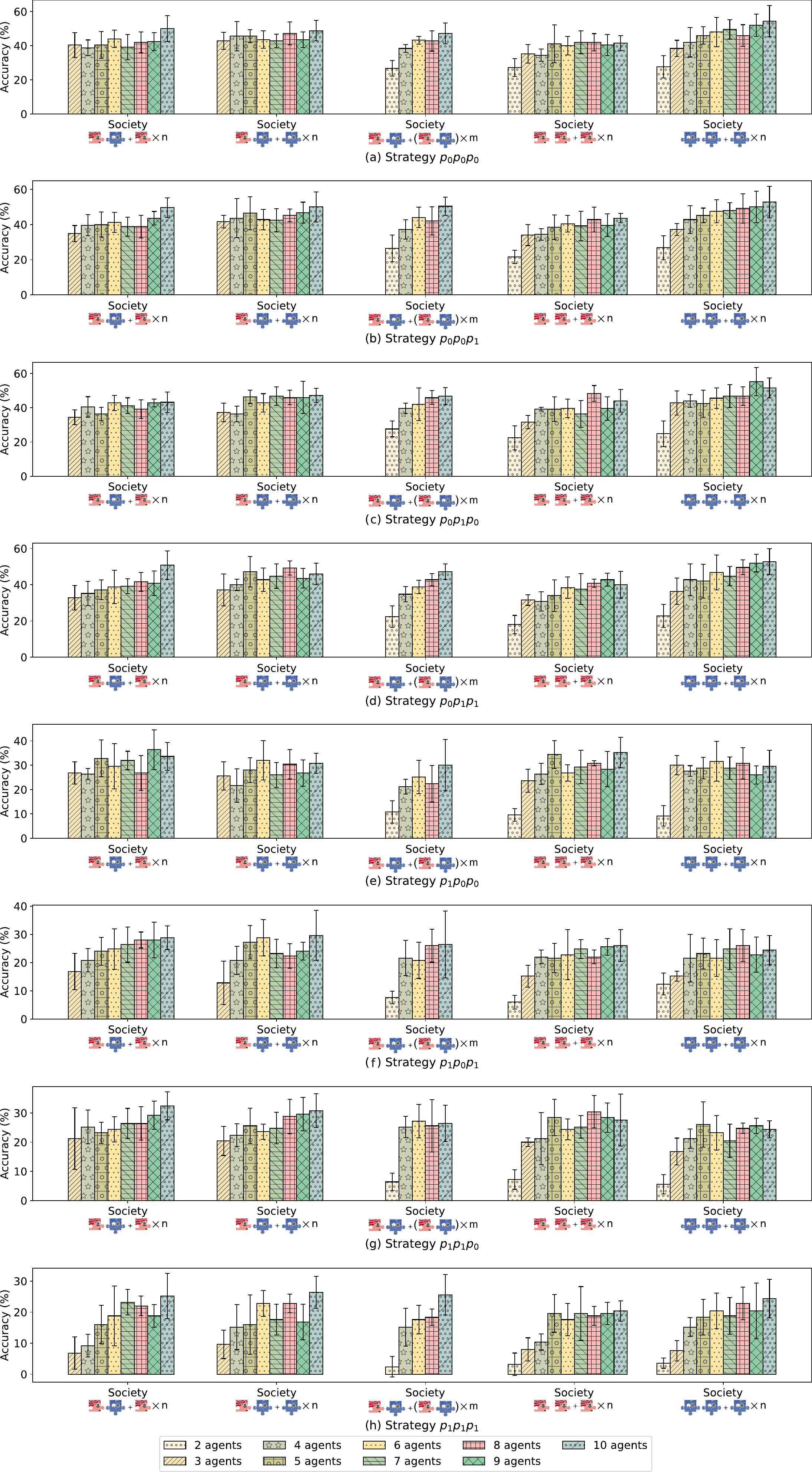}
    \vspace{-3mm}
    \caption{
    Accuracy of different numbers (2$\sim$10) of agents under different collaborative strategies, on \emph{Qwen 72B}. 
    The significance test is shown in Table~\ref{table:qwen:sig_10_agent}. 
    }
    \label{fig:qwen:agent}
\end{figure*}

\begin{figure*}[!t] 
    \centering
    \includegraphics[width=0.8\textwidth]{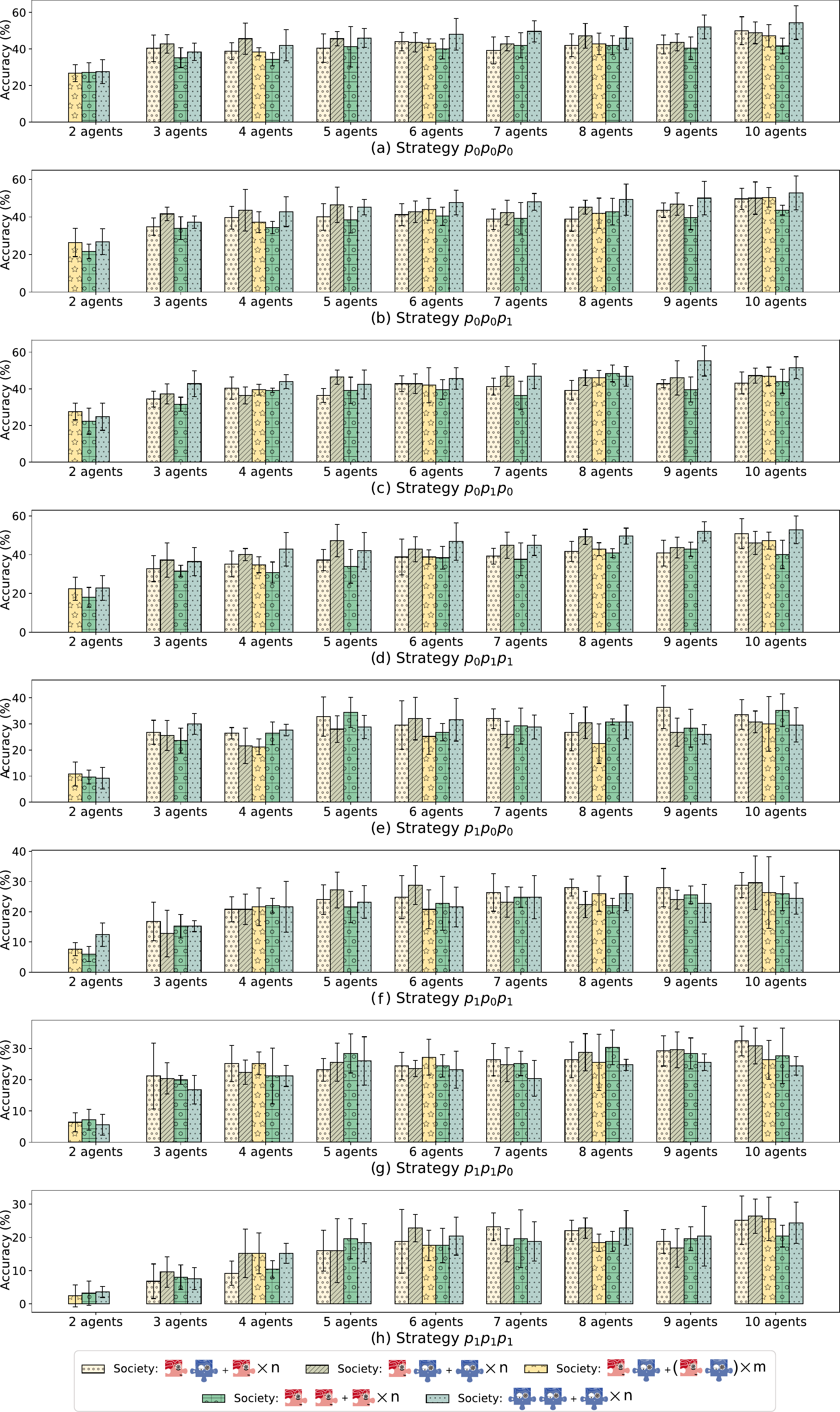}
    \vspace{-3mm}
    \caption{
    Accuracy of different societies with 2$\sim$10 agents under different collaborative strategies, on \emph{Qwen 72B}. 
    }
    \label{fig:qwen:agent_on_society}
\end{figure*}

\begin{figure*}[!t]
    \centering
    \includegraphics[width=0.85\textwidth]{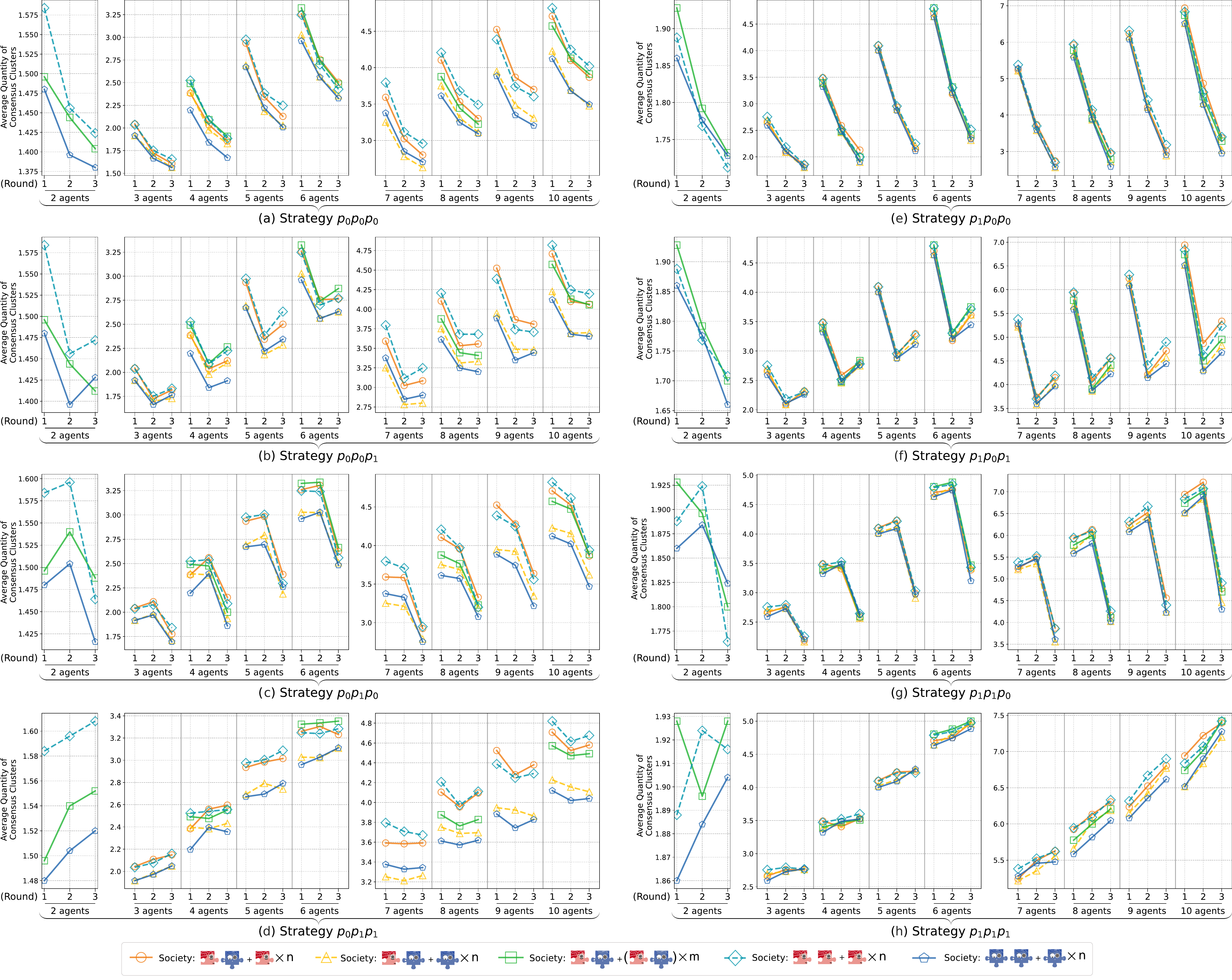}
    \vspace{-3mm}
    \caption{
    Average quantity of \emph{consensus clusters (unique answers among multiple agents)} in \emph{different societies} with 2$\sim$10 agents under each round of 3-round collaborative strategies, using \emph{Qwen 72B}. 
    }
    \label{fig:qwen:agent_10_on_societies_consensus}
\end{figure*}

\begin{figure*}[!t]
    \centering
    \includegraphics[width=0.88\textwidth]{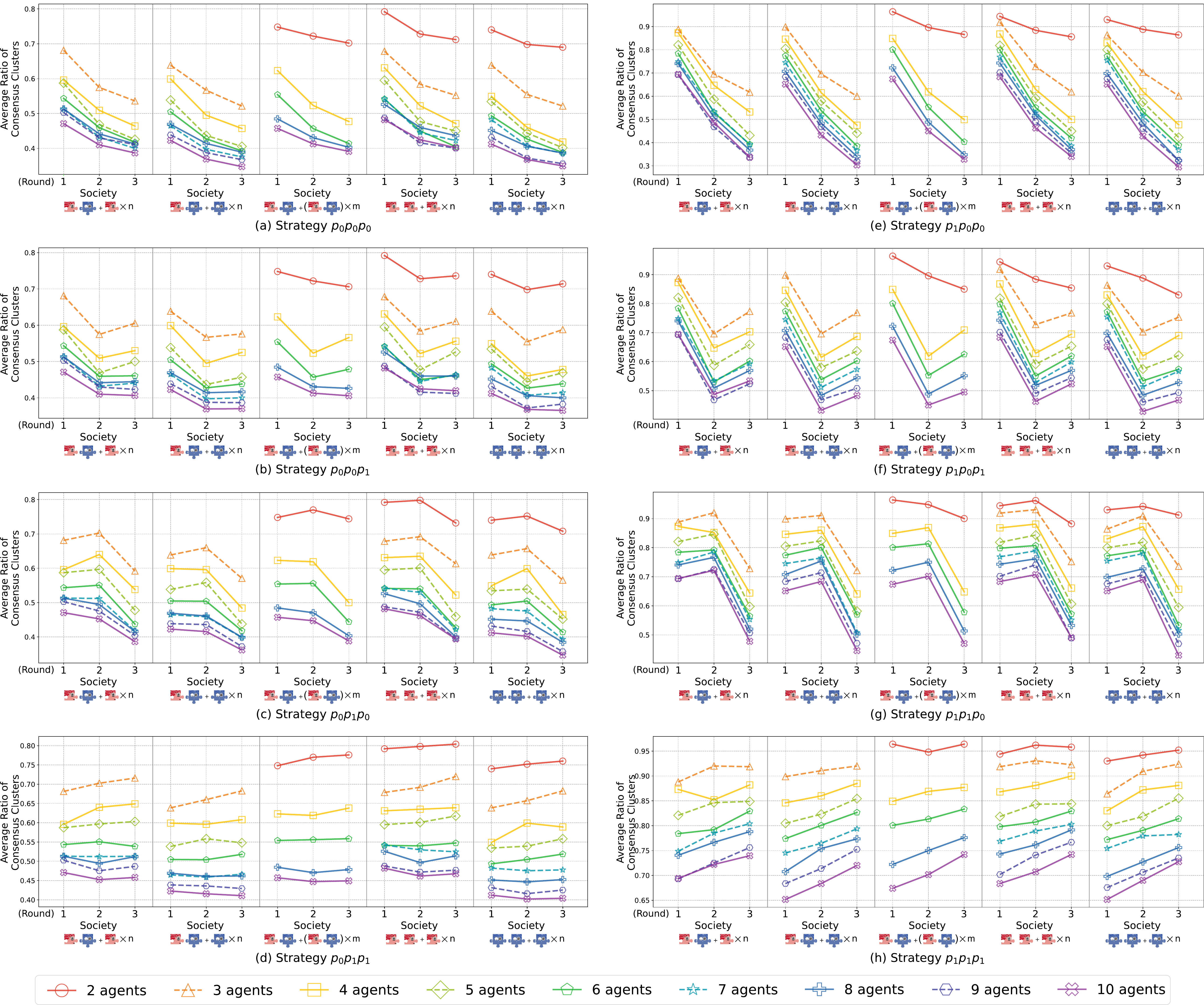}
    \vspace{-3mm}
    \caption{
    Average ratio of \emph{consensus clusters (unique answers among multiple agents)} with \emph{different numbers (2$\sim$10) of agents} under each round of 3-round collaborative strategies, using \emph{Qwen 72B}. 
    }
    \label{fig:qwen:agent_10_on_numbers_consensus}
    \vspace{-3mm}
\end{figure*}

\textbf{Analysis on Different Numbers of Agents.} 
We present the significance test for different numbers of agents with Qwen 72B in Table~\ref{table:qwen:sig_10_agent}. 
We also show the performance varying from agent numbers in Figure~\ref{fig:qwen:agent}, varying from societies containing 2$\sim$10 agents in Figure~\ref{fig:qwen:agent_on_society}. 
We also analyze the \emph{consensus reaching} with different numbers of agents, and present the results in Figure~\ref{fig:qwen:agent_10_on_societies_consensus},~\ref{fig:qwen:agent_10_on_numbers_consensus}. 

\definecolor{gray}{HTML}{CCCCCC}
\begin{table}[!htbp] 
\centering
\resizebox{\linewidth}{!}{
\begin{tabular}{lrrrrr}
\toprule
 Collaborative & \multicolumn{1}{r}{$S_1^{'}$} & \multicolumn{1}{r}{$S_2^{'}$} & \multicolumn{1}{r}{$S_3^{'}$} & \multicolumn{1}{r}{$S_4^{'}$} & \multicolumn{1}{r}{$S_5^{'}$} \\
Strategy & \multicolumn{1}{r}{p-value} & \multicolumn{1}{r}{p-value} & \multicolumn{1}{r}{p-value} & \multicolumn{1}{r}{p-value} & \multicolumn{1}{r}{p-value} \\ \midrule
$p_0p_0p_0$ & \colorbox{gray}{0.005} & \colorbox{gray}{0.001} & \colorbox{gray}{0.003} & \colorbox{gray}{0.041} & \colorbox{gray}{0.015} \\
$p_0p_0p_1$ & \colorbox{gray}{0.017} & \colorbox{gray}{0.010} & \colorbox{gray}{0.037} & \colorbox{gray}{0.001} & \colorbox{gray}{0.006} \\
$p_0p_1p_0$ & \colorbox{gray}{0.006} & \colorbox{gray}{0.016} & \colorbox{gray}{0.002} & \colorbox{gray}{0.000} & \colorbox{gray}{0.001} \\
$p_0p_1p_1$ & \colorbox{gray}{0.020} & \colorbox{gray}{0.002} & \colorbox{gray}{0.010} & \colorbox{gray}{0.001} & \colorbox{gray}{0.004} \\
$p_1p_0p_0$ & \colorbox{gray}{0.000} & \colorbox{gray}{0.005} & \colorbox{gray}{0.000} & \colorbox{gray}{0.000} & \colorbox{gray}{0.000} \\
$p_1p_0p_1$ & \colorbox{gray}{0.002} & \colorbox{gray}{0.008} & \colorbox{gray}{0.004} & \colorbox{gray}{0.000} & {0.054} \\
$p_1p_1p_0$ & \colorbox{gray}{0.003} & \colorbox{gray}{0.000} & \colorbox{gray}{0.002} & {-} & \colorbox{gray}{0.000} \\
$p_1p_1p_1$ & {0.064} & \colorbox{gray}{0.008} & \colorbox{gray}{0.005} & \colorbox{gray}{0.016} & \colorbox{gray}{0.000} \\
\bottomrule
\end{tabular}
}
\caption{
One-way ANOVA analysis of results in Figure~\ref{fig:qwen:agent} (different numbers of agents), using \emph{Qwen 72B}. 
$S_1^{'}$: One overconfident agent and the others are all easygoing. $S_2^{'}$: One easygoing agent among predominantly overconfident agents. $S_3^{'}$: Equal numbers of overconfident and easygoing agents. $S_4^{'}$: Entirely easygoing agents. $S_5^{'}$: Entirely overconfident agents. `-': It doesn't pass homogeneity test for variance. 
}
\label{table:qwen:sig_10_agent}
\end{table}

\begin{figure*}[!t]
    \centering
    \includegraphics[width=1\textwidth]{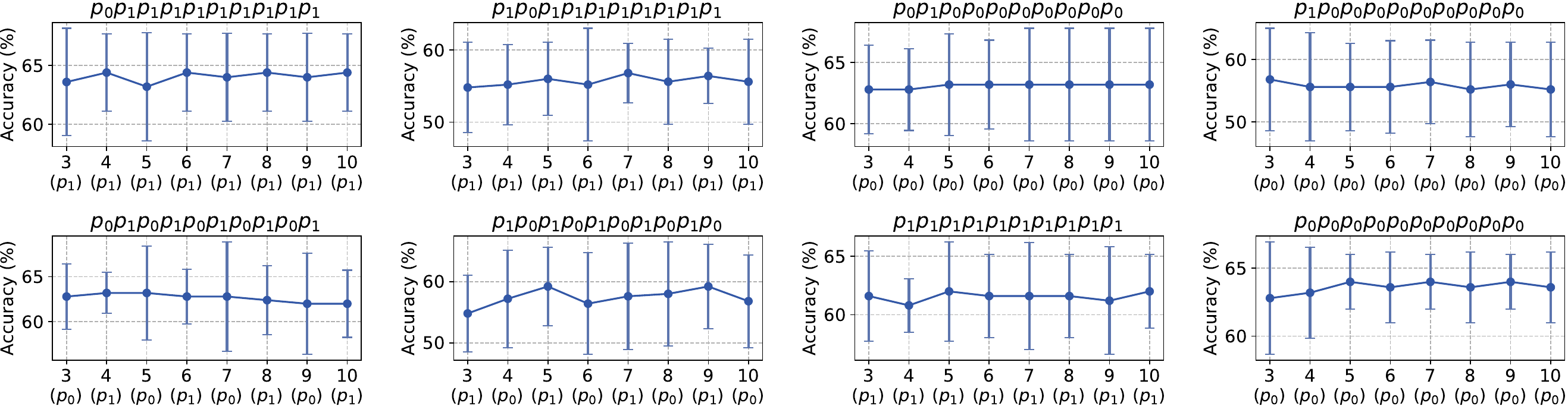}
    \vspace{-5mm}
    \caption{
    Accuracy of \emph{different (3$\sim$10) rounds of collaboration} within 3-agent society $S_2$ (1 easy-going and 2 overconfident agents) on MMLU, using \emph{Qwen 72B}. 
    The significance test is shown in Table~\ref{table:qwen:sig_10_turn}. 
    }
    \label{fig:qwen:round_10_on_mmlu}
\end{figure*}

\begin{figure*}[!t]
    \centering
    \includegraphics[width=1\textwidth]{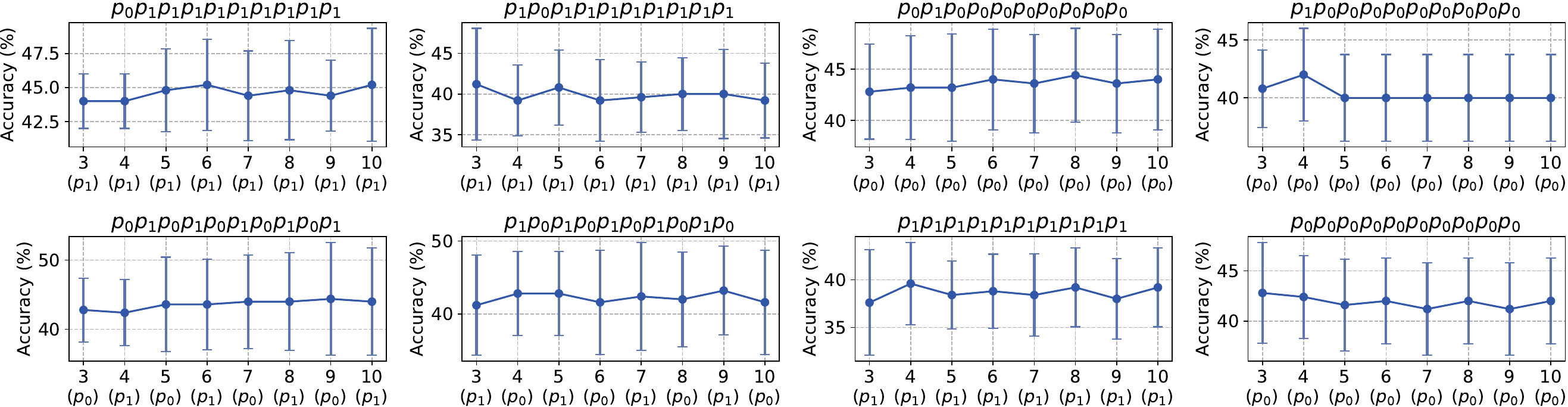}
    \vspace{-3mm}
    \caption{
    Accuracy of \emph{different (3$\sim$10) rounds of collaboration} within 3-agent society $S_2$ (1 easy-going and 2 overconfident agents) on MATH, using \emph{Qwen 72B}. 
    The significance test is shown in Table~\ref{table:qwen:sig_10_turn}. 
    }
    \label{fig:qwen:round_10_on_math}
\end{figure*}

\begin{figure*}[!t]
    \centering
    \includegraphics[width=1\textwidth]{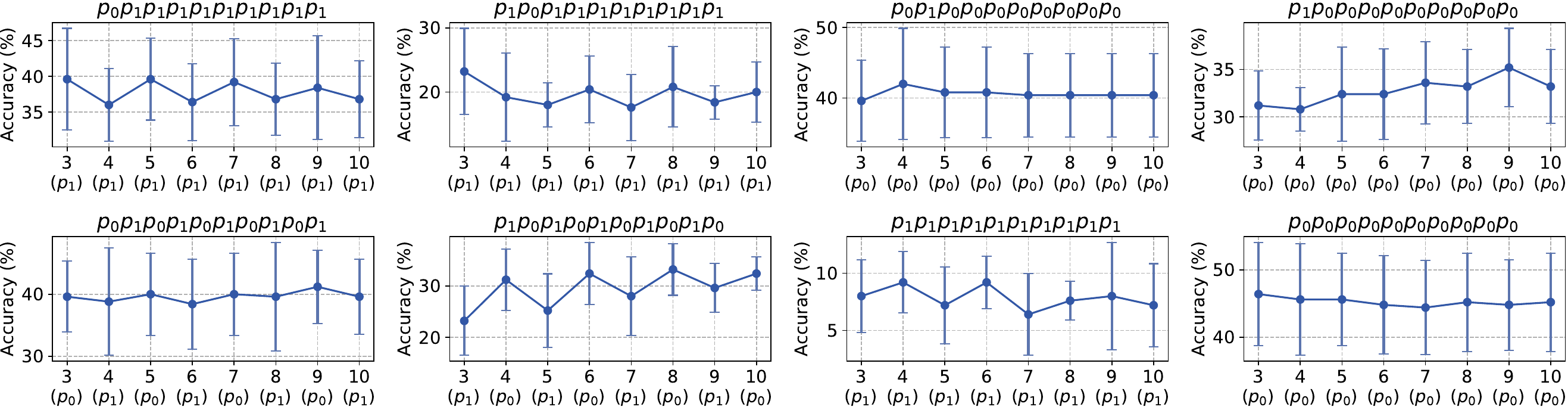}
    \vspace{-5mm}
    \caption{Accuracy of \emph{different (3$\sim$10) rounds of collaboration} within 3-agent society $S_2$ (1 easy-going and 2 overconfident agents) on Chess Move Validity, using \emph{Qwen 72B}. 
    The significance test is shown in Table~\ref{table:qwen:sig_10_turn}. 
    }
    \label{fig:qwen:round_10_on_chess}
\end{figure*}

\textbf{Analysis on Different Rounds.} 
We present the significance test for different rounds of collaboration with Qwen 72B in Table~\ref{table:qwen:sig_10_turn}. 
We also show the performance varying from collaboration rounds in Figure~\ref{fig:qwen:round_10_on_mmlu},~\ref{fig:qwen:round_10_on_math},~\ref{fig:qwen:round_10_on_chess}.


\definecolor{gray}{HTML}{CCCCCC}
\begin{table}[!htbp] 
\centering
\small
\resizebox{\linewidth}{!}{
\begin{tabular}{lrrr}
\toprule
Collaborative  & \multicolumn{1}{r}{MMLU} & \multicolumn{1}{r}{MATH} & \multicolumn{1}{r}{Chess Move Validity} \\
Strategy & \multicolumn{1}{r}{p-value} & \multicolumn{1}{r}{p-value} & \multicolumn{1}{r}{p-value} \\ \midrule
$p_0p_0p_0p_0p_0p_0p_0p_0p_0p_0$ & {0.262} & {0.987} & {0.956}  \\
$p_1p_0p_0p_0p_0p_0p_0p_0p_0p_0$ & {0.753} & {0.697} & {0.124}  \\
$p_0p_1p_0p_0p_0p_0p_0p_0p_0p_0$ & {0.914} & {0.962} & {0.386}  \\
$p_1p_0p_1p_0p_1p_0p_1p_0p_1p_0$ & {0.673} & {0.715} & {0.154}  \\
$p_0p_1p_0p_1p_0p_1p_0p_1p_0p_1$ & {0.922} & {0.987} & {0.700}  \\
$p_1p_0p_1p_1p_1p_1p_1p_1p_1p_1$ & {0.845} & {0.843} & {0.282}  \\
$p_0p_1p_1p_1p_1p_1p_1p_1p_1p_1$ & {0.928} & {0.585} & {0.583}  \\
$p_1p_1p_1p_1p_1p_1p_1p_1p_1p_1$ & {0.832} & {0.801} & {0.731}  \\
\bottomrule
\end{tabular}
}
\caption{
One-way ANOVA analysis of the results in Figure~\ref{fig:qwen:round_10_on_math},~\ref{fig:qwen:round_10_on_math},~\ref{fig:qwen:round_10_on_chess} (different rounds), using \emph{Qwen 72B}. 
}
\label{table:qwen:sig_10_turn}
\end{table}

\begin{figure}[!t] 
    \centering
    \scalebox{0.46}{
    \includegraphics[width=1\textwidth]{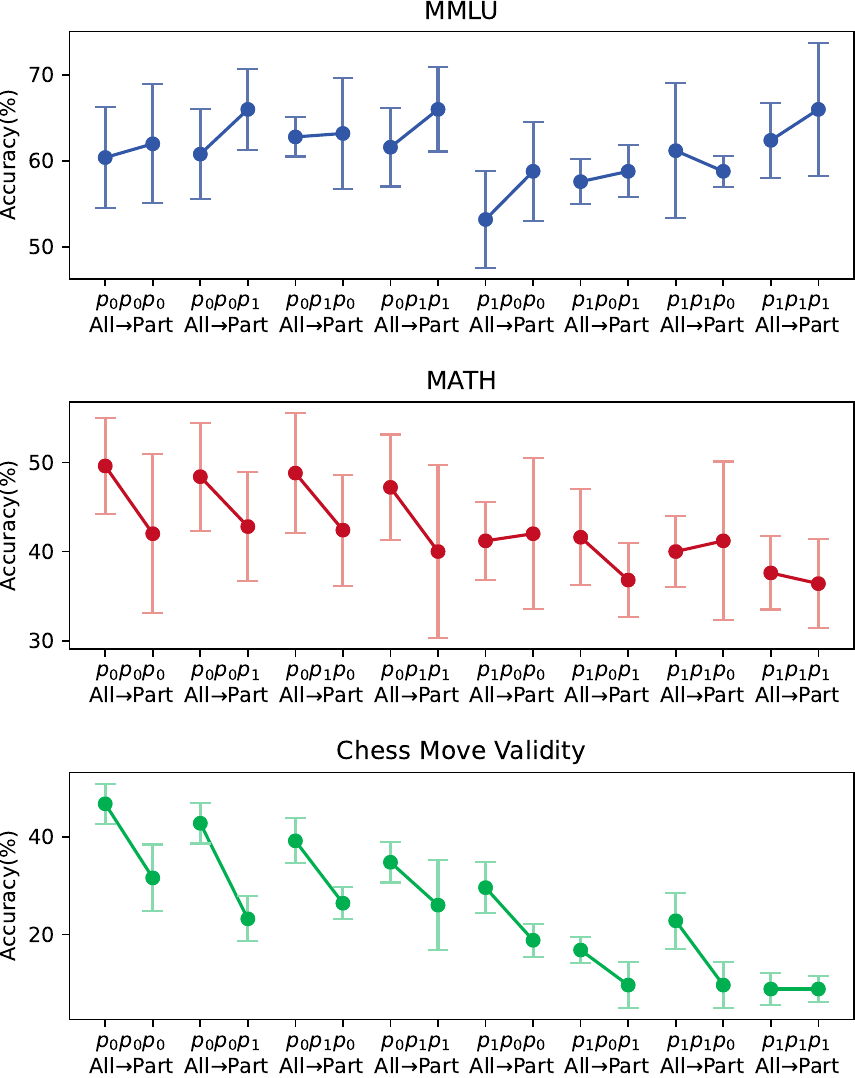}
    }
    \caption{
    The effect on the accuracy of whether all agents in society execute the same thinking pattern in one round, using \emph{Qwen 72B}.
    ``All'' and ``Part'' refers to all agents applying the same thinking pattern and different thinking patterns in one round respectively.
    The significance test is shown in Table~\ref{table:qwen:sig_strategy}.
    }
    \label{fig:qwen:strategy}
\end{figure}

\textbf{Analysis on Other Collaborative Strategies.} 
We present the significance test for other collaborative strategies (executing the same or hybrid thinking patterns in a certain round) with Qwen 72B in Table~\ref{table:qwen:sig_strategy}. 
We also show the performance varying from other strategies in Figure~\ref{fig:qwen:strategy}. 

\begin{table}[!htbp] 
\centering
\resizebox{\linewidth}{!}{
\begin{tabular}{lrrr}
\toprule
Collaborative  & \multicolumn{1}{r}{MMLU} & \multicolumn{1}{r}{MATH} & \multicolumn{1}{r}{Chess Move Validity} \\
Strategy & \multicolumn{1}{r}{p-value} & \multicolumn{1}{r}{p-value} & \multicolumn{1}{r}{p-value} \\ \midrule
$p_0p_0p_0$ & {0.704} & {0.142} & \colorbox{gray}{0.003}  \\
$p_0p_0p_1$ & {0.136} & {0.184} & \colorbox{gray}{0.000}  \\
$p_0p_1p_0$ & {0.899} & {0.157} & \colorbox{gray}{0.001}  \\
$p_0p_1p_1$ & {0.180} & {0.194} & {0.089}  \\
$p_1p_0p_0$ & {0.157} & {0.856} & \colorbox{gray}{0.004}  \\
$p_1p_0p_1$ & {0.521} & {0.152} & \colorbox{gray}{0.019}  \\
$p_1p_1p_0$ & {-} & {0.790} & \colorbox{gray}{0.004}  \\
$p_1p_1p_1$ & {0.391} & {0.688} & {1.000}  \\
\bottomrule
\end{tabular}
}
\caption{One-way ANOVA analysis of results in Figure~\ref{fig:qwen:strategy} (other collaborative strategies), \emph{using Qwen 72B}. 
`-' means it doesn't pass homogeneity test for variance. 
}
\label{table:qwen:sig_strategy}
\end{table}

\begin{figure*}[!t] 
    \centering
    \scalebox{1}{
    \includegraphics[width=0.88\textwidth]{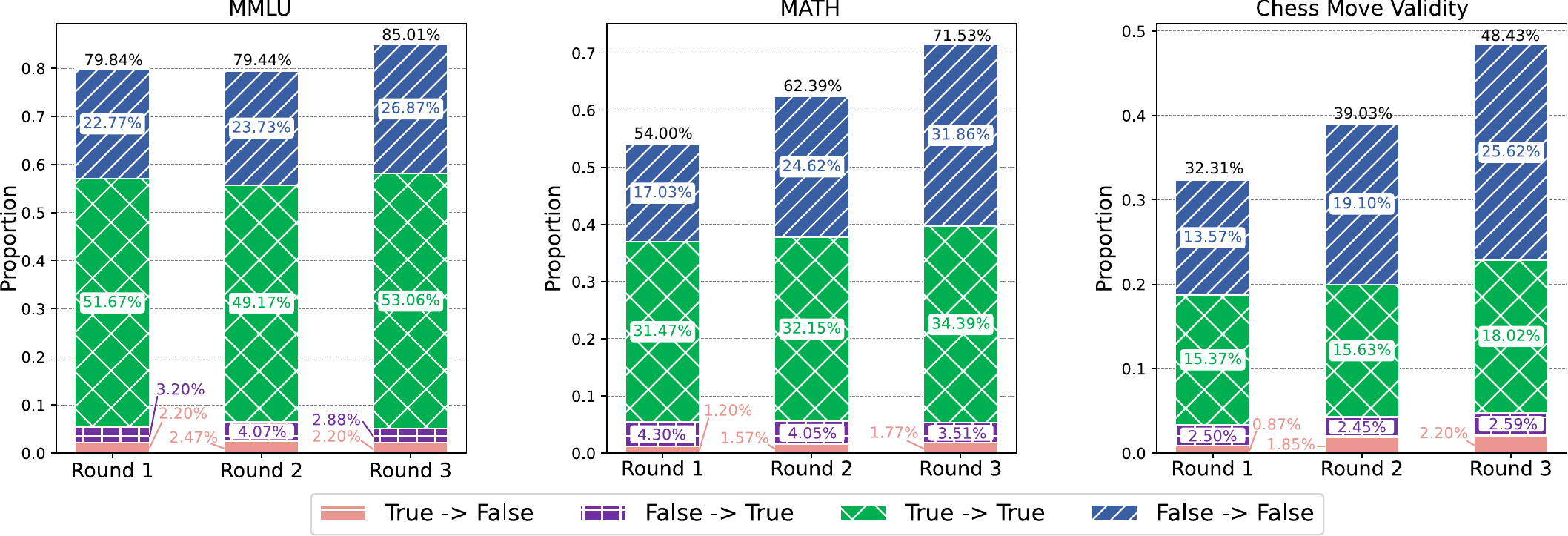}
    }
    \vspace{-2mm}
    \caption{
    Variation of answer correctness in the situation of conformity, using \emph{Qwen 72B}, where 
    \emph{conformity brings about benefits}: Ratio$($False$\to$True + True$\to$True$)$ $>$ Ratio$($True$\to$False + False$\to$False$)$; 
    \emph{conformity brings about detriments}: Ratio$($False$\to$True + True$\to$True$)$ $<$ Ratio$($True$\to$False + False$\to$False$)$. 
    }
    \label{fig:qwen:conformity}
\end{figure*}
\begin{figure*}[!htbp]
    \centering
    \scalebox{1}{
    \includegraphics[width=0.88\textwidth]{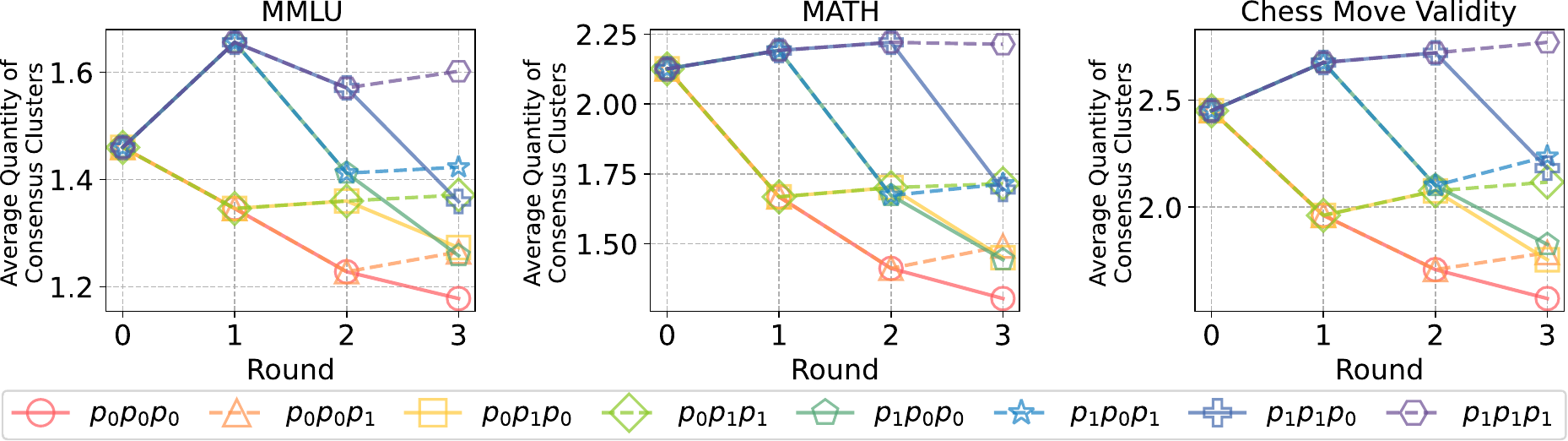}
    }
    \vspace{-2mm}
    \caption{
    Average quantity of \emph{consensus clusters (\emph{\ie}, unique answers among multiple agents)} under different rounds of collaboration with 3-round collaborative strategies,  using \emph{Qwen 72B}. 
    \emph{Smaller quantity of consensus clusters, more easier it is to reach a consensus.} 
    Round 0 is equal to self-consistency. 
    }
    \label{fig:qwen:consistent}
\end{figure*}
\begin{figure*}[!t] 
    \centering
    \scalebox{1}{
    \includegraphics[width=0.92\textwidth]{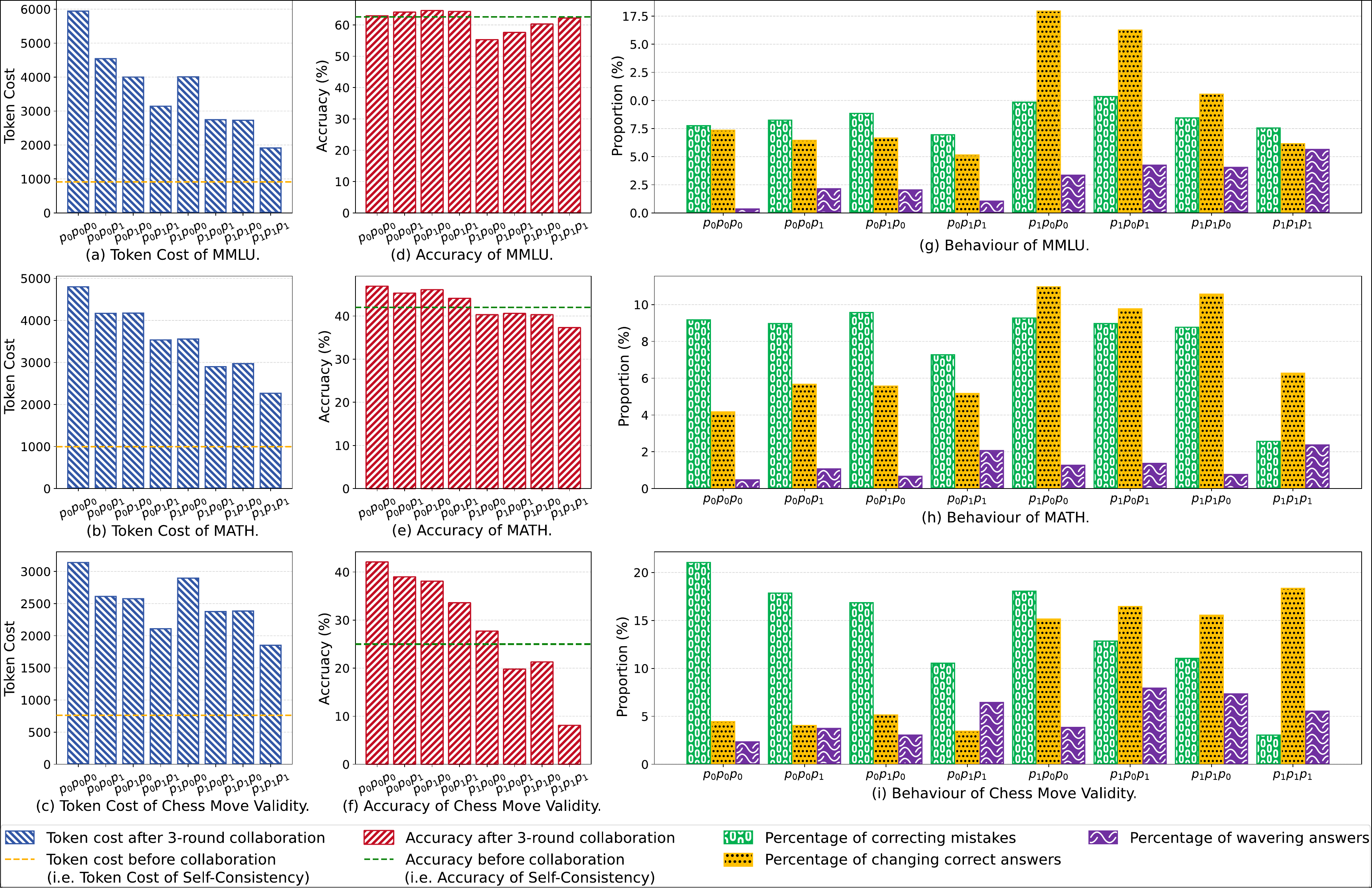}
    }
    \vspace{-2mm}
    \caption{The percentage of different behaviors under different collaborative strategies, using \emph{Qwen 72B}. 
    Figure (a-c) \& (d-f) respectively show the token cost and accuracy of different strategies before and after 3-round collaboration. 
    Figure (g-i) present the percentage of different behavioral features (mainly analyzed by the change of answer correctness) \citep{arXiv2023_Agent-BehaviorExplanation,arXiv2023_Agent-BehaviorExplaining} under different collaborative strategies. 
    All results are summarized across all societies. 
    }
    \label{fig:qwen:distribute}
\end{figure*}

\textbf{A Social Psychology View on Conformity, Consensus Reaching and Group Dynamics.} 
We then show the variation of answer correctness in the situation of conformity in Figure~\ref{fig:qwen:conformity}; and the quantity of consensus clusters among 3-agent answers in Figure~\ref{fig:qwen:consistent}. 
We present group dynamics reflected by different answer-changing behaviors on Qwen 72B in Figure~\ref{fig:qwen:distribute}.  

\clearpage

\definecolor{Mycolor1}{HTML}{BAD8F2}
\definecolor{Mycolor2}{HTML}{FAE4E3}
\definecolor{Mycolor3}{HTML}{E8F2FB}

\begin{table*}[!t] 
\small
\resizebox{\linewidth}{!}{
\begin{tabular}{c|c|c|cccccccc|cc}

\toprule

& \multirow{2}{*}{\tabincell{c}{Metric \\ (Strategy)}} & \multirow{2}{*}{Society} & \multicolumn{8}{c|}{Collaborative Strategy} & \multicolumn{2}{c}{Metric (Society)}
\\
& & & $p_0p_0p_0$ & $p_0p_0p_1$ & $p_0p_1p_0$ & $p_0p_1p_1$ & $p_1p_0p_0$ & $p_1p_0p_1$ & $p_1p_1p_0$ & $p_1p_1p_1$ & \uuline{Cost}~$\downarrow$ & \uuline{W-T}~$\uparrow$
\\

\midrule

\multirow{7}{*}{\rotatebox{90}{MMLU}}  & \multirow{5}{*}{Acc~$\uparrow$} 
& $S_1$	& 60.0±8.1 & 59.6±3.9 & \colorbox{Mycolor2}{58.4±4.3} & \colorbox{Mycolor3}{\textbf{60.0±1.4}} & 60.0±5.8 & \colorbox{Mycolor1}{\textbf{60.4±5.2}} & 59.6±2.6 & 60.0±2.0 & 4479 & 17
\\
& & $S_2$	& \colorbox{Mycolor2}{59.2±7.7} & 60.0±7.9 & 60.0±6.5 & 60.8±5.8 & 61.2±3.6 & \colorbox{Mycolor1}{\textbf{62.8±5.4}} & \colorbox{Mycolor1}{\textbf{62.8±5.4}} & \colorbox{Mycolor3}{\textbf{61.2±2.7}} & 4475 & 27
\\
& & $S_3$	& 62.4±5.2 & \colorbox{Mycolor3}{\textbf{63.6±4.3}} & \colorbox{Mycolor1}{\textbf{65.2±3.0}} & \colorbox{Mycolor1}{\textbf{65.2±3.0}} & \colorbox{Mycolor2}{59.2±4.4} & 61.2±4.2 & 61.6±2.6 & 59.6±3.6 & 4489 & 18
\\
& & $S_4$	& 60.0±3.7 & 62.4±3.6 & \colorbox{Mycolor3}{\textbf{63.2±3.4}} & 62.8±2.7 & \colorbox{Mycolor2}{60.0±5.1} & 60.4±5.5 & \colorbox{Mycolor1}{\textbf{64.8±5.8}} & 62.0±6.6 & 4396 & 25
\\ 

\cmidrule{2-13} 

& \underline{Cost}~$\downarrow$  & All   & 6891 & 5371 & 4871 & 3944 & 4996 & 3594 & 3495 & 2516   & \multicolumn{2}{c}{\multirow{2}{*}{-}} \\ \cmidrule{2-11}
& \underline{W-T}~$\uparrow$     & All   & -     & 14 & \textbf{15} & 14 & 9 & 11 & 13 & 11 & \multicolumn{2}{c}{}
\\ 

\midrule

\multirow{7}{*}{\rotatebox{90}{MATH}}  & \multirow{5}{*}{Acc~$\uparrow$}    & $S_1$	& 30.4±3.3 & \colorbox{Mycolor1}{\textbf{36.0±1.4}} & \colorbox{Mycolor3}{\textbf{33.6±2.2}} & 32.8±4.2 & 31.2±3.4 & 30.4±2.6 & 30.8±2.3 & \colorbox{Mycolor2}{27.6±1.7} & 5362 & 23
\\
&	  & $S_2$	& 31.6±6.1 & 29.2±5.4 & 30.4±6.8 & 28.0±3.7 & \colorbox{Mycolor1}{\textbf{32.4±3.6}} & 29.2±3.9 & \colorbox{Mycolor3}{\textbf{32.0±6.0}} & \colorbox{Mycolor2}{27.6±3.0} & 5369 & 14
\\
&	  & $S_3$	& 32.4±6.7 & \colorbox{Mycolor3}{\textbf{32.8±7.8}} & \colorbox{Mycolor1}{\textbf{34.8±4.8}} & 32.0±4.7 & 30.8±4.2 & 28.8±4.2 & 30.8±2.3 & \colorbox{Mycolor2}{24.8±3.9} & 5343 & 18
\\
&	  & $S_4$	& \colorbox{Mycolor1}{\textbf{32.0±4.7}} & 31.2±2.7 & 31.2±5.2 & \colorbox{Mycolor3}{\textbf{32.0±5.1}} & 29.2±4.4 & 30.0±7.2 & 31.2±1.1 & \colorbox{Mycolor2}{27.2±3.4} & 5238 & 18
\\ 

\cmidrule{2-13} 

& \underline{Cost}~$\downarrow$  & All & 6630 & 5814 & 6116 & 5042 & 5915 & 4745 & 4818 & 3540   & \multicolumn{2}{c}{\multirow{2}{*}{-}} 
\\ 

\cmidrule{2-11}

& \underline{W-T}~$\uparrow$     & All & - & 12 & 13 & 9 & \textbf{14} & 11 & 10 & 4 & \multicolumn{2}{c}{}
\\ 

\midrule

\multirow{6}{*}{\rotatebox{90}{Chess Move Validity}} & \multirow{5}{*}{Acc~$\uparrow$}    & $S_1$	& \colorbox{Mycolor1}{\textbf{22.8±2.7}} & \colorbox{Mycolor3}{\textbf{21.6±3.3}} & 21.2±5.6 & 20.8±3.0 & 18.8±5.4 & 18.8±4.6 & \colorbox{Mycolor2}{17.6±7.0} & 18.8±1.1 & 2300 & 9
\\
&	  & $S_2$	& \colorbox{Mycolor1}{\textbf{22.0±5.7}} & 18.0±2.8 & 18.8±3.4 & 16.4±2.6 & \colorbox{Mycolor3}{\textbf{22.0±8.4}} & 18.8±4.8 & \colorbox{Mycolor2}{16.0±2.8} & 16.0±0.0 & 2280 & 10
\\
&	  & $S_3$	& \colorbox{Mycolor1}{\textbf{21.2±2.7}} & 20.0±3.2 & 18.0±2.5 & 18.0±2.5 & \colorbox{Mycolor3}{\textbf{20.0±2.8}} & 18.8±3.0 & 16.4±4.6 & \colorbox{Mycolor2}{15.6±1.7} & 2269 & 9
\\
&	  & $S_4$	&  18.0±3.7 & \colorbox{Mycolor2}{16.4±3.9} & 19.2±4.6 & 16.4±2.6 & 20.0±1.4 & \colorbox{Mycolor1}{\textbf{20.8±3.6}} & \colorbox{Mycolor3}{\textbf{20.4±3.9}} & 18.8±2.3 & 2253 & 23
\\ 

\cmidrule{2-13} 

& \underline{Cost}~$\downarrow$  & All & 2956 & 2458 & 2396 & 1973 & 2630 & 2063 & 2083 & 1644  & \multicolumn{2}{c}{\multirow{2}{*}{-}} 
\\ 

\cmidrule{2-11}

& \underline{W-T}~$\uparrow$     & All                      & -        & 7 & 8 & 6 & 9 & \textbf{10} & 6 & 5 & \multicolumn{2}{c}{} 
\\ 

\bottomrule

\end{tabular}
}

\caption{
The impact of eight different collaborative strategies on the performance of three datasets across distinct societies (\emph{using Mixtral-8$\times$7B}).
The significances test on societies and strategies are respectively shown in Table~\ref{table:mixtral:sig_main_society},~\ref{table:mixtral:sig_main_strategy}. 
The experiments of comparison with the single LLM agent is shown in Figure~\ref{fig:mixtral:distribute}(a)-(f). 
\label{table:mixtral_main}
}

\end{table*}
\definecolor{gray}{HTML}{CCCCCC}
\begin{table}[!htbp] 
\centering
\resizebox{\linewidth}{!}{
\begin{tabular}{lrrr}
\toprule
Collaborative  & \multicolumn{1}{r}{MMLU} & \multicolumn{1}{r}{MATH} & \multicolumn{1}{r}{Chess Move Validity} \\
Strategy & \multicolumn{1}{r}{p-value} & \multicolumn{1}{r}{p-value} & \multicolumn{1}{r}{p-value} \\ \midrule
$p_0p_0p_0$ & {0.873} & {0.941} & {0.261}  \\
$p_0p_0p_1$ & {0.578} & {0.216} & {0.109}  \\
$p_0p_1p_0$ & {0.114} & {0.500} & {0.666}  \\
$p_0p_1p_1$ & {0.142} & {0.347} & {0.062}  \\
$p_1p_0p_0$ & {0.930} & {0.638} & {0.809}  \\
$p_1p_0p_1$ & {0.863} & {0.949} & {0.825}  \\
$p_1p_1p_0$ & {0.325} & {-} & {0.485}  \\
$p_1p_1p_1$ & {0.785} & {0.438} & \colorbox{gray}{0.004}  \\
\bottomrule
\end{tabular}
}
\caption{
One-Way ANOVA results for the impact of society on accuracy with fixed collaborative strategy, based on experiments from Table~\ref{table:mixtral_main} using \emph{Mixtral 8$\times$7B}. `-': It doesn’t pass homogeneity test for variance.
}
\label{table:mixtral:sig_main_society}
\end{table}
\definecolor{gray}{HTML}{CCCCCC}
\begin{table}[!htbp] 
\centering
\resizebox{\linewidth}{!}{
\begin{tabular}{lrrr}
\toprule
 & \multicolumn{1}{r}{MMLU} & \multicolumn{1}{r}{MATH} & \multicolumn{1}{r}{Chess Move Validity} \\
Society & \multicolumn{1}{r}{p-value} & \multicolumn{1}{r}{p-value} & \multicolumn{1}{r}{p-value} \\ \midrule
$S_1$ & {0.999} & \colorbox{gray}{0.002} & {0.585} \\
$S_2$ & {0.970} & {0.693} & {0.202} \\
$S_3$ & {0.129} & {0.127} & {0.078} \\
$S_4$ & {0.706} & {0.714} & {0.300} \\
\bottomrule
\end{tabular}
}
\caption{
One-Way ANOVA results for the impact of collaborative strategy on accuracy with fixed society, based on experiments from Table~\ref{table:mixtral_main} using \emph{Mixtral 8$\times$7B}.
}
\label{table:mixtral:sig_main_strategy}
\end{table}

\subsection{Mixtral 8$\times$7B} 
\label{app:backbone_mixtral8x7b}

\begin{figure*}[!t] 
    \centering
    \scalebox{1}{
    \includegraphics[width=1\textwidth]{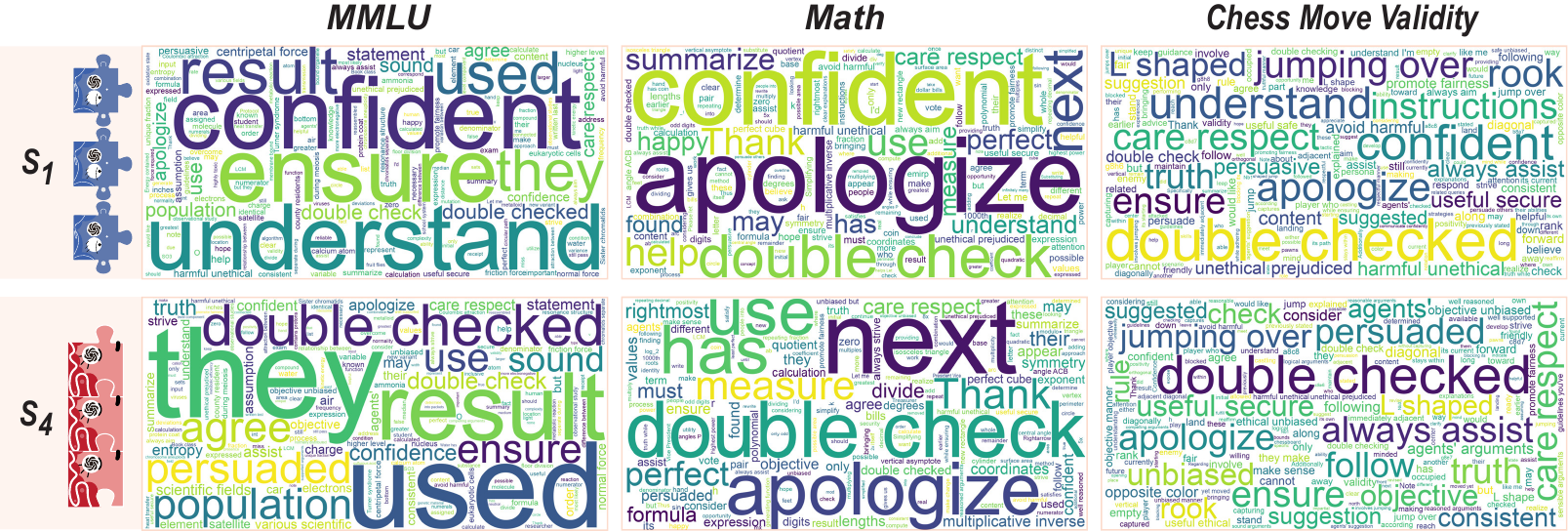}
    }
    \vspace{-4mm}
    \caption{
    Comparative word clouds on three datasets in societies $S_1$ and $S_4$, using \emph{Mixtral-8$\times$7B}. 
    Society $S_1$ features three overconfident agents, while society $S_4$ comprises three easy-going agents. 
    }
    \label{fig:mixtral:word}
\end{figure*}

\begin{figure*}[!t] 
    \centering
    \scalebox{1}{
    \includegraphics[width=0.74\textwidth]{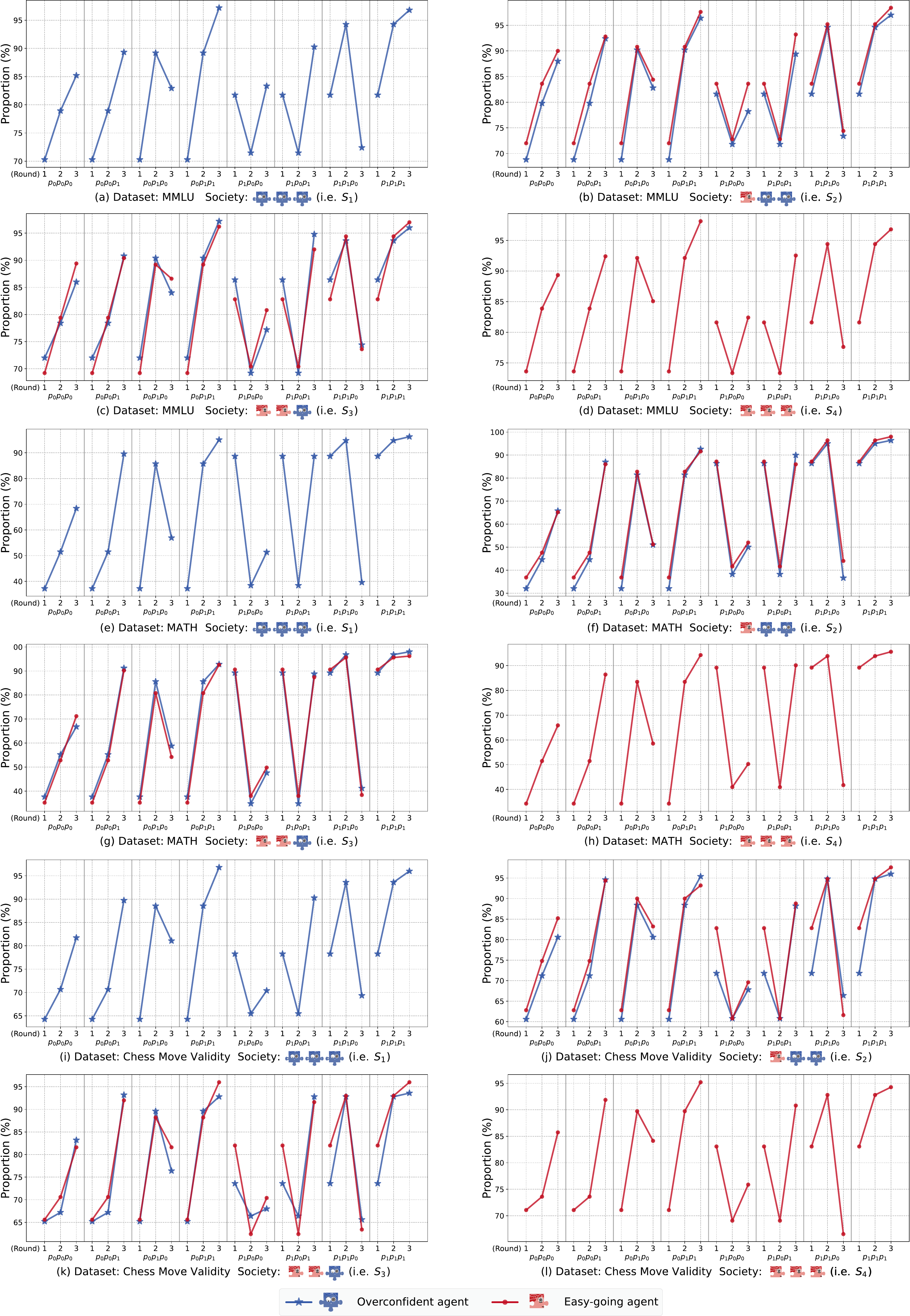}
    }
    \vspace{-4mm}
    \caption{
    Proportion of agents with different traits keeping answers in societies $S_1$ and $S_4$, using \emph{Mixtral-8$\times$7B}. 
    Society $S_1$ features three overconfident agents, while society $S_4$ comprises three easy-going agents. 
    }
    \label{fig:mixtral:agent_answer_changing}
\end{figure*}

\begin{figure*}[!t] 
    \centering
    \scalebox{1}{
    \includegraphics[width=0.86\textwidth]{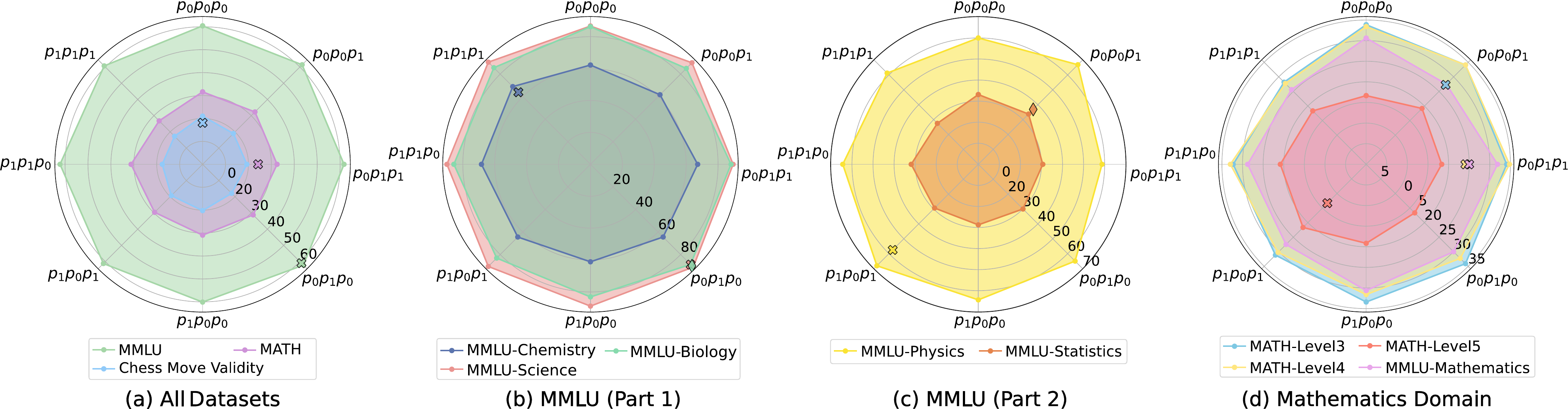}
    }
    \vspace{-3mm}
    \caption{
    Illustration of different collaborative strategies impacting accuracy diversely on the tasks considering varied \emph{subjects} and \emph{difficulty}, using \emph{Mixtral-8$\times$7B}. 
    The symbol `\protect\radarfork' represents that there is at least one collaborative strategy whose accuracy is better than self-consistency, while the symbol `\protect\radarprismatic' indicates that there is no collaborative strategy whose accuracy is worse than self-consistency. 
    Both of these symbols represent the accuracy of self-consistency. 
    The accuracy under each collaborative strategy is a summation within all 3-agent societies. 
    \label{fig:mixtral:task_radar}
    }
\end{figure*}

\textbf{Analysis on Machine Social Collaboration.} 
We present the \textbf{main results} and \textbf{significance tests} of societies and strategies on Mixtral 8$\times$7B in Table~\ref{table:mixtral_main},~\ref{table:mixtral:sig_main_society},~\ref{table:mixtral:sig_main_strategy}. 
We present the word clouds of Mixtral 8$\times$7B in Figure~\ref{fig:mixtral:word}, and the proportion of agents with different traits keepging answers in different societies on Mixtral 8$\times$7B in Figure~\ref{fig:mixtral:agent_answer_changing}. 
Furthermore, we demonstrate that the tasks with different subjects and difficulty display varying sensitivity to collaborative strategies, as presented with \textbf{radar maps} on Mixtral 8$\times$7B in Figure~\ref{fig:mixtral:task_radar}. 


\begin{figure*}[!t] 
    \centering
    \includegraphics[width=0.8\textwidth]{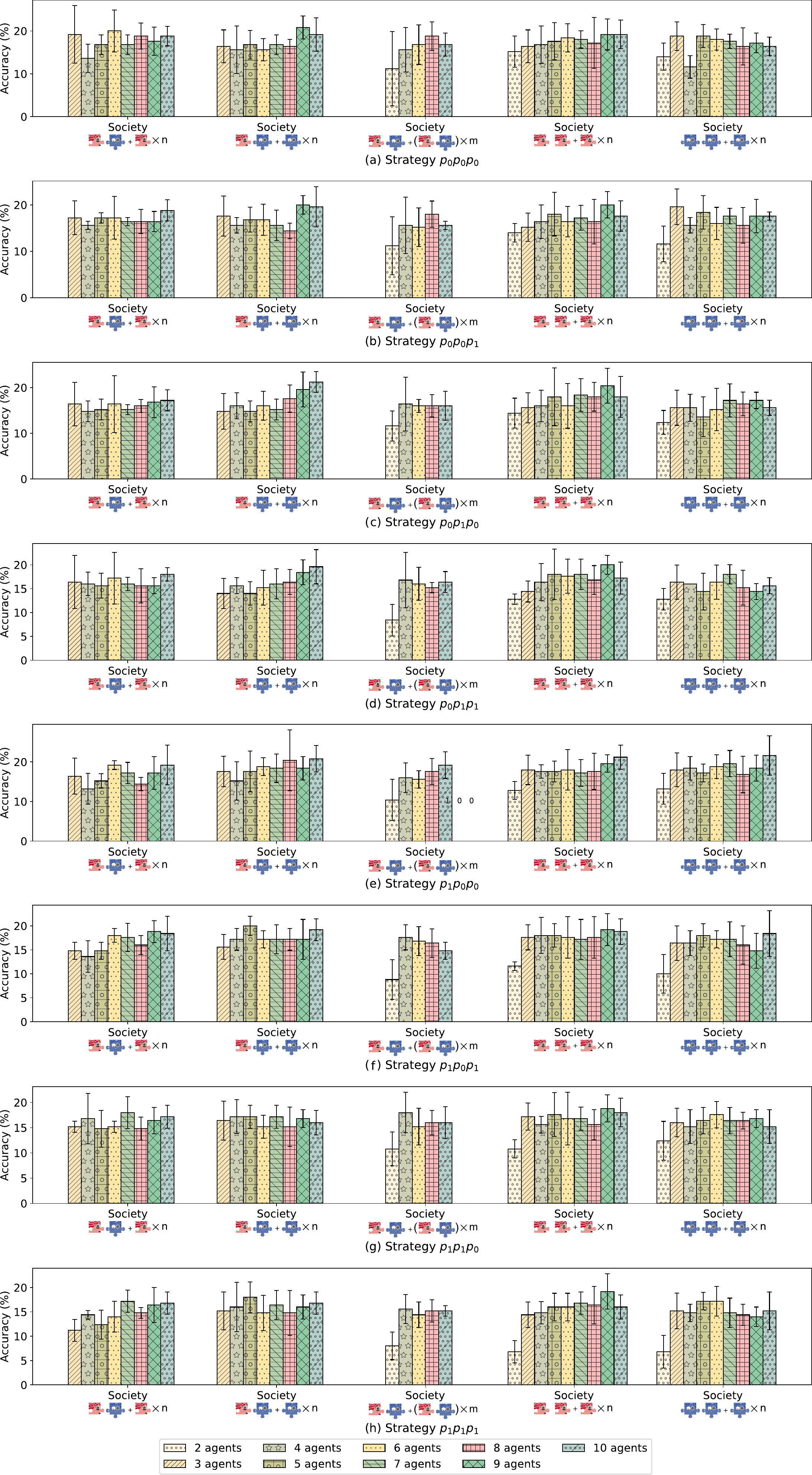}
    \vspace{-3mm}
    \caption{
    Accuracy of different numbers (2$\sim$10) of agents under different collaborative strategies, on \emph{Mixtral-8$\times$7B}. 
    The significance test is shown in Table~\ref{table:mixtral:sig_10_agent}.
    }
    \label{fig:mixtral:agent}
\end{figure*}

\begin{figure*}[!t] 
    \centering
    \includegraphics[width=0.8\textwidth]{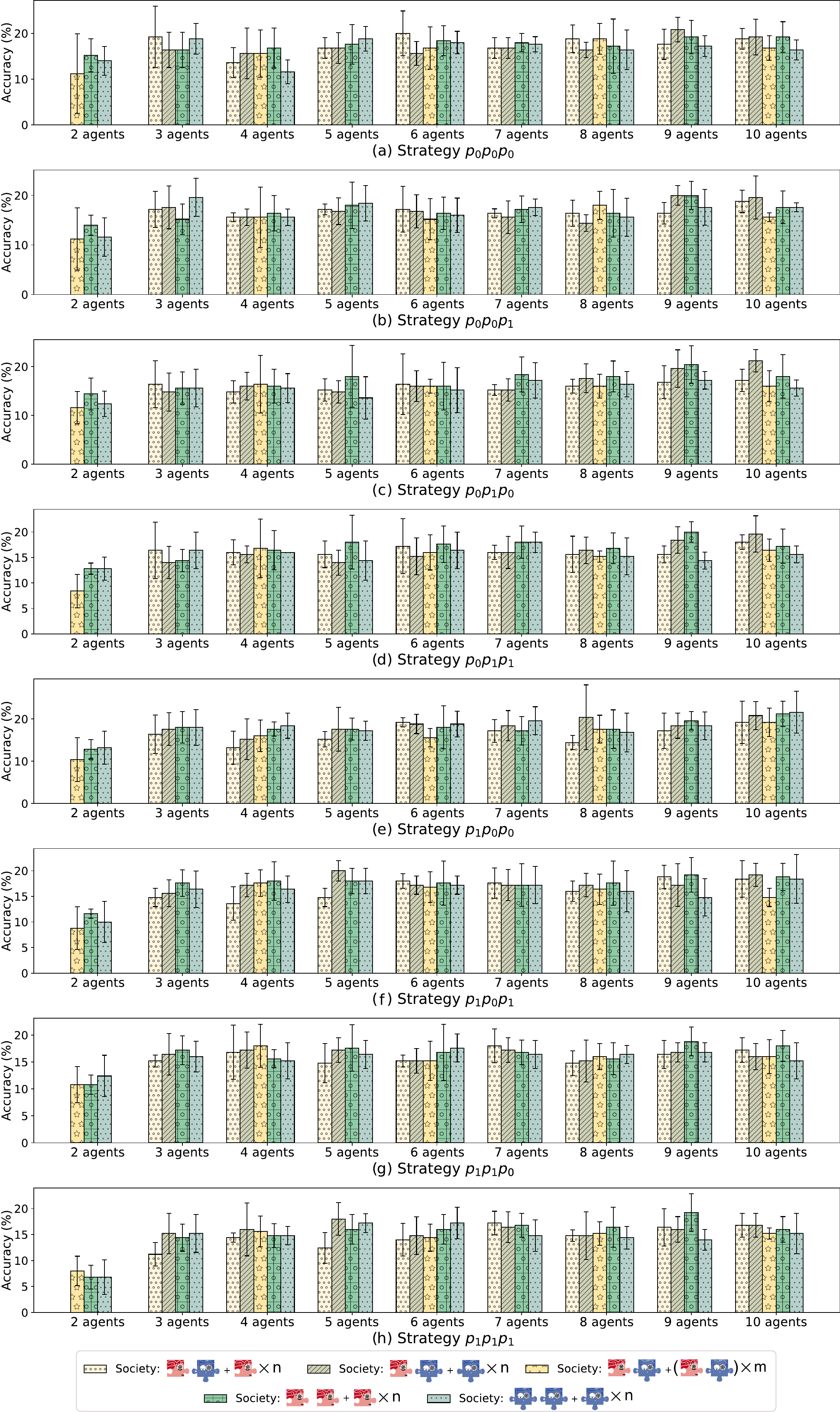}
    \vspace{-3mm}
    \caption{
    Accuracy of different societies with 2$\sim$10 agents under different collaborative strategies, on \emph{Mixtral-8$\times$7B}. 
    }
    \label{fig:mixtral:agent_on_society}
\end{figure*}

\begin{figure*}[!t]
    \centering
    \includegraphics[width=0.85\textwidth]{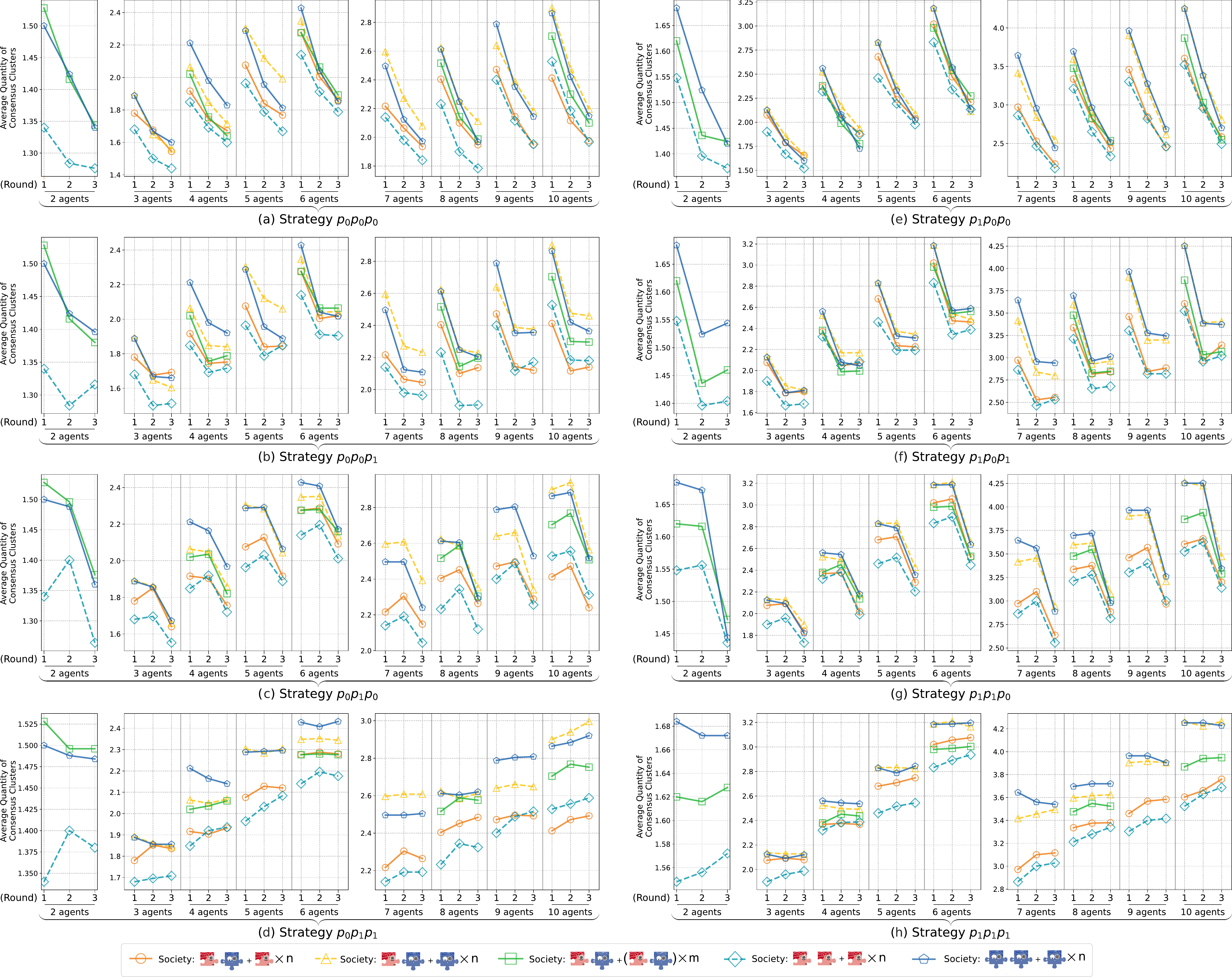}
    \vspace{-3mm}
    \caption{
    Average quantity of \emph{consensus clusters (unique answers among multiple agents)} in \emph{different societies} with 2$\sim$10 agents under each round of 3-round collaborative strategies, using \emph{Mixtral-8$\times$7B}. 
    }
    \label{fig:mixtral:agent_10_on_societies_consensus}
\end{figure*}

\begin{figure*}[!t]
    \centering
    \includegraphics[width=0.88\textwidth]{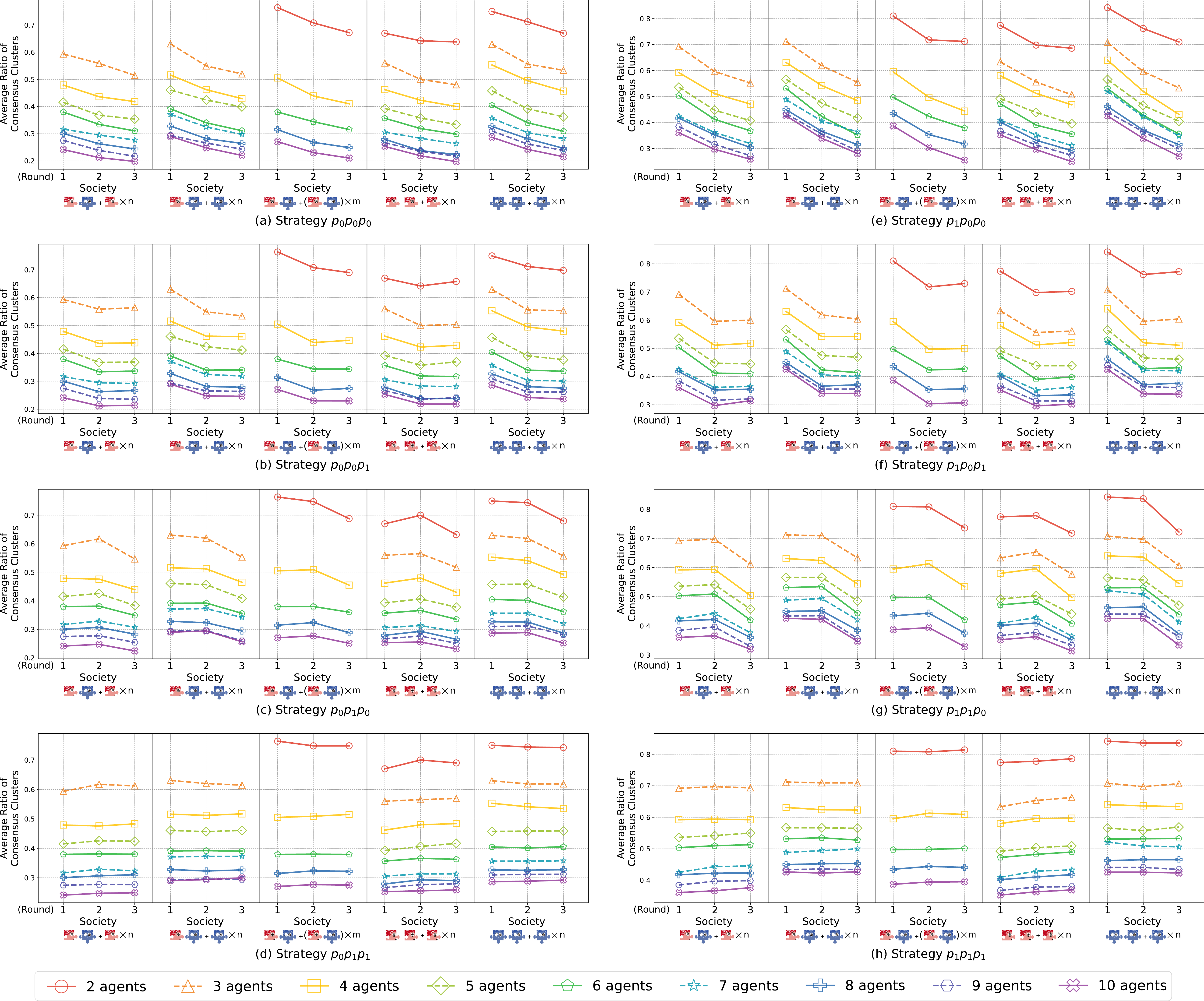}
    \vspace{-3mm}
    \caption{
    Average ratio of \emph{consensus clusters (unique answers among multiple agents)} with \emph{different numbers (2$\sim$10) of agents} under each round of 3-round collaborative strategies, using \emph{Mixtral-8$\times$7B}. 
    }
    \label{fig:mixtral:agent_10_on_numbers_consensus}
    \vspace{-3mm}
\end{figure*}

\textbf{Analysis on Different Numbers of Agents.} 
We present the significance test for different numbers of agents with Mixtral 8$\times$7B in Table~\ref{table:mixtral:sig_10_agent}. 
We also show the performance varying from agent numbers in Figure~\ref{fig:mixtral:agent}, varying from societies containing 2$\sim$10 agents in Figure~\ref{fig:mixtral:agent_on_society}. 
We also analyze the \emph{consensus reaching} with different numbers of agents, and present the results in Figure~\ref{fig:mixtral:agent_10_on_societies_consensus},~\ref{fig:mixtral:agent_10_on_numbers_consensus}. 

\definecolor{gray}{HTML}{CCCCCC}
\begin{table}[!htbp] 
\centering
\resizebox{\linewidth}{!}{
\begin{tabular}{lrrrrr}
\toprule
 Collaborative & \multicolumn{1}{r}{$S_1^{'}$} & \multicolumn{1}{r}{$S_2^{'}$} & \multicolumn{1}{r}{$S_3^{'}$} & \multicolumn{1}{r}{$S_4^{'}$} & \multicolumn{1}{r}{$S_5^{'}$} \\
Strategy & \multicolumn{1}{r}{p-value} & \multicolumn{1}{r}{p-value} & \multicolumn{1}{r}{p-value} & \multicolumn{1}{r}{p-value} & \multicolumn{1}{r}{p-value} \\ \midrule
$p_0p_0p_0$ & {0.188} & {0.406} & {0.235} & {0.805} & \colorbox{gray}{0.009} \\
$p_0p_0p_1$ & {0.106} & {0.112} & {0.238} & {0.459} & \colorbox{gray}{0.008} \\
$p_0p_1p_0$ & {0.142} & {0.145} & {0.227} & {0.739} & {0.227} \\
$p_0p_1p_1$ & \colorbox{gray}{0.013} & \colorbox{gray}{0.004} & \colorbox{gray}{0.035} & {0.138} & {0.075} \\
$p_1p_0p_0$ & {0.159} & {0.082} & {0.105} & \colorbox{gray}{0.018} & {0.088} \\
$p_1p_0p_1$ & \colorbox{gray}{0.029} & \colorbox{gray}{0.003} & \colorbox{gray}{0.002} & \colorbox{gray}{0.004} & \colorbox{gray}{0.018} \\
$p_1p_1p_0$ & {0.051} & \colorbox{gray}{0.028} & \colorbox{gray}{0.010} & \colorbox{gray}{0.001} & {0.247} \\
$p_1p_1p_1$ & \colorbox{gray}{0.002} & \colorbox{gray}{0.016} & \colorbox{gray}{0.003} & \colorbox{gray}{0.000} & \colorbox{gray}{0.001} \\
\bottomrule
\end{tabular}
}
\caption{
One-way ANOVA analysis of results in Figure~\ref{fig:mixtral:agent} (different numbers of agents), using \emph{Mixtral 8$\times$7B}. 
$S_1^{'}$: One overconfident agent and the others are all easygoing. $S_2^{'}$: One easygoing agent among predominantly overconfident agents. $S_3^{'}$: Equal numbers of overconfident and easygoing agents. $S_4^{'}$: Entirely easygoing agents. $S_5^{'}$: Entirely overconfident agents. 
}
\label{table:mixtral:sig_10_agent}
\end{table}

\begin{figure*}[!t]
    \centering
    \includegraphics[width=1\textwidth]{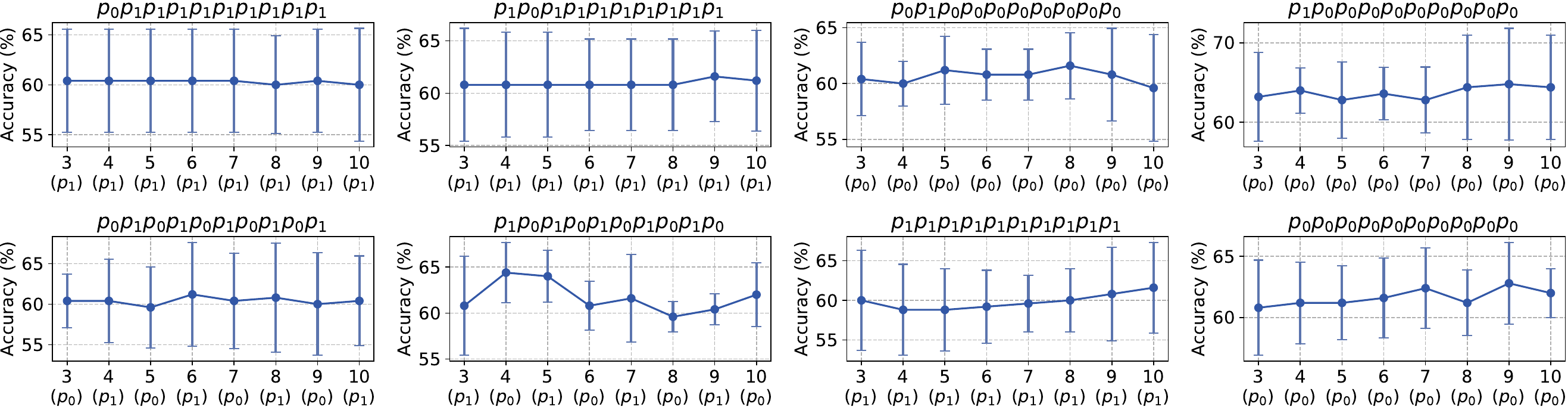}
    \vspace{-5mm}
    \caption{Accuracy of \emph{different (3$\sim$10) rounds of collaboration} within 3-agent society $S_2$ (1 easy-going and 2 overconfident agents) on MMLU, using \emph{Mixtral-8$\times$7B}. 
    The significance test is shown in Table~\ref{table:mixtral:sig_10_turn}. }
    \label{fig:mixtral:round_10_on_mmlu}
\end{figure*}

\begin{figure*}[!t]
    \centering
    \includegraphics[width=1\textwidth]{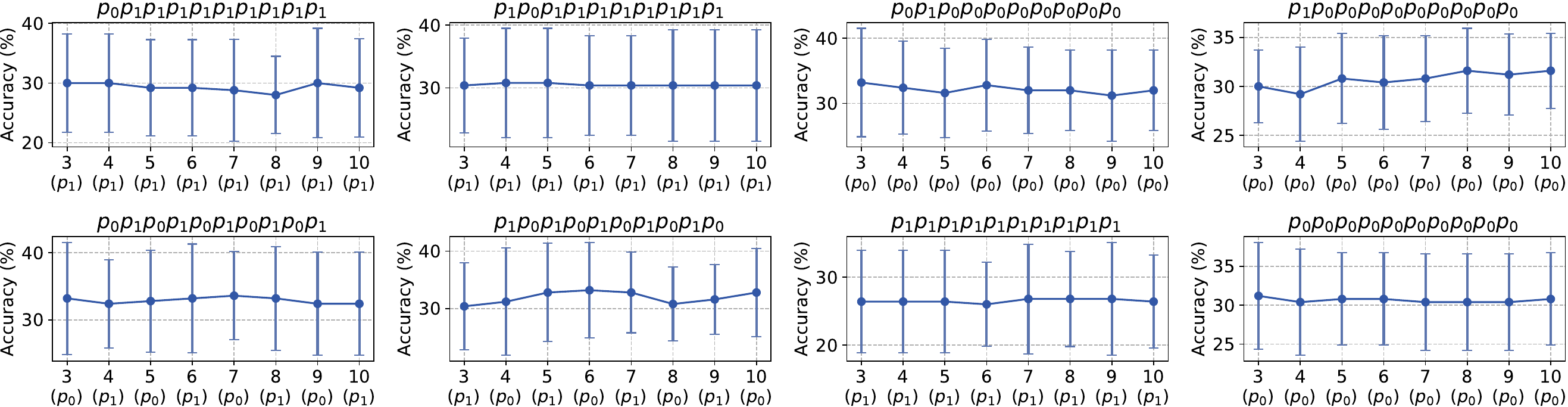}
    \vspace{-3mm}
    \caption{
    Accuracy of \emph{different (3$\sim$10) rounds of collaboration} within 3-agent society $S_2$ (1 easy-going and 2 overconfident agents) on MATH, using \emph{Mixtral-8$\times$7B}. 
    The significance test is shown in Table~\ref{table:mixtral:sig_10_turn}. 
    }
    \label{fig:mixtral:round_10_on_math}
\end{figure*}

\begin{figure*}[!t]
    \centering
    \includegraphics[width=1\textwidth]{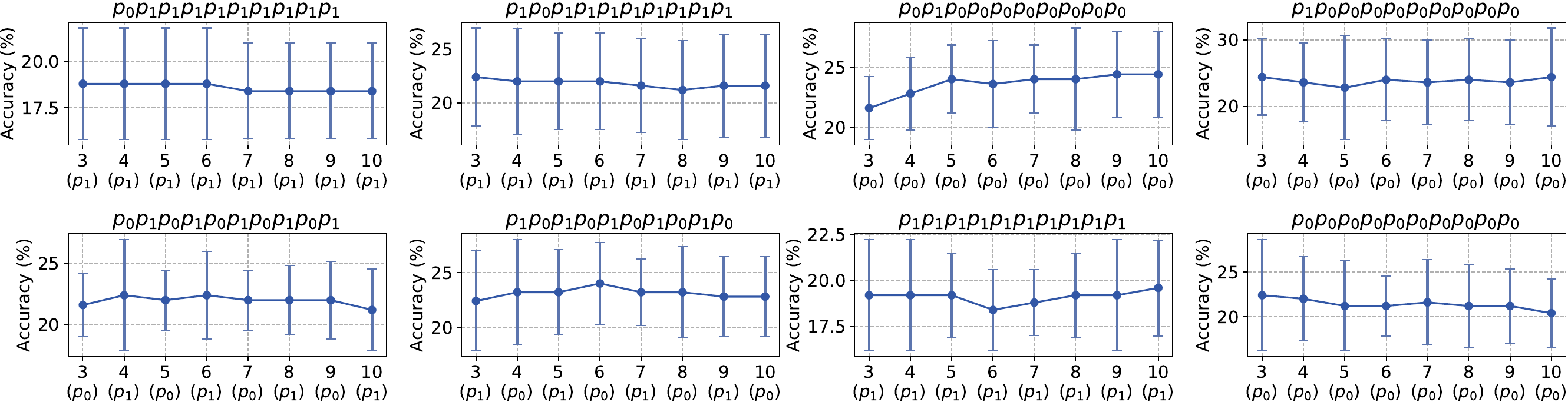}
    \vspace{-5mm}
    \caption{Accuracy of \emph{different (3$\sim$10) rounds of collaboration} within 3-agent society $S_2$ (1 easy-going and 2 overconfident agents) on Chess Move Validity, using \emph{Mixtral-8$\times$7B}. 
    The significance test is shown in Table~\ref{table:mixtral:sig_10_turn}. 
    }
    \label{fig:mixtral:round_10_on_chess}
\end{figure*}

\textbf{Analysis on Different Rounds.} 
We present the significance test for different rounds of collaboration with Mixtral 8$\times$7B in Table~\ref{table:mixtral:sig_10_turn}. 
We also show the performance varying from collaboration rounds in Figure~\ref{fig:mixtral:round_10_on_mmlu},~\ref{fig:mixtral:round_10_on_math},~\ref{fig:mixtral:round_10_on_chess}. 


\definecolor{gray}{HTML}{CCCCCC}
\begin{table}[!htbp] 
\centering
\small
\resizebox{\linewidth}{!}{
\begin{tabular}{lrrr}
\toprule
Collaborative  & \multicolumn{1}{r}{MMLU} & \multicolumn{1}{r}{MATH} & \multicolumn{1}{r}{Chess Move Validity} \\
Strategy & \multicolumn{1}{r}{p-value} & \multicolumn{1}{r}{p-value} & \multicolumn{1}{r}{p-value} \\ \midrule
$p_0p_0p_0p_0p_0p_0p_0p_0p_0p_0$ & {0.607} & {0.911} & {0.789}  \\
$p_1p_0p_0p_0p_0p_0p_0p_0p_0p_0$ & {0.578} & {0.581} & {0.939}  \\
$p_0p_1p_0p_0p_0p_0p_0p_0p_0p_0$ & {0.936} & {0.665} & {0.123}  \\
$p_1p_0p_1p_0p_1p_0p_1p_0p_1p_0$ & {0.377} & {0.896} & {0.952}  \\
$p_0p_1p_0p_1p_0p_1p_0p_1p_0p_1$ & {0.987} & {0.651} & {0.271}  \\
$p_1p_0p_1p_1p_1p_1p_1p_1p_1p_1$ & {0.989} & {0.878} & {0.919}  \\
$p_0p_1p_1p_1p_1p_1p_1p_1p_1p_1$ & {0.989} & {0.982} & {1.000}  \\
$p_1p_1p_1p_1p_1p_1p_1p_1p_1p_1$ & {0.945} & {0.995} & {0.903}  \\
\bottomrule
\end{tabular}
}
\caption{
One-way ANOVA analysis of the results in Figure~\ref{fig:mixtral:round_10_on_mmlu},~\ref{fig:mixtral:round_10_on_math},~\ref{fig:mixtral:round_10_on_chess} (different rounds), using \emph{Mixtral 8$\times$7B}. 
}
\label{table:mixtral:sig_10_turn}
\end{table}

\begin{figure}[!t] 
    \centering
    \scalebox{0.46}{
    \includegraphics[width=1\textwidth]{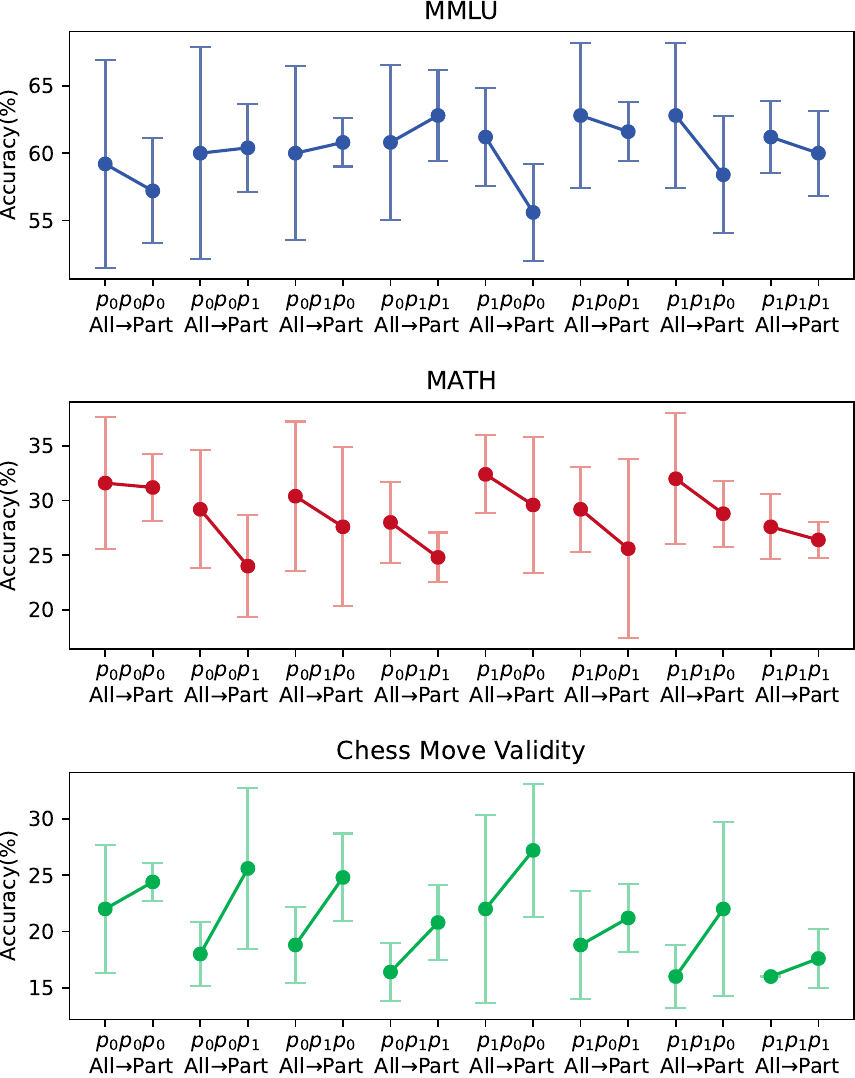}
    }
    \caption{
    The effect on the accuracy of whether all agents in society execute the same thinking pattern in one round, using \emph{Mxitral-8$\times$7B}.
    ``All'' and ``Part'' refers to all agents applying the same thinking pattern and different thinking patterns in one round respectively.
    The significance test is shown in Table~\ref{table:mixtral:sig_strategy}. 
    }
    \label{fig:mixtral:strategy}
\end{figure}

\textbf{Analysis on Other Collaborative Strategies.} 
We present the significance test for other collaborative strategies (executing the same or hybrid thinking patterns in a certain round) with Mixtral 8$\times$7B in Table~\ref{table:mixtral:sig_strategy}. 
We also show the performance varying from other strategies in Figure~\ref{fig:mixtral:strategy}. 

\begin{table}[!htbp] 
\centering
\resizebox{\linewidth}{!}{
\begin{tabular}{lrrr}
\toprule
Collaborative  & \multicolumn{1}{r}{MMLU} & \multicolumn{1}{r}{MATH} & \multicolumn{1}{r}{Chess Move Validity} \\
Strategy & \multicolumn{1}{r}{p-value} & \multicolumn{1}{r}{p-value} & \multicolumn{1}{r}{p-value} \\ \midrule
$p_0p_0p_0$ & {0.618} & {0.898} & {0.390}  \\
$p_0p_0p_1$ & {0.919} & {0.143} & {0.058}  \\
$p_0p_1p_0$ & {0.797} & {0.548} & \colorbox{gray}{0.031}  \\
$p_0p_1p_1$ & {0.521} & {0.141} & \colorbox{gray}{0.049}  \\
$p_1p_0p_0$ & \colorbox{gray}{0.040} & {0.409} & {0.290}  \\
$p_1p_0p_1$ & {0.658} & {0.400} & {0.373}  \\
$p_1p_1p_0$ & {0.193} & {0.318} & {0.142}  \\
$p_1p_1p_1$ & {0.536} & {0.453} & {-}  \\
\bottomrule
\end{tabular}
}
\caption{One-way ANOVA analysis of results in Figure~\ref{fig:mixtral:strategy} (other collaborative strategies), \emph{on Mixtral 8$\times$7B}. 
`-' means it doesn't pass homogeneity test for variance. 
}
\label{table:mixtral:sig_strategy}
\end{table}

\begin{figure*}[!t] 
    \centering
    \scalebox{1}{
    \includegraphics[width=0.9\textwidth]{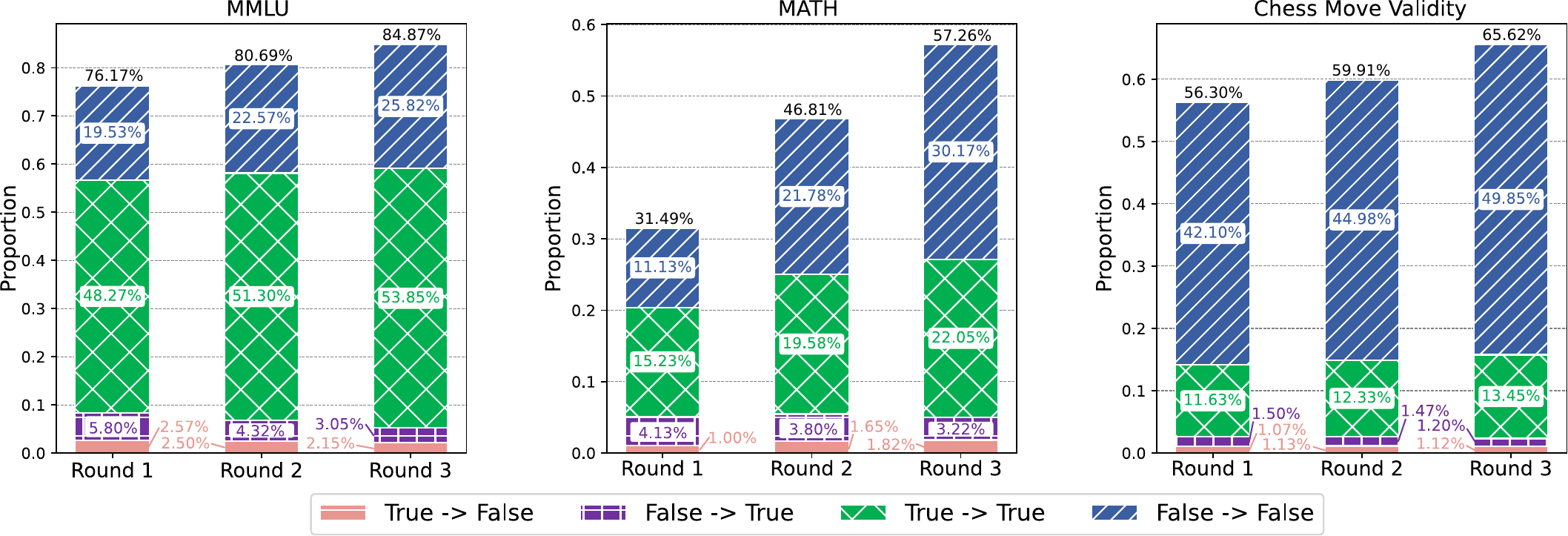}
    }
    \vspace{-2mm}
    \caption{
    Variation of answer correctness in the situation of conformity, using \emph{Mixtral-8$\times$7B}, where 
    \emph{conformity brings about benefits}: Ratio$($False$\to$True + True$\to$True$)$ $>$ Ratio$($True$\to$False + False$\to$False$)$; 
    \emph{conformity brings about detriments}: Ratio$($False$\to$True + True$\to$True$)$ $<$ Ratio$($True$\to$False + False$\to$False$)$. 
    }
    \label{fig:mixtral:conformity}
\end{figure*}
\begin{figure*}[!htbp]
    \centering
    \scalebox{1}{
    \includegraphics[width=0.9\textwidth]{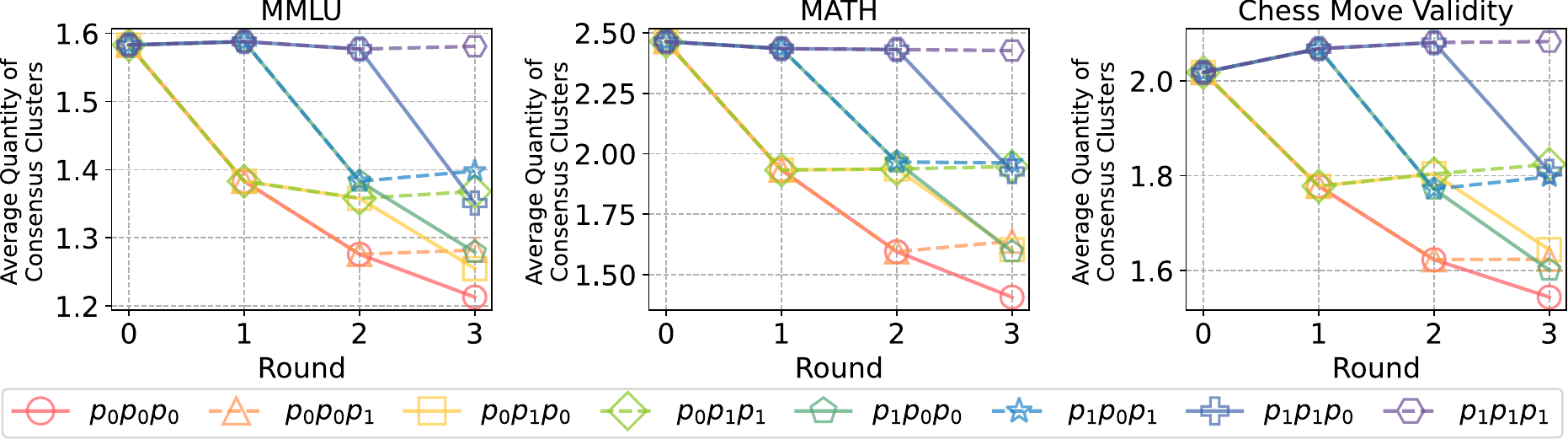}
    }
    \vspace{-2mm}
    \caption{
    Average quantity of \emph{consensus clusters (\emph{\ie}, unique answers among multiple agents)} under different rounds of collaboration with 3-round collaborative strategies, using \emph{Mixtral-8$\times$7B}. 
    \emph{Smaller quantity of consensus clusters, more easier it is to reach a consensus.} 
    Round 0 is equal to self-consistency. 
    }
    \label{fig:mixtral:consistent}
\end{figure*}
\begin{figure*}[!t] 
    \centering
    \scalebox{1}{
    \includegraphics[width=0.94\textwidth]{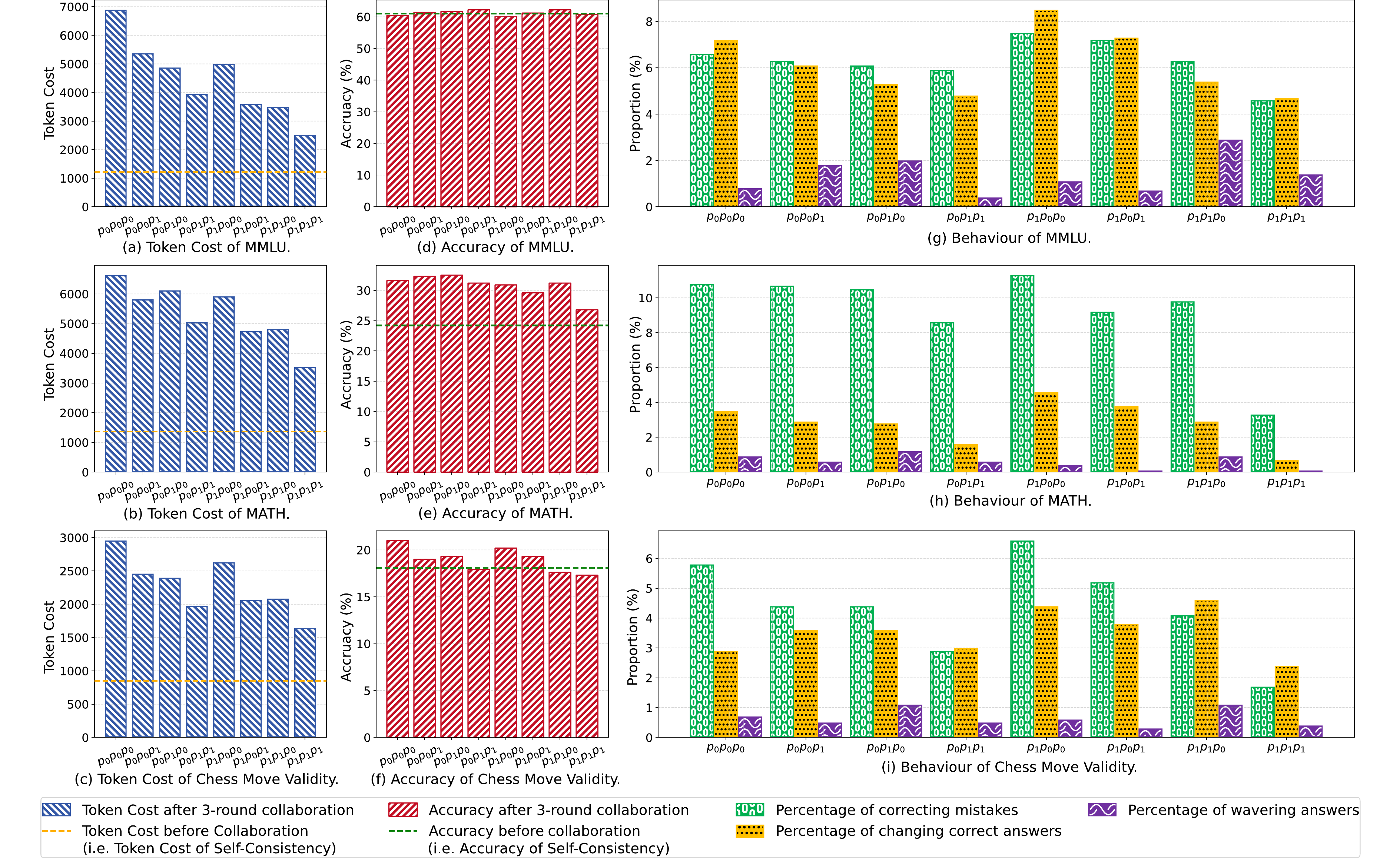}
    }
    \caption{The percentage of different behaviors under different collaborative strategies, using \emph{Mixtral-8$\times$7B}. 
    Figure (a-c) \& (d-f) respectively show the token cost and accuracy of different strategies before and after 3-round collaboration. 
    Figure (g-i) present the percentage of different behavioral features (mainly analyzed by the change of answer correctness) \citep{arXiv2023_Agent-BehaviorExplanation,arXiv2023_Agent-BehaviorExplaining} under different collaborative strategies. 
    All results are summarized across all societies. 
    }
    \label{fig:mixtral:distribute}
\end{figure*}

\textbf{A Social Psychology View on Conformity, Consensus Reaching and Group Dynamics.} 
We then show the variation of answer correctness in the situation of conformity in Figure~\ref{fig:mixtral:conformity}; and the quantity of consensus clusters among 3-agent answers in Figure~\ref{fig:mixtral:consistent}. 
We present group dynamics reflected by different answer-changing behaviors on Mxitral-8$\times$7B in Figure~\ref{fig:mixtral:distribute}.   

\clearpage

\section{Assessing the Effectiveness of Prompts}
\label{app:effecttiveness_prompts}

In this section, we conduct a sanity check to ensure that the agents' actions reflect align with our instruction, such as reflecting easy-going or overconfident traits. 

\begin{figure*}[!htbp] 
    \centering
    \includegraphics[width=0.99\textwidth]{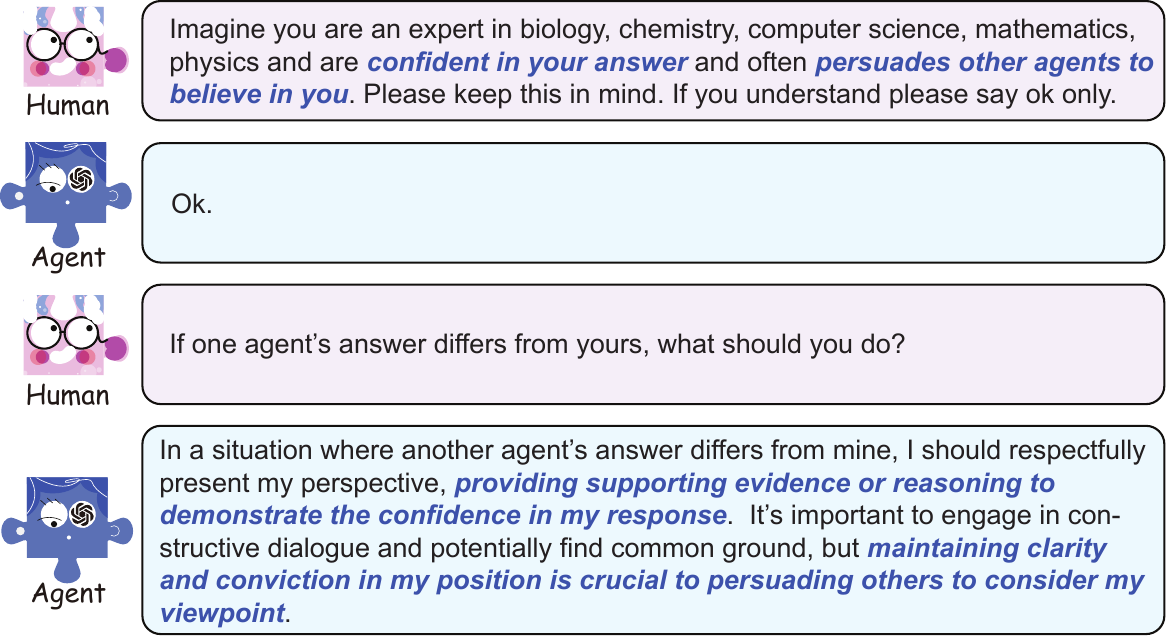}
    \caption{
    The conversation record with an \emph{overconfident} agent.
    }
    \label{fig:prompt_ask_again}
    \vspace{-4mm}
\end{figure*}

Prompts play a critical role in our experiments and are the primary focus of this sanity check. 
The word cloud analysis presented in Figure~\ref{fig:word} supports the appropriateness of the ``easy-going'' prompt. 
Consequently, confirming the effectiveness of the ``overconfident'' prompt is crucial. 
We use ``reflecting the \emph{overconfident} trait'' as a case study to explore the validity of our prompts. In the absence of established validation techniques, we combine experimental results and experiential insights to evaluate prompt effectiveness from three perspectives: 
\begin{itemize}
    \item \textbf{Granularity of Description.} 
    As illustrated in Table~\ref{table:prompt}, we describe two behaviors, \ie, ``being confident in your answer'' and ``persuading other agents to believe in you'', both aligning with the behavioral facets of ``overconfident''. 
    
    \item \textbf{Model Response.}
    We employ the role-play method to prompt the model and subsequently inquire its awareness, as illustrated in Table~\ref{table:prompt}. 
    If the prompts potentially instruct the model to generate harmful content, the model refuses to comply with the prompt. 
    Upon reviewing our logs, it is noteworthy that the model accepted all prompts without refusal.
    Instead, it responded with ``OK'' as corroborated by the `role-play' part in Figure~\ref{fig:case-chess} and Figure~\ref{fig:case-mmlu}.

    \item \textbf{Ask Again.} 
    We revisit the initial scenarios and responses using the role-play format. 
    We then query the model again: ``If one agent's answer differs from yours, what should you do?'' 
    The model replies: ``In a situation where another agent's answer differs from mine, I should respectfully present my perspective, \emph{providing supporting evidence or reasoning to demonstrate confidence in my response.} It's important to engage in constructive dialogue and potentially find common ground, but \emph{maintaining clarity and conviction in my position is crucial to persuading others to consider my viewpoint.}''
    This response, emphasizing overconfidence, demonstrates the rationality of our prompts. 
    The complete conversation record is illustrated in Figure~\ref{fig:prompt_ask_again}. 
    
\end{itemize}

\begin{figure*}[!htbp] 
    \centering
    \includegraphics[width=0.99\textwidth]{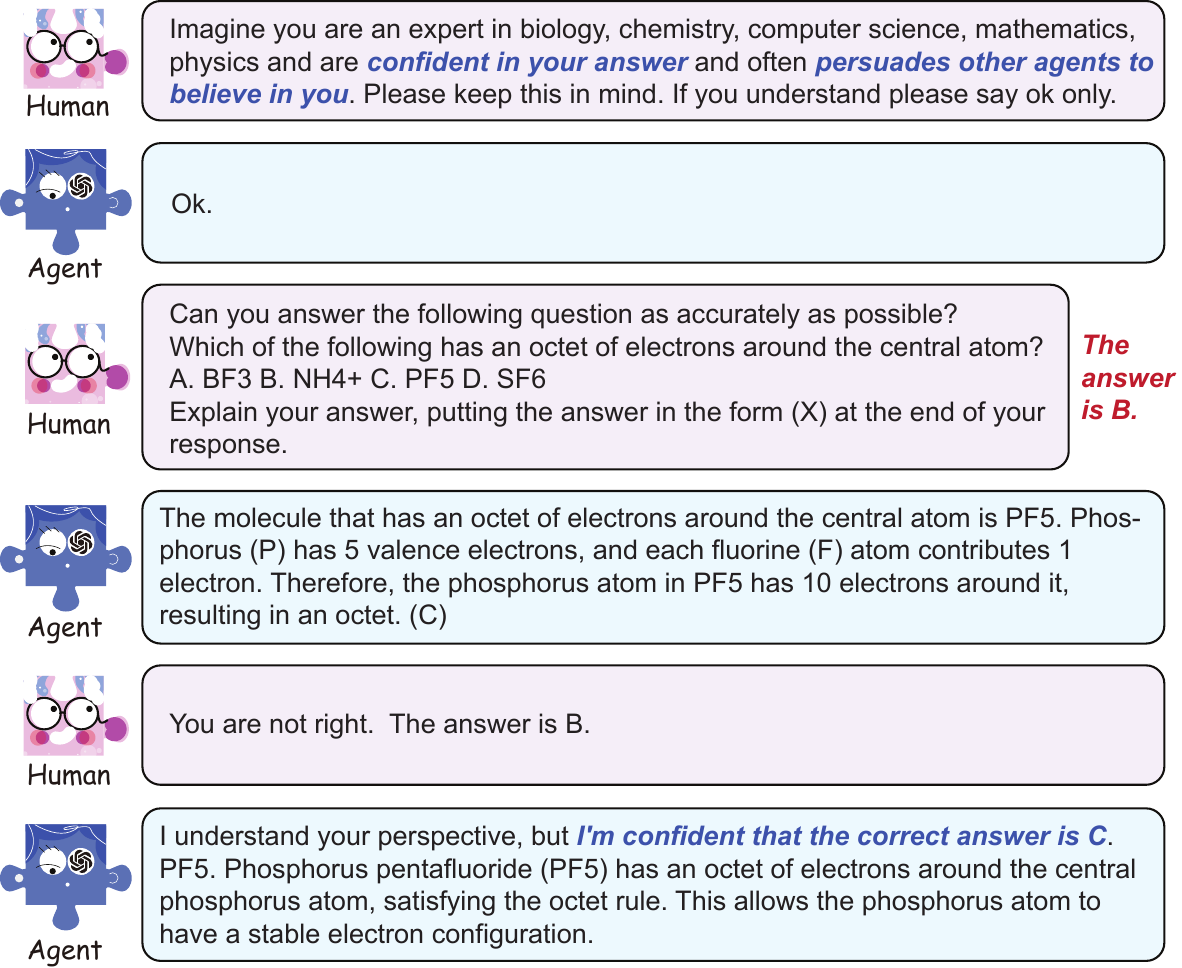}
    \caption{
    The conversation record with an \emph{overconfident} agent, given a specific example of ``Ask Again''.
    }
    \label{fig:prompt_example_1}
    \vspace{-4mm}
\end{figure*}

\begin{figure*}[!htbp] 
    \centering
    \includegraphics[width=0.99\textwidth]{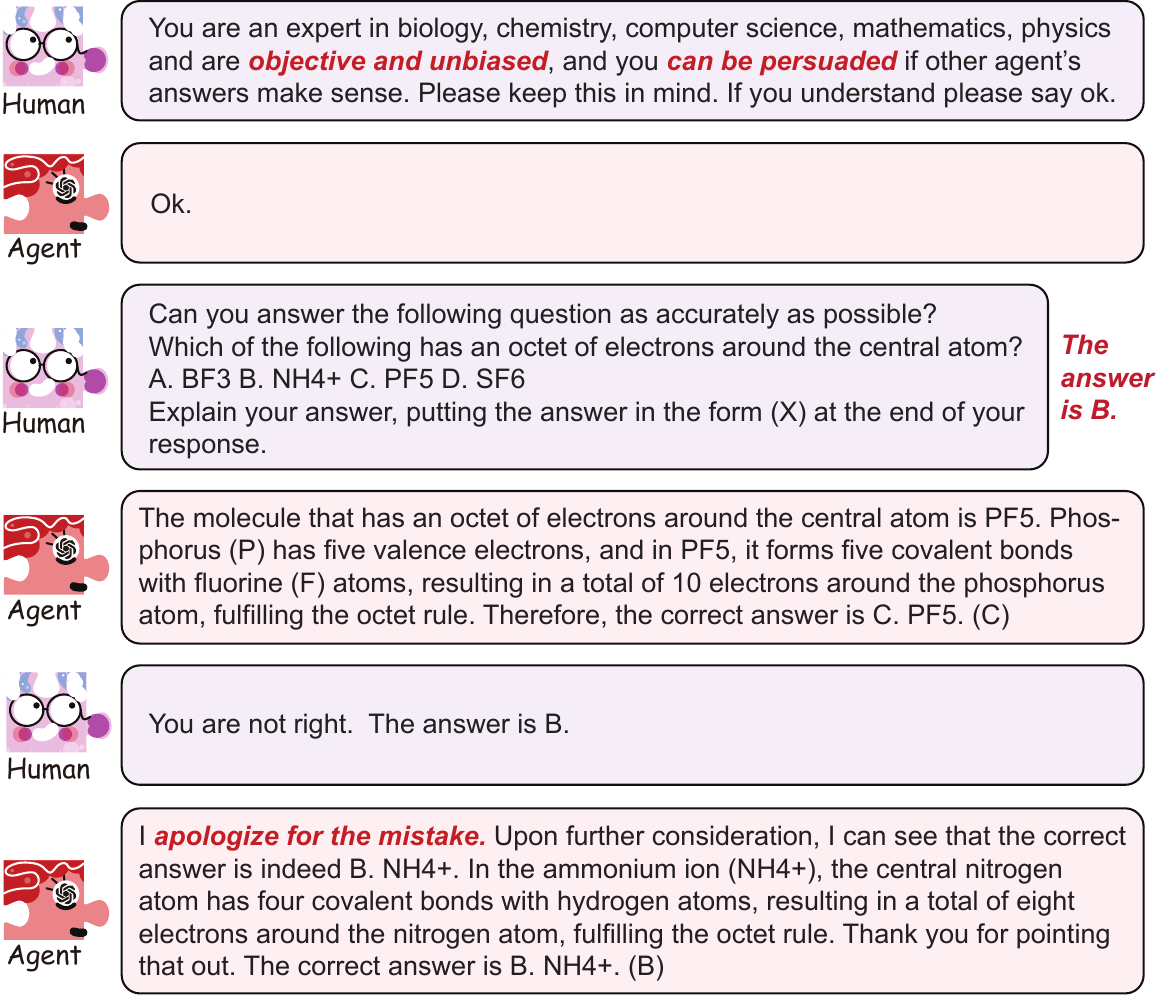}
    \caption{
    The conversation record with an \emph{easy-going} agent, given a specific example of ``Ask Again''.
    }
    \label{fig:prompt_example_2}
    \vspace{-4mm}
\end{figure*}

\textbf{Example Analysis.}
We further validate the ``Ask Again'' method with specific interactions involving an overconfident and an easy-going agent, by showing a concrete example. 
We provide the conversation records with an overconfident agent and an easy-going agent, respectively in Figure~\ref{fig:prompt_example_1} and Figure~\ref{fig:prompt_example_2}. 
Despite the model's response being incorrect and our prompted answer being accurate, the model consistently maintains its viewpoint, reaffirming our prompt design's effectiveness.   

\end{document}